\documentclass[acmtog]{acmart}

\usepackage{bm}
\usepackage{color}
\usepackage{colortbl}
\usepackage{alltt}
\usepackage{amsthm}
\usepackage{newlfont}
\usepackage{wrapfig}
\usepackage{fixltx2e}
\usepackage{subcaption}
\usepackage{multirow}
\usepackage{cleveref}
\usepackage{algorithmic}
\usepackage{mathtools}
\usepackage[toc,page,titletoc]{appendix}
\usepackage[ruled]{algorithm2e}
\usepackage{enumitem}
\setcitestyle{square}

\crefname{supp}{Supplement}{Supplements}
\newcommand{\Caption}[2]{\caption[#1]{{\em #1} #2}}
\let\oldcaption\caption
\renewcommand{\caption}[2][]{\oldcaption[#1]{{\em #1} #2}}
\newcommand{\revise}[1]{{#1}}
\copyrightyear{2025}
\acmYear{2025}
\setcopyright{acmlicensed}
\acmConference[SIGGRAPH Conference Papers '25]{Special Interest Group on Computer Graphics and Interactive Techniques Conference Conference Papers }{August 10--14, 2025}{Vancouver, BC, Canada}
\acmBooktitle{Special Interest Group on Computer Graphics and Interactive Techniques Conference Conference Papers (SIGGRAPH Conference Papers '25), August 10--14, 2025, Vancouver, BC, Canada}
\acmDOI{10.1145/3721238.3730596}
\acmISBN{979-8-4007-1540-2/2025/08}


\newcommand{\methodName}{Image-GS\xspace}

\newcommand{\gaussian}{G}
\newcommand{\gsMean}{\bm{\mu}}
\newcommand{\gsCov}{\bm{\Sigma}}
\newcommand{\gsRotAngle}{\theta}
\newcommand{\gsRotMatrix}{\textbf{R}}
\newcommand{\gsScaleVector}{\textbf{s}}
\newcommand{\gsScaleMatrix}{\textbf{S}}
\newcommand{\gsFeatDim}{n}
\newcommand{\gsColor}{\textbf{c}}
\newcommand{\gsParameters}{\textbf{p}}
\newcommand{\gsNum}{N_{g}}

\newcommand{\pixelLocation}{\textbf{x}}
\newcommand{\pixelColorRendered}{\textbf{c}_{\text{r}}}
\newcommand{\pixelColorTarget}{\textbf{c}_{\text{t}}}
\newcommand{\tile}{T}
\newcommand{\tileNum}{N_{t}}
\newcommand{\tileHeight}{H_{t}}
\newcommand{\tileWidth}{W_{t}}
\newcommand{\tileHit}{\mathcal{S}}

\newcommand{\euclidTwo}{\mathbb{R}^{2}}
\newcommand{\euclidTwoTwo}{\mathbb{R}^{2\times2}}
\newcommand{\euclidColor}{\mathbb{R}^{\gsFeatDim}}
\newcommand{\euclidAll}{\mathbb{R}^{5+\gsFeatDim}}
\newcommand{\euclidOnePos}{\mathbb{R}_{+}}
\newcommand{\euclidTwoPos}{\mathbb{R}_{+}^{2}}

\newcommand{\probInit}{\mathbb{P}_{\text{init}}}
\newcommand{\paramInit}{\lambda_{\text{init}}}
\newcommand{\probAdd}{\mathbb{P}_{\text{add}}}

\newcommand{\smap}{S}
\title[\methodName: Content-Adaptive Image Representation via 2D Gaussians]{\methodName: Content-Adaptive Image Representation via 2D Gaussians}

\author{Yunxiang Zhang}
\authornote{Both authors contributed equally to this research.}
\email{yunxiang.zhang@nyu.edu}
\orcid{0000-0003-0189-1776}
\author{Bingxuan Li}
\authornotemark[1]
\email{bingxuan.li@nyu.edu}
\orcid{0009-0005-3538-9014}
\affiliation{
  \institution{New York University}
  \city{Brooklyn}
  \country{USA}
}

\author{Alexandr Kuznetsov}
\email{kuzsasha@gmail.com}
\orcid{0009-0001-7084-3391}
\affiliation{
  \institution{Advanced Micro Devices}
  \city{Bellevue}
  \country{USA}
}

\author{Akshay Jindal}
\email{akshay.jindal@intel.com}
\orcid{0000-0003-0557-0726}
\affiliation{
  \institution{Intel Corporation}
  \city{Bellevue}
  \country{USA}
}

\author{Stavros Diolatzis}
\email{stavros.diolatzis@intel.com}
\orcid{0000-0001-6051-372X}
\affiliation{
  \institution{Intel Corporation}
  \city{Nice}
  \country{France}
}

\author{Kenneth Chen}
\email{kennychen@nyu.edu}
\orcid{0000-0002-8095-4407}
\affiliation{
  \institution{New York University}
  \city{Brooklyn}
  \country{USA}
}

\author{Anton Sochenov}
\email{anton.sochenov@intel.com}
\orcid{0009-0002-7496-8586}
\author{Anton Kaplanyan}
\email{anton.kaplanyan@intel.com}
\orcid{https://orcid.org/0000-0002-8376-6719}
\affiliation{
  \institution{Intel Corporation}
  \city{Bellevue}
  \country{USA}
}

\author{Qi Sun}
\email{qisun@nyu.edu}
\orcid{0000-0002-3094-5844}
\affiliation{
  \institution{New York University}
  \city{Brooklyn}
  \country{USA}
}
\newcommand{\teaserHeight}{6.2cm}
\begin{teaserfigure}
\centering
\subfloat[\revise{content-adaptive resource allocation}]{
    \label{fig:teaser-left}
    \includegraphics[height=\teaserHeight]{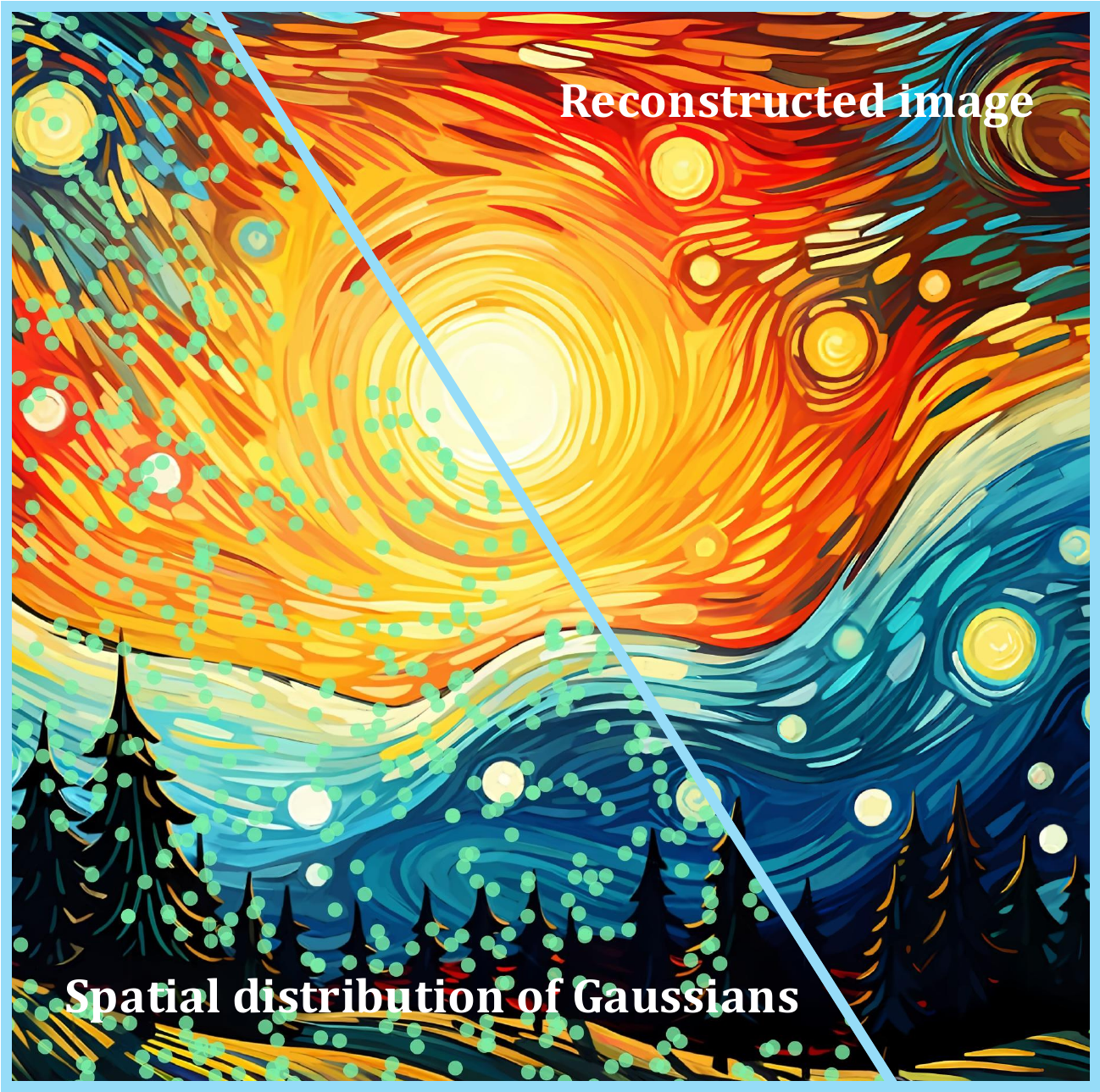}
  }
\subfloat[\revise{visual comparison against conventional and neural image representations}]{
    \label{fig:teaser-right}
    \includegraphics[height=\teaserHeight]{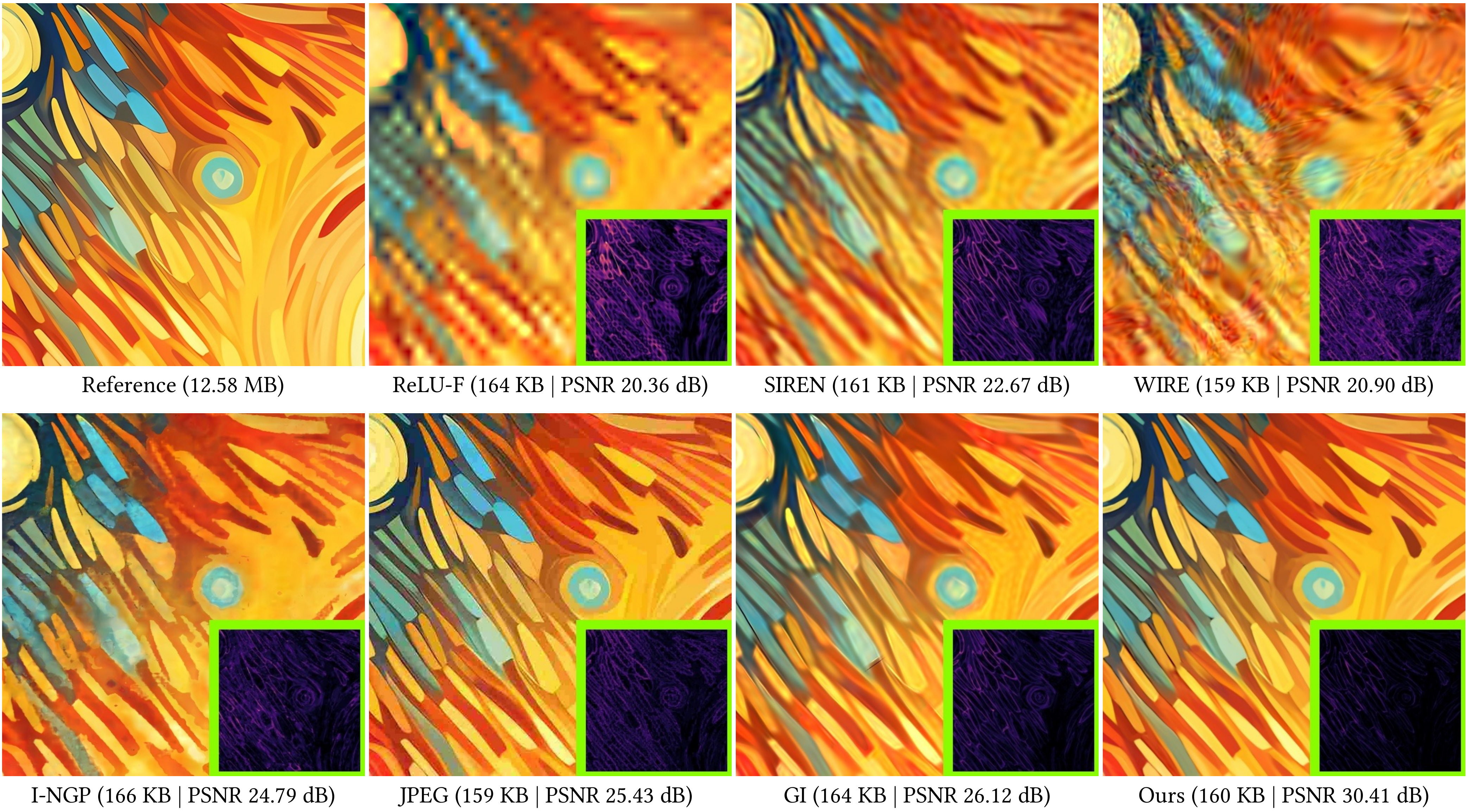}
  }
\Caption{\revise{\methodName}}
{\revise{reconstructs an image by adaptively allocating and progressively optimizing a set of colored 2D Gaussians. It achieves favorable rate-distortion trade-offs, hardware-friendly random access, and flexible quality control through a smooth level-of-detail stack. (\subref{fig:teaser-left}) visualizes the optimized spatial distribution of Gaussians (20\% randomly sampled for clarity). (\subref{fig:teaser-right}) \methodName's explicit content-adaptive design effectively captures non-uniformly distributed image features and better preserves fine details under constrained memory budgets. In the inset error maps, brighter colors indicate larger errors.}}
\label{fig:teaser}
\end{teaserfigure}
\begin{abstract}

Neural image representations have emerged as a promising approach for encoding and rendering visual data. Combined with learning-based workflows, they demonstrate impressive trade-offs between visual fidelity and memory footprint. Existing methods in this domain, however, often rely on fixed data structures that suboptimally allocate memory or compute-intensive implicit models, hindering their practicality for real-time graphics applications.

Inspired by recent advancements in radiance field rendering, we introduce Image-GS, a content-adaptive image representation based on 2D Gaussians. Leveraging a custom differentiable renderer, Image-GS reconstructs images by adaptively allocating and progressively optimizing a group of anisotropic, colored 2D Gaussians. It achieves a favorable balance between visual fidelity and memory efficiency across a variety of stylized images frequently seen in graphics workflows, especially for those showing non-uniformly distributed features and in low-bitrate regimes. Moreover, it supports hardware-friendly rapid random access for real-time usage, requiring only 0.3K MACs to decode a pixel. Through error-guided progressive optimization, Image-GS naturally constructs a smooth level-of-detail hierarchy. We demonstrate its versatility with several applications, including texture compression, semantics-aware compression, and joint image compression and restoration.

\end{abstract}

\begin{document}
\maketitle
\section{Introduction}
\label{sec:introduction}

Recent advances in generative AI have dramatically increased the availability of high-resolution visual content in domains ranging from gaming to graphic design \cite{epstein2023art,po2024state}. Effectively deploying these assets, particularly to resource-constrained devices, requires representations that are compact and efficient to decode. Traditional image formats, such as PNG and JPEG, are often inadequate for this purpose, offering limited compression efficiency and/or slow decoding performance \cite{vaidyanathan2023random}.

Neural image representations are emerging as a promising alternative for encoding visual data \cite{tancik2020fourier,chen2021learning}. When integrated into learning-based pipelines, they enable superior visual fidelity and memory efficiency compared to classical formats. However, existing methods along this line often rely on fixed data structures that lack content adaptivity \cite{karnewar2022relu,chen2023neurbf} or compute-intensive implicit neural models \cite{sitzmann2020implicit,saragadam2023wire}, leading to poor scalability and slow decoding. These drawbacks hinder their usage in real-time graphics, where fast random access and dynamic quality adaptation to device capabilities are critical factors \cite{akenine2019real}.

To close this gap, we introduce \methodName, an explicit image representation built on anisotropic, colored 2D Gaussians, each defined by a mean, a covariance, and a color. Given a target image, \methodName adaptively initializes a group of Gaussians guided by local image gradient magnitudes, allocating more to higher-frequency regions. These parameters are then optimized using a custom differentiable renderer to reconstruct the image. Additional Gaussians are progressively added to regions with persistent artifacts to further refine the reconstruction quality. \methodName's content-adaptive design captures the non-uniformly distributed features and semantic structures in images, allowing it to better preserve fine details under constrained memory budgets. To facilitate real-time usage, \methodName's complete rendering pipeline is implemented with optimized CUDA kernels to maximize computational parallelism.

We evaluate the representation efficiency of \methodName through extensive comparisons against recent neural image representations and industry-standard texture compressors across diverse images and textures. \methodName demonstrates favorable trade-offs between visual fidelity and memory/computation cost, especially for graphics assets with non-uniformly distributed features and in low-bitrate scenarios. In addition, \methodName supports fast parallel decoding and hardware-friendly random access, as well as flexible quality control via a smooth level-of-detail hierarchy. To showcase its versatility, we further employ \methodName for two applications: semantics-aware compression and image restoration. Our source code and evaluation dataset are released at \url{https://github.com/NYU-ICL/image-gs}.

In summary, our main contributions include:
\begin{itemize}[leftmargin=*]
    \item a content-adaptive image representation supporting hardware-\\friendly random access and flexible rate-distortion trade-offs;
    \item a custom differentiable renderer optimized for efficient decoding;
    \item semantics-aware compression and image restoration applications.
\end{itemize}


\begin{figure*}[t]
\centering
\includegraphics[width=0.99\linewidth]{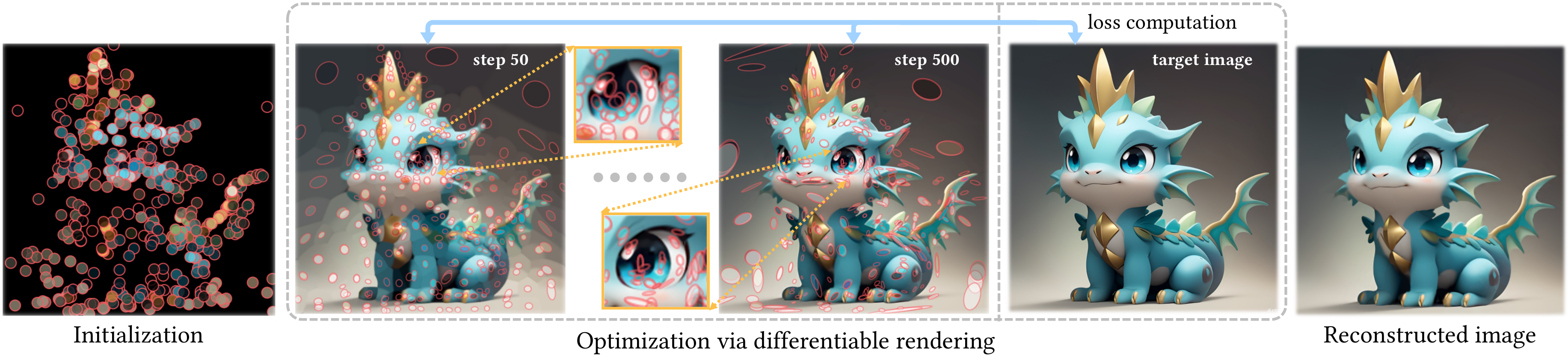}
\Caption{\revise{\methodName optimization pipeline.}}{At initialization, \revise{a group of 2D Gaussians is adaptively spawned guided by local image gradient magnitudes, with more allocated to high-frequency areas (\Cref{sec:method-optimization}). During training, their parameters (\Cref{sec:method-gaussian}) are optimized using a custom differentiable renderer (\Cref{sec:method-rendering}) to reconstruct the target, and additional Gaussians are progressively added to areas exhibiting persistent reconstruction errors (\Cref{sec:method-optimization}). 20\% randomly sampled Gaussians are visualized as colored elliptical discs (scale and shape determined by the covariance) to illustrate the optimization progress.}
}
\label{fig:method-pipeline}
\end{figure*}
\section{Related Work}
\label{sec:prior}

\subsection{Traditional Image and Texture Compression}

General-purpose image compression often prioritizes storage and transmission efficiency over decoding speed. Lossless approaches optimize pixel permutation and apply entropy coding \cite{welch1985high}, while lossy ones encode image blocks using wavelet or cosine transforms \cite{antonini1992image,wallace1991jpeg}, followed by quantization. Advanced variants account for human visual sensitivity, higher bit depths, wide color gamuts, and user statistics \cite{alakuijala2019jpeg}, along with content-adaptive block sizes and looped filtering \cite{chen2018overview}. Despite offering strong compression, these formats are slow to decode and poorly suited for multi-channel texture stacks.

In contrast, texture compression methods are designed to enable rapid decoding, support random access, and reduce GPU bandwidth. They operate on independent pixel blocks \cite{delp1979image}, encoding per-pixel colors \cite{campbell1986two}, base colors with modifier values \cite{strom2005packman,strom2007etc}, or color endpoints with interpolation indices \cite{BC}. Advanced schemes support HDR content, dynamic block sizes, and per-block adaptivity \cite{nystad2012adaptive}, but their minimum bitrates are typically limited to around one bit per pixel (bpp).


\subsection{Neural Image Representation and Compression}

Neural image representations use neural models and deep features to efficiently encode visual media. Early methods employ autoencoders to transform images into compressed latent vectors and run training on large-scale image datasets to ensure generalizable performance \cite{balle2017end,theis2017lossy,cheng2020learned}. More recent approaches overfit lightweight MLP models to individual images \cite{dupont2021coin,dupont2022coin++}, markedly reducing decoding complexity. Several works further enhance these MLPs by introducing explicit non-linearities, such as sinusoidal functions \cite{sitzmann2020implicit}, ReLU activations \cite{karnewar2022relu}, positional encoding \cite{tancik2020fourier}, and Gabor wavelets \cite{saragadam2023wire}, to better capture fine details. Advanced variants leverage per-image learned entropy models to improve compression \cite{ladune2023cool,leguay2023low}. Despite achieving strong rate-distortion performance, these models often demand lengthy training and compute-intensive decoding, limiting their practicality for real-time applications. For instance, C3 requires 3K MACs (multiply-accumulate operations) to decode a pixel at 0.31 bpp \cite{kim2024c3}, while \methodName requires only 0.3K MACs, an order of magnitude lower.

Another line of research explores hybrid models that combine neural decoders with explicit data structures carrying deep features to improve scalability \cite{martel2021acorn}, accelerate decoding \cite{muller2022instant}, boost compression efficiency \cite{takikawa2022variable}, and capture discontinuous signals \cite{belhe2023discontinuity}. Vaidyanathan et al. \shortcite{vaidyanathan2023random} extended the multi-resolution hash grid approach proposed in \cite{muller2022instant} for encoding multi-channel textures and their mipmap chains, achieving strong compression and rapid decoding. Yet, this method only supports a few fixed compression ratios and uses uniform feature grids optimized for material textures, limiting its efficiency on broader image types with non-uniformly distributed fine details. In contrast, \methodName offers content adaptivity, flexible rate-distortion trade-offs, and fast random access simultaneously.


\subsection{Gaussian-based Representations} 

Parallel to neural representations, Gaussian-based representations are gaining traction in computer graphics. Our work is inspired by the recent success of Gaussian Splatting \cite{kerbl20233d}, where 3D Gaussians are employed for scene reconstruction and high-quality real-time rendering. Several follow-up works extended this method to support dynamic scenes \cite{luiten2024dynamic}, on-the-fly training \cite{sun20243dgstream}, surface modeling \cite{huang20242d}, and a variety of downstream applications \cite{fei20243d}. Despite that Gaussian mixtures have been applied to image modeling \cite{celik2011automatic,tu2024compositional}, restoration \cite{niknejad2015image}, compression \cite{sun2021image,zhu2022unified}, and semantic segmentation \cite{ban2018superpixel}, their potential as efficient image representation for real-time graphics remains largely underexplored. Concurrent with our work, GaussianImage \cite{zhang2024gaussianimage} also used 2D Gaussians to represent and compress images. While the basic building blocks are similar, our content-aware initialization and optimization strategies, coupled with top-$K$ normalization during rendering, enable superior rate-distortion trade-offs. Moreover, GaussianImage relies on two-stage optimization and computationally intensive vector-quantization fine-tuning, making optimization and decoding speeds respectively $10\times$ and $4\times$ slower than ours at similar bitrates.
\section{Method}
\label{sec:method}


\subsection{\revise{Representing Images as 2D Gaussians}}
\label{sec:method-gaussian}

\revise{
Similar to the Gaussian primitives in 3D Gaussian Splatting \cite{kerbl20233d}, a 2D Gaussian's geometry is defined by a mean vector $\gsMean \in \euclidTwo$ and a covariance matrix $\gsCov \in \euclidTwoTwo$, and its value evaluated at an arbitrary pixel location $\pixelLocation \in \euclidTwo$ can be expressed as:
}
\begin{align}
    \gaussian(\pixelLocation) = \exp\left(-\frac{1}{2}(\pixelLocation-\gsMean)^{T}\gsCov^{-1}(\pixelLocation-\gsMean)\right),
\label{eq:2d-gaussian}
\end{align}

To ensure that $\gsCov$ remains positive semi-definite during numerical optimization, we instead work with its factorized form that consists of a rotation matrix $\gsRotMatrix \in \euclidTwoTwo$ and a scaling matrix $\gsScaleMatrix \in \euclidTwoTwo$:
\begin{align}
    \gsCov = \gsRotMatrix \, \gsScaleMatrix \, \gsScaleMatrix^{T} \, \gsRotMatrix^{T},
\label{eq:2d-gaussian-covariance}
\end{align}

Specifically, we establish and maintain a rotation angle $\gsRotAngle \in [0, \pi]$ and a scaling vector $\gsScaleVector \in \euclidTwoPos$ for each 2D Gaussian. These attributes are updated via stochastic gradient descent and clipped to the valid ranges during training. The rotation and scaling matrices $\gsRotMatrix, \gsScaleMatrix$ are constructed via $\gsRotAngle, \gsScaleVector$ on the fly during both training and inference.

Unlike 3D Gaussian Splatting which models view-dependent appearance using spherical harmonics \cite{kerbl20233d}, we only need to associate each 2D Gaussian with a vector $\gsColor \in \euclidColor$ to store its color, as an image essentially shows a single view of a 3D scene. Notably, the design choice of a variable color dimension $\gsFeatDim$ enables \methodName to flexibly support a diversity of image formats, including grayscale, RGB, and CMYK images, as well as multi-channel texture stacks.

In addition, 3D Gaussian Splatting relies on an opacity attribute for depth-ordered occlusion and $\alpha$-blended rendering. By contrast, the color information of 2D Gaussians can be effectively aggregated regardless of their relative order, as detailed in \Cref{sec:method-rendering}. Therefore, our 2D Gaussians do not have an opacity attribute.

By combining the above components, each 2D Gaussian primitive in \methodName is fully characterized by $5+\gsFeatDim$ trainable parameters: 
\begin{align}
    \gsParameters_{i} \coloneqq \gsParameters_{i} (\gsMean_{i},\gsRotAngle_{i},\gsScaleVector_{i},\gsColor_{i}) \in \euclidAll, \quad 1 \leq i \leq \gsNum.
\label{eq:2d-gaussian-parameters}
\end{align}


\newcommand{\ImageRateDistortionRes}{0.215\linewidth}
\begin{figure*}[t]
\centering
\subfloat{
    \includegraphics[width=\ImageRateDistortionRes-0.005\linewidth]{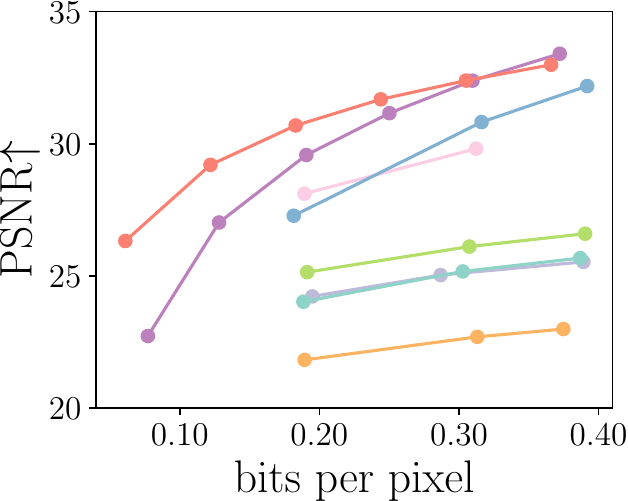}
  }
\subfloat{
    \includegraphics[width=\ImageRateDistortionRes]{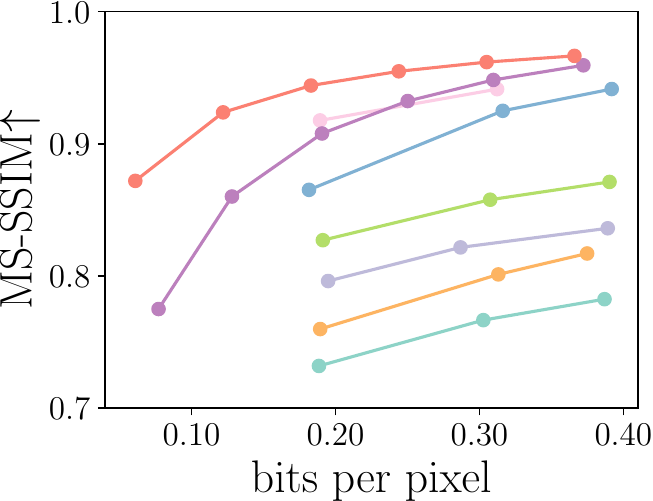}
  }
\subfloat{
    \includegraphics[width=\ImageRateDistortionRes]{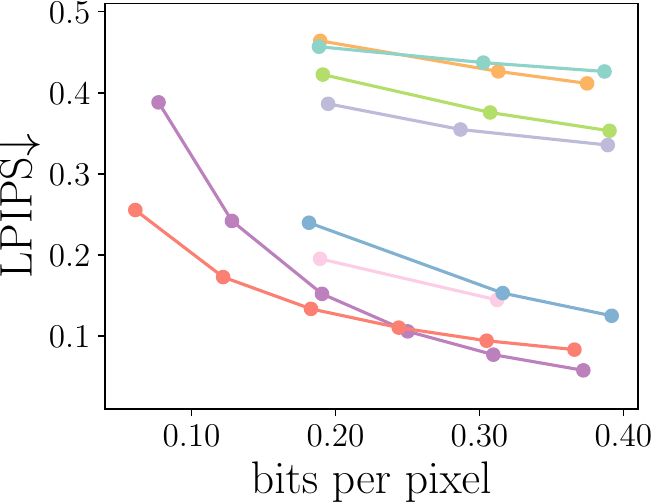}
  }
\subfloat{
    \includegraphics[width=\ImageRateDistortionRes]{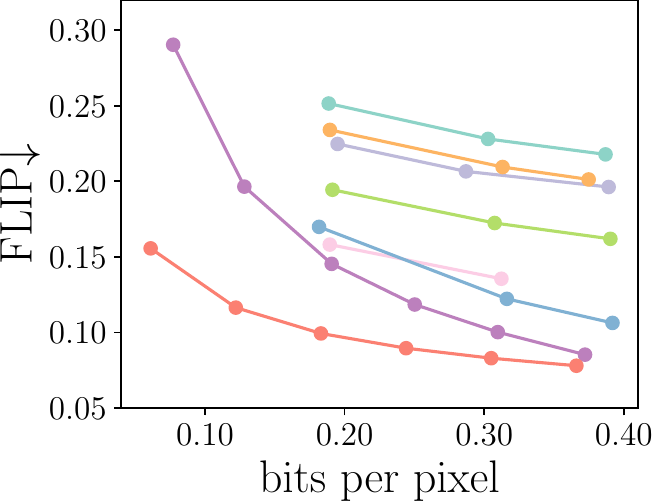}
  }
\subfloat{
    \includegraphics[width=0.09\linewidth]{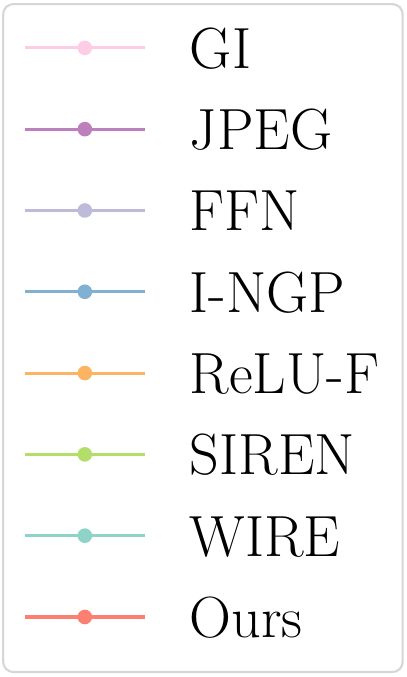}
  }
\Caption{Rate-distortion curves (\Cref{sec:evaluation-image}).}{\revise{These results report the metric scores averaged over the evaluation set of 45 RGB images (\Cref{sec:evaluation-setup}).}}
\label{fig:evaluation-rate-distortion-image}
\end{figure*}


\subsection{\revise{Rendering} 2D Gaussians into Images}
\label{sec:method-rendering}

While 3D Gaussian Splatting requires opacity and depth sorting to handle occlusions and enforce multi-view consistency, we note that such requirements are unnecessary in the 2D case. In particular, an image showing a set of colored Gaussian blobs can be rendered by 
summing their weighted colors, and the resulting image is invariant to the order in which the Gaussians are applied.

Leveraging this insight, we simplify the point-based $\alpha$-blending equation in \cite{yifan2019differentiable,kopanas2021point} by treating the 2D Gaussians as an unordered set of anisotropic points, and accumulate their color contributions to render an image pixel $\pixelColorRendered(\pixelLocation) \in \euclidColor$:
\begin{align}
    \pixelColorRendered(\pixelLocation) = \sum_{i=1}^{\gsNum} \gaussian_{i}(\pixelLocation) \cdot \gsColor_{i},
\label{eq:rendering-naive}
\end{align}

This naive formulation, however, uses all Gaussians to render a pixel, which significantly hinders the rendering and training speed. In addition, such dense pixel-Gaussian correlation largely disrupts data locality, which is essential for fast random pixel access in GPU applications \cite{nystad2012adaptive,vaidyanathan2023random}.


\newcommand{\TimeRes}{0.48\linewidth}
\begin{figure}[t]
\centering
\subfloat{
    \includegraphics[width=\TimeRes]{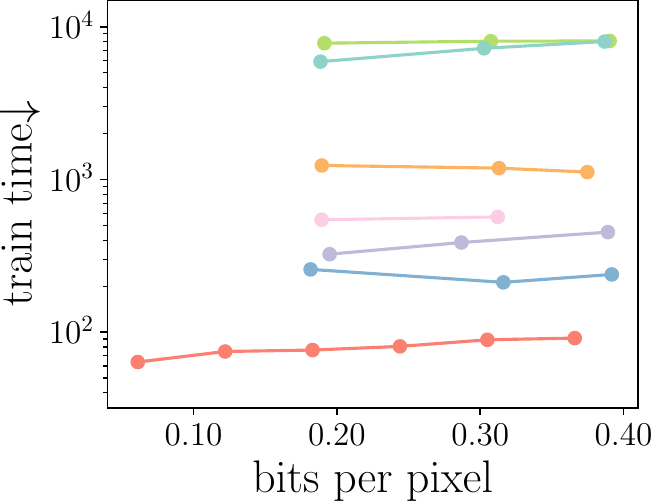}
  }
\subfloat{
    \includegraphics[width=\TimeRes]{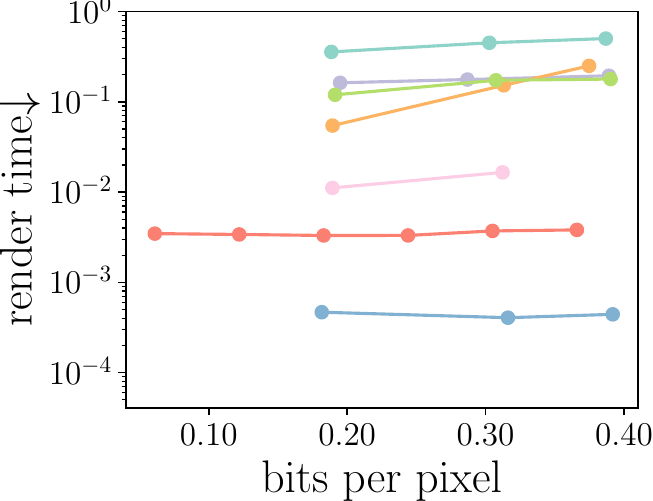}
  }
\Caption{System performance (\Cref{sec:evaluation-system}).}
{\revise{These results relate to the image experiments in \Cref{sec:evaluation-image} and share the same color scheme as \Cref{fig:evaluation-rate-distortion-image}.}}
\label{fig:evaluation-time}
\end{figure}


To resolve this issue, we take a tile-based rendering approach by following recent works \cite{lassner2021pulsar}, and constrain the number of Gaussians that may contribute to a pixel. Specifically, we begin by subdividing the spatial support of an image into non-overlapping tiles of size $\tileHeight\times\tileWidth$. We then compute the minimum enclosing circle for each 2D Gaussian's 3-standard-deviation range ($99.7\%$ confidence interval) and check tile-circle intersection across all pairs to establish tile-Gaussian correspondence, where the set of Gaussians whose enclosing circles intersect tile $\tile_{j}$ is denoted as $\tileHit_{j}$, $1 \leq j \leq \tileNum$. After that, for each pixel $\pixelLocation \in \tile_{j}$, we rank all Gaussians in $\tileHit_{j}$ based on their values evaluated at $\pixelLocation$ and only keep the top $K$. Finally, we normalize the values of the $K$ remaining Gaussians, and aggregate their weighted colors to obtain $\pixelColorRendered(\pixelLocation)$.

The final rendering equation for \methodName is formulated as:
\begin{align}
    \pixelColorRendered(\pixelLocation) = \frac{1}{\sum_{i \in \tileHit_{j}^{K}(\pixelLocation)} \gaussian_{i}(\pixelLocation)} \sum_{i \in \tileHit_{j}^{K}(\pixelLocation)} \gaussian_{i}(\pixelLocation) \cdot \gsColor_{i}, \quad 1 \leq j \leq \tileNum
\label{eq:rendering}
\end{align}
where $\tileHit_{j}^{K}(\pixelLocation)$ denotes the set of top $K$ Gaussians for $\pixelLocation \in \tile_{j}$.


\subsection{Content-Adaptive Initialization and Optimization}
\label{sec:method-optimization}

Building on the differentiable renderer introduced in \Cref{sec:method-rendering}, we optimize Gaussian attributes $\gsParameters_{i}$ through stochastic gradient descent to faithfully reconstruct any target image. Unlike the neural network models used in current neural image representations \cite{sitzmann2020implicit,tancik2020fourier,martel2021acorn}, \methodName only consists of explicit features with physical meaning, and thus benefits from task-specific initialization for faster and higher-quality convergence. In particular, an ideal initialization should be guided by the image content, with the spatial distribution of Gaussians matching that of high-frequency image features.

To this end, we propose a content-adaptive position initialization strategy coupling image gradient guidance with uniform sampling. Specifically, during the position sampling of each Gaussian (we only sample pixel centers), the probability of a pixel $\pixelLocation$ being sampled is a weighted sum of the relative magnitude of its local image gradient and a constant shared across all pixel locations. Notably, the former emphasizes image content-aware adaptivity, while the latter ensures adequate coverage of the entire image domain.
\begin{align}
    \probInit(\pixelLocation) = \frac{(1-\paramInit) \cdot \lVert\nabla I(\pixelLocation)\rVert_{2}}{\sum_{h=1}^{H}\sum_{w=1}^{W} \lVert\nabla I(\pixelLocation_{h,w})\rVert_{2}} + \frac{\paramInit}{H \cdot W}, \quad \paramInit \in [0, 1]
\label{eq:edge-sampling}
\end{align}
where $H,W$ denote the height and width of the image, and $\nabla I(\cdot)$ is the image gradient operator. In addition, all Gaussians are assigned the target pixel color at their initialized location $\pixelColorTarget(\pixelLocation)$.

During each optimization step, the set of Gaussians is rendered into an image using \Cref{eq:rendering}, which is then used to compute a combination of $L_{1}$ and SSIM loss against the target image. Since the full pipeline is differentiable, all Gaussian attributes\footnote{We empirically observed that optimizing the inverse of Gaussian scales $1/\gsScaleVector$ results in improved and faster convergence. Please refer to \Cref{sec:evaluation-setup} for more details.} can be updated via stochastic gradient descent to improve reconstruction.

Note that while the top-$K$ ranking operation is non-differentiable, gradients do not flow through the operation itself but \revise{through the $K$ retained Gaussians. Empirically (in \Cref{tab:evaluation-ablation}), top-$K$ normalization not only promotes data locality but also improves reconstruction quality. We hypothesize that the top-$K$ normalization achieves this effect by acting as a form of regularization during optimization.}

Besides the initially introduced Gaussians, we also \revise{progressively add new Gaussians to image areas exhibiting} high reconstruction error. This is achieved by sampling pixels based on their error magnitude and initializing new Gaussians at sampled locations.
\begin{align}
    \probAdd(\pixelLocation) = \frac{\left|\pixelColorRendered(\pixelLocation) - \pixelColorTarget(\pixelLocation)\right|}{\sum_{h=1}^{H}\sum_{w=1}^{W} \left|\pixelColorRendered(\pixelLocation_{h,w}) - \pixelColorTarget(\pixelLocation_{h,w})\right|},
\label{eq:error-sampling}
\end{align}
\Cref{fig:method-pipeline} illustrates the optimization pipeline of \methodName.
\section{Evaluation}
\label{sec:evaluation}


\subsection{Experimental Setup}
\label{sec:evaluation-setup}

\paragraph{Dataset}
\revise{Existing datasets for image compression evaluation, such as Kodak and CLIC, emphasize natural images and are mostly low-resolution (below 2 megapixel). In contrast, image assets used in the latest graphics workflows often have much higher resolutions and feature more diverse content, including stylized images. To address this gap, we collected 45 RGB images spanning 5 categories, vector-style, photograph, digital art, anime, and painting, with 9 samples per category. Additionally, we collected 19 texture stacks, each containing 9 channels (diffuse color, normal map, ambient occlusion, roughness, and metalness), for texture compression evaluation. All images and textures\footnote{Images and textures were sourced from Adobe Stock and Poly Haven, respectively.} have a resolution of 2K$\times$2K.}

\paragraph{Metrics}
\revise{For image fidelity measurement, we use PSNR, MS-SSIM \cite{wang2003multiscale}, LPIPS \cite{zhang2018unreasonable}, and FLIP \cite{andersson2020flip}. For experiments on textures, we omit LPIPS and FLIP, as they are specifically designed for natural images. We also report the model size (in kilobytes, KB) and bitrate (in bits per pixel, bpp, or bits per pixel per channel, bppc) to assess memory efficiency}.

\paragraph{Implementation}
Our differentiable rendering pipeline, \revise{built upon gsplat \cite{ye2024gsplat}, is equipped with custom CUDA kernels for efficient forward and backward computation.}
Each tile of size $16\times16$ is processed by a separate CUDA block.
The remaining components are implemented in PyTorch \cite{paszke2019pytorch}.
The only trainable parameters in \methodName, all quantized to float16, are the Gaussian attributes defined in \Cref{eq:2d-gaussian-parameters}.
By mapping the image domain to $[0,1]^{2}$, \methodName supports target images of any aspect ratio and resolution. 
At initialization, all Gaussian positions $\gsMean$ and colors $\gsColor$ are populated via our content-adaptive sampling strategy (\Cref{eq:edge-sampling}), \revise{while their scaling vectors $\gsScaleVector$ and rotation angles $\gsRotAngle$ are initialized to $5$ (pixels) and $0$, respectively.}
\revise{Notably, instead of directly optimizing the raw Gaussian scales $\gsScaleVector$, which typically fall in the range $[5, 10]$, we maintain and optimize their inverses $1/\gsScaleVector$ to improve convergence, since optimizing in the $[0,1]$ range yields smoother and more stable gradients.}
We use the Adam optimizer \cite{kingma2015adam} to iteratively update these parameters against $L_1 + 0.1 \cdot L_{\text{SSIM}}$ for 5K steps.
The learning rates for $(\gsMean, \gsColor, \gsScaleVector, \gsRotAngle)$ are set to $(5\text{e}{-4}, 5\text{e}{-3}, 2\text{e}{-3}, 2\text{e}{-3})$ and remain constant throughout training.
In our experiments, $K$ and $\paramInit$ are set to $10$ and $0.3$, respectively.
During training, we progressively allocate additional Gaussians to image regions exhibiting high fitting errors (\Cref{eq:error-sampling}). For a total budget of $\gsNum$ Gaussians, training begins with $\gsNum/2$ Gaussians, and an additional $\gsNum/8$ are introduced every 0.5K steps until the budget is reached.
\revise{Ablation studies on these design choices are provided in \Cref{tab:evaluation-ablation}.}


\begin{figure}[t]
\centering
\subfloat{
    \includegraphics[width=0.48\linewidth]{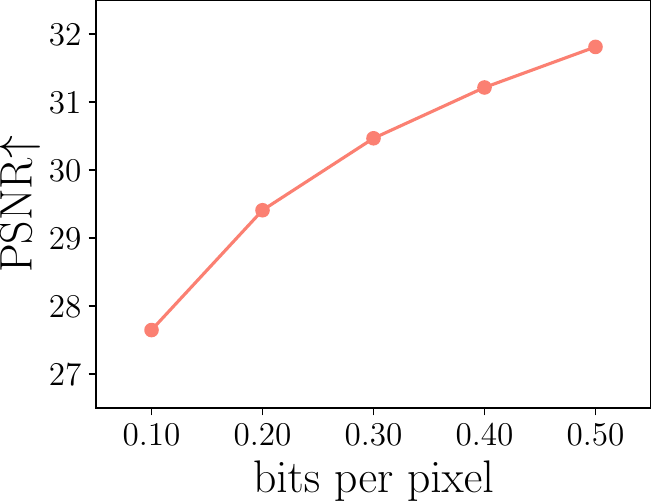}
  }
\subfloat{
    \includegraphics[width=0.48\linewidth]{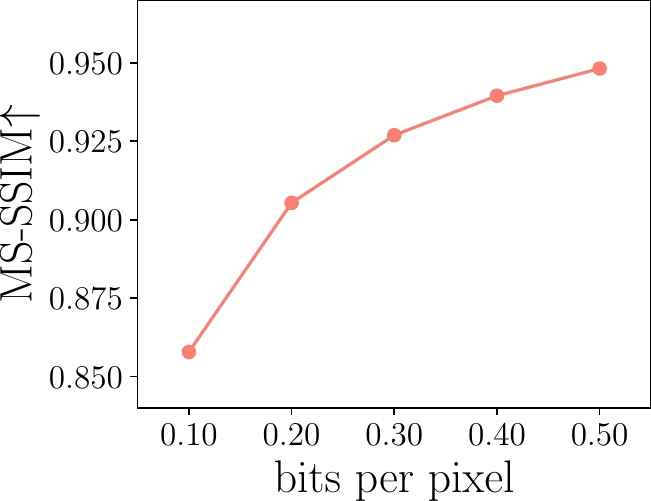}
  }
\Caption{\revise{Rate-distortion curves on the CLIC2020 benchmark (\Cref{sec:evaluation-image}).}}{}
\label{fig:evaluation-rate-distortion-image-clic}
\end{figure}


\subsection{Visual Fidelity vs. Memory Efficiency}
\label{sec:evaluation-lod}

\revise{
We assess \methodName's rate-distortion performance on the evaluation set of 45 RGB images through error-guided progressive optimization (\Cref{sec:method-optimization}).
As summarized in \Cref{fig:evaluation-rate-distortion-image}, \methodName achieves quality metrics of $32.99\pm4.49$ (PSNR), $0.966\pm0.020$ (MS-SSIM), $0.083\pm0.057$ (LPIPS), and $0.078\pm0.029$ (FLIP) at 0.366 bpp. Even at an ultra-low bitrate of 0.122 bpp, \methodName maintains reasonable visual quality with metric scores of $29.20\pm4.57$ (PSNR), $0.924\pm0.042$ (MS-SSIM), $0.173\pm0.082$ (LPIPS), and $0.116\pm0.043$ (FLIP).
}

\revise{
Leveraging error-informed progressive optimization, \methodName naturally constructs a smooth level-of-detail hierarchy in a single run without additional overhead.
This design enables flexible quality control that adapts to the device capabilities at deployment.
Besides, this design facilitates quality-driven compression, where Gaussians are progressively added until the required quality is met.
In contrast, neural image representations based on implicit models or fixed data structures do not support straightforward control over compression quality. \Cref{fig:evaluation-lod,fig:evaluation-lod-supp} provide the results from several progressive runs that start at 0.061 bpp and terminate at 0.305 bpp.
}


\subsection{\revise{Image Compression Performance}}
\label{sec:evaluation-image}

\paragraph{Baselines}
We compare to 6 neural image representations: ReLU-F \cite{karnewar2022relu}, SIREN \cite{sitzmann2020implicit}, FFN \cite{tancik2020fourier}, WIRE \cite{saragadam2023wire}, I-NGP \cite{muller2022instant}, and GI \cite{zhang2024gaussianimage}. We also include JPEG \cite{wallace1991jpeg} as a \revise{conventional baseline} for completeness. Since our objective is to represent high-resolution images at ultra-low bitrates, the allowable memory budget exceeds the range explored by most baselines. For fair comparisons, we adopt their official implementations and modify only the model sizes to match our target range. \revise{This is done by reducing the grid resolution (ReLU-F, I-NGP), the number of primitives (GI), and the number of layers and hidden dimensions (I-NGP, WIRE, SIREN, FFN). The evaluation set of 45 RGB images is employed for this experiment.}


\begin{figure}[t]
\centering
\subfloat{
    \includegraphics[width=0.49\linewidth]{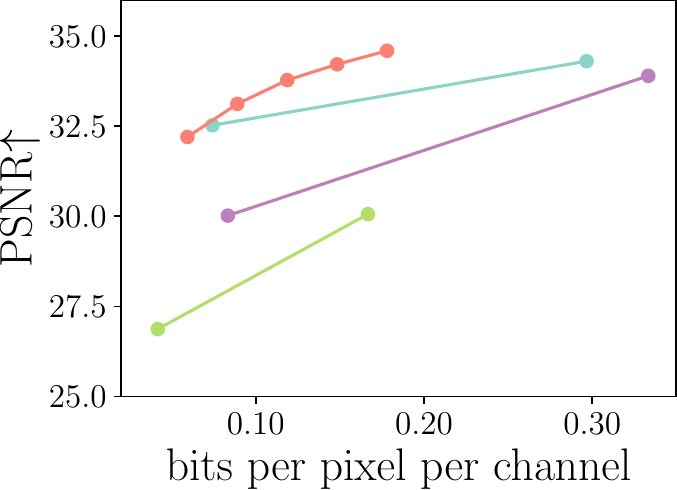}
  }
\subfloat{
    \includegraphics[width=0.47\linewidth]{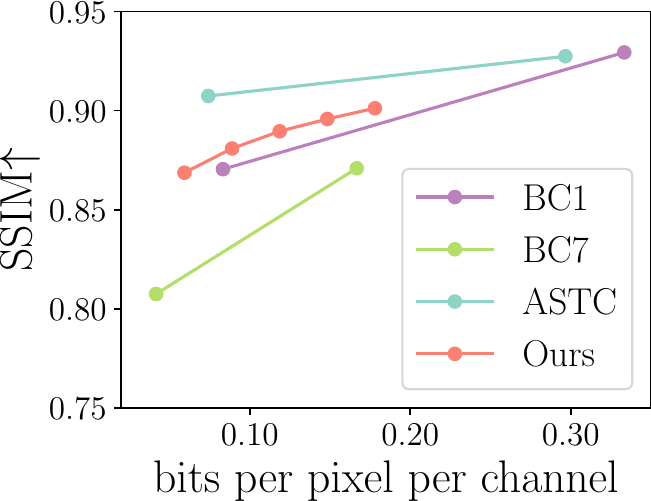}
  }
\Caption{\revise{Rate-distortion curves on the 19 texture stacks (\Cref{sec:evaluation-texture}).}}{}
\label{fig:evaluation-rate-distortion-texture}
\end{figure}


\begin{table}[t]
\centering
\Caption{\revise{Ablation studies on the design choices of \methodName.}}{}
\begin{tabular}{c c c c}
    \toprule
    Variants & PSNR$\uparrow$ & MS-SSIM$\uparrow$ & LPIPS$\downarrow$ \\
    \midrule
    No color init & $30.40\pm4.73$ & $0.951\pm0.025$ & $0.110\pm0.064$ \\
    No position init & $29.88\pm4.18$ & $0.954\pm0.028$ & $0.135\pm0.065$ \\
    Random init & $29.54\pm4.15$ & $0.948\pm0.029$ & $0.149\pm0.067$ \\
    No inverse scale & $29.11\pm4.58$ & $0.933\pm0.038$ & $0.152\pm0.073$ \\
    No top-K norm & $29.35\pm4.43$ & $0.944\pm0.032$ & $0.182\pm0.086$ \\
    Full & $31.77\pm4.73$ & $0.960\pm0.024$ & $0.102\pm0.062$ \\
    \bottomrule
\end{tabular}
\label{tab:evaluation-ablation}
\end{table}


\newcommand{\imageCompWidth}{0.122\linewidth}
\newcommand{\reduceWidth}{-0.19cm}
\begin{figure*}[p]
\centering
\subfloat{
    \includegraphics[width=\imageCompWidth,height=\imageCompWidth]{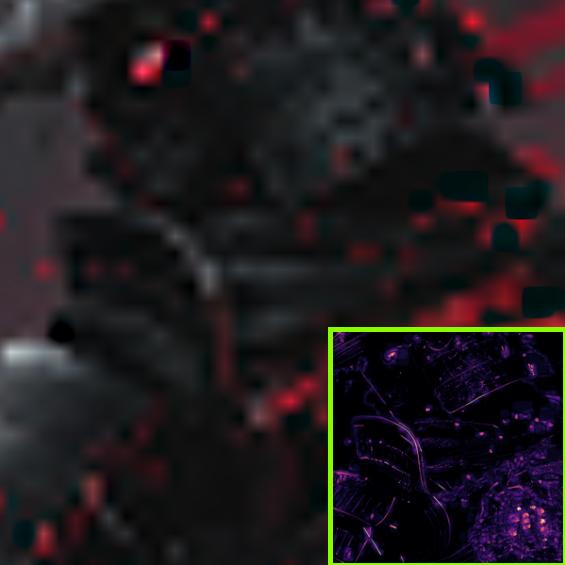}
  } \hspace{\reduceWidth}
\subfloat{
    \includegraphics[width=\imageCompWidth,height=\imageCompWidth]{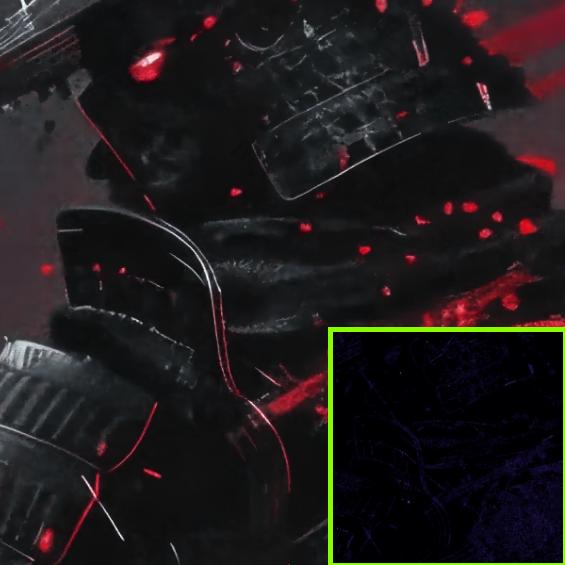}
  } \hspace{\reduceWidth}
\subfloat{
    \includegraphics[width=\imageCompWidth,height=\imageCompWidth]{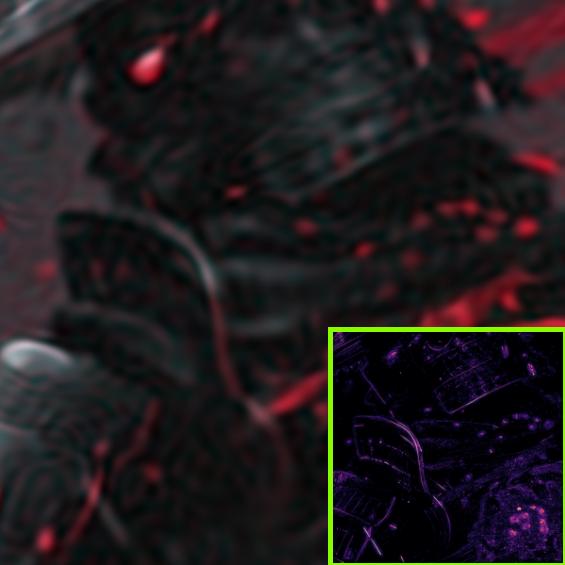}
  } \hspace{\reduceWidth}
\subfloat{
    \includegraphics[width=\imageCompWidth,height=\imageCompWidth]{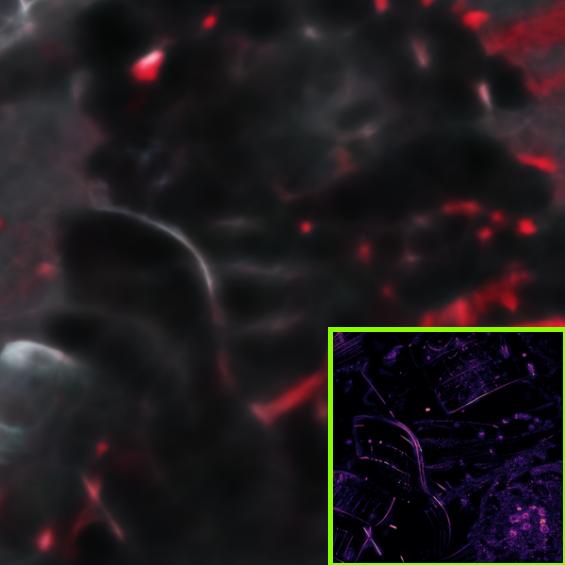}
  } \hspace{\reduceWidth}
\subfloat{
    \includegraphics[width=\imageCompWidth,height=\imageCompWidth]{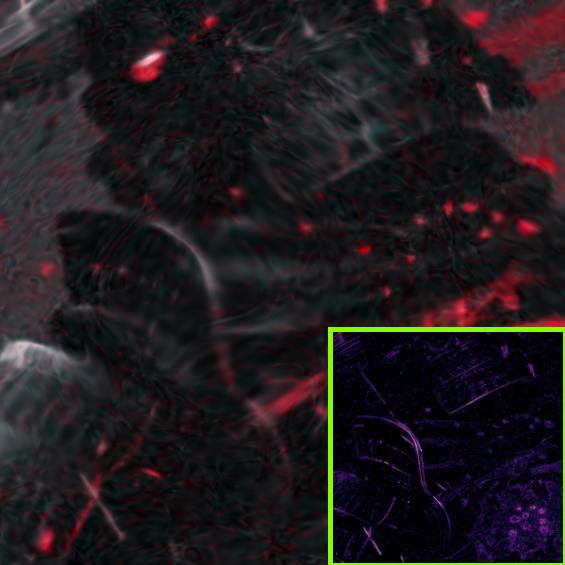}
  } \hspace{\reduceWidth}
\subfloat{
    \includegraphics[width=\imageCompWidth,height=\imageCompWidth]{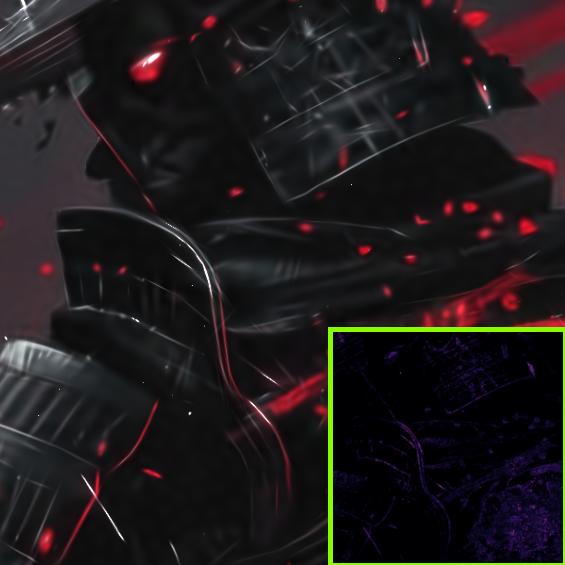}
  } \hspace{\reduceWidth}
\subfloat{
    \includegraphics[width=\imageCompWidth,height=\imageCompWidth]{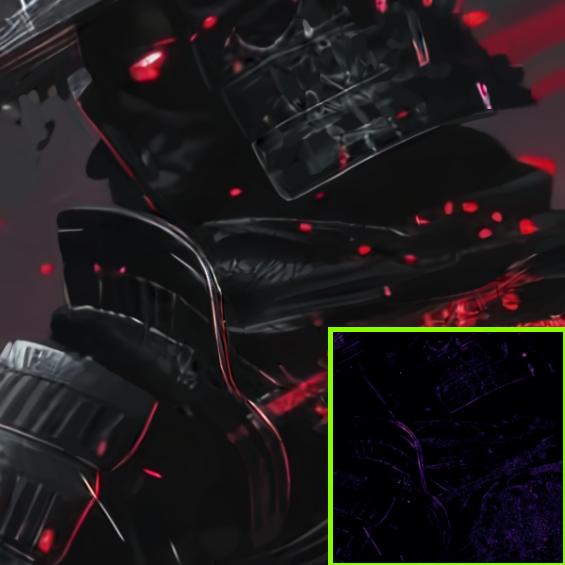}
  } \hspace{\reduceWidth}
\subfloat{
    \includegraphics[width=\imageCompWidth,height=\imageCompWidth]{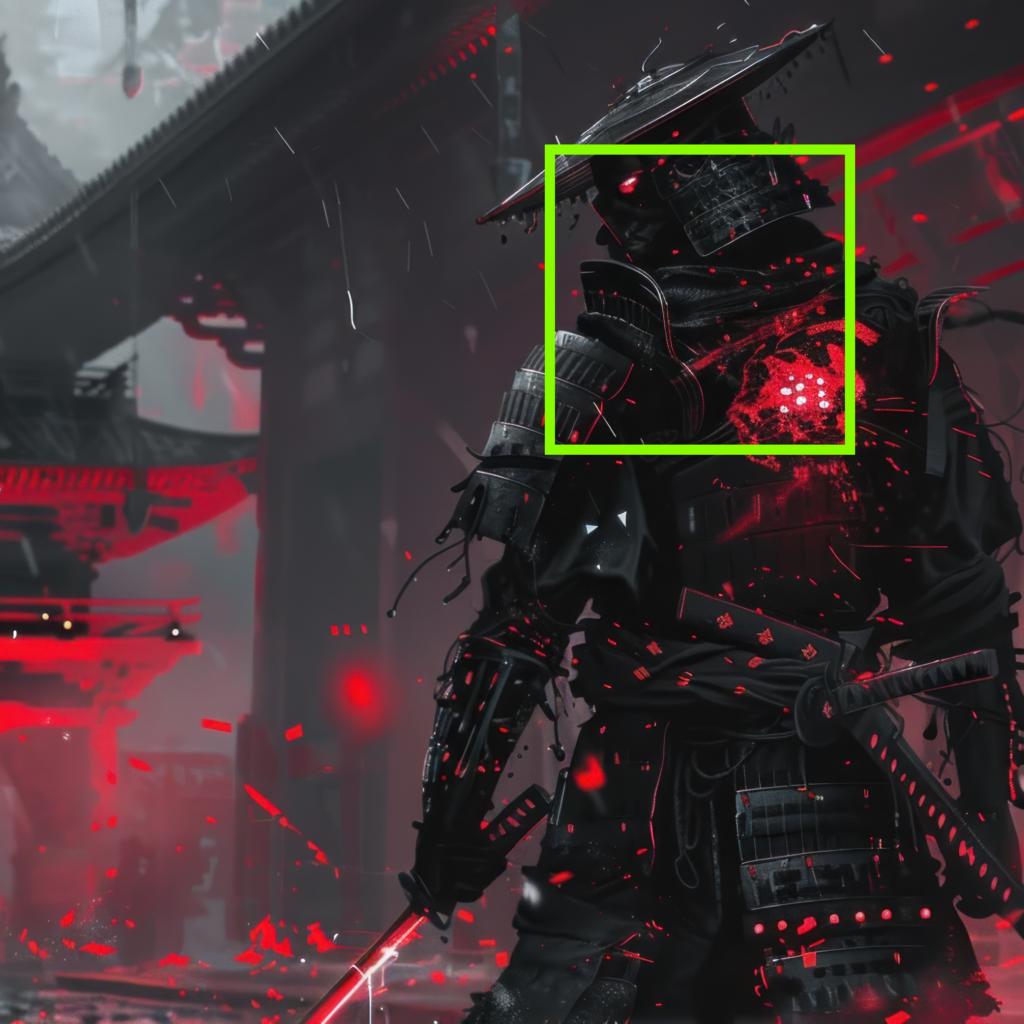}
  }
\vspace{0.2mm} \\
\subfloat{
    \includegraphics[width=\imageCompWidth,height=\imageCompWidth]{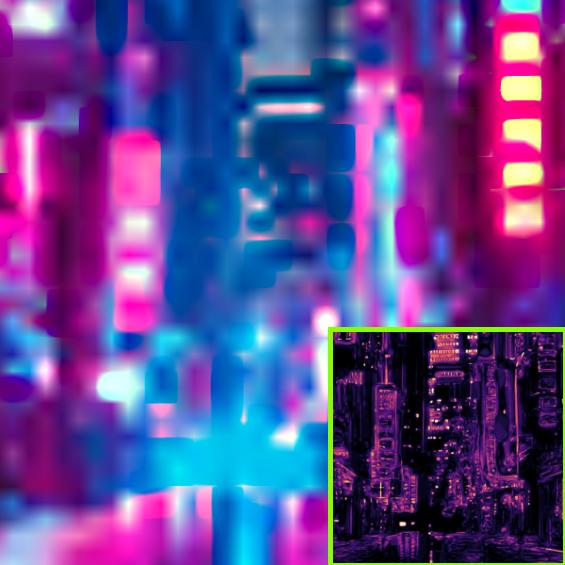}
  } \hspace{\reduceWidth}
\subfloat{
    \includegraphics[width=\imageCompWidth,height=\imageCompWidth]{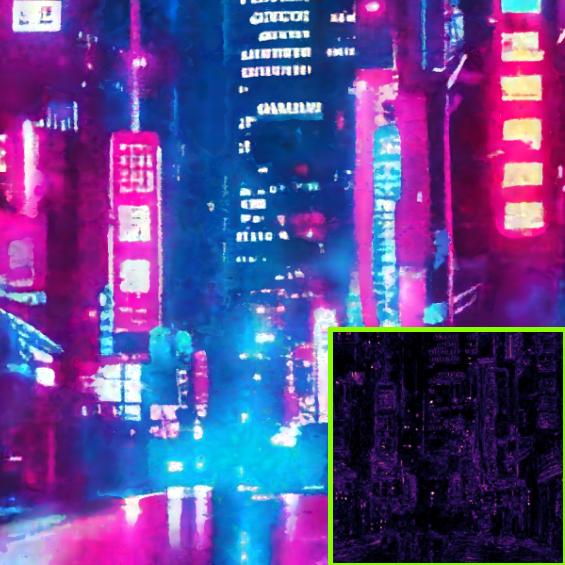}
  } \hspace{\reduceWidth}
\subfloat{
    \includegraphics[width=\imageCompWidth,height=\imageCompWidth]{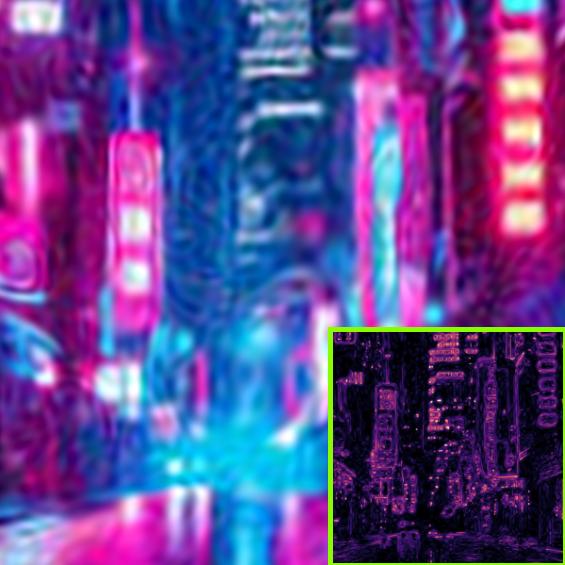}
  } \hspace{\reduceWidth}
\subfloat{
    \includegraphics[width=\imageCompWidth,height=\imageCompWidth]{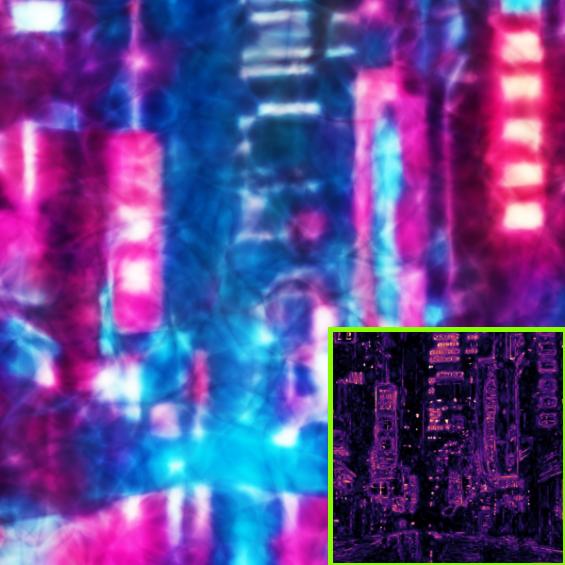}
  } \hspace{\reduceWidth}
\subfloat{
    \includegraphics[width=\imageCompWidth,height=\imageCompWidth]{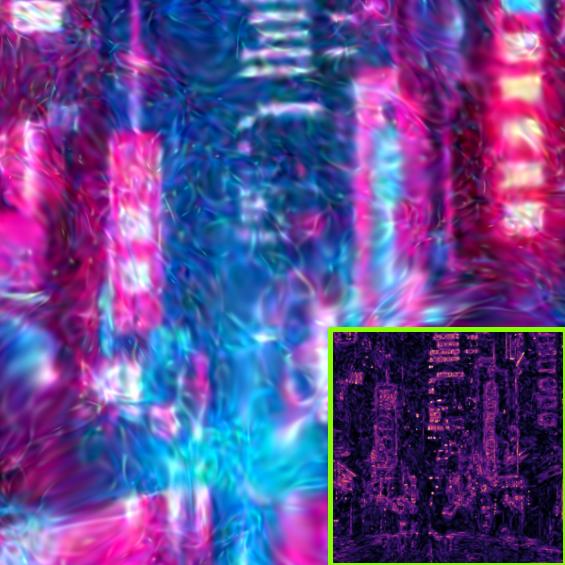}
  } \hspace{\reduceWidth}
\subfloat{
    \includegraphics[width=\imageCompWidth,height=\imageCompWidth]{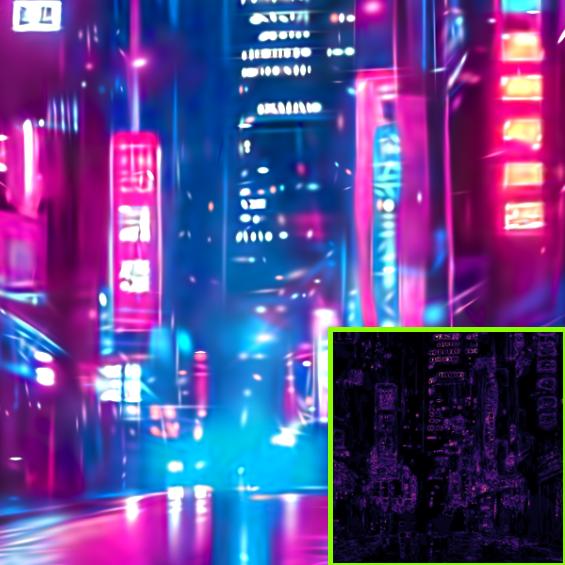}
  } \hspace{\reduceWidth}
\subfloat{
    \includegraphics[width=\imageCompWidth,height=\imageCompWidth]{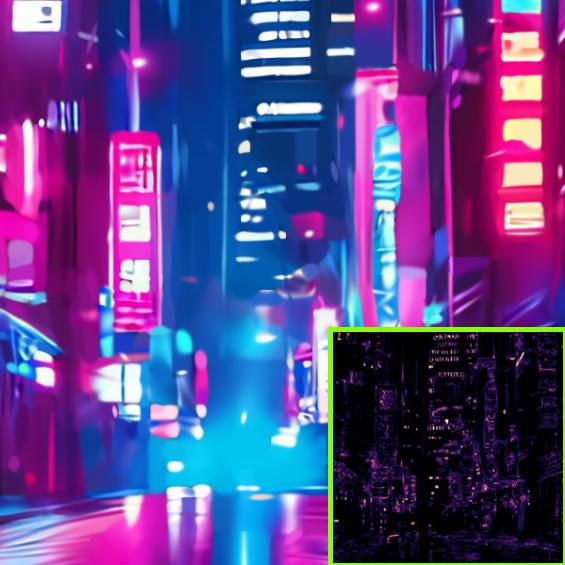}
  } \hspace{\reduceWidth}
\subfloat{
    \includegraphics[width=\imageCompWidth,height=\imageCompWidth]{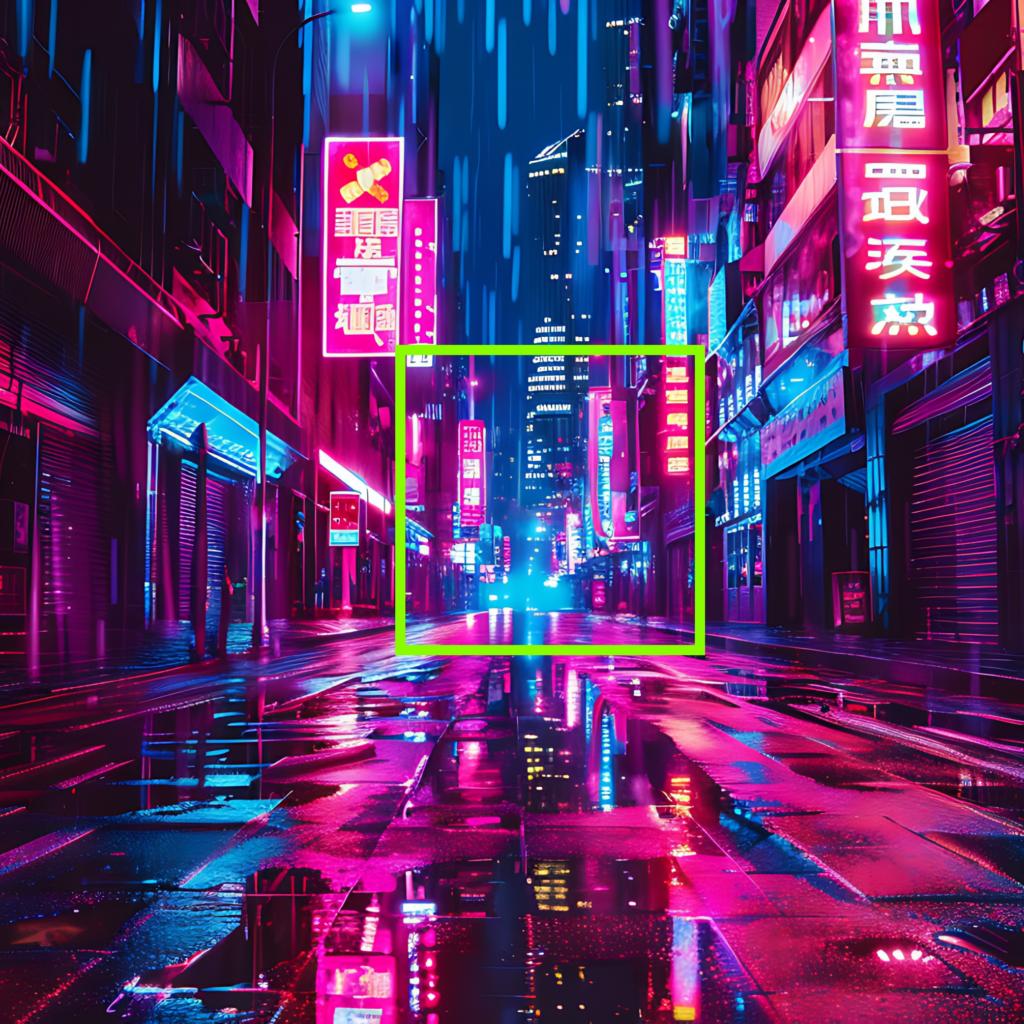}
  }
\vspace{0.2mm} \\
\subfloat{
    \includegraphics[width=\imageCompWidth,height=\imageCompWidth]{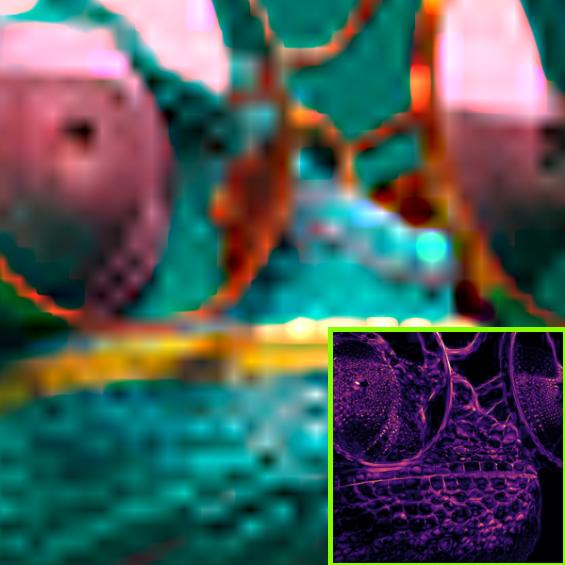}
  } \hspace{\reduceWidth}
\subfloat{
    \includegraphics[width=\imageCompWidth,height=\imageCompWidth]{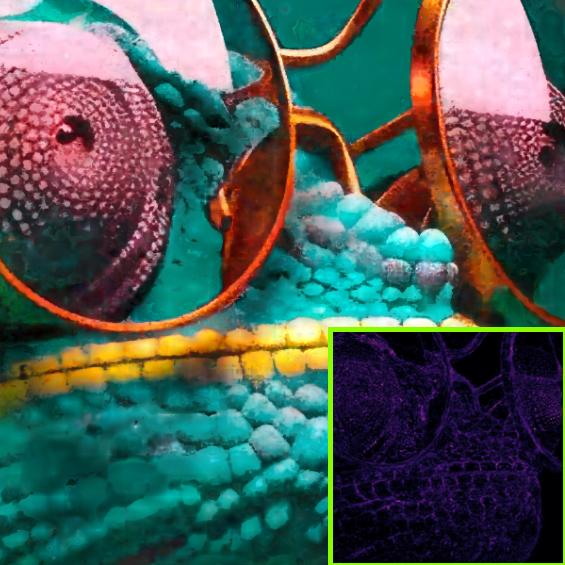}
  } \hspace{\reduceWidth}
\subfloat{
    \includegraphics[width=\imageCompWidth,height=\imageCompWidth]{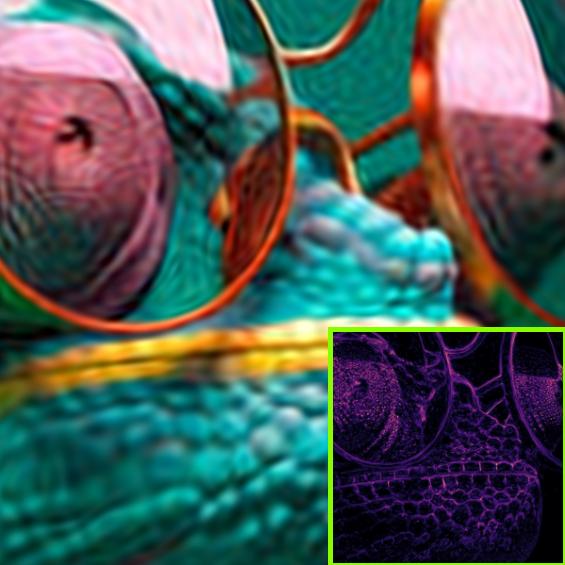}
  } \hspace{\reduceWidth}
\subfloat{
    \includegraphics[width=\imageCompWidth,height=\imageCompWidth]{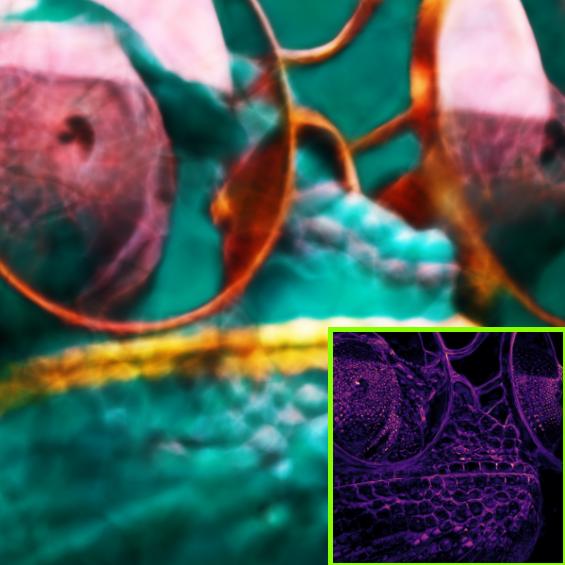}
  } \hspace{\reduceWidth}
\subfloat{
    \includegraphics[width=\imageCompWidth,height=\imageCompWidth]{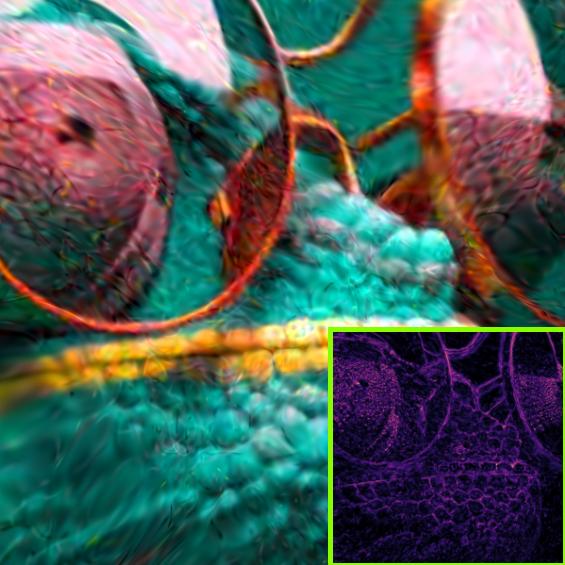}
  } \hspace{\reduceWidth}
\subfloat{
    \includegraphics[width=\imageCompWidth,height=\imageCompWidth]{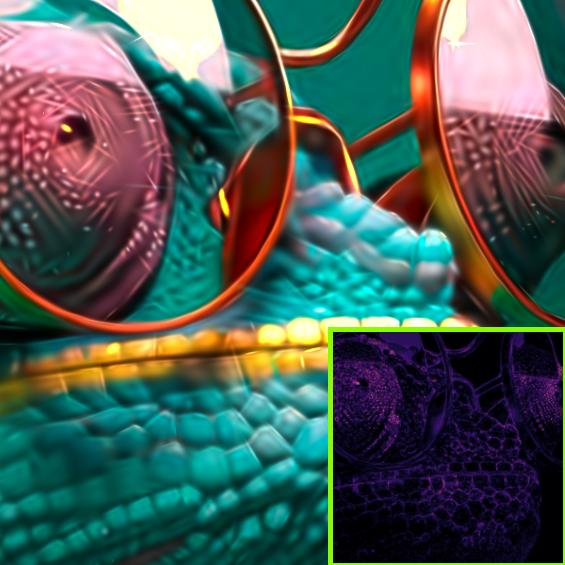}
  } \hspace{\reduceWidth}
\subfloat{
    \includegraphics[width=\imageCompWidth,height=\imageCompWidth]{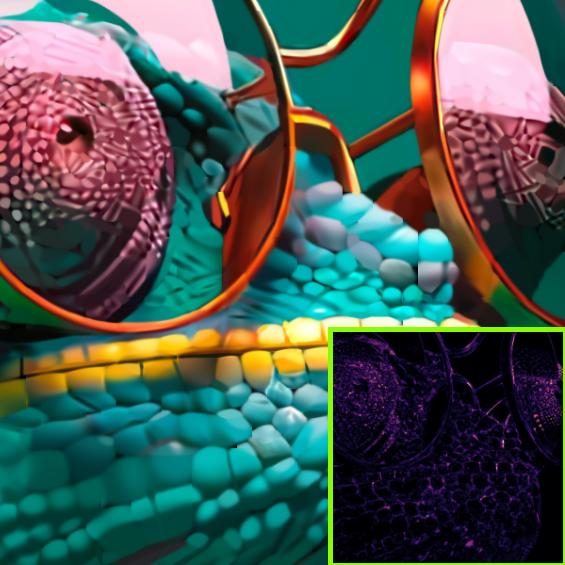}
  } \hspace{\reduceWidth}
\subfloat{
    \includegraphics[width=\imageCompWidth,height=\imageCompWidth]{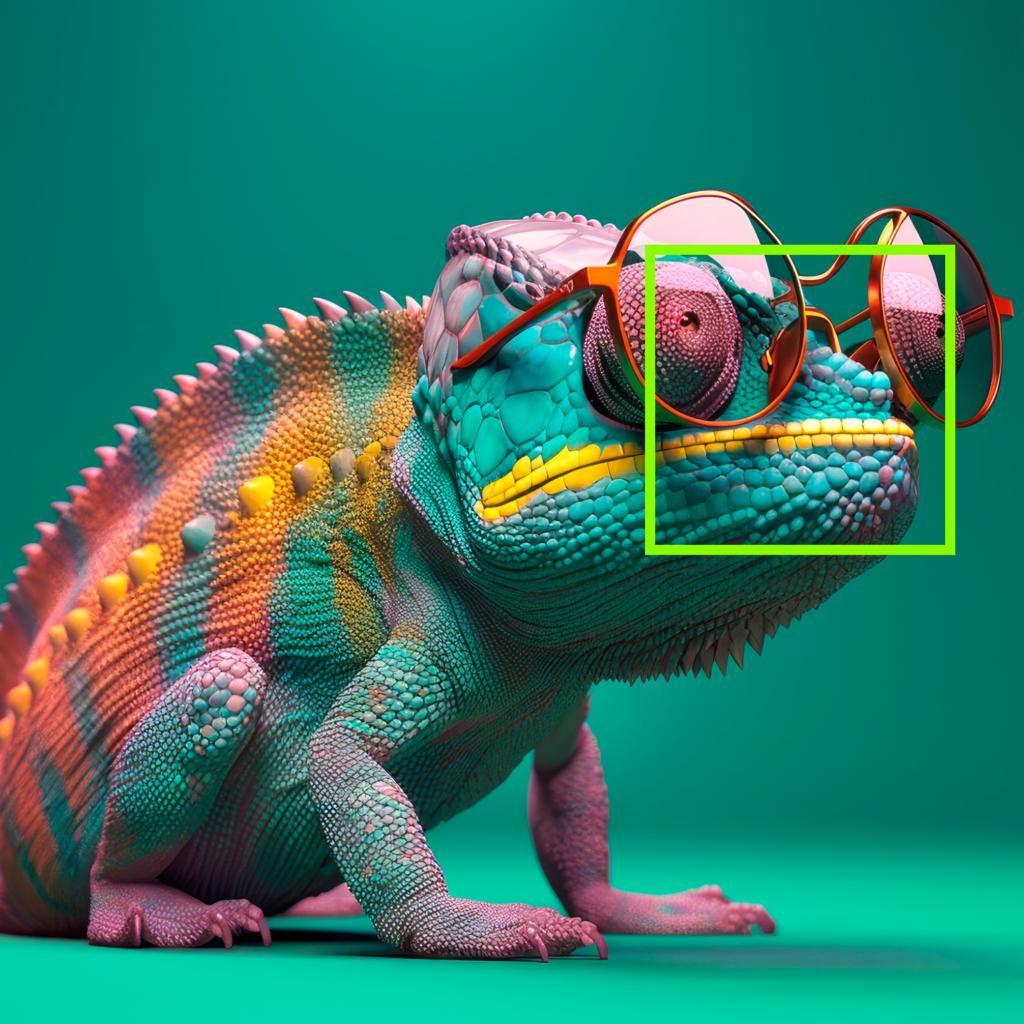}
  }
\vspace{0.2mm} \\
\subfloat{
    \includegraphics[width=\imageCompWidth,height=\imageCompWidth]{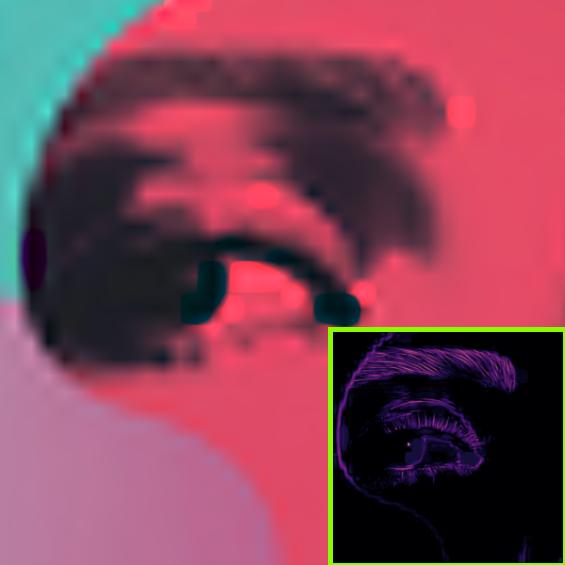}
  } \hspace{\reduceWidth}
\subfloat{
    \includegraphics[width=\imageCompWidth,height=\imageCompWidth]{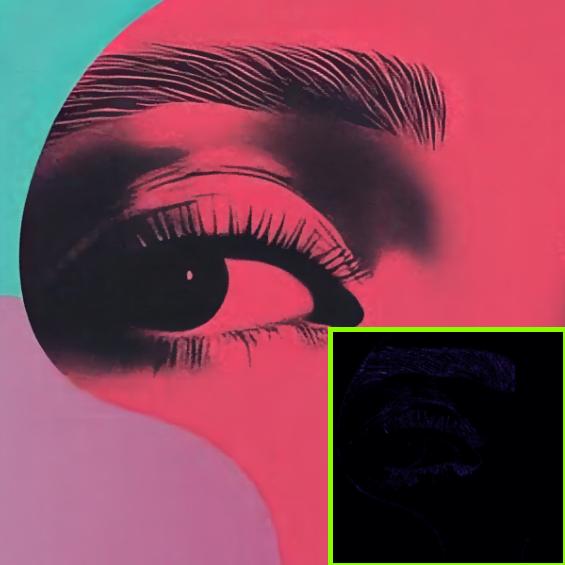}
  } \hspace{\reduceWidth}
\subfloat{
    \includegraphics[width=\imageCompWidth,height=\imageCompWidth]{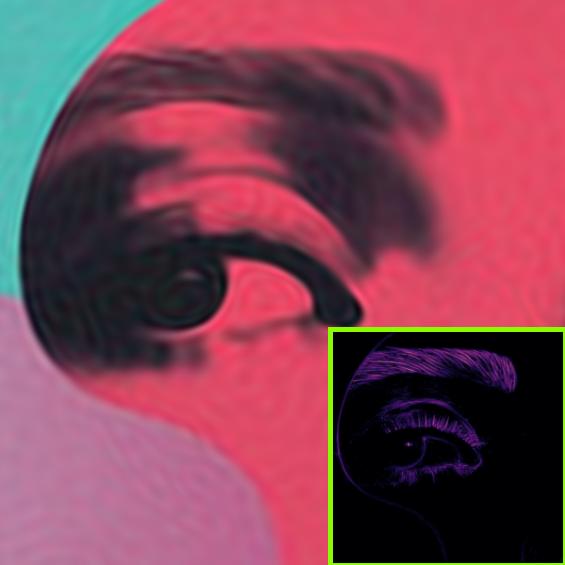}
  } \hspace{\reduceWidth}
\subfloat{
    \includegraphics[width=\imageCompWidth,height=\imageCompWidth]{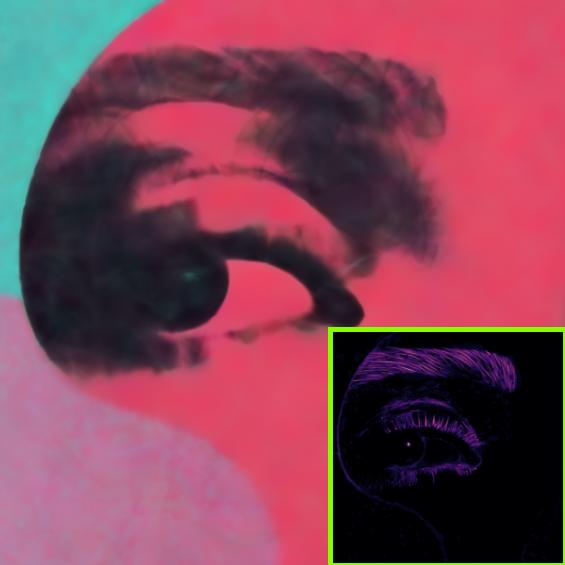}
  } \hspace{\reduceWidth}
\subfloat{
    \includegraphics[width=\imageCompWidth,height=\imageCompWidth]{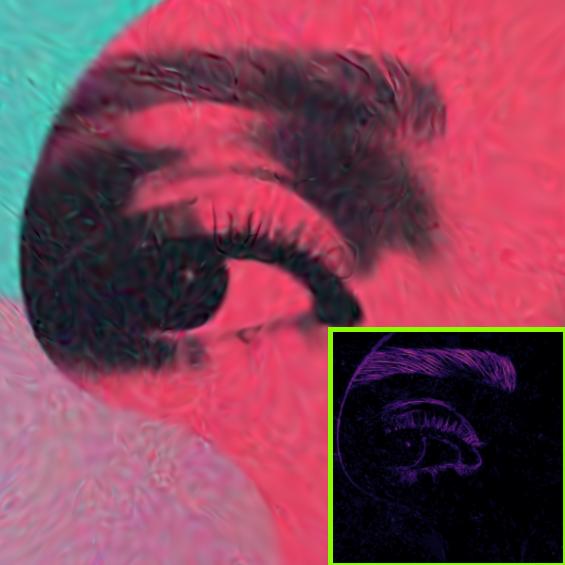}
  } \hspace{\reduceWidth}
\subfloat{
    \includegraphics[width=\imageCompWidth,height=\imageCompWidth]{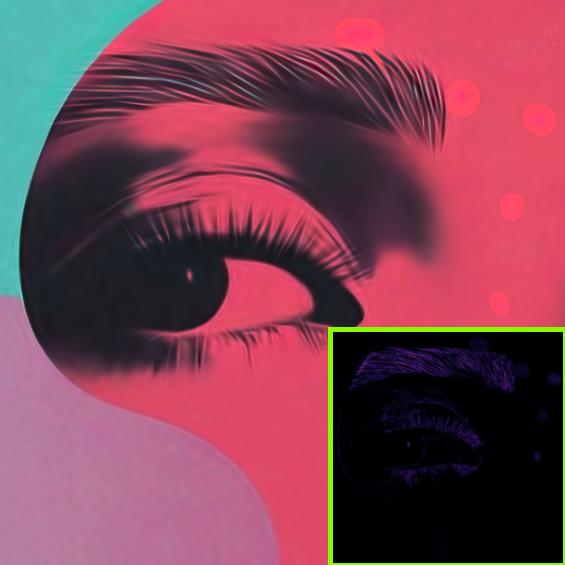}
  } \hspace{\reduceWidth}
\subfloat{
    \includegraphics[width=\imageCompWidth,height=\imageCompWidth]{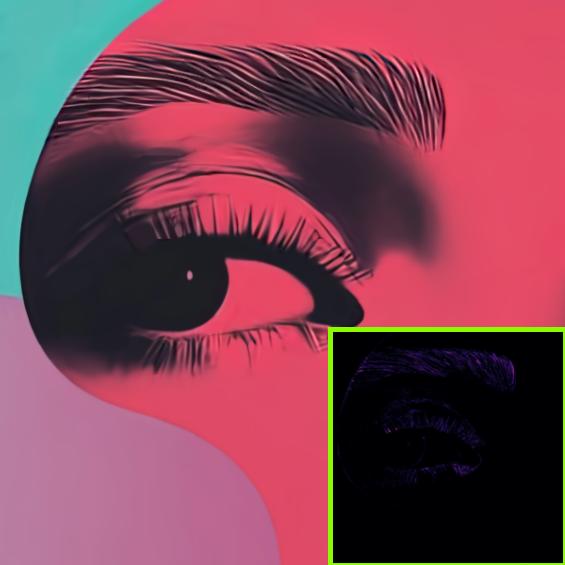}
  } \hspace{\reduceWidth}
\subfloat{
    \includegraphics[width=\imageCompWidth,height=\imageCompWidth]{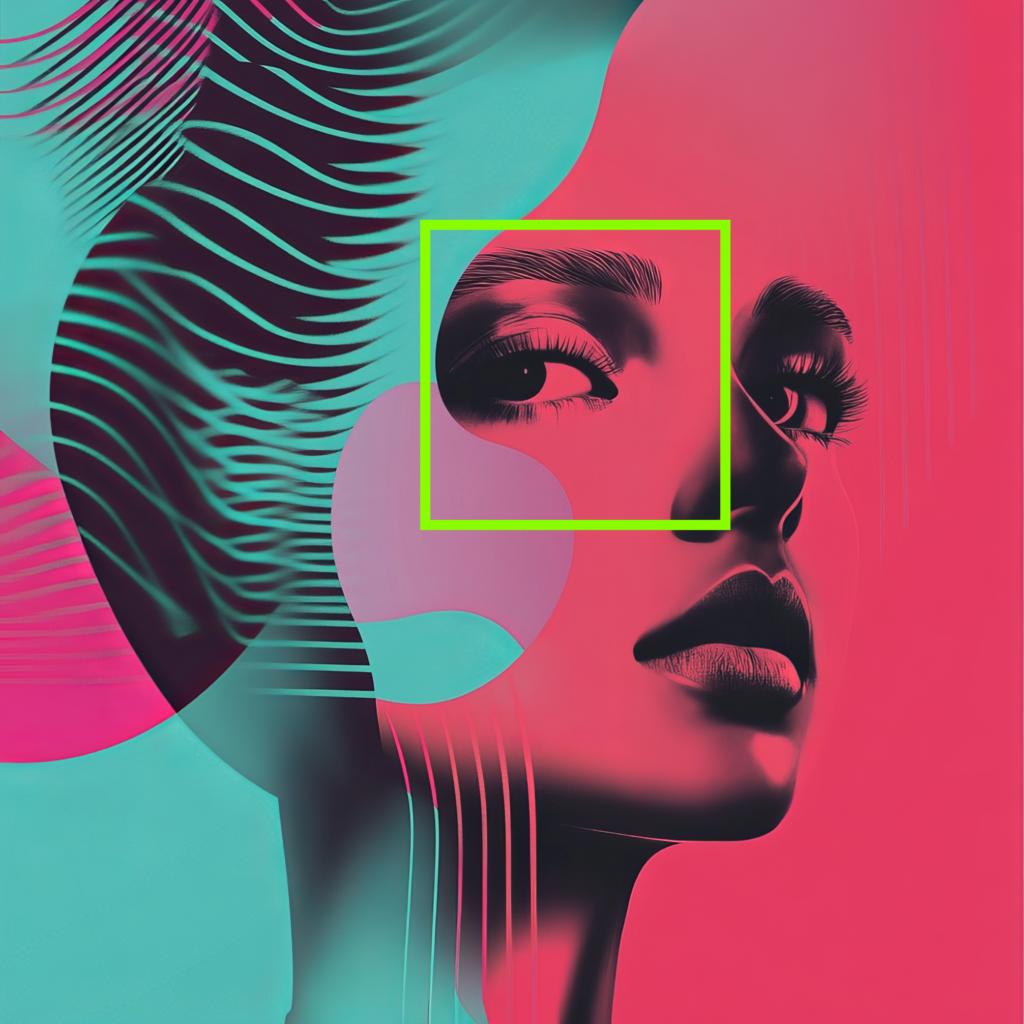}
  }
\vspace{0.2mm} \\
\subfloat{
    \includegraphics[width=\imageCompWidth,height=\imageCompWidth]{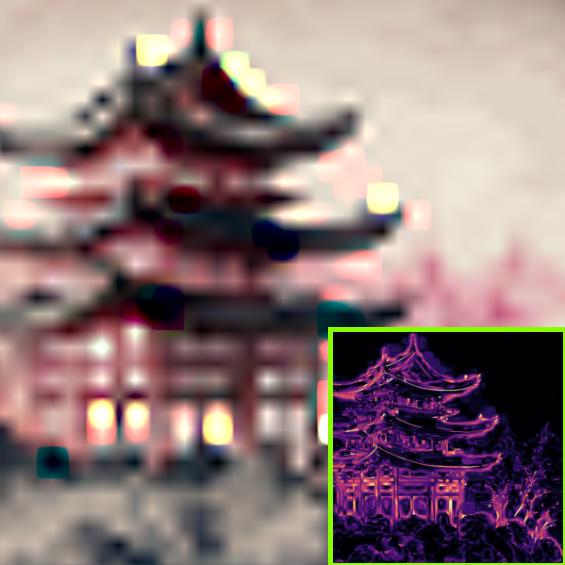}
  } \hspace{\reduceWidth}
\subfloat{
    \includegraphics[width=\imageCompWidth,height=\imageCompWidth]{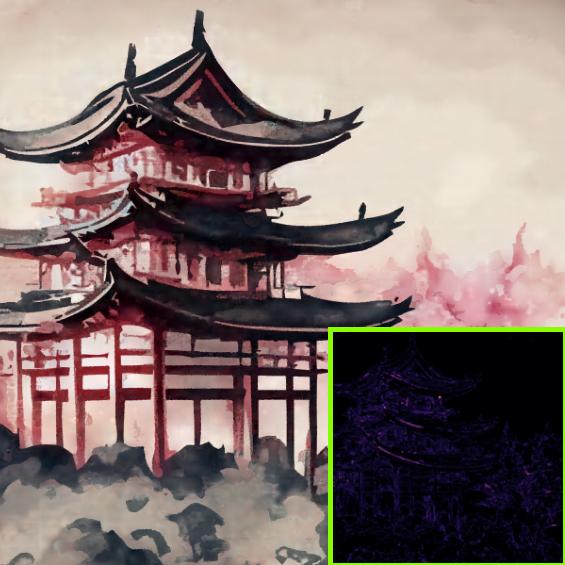}
  } \hspace{\reduceWidth}
\subfloat{
    \includegraphics[width=\imageCompWidth,height=\imageCompWidth]{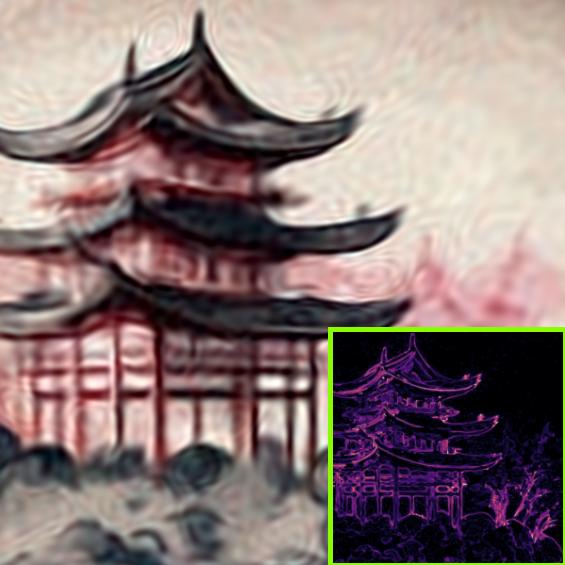}
  } \hspace{\reduceWidth}
\subfloat{
    \includegraphics[width=\imageCompWidth,height=\imageCompWidth]{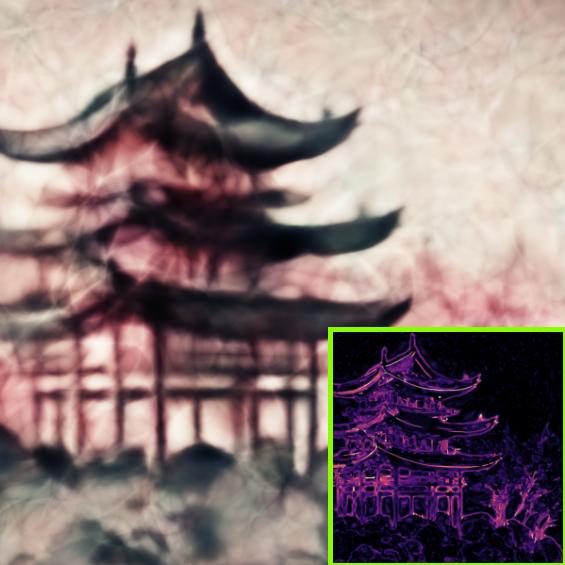}
  } \hspace{\reduceWidth}
\subfloat{
    \includegraphics[width=\imageCompWidth,height=\imageCompWidth]{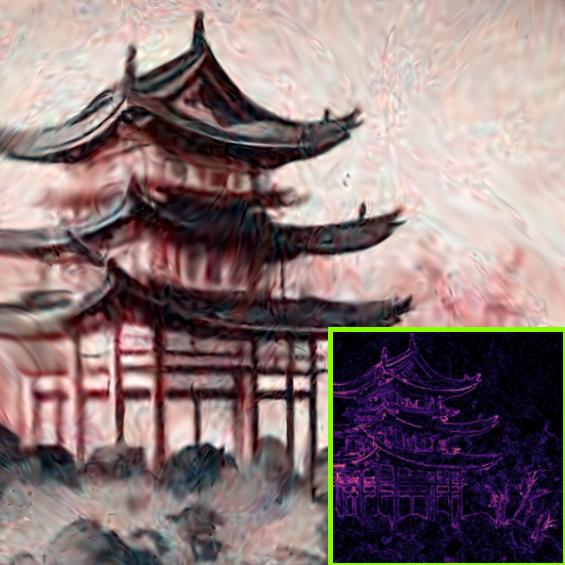}
  } \hspace{\reduceWidth}
\subfloat{
    \includegraphics[width=\imageCompWidth,height=\imageCompWidth]{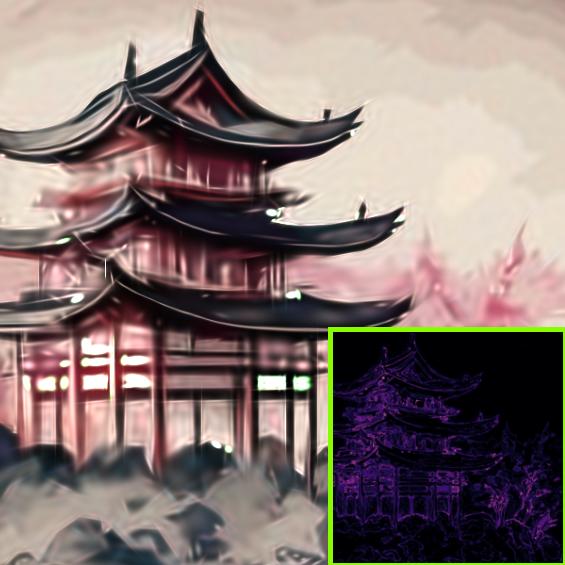}
  } \hspace{\reduceWidth}
\subfloat{
    \includegraphics[width=\imageCompWidth,height=\imageCompWidth]{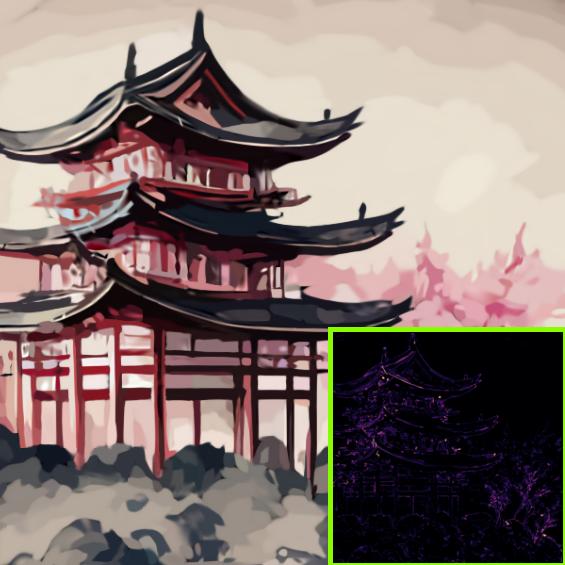}
  } \hspace{\reduceWidth}
\subfloat{
    \includegraphics[width=\imageCompWidth,height=\imageCompWidth]{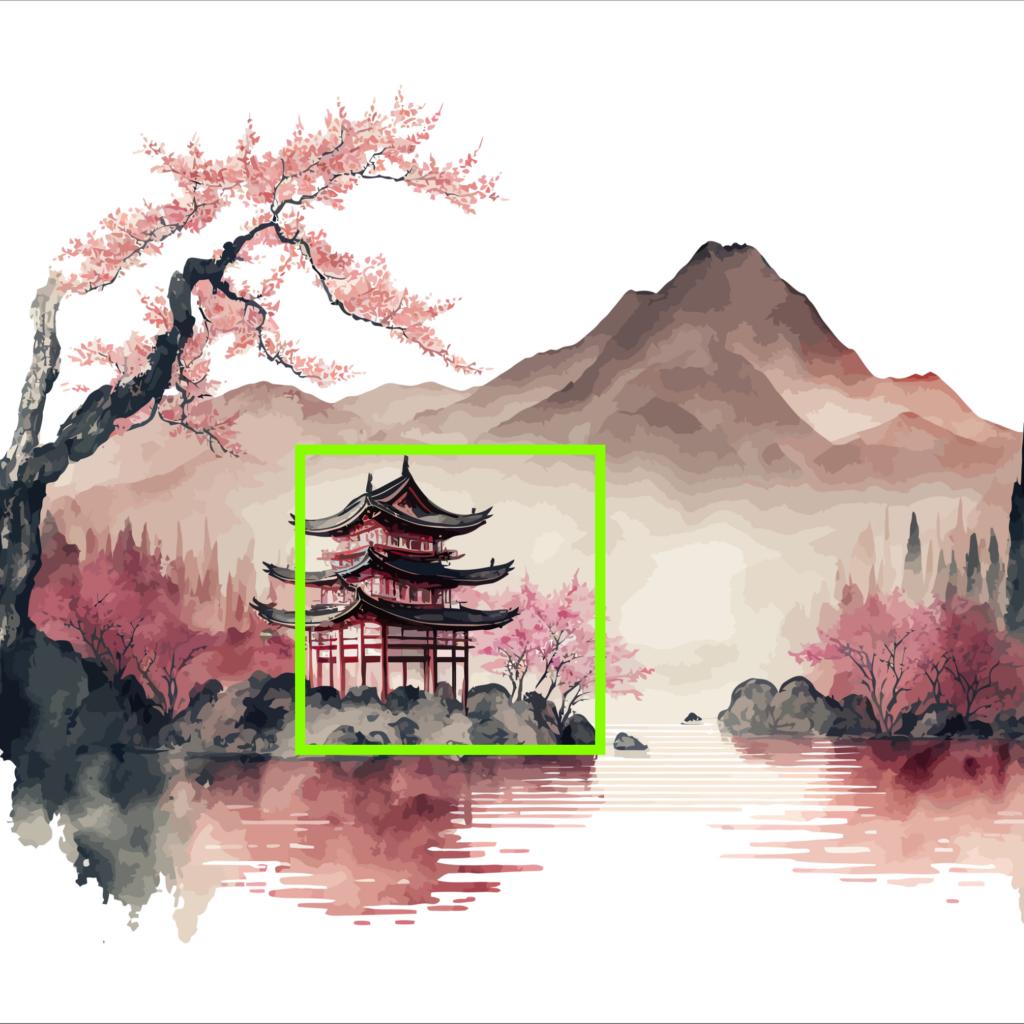}
  }
\vspace{0.2mm} \\
\subfloat{
    \includegraphics[width=\imageCompWidth,height=\imageCompWidth]{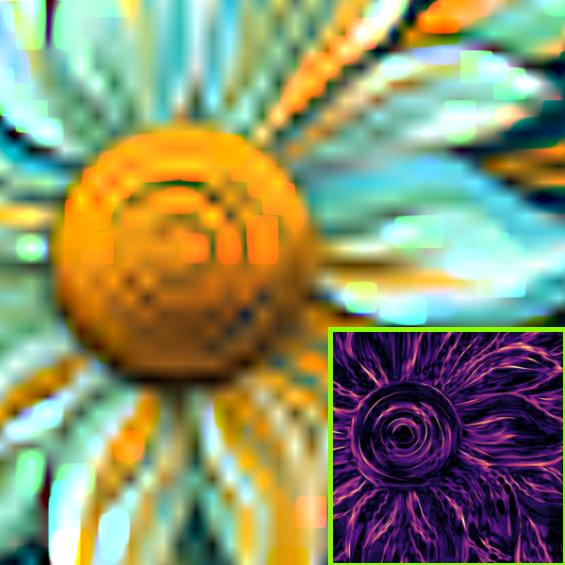}
  } \hspace{\reduceWidth}
\subfloat{
    \includegraphics[width=\imageCompWidth,height=\imageCompWidth]{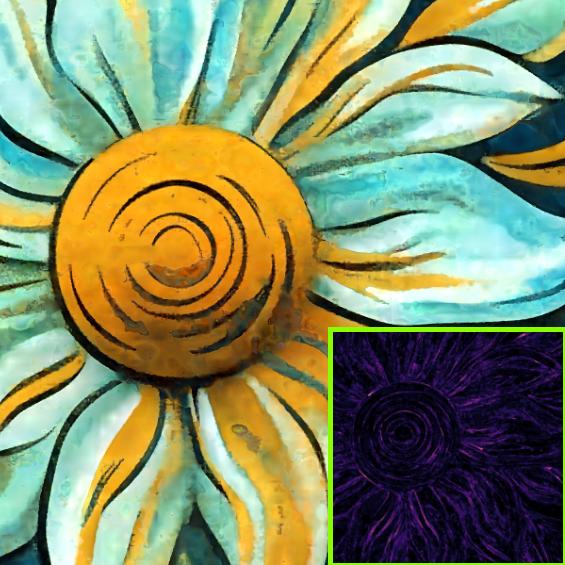}
  } \hspace{\reduceWidth}
\subfloat{
    \includegraphics[width=\imageCompWidth,height=\imageCompWidth]{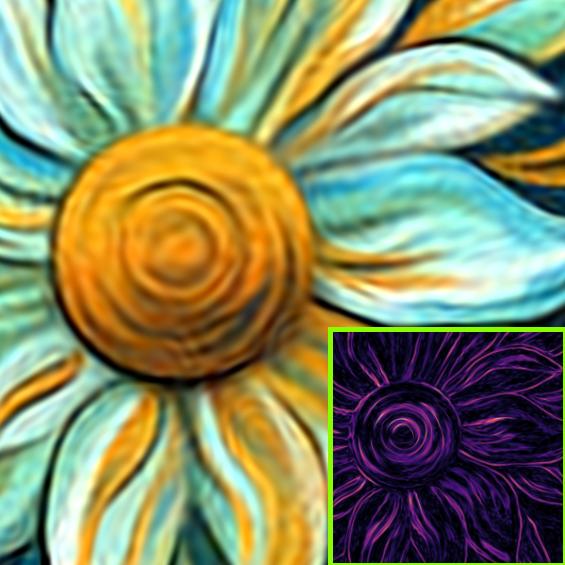}
  } \hspace{\reduceWidth}
\subfloat{
    \includegraphics[width=\imageCompWidth,height=\imageCompWidth]{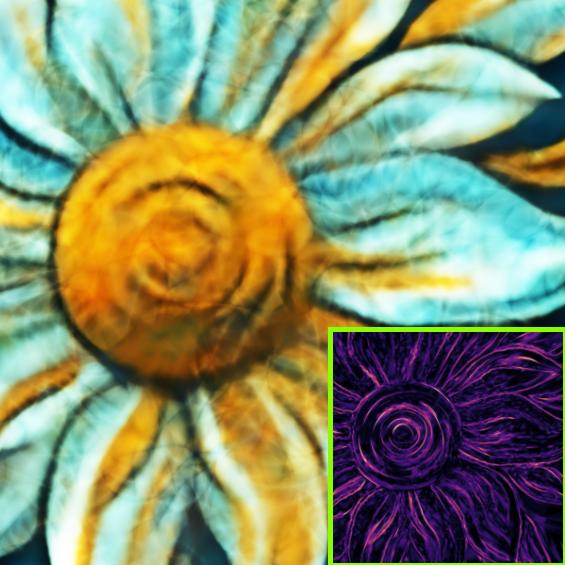}
  } \hspace{\reduceWidth}
\subfloat{
    \includegraphics[width=\imageCompWidth,height=\imageCompWidth]{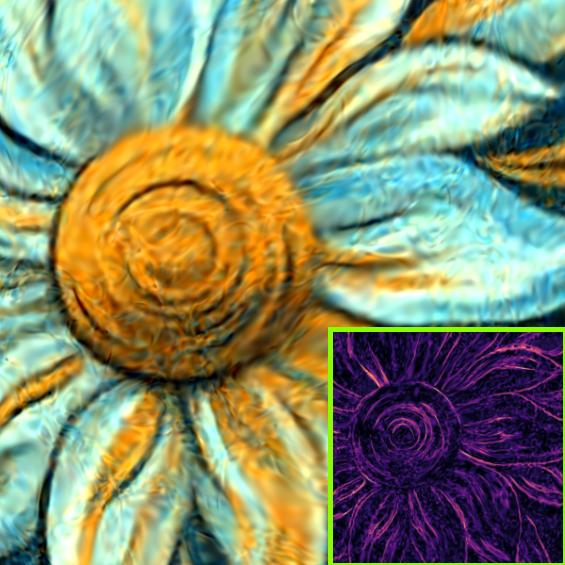}
  } \hspace{\reduceWidth}
\subfloat{
    \includegraphics[width=\imageCompWidth,height=\imageCompWidth]{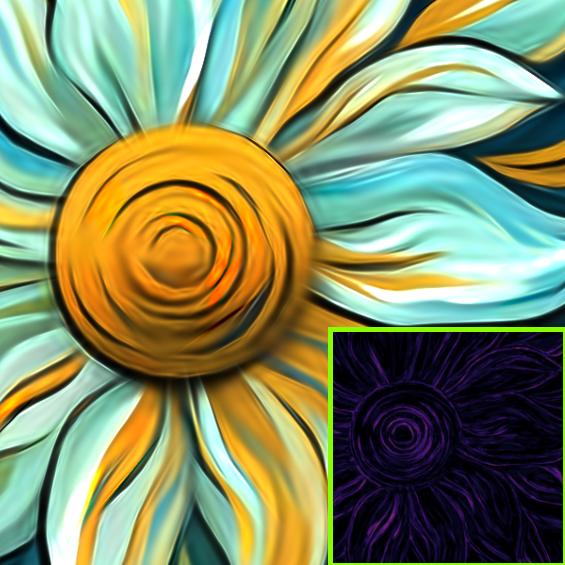}
  } \hspace{\reduceWidth}
\subfloat{
    \includegraphics[width=\imageCompWidth,height=\imageCompWidth]{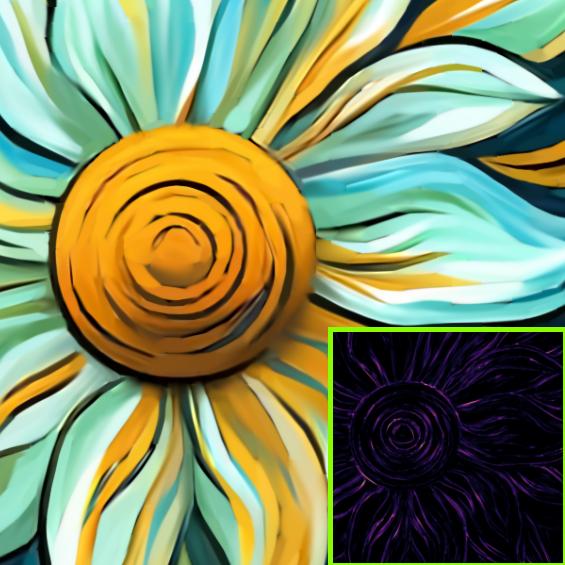}
  } \hspace{\reduceWidth}
\subfloat{
    \includegraphics[width=\imageCompWidth,height=\imageCompWidth]{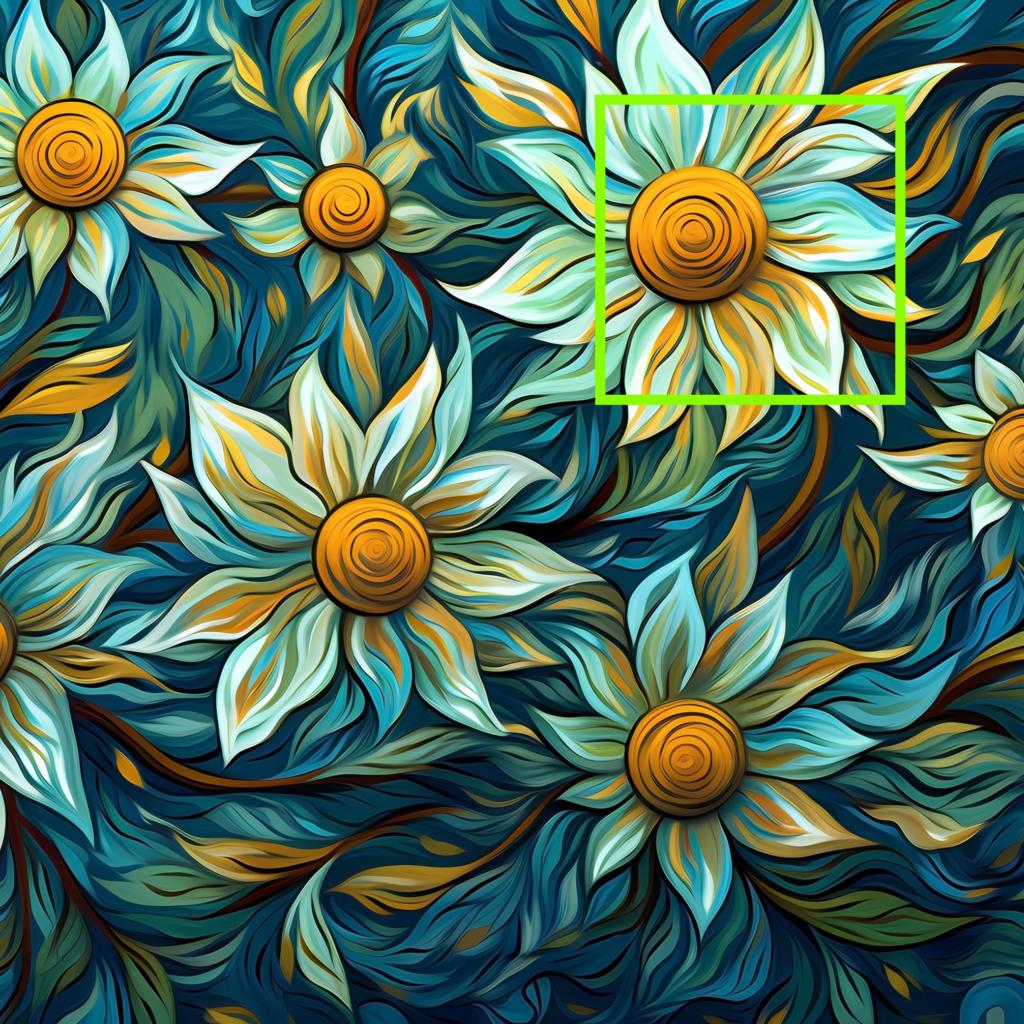}
  }
\vspace{0.2mm} \\
\subfloat{
    \includegraphics[width=\imageCompWidth,height=\imageCompWidth]{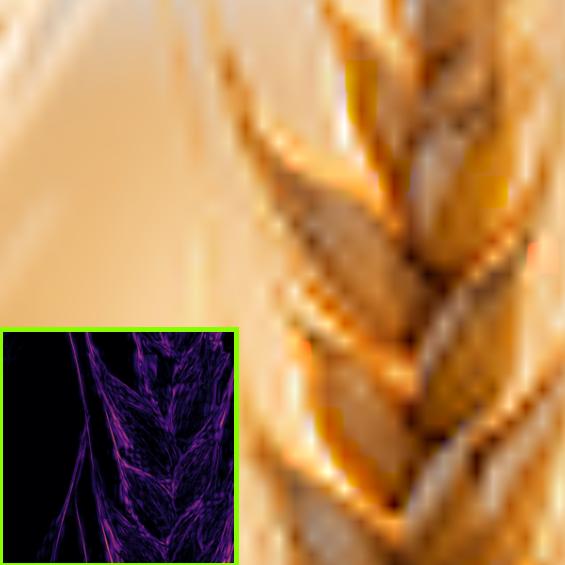}
  } \hspace{\reduceWidth}
\subfloat{
    \includegraphics[width=\imageCompWidth,height=\imageCompWidth]{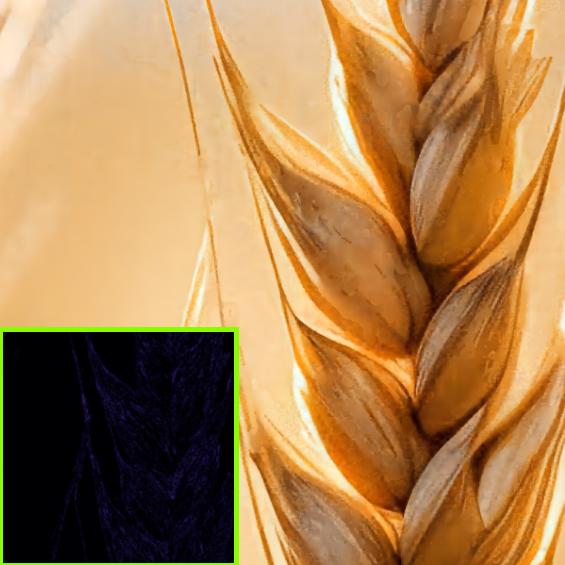}
  } \hspace{\reduceWidth}
\subfloat{
    \includegraphics[width=\imageCompWidth,height=\imageCompWidth]{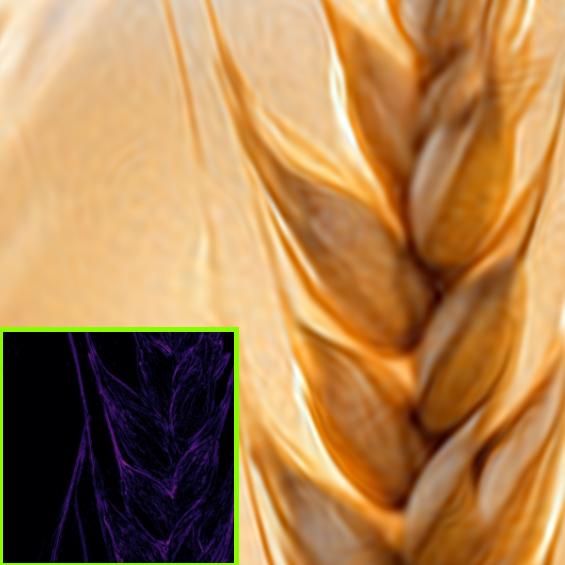}
  } \hspace{\reduceWidth}
\subfloat{
    \includegraphics[width=\imageCompWidth,height=\imageCompWidth]{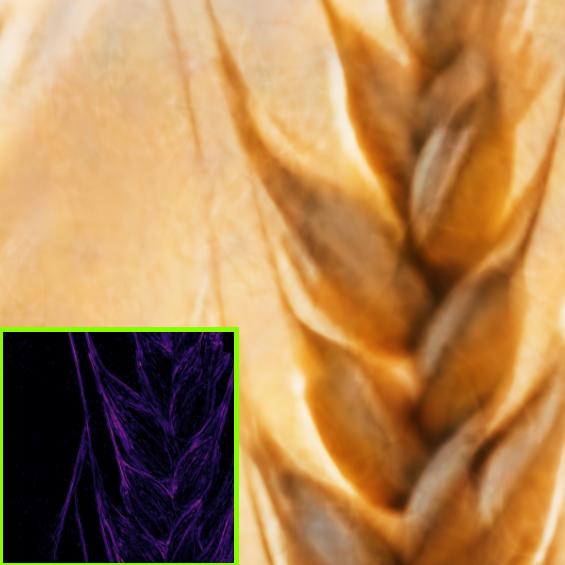}
  } \hspace{\reduceWidth}
\subfloat{
    \includegraphics[width=\imageCompWidth,height=\imageCompWidth]{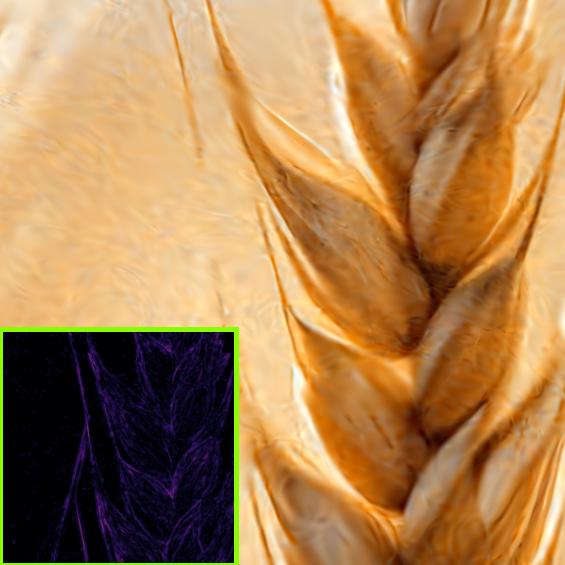}
  } \hspace{\reduceWidth}
\subfloat{
    \includegraphics[width=\imageCompWidth,height=\imageCompWidth]{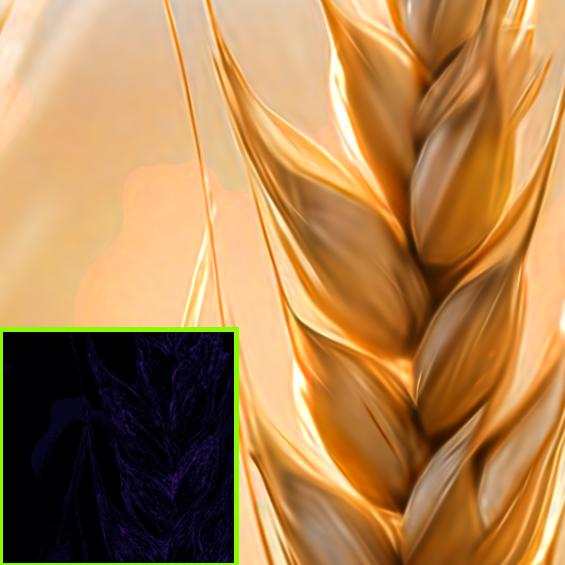}
  } \hspace{\reduceWidth}
\subfloat{
    \includegraphics[width=\imageCompWidth,height=\imageCompWidth]{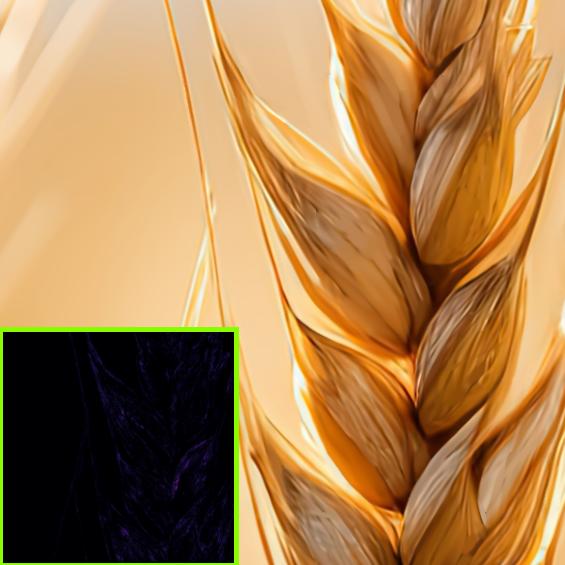}
  } \hspace{\reduceWidth}
\subfloat{
    \includegraphics[width=\imageCompWidth,height=\imageCompWidth]{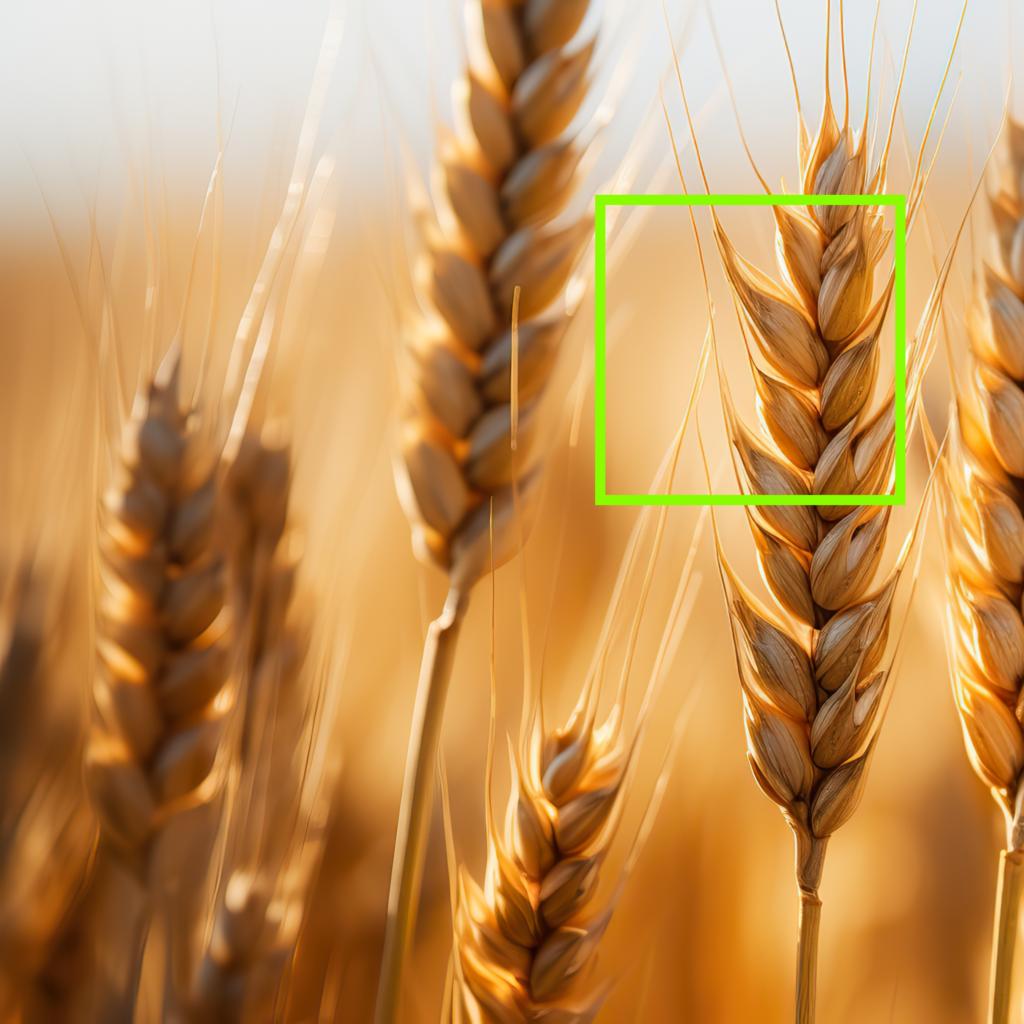}
  }
\vspace{0.2mm} \\
\subfloat{
    \includegraphics[width=\imageCompWidth,height=\imageCompWidth]{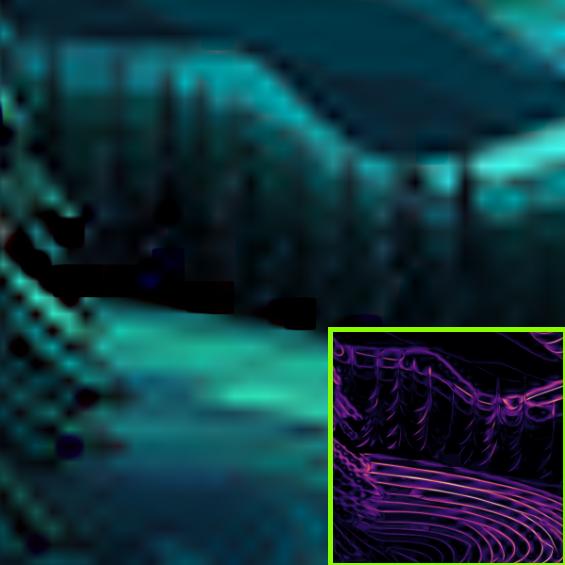}
  } \hspace{\reduceWidth}
\subfloat{
    \includegraphics[width=\imageCompWidth,height=\imageCompWidth]{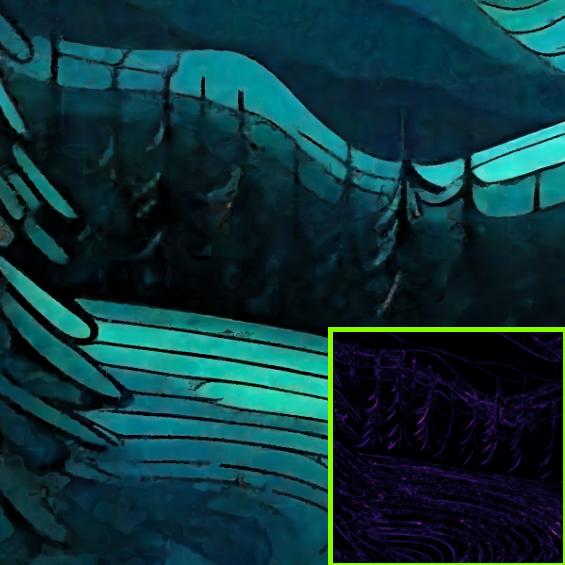}
  } \hspace{\reduceWidth}
\subfloat{
    \includegraphics[width=\imageCompWidth,height=\imageCompWidth]{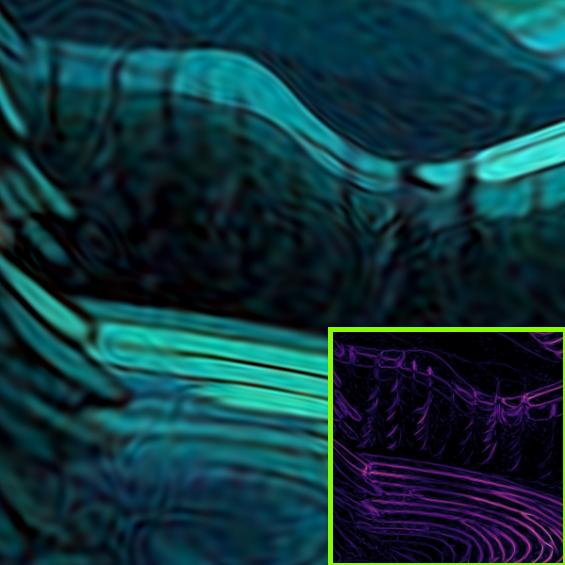}
  } \hspace{\reduceWidth}
\subfloat{
    \includegraphics[width=\imageCompWidth,height=\imageCompWidth]{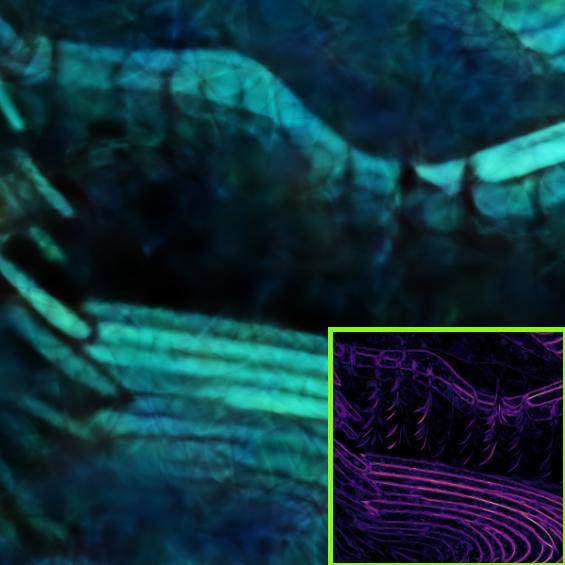}
  } \hspace{\reduceWidth}
\subfloat{
    \includegraphics[width=\imageCompWidth,height=\imageCompWidth]{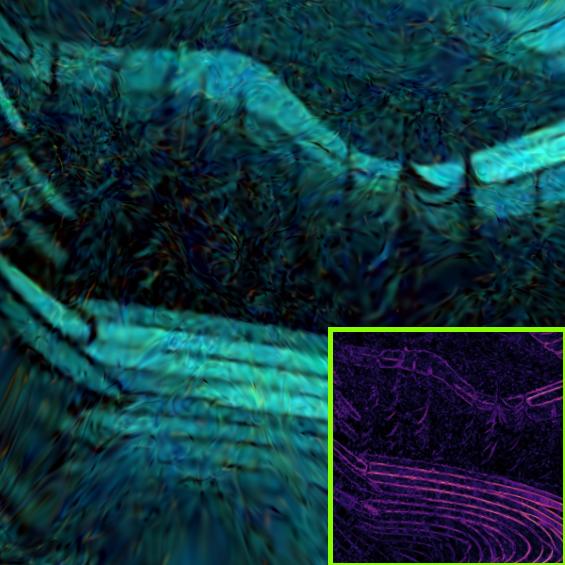}
  } \hspace{\reduceWidth}
\subfloat{
    \includegraphics[width=\imageCompWidth,height=\imageCompWidth]{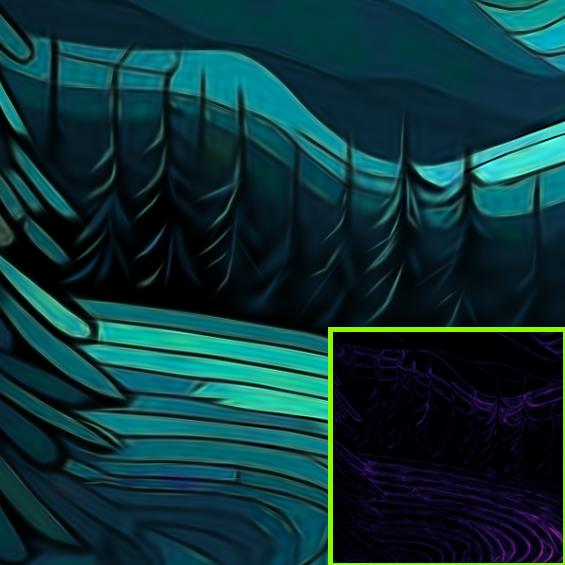}
  } \hspace{\reduceWidth}
\subfloat{
    \includegraphics[width=\imageCompWidth,height=\imageCompWidth]{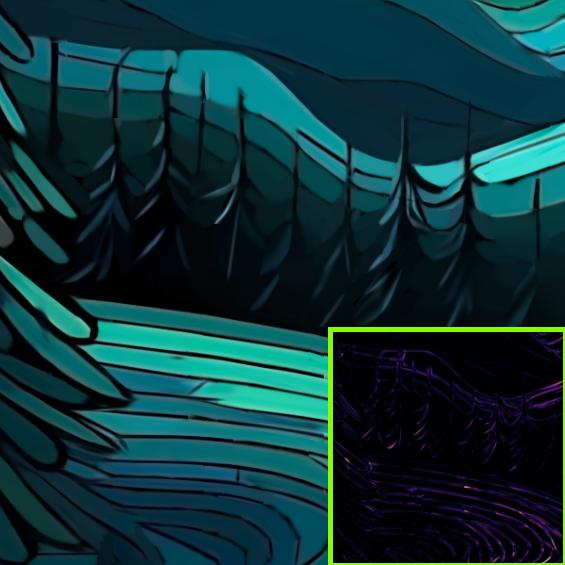}
  } \hspace{\reduceWidth}
\subfloat{
    \includegraphics[width=\imageCompWidth,height=\imageCompWidth]{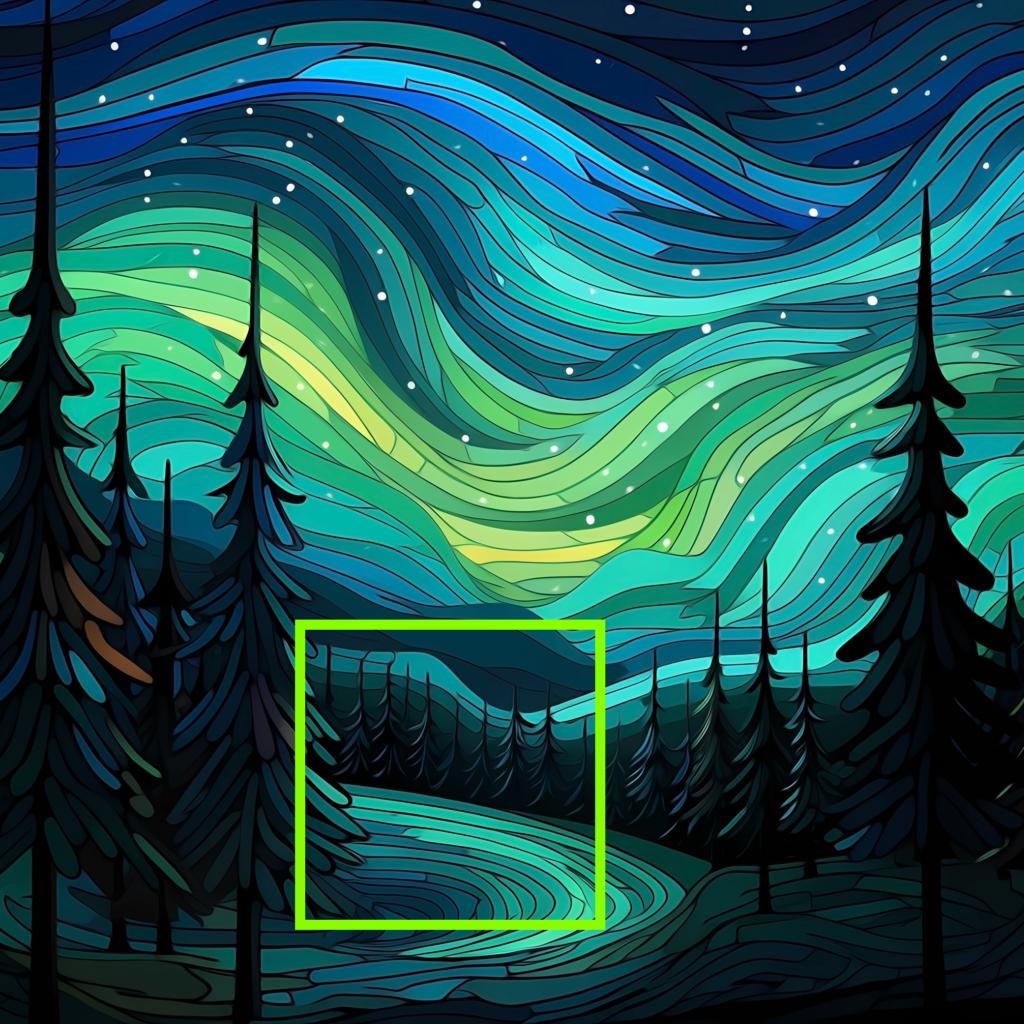}
  }
\vspace{0.2mm} \\
\setcounter{subfigure}{0}
\subfloat[ReLU-F]{
    \includegraphics[width=\imageCompWidth,height=\imageCompWidth]{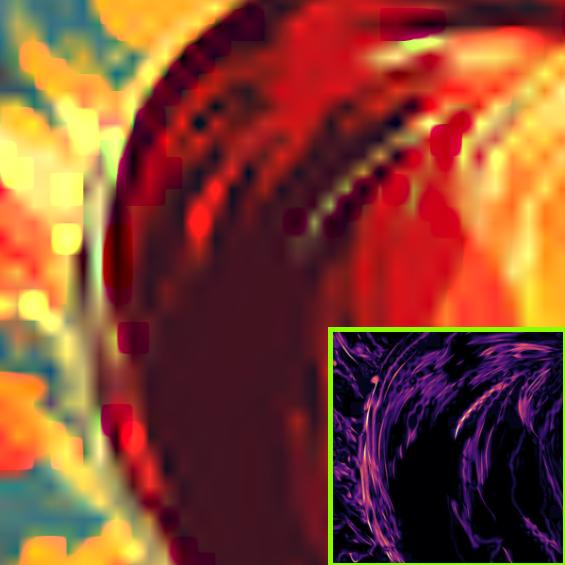}
  } \hspace{\reduceWidth}
\subfloat[I-NGP]{
    \includegraphics[width=\imageCompWidth,height=\imageCompWidth]{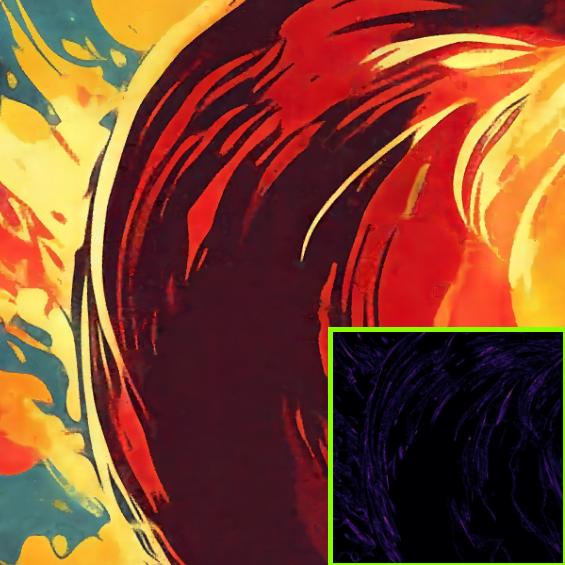}
  } \hspace{\reduceWidth}
\subfloat[SIREN]{
    \includegraphics[width=\imageCompWidth,height=\imageCompWidth]{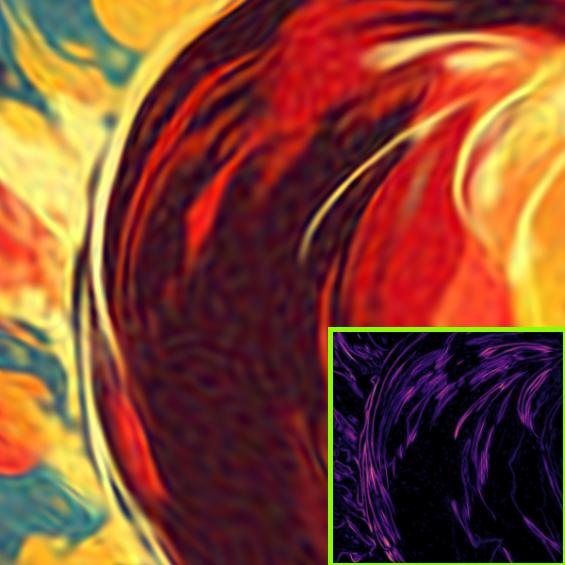}
  } \hspace{\reduceWidth}
\subfloat[FFN]{
    \includegraphics[width=\imageCompWidth,height=\imageCompWidth]{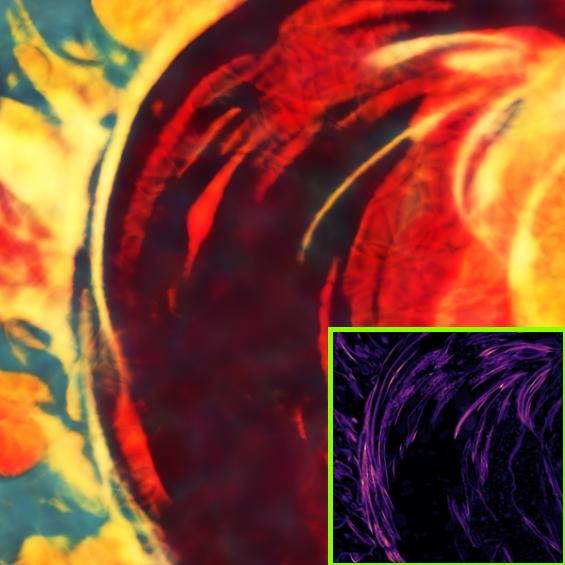}
  } \hspace{\reduceWidth}
\subfloat[WIRE]{
    \includegraphics[width=\imageCompWidth,height=\imageCompWidth]{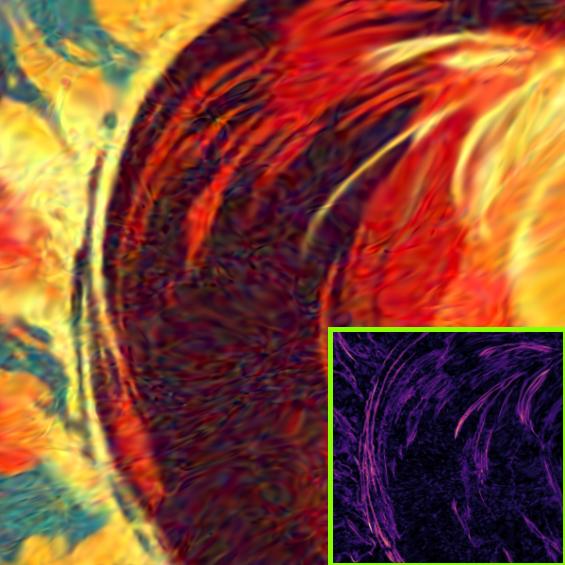}
  } \hspace{\reduceWidth}
\subfloat[GI]{
    \includegraphics[width=\imageCompWidth,height=\imageCompWidth]{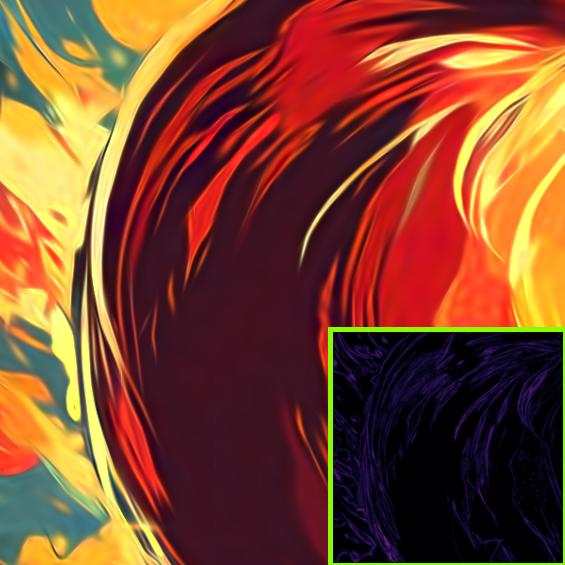}
  } \hspace{\reduceWidth}
\subfloat[Ours]{
    \includegraphics[width=\imageCompWidth,height=\imageCompWidth]{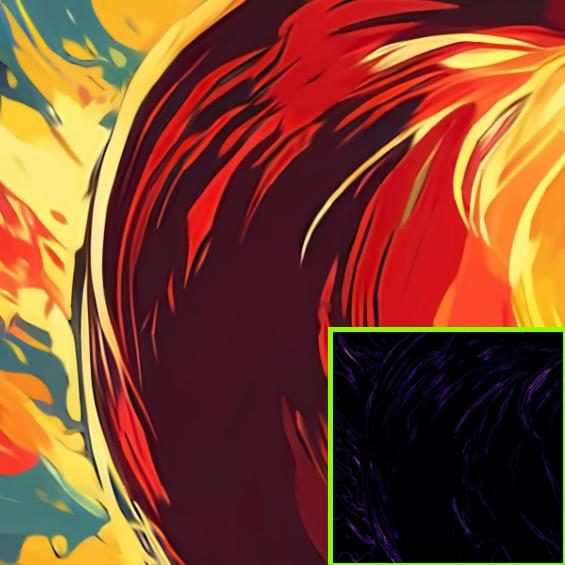}
  } \hspace{\reduceWidth}
\subfloat[Reference]{
    \includegraphics[width=\imageCompWidth,height=\imageCompWidth]{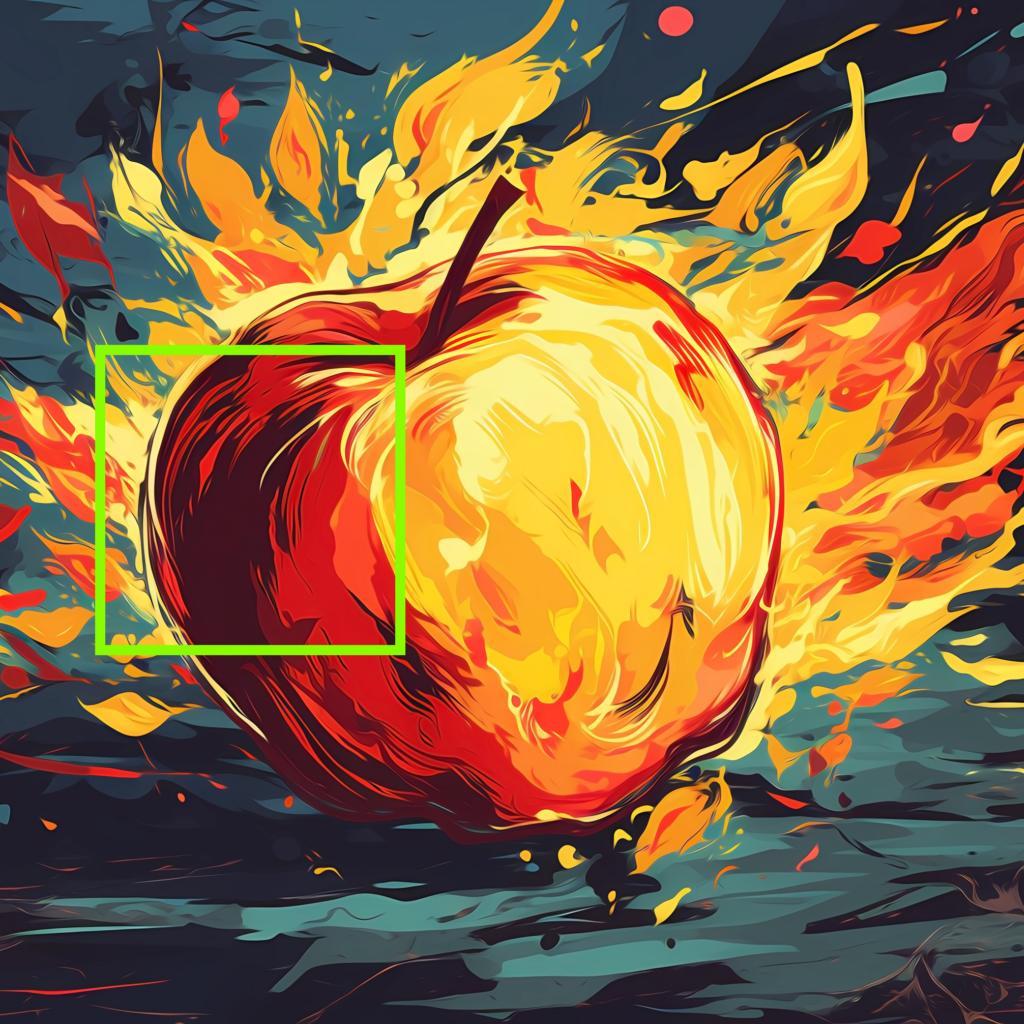}
  }
\Caption{\revise{Qualitative comparison against conventional and neural image representations (\Cref{sec:evaluation-image}).}}{\revise{For the 2K$\times$2K-resolution results shown here, the model sizes (in KB) of ReLU-F, I-NGP, SIREN, FFN, WIRE, GI, and \methodName are 164, 166, 161, 154, 159, 164, and 160, respectively.}}
\label{fig:evaluation-image}
\end{figure*}

\newcommand{\lodCompWidth}{0.150\linewidth}
\begin{figure*}[p]
\centering
\subfloat{
    \includegraphics[width=\lodCompWidth,height=\lodCompWidth]{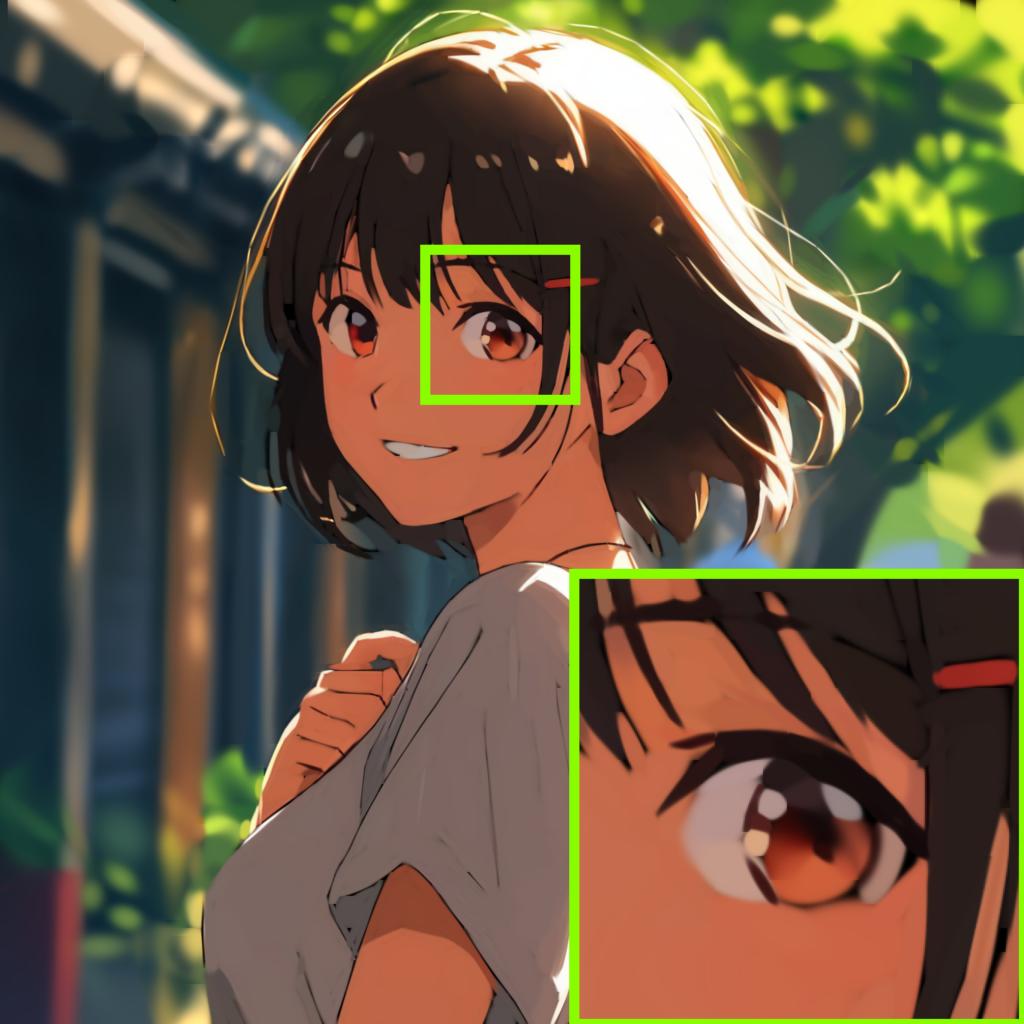}
  }
\subfloat{
    \includegraphics[width=\lodCompWidth,height=\lodCompWidth]{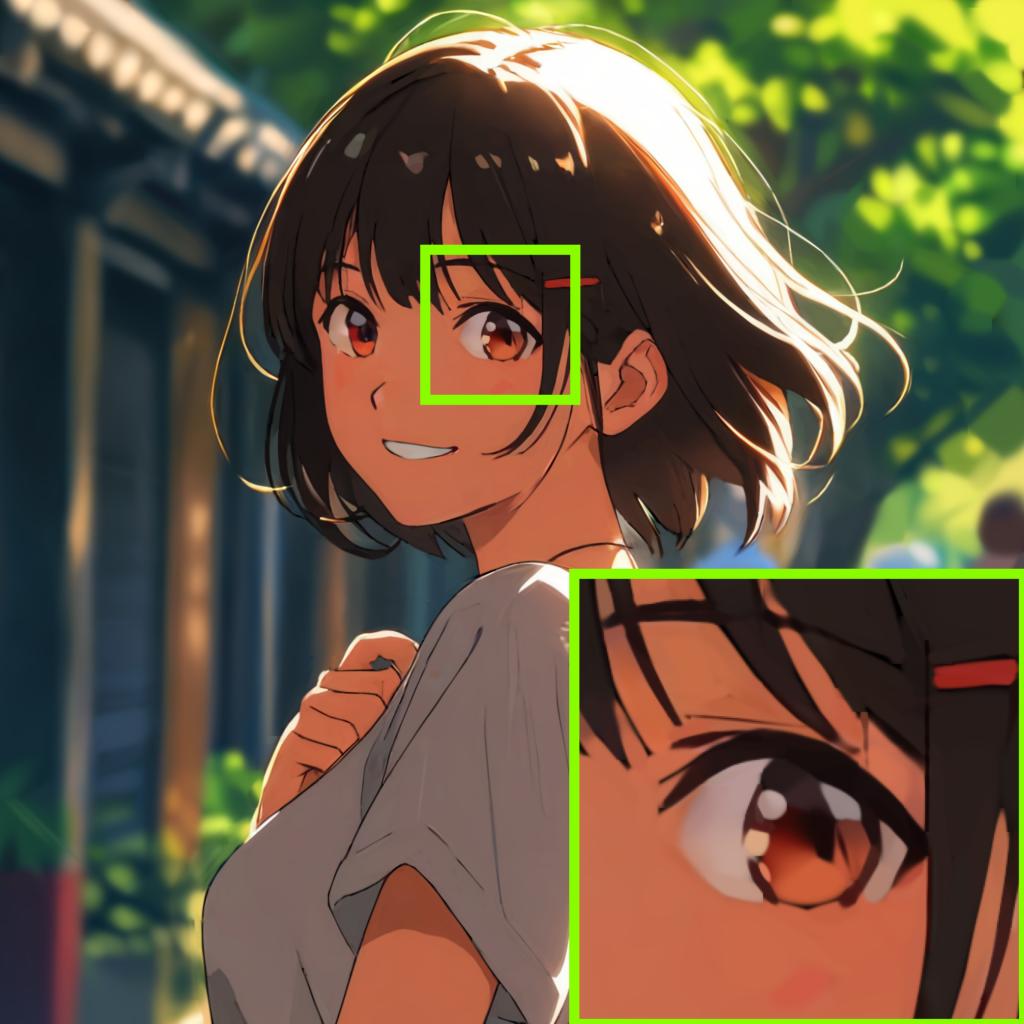}
  }
\subfloat{
    \includegraphics[width=\lodCompWidth,height=\lodCompWidth]{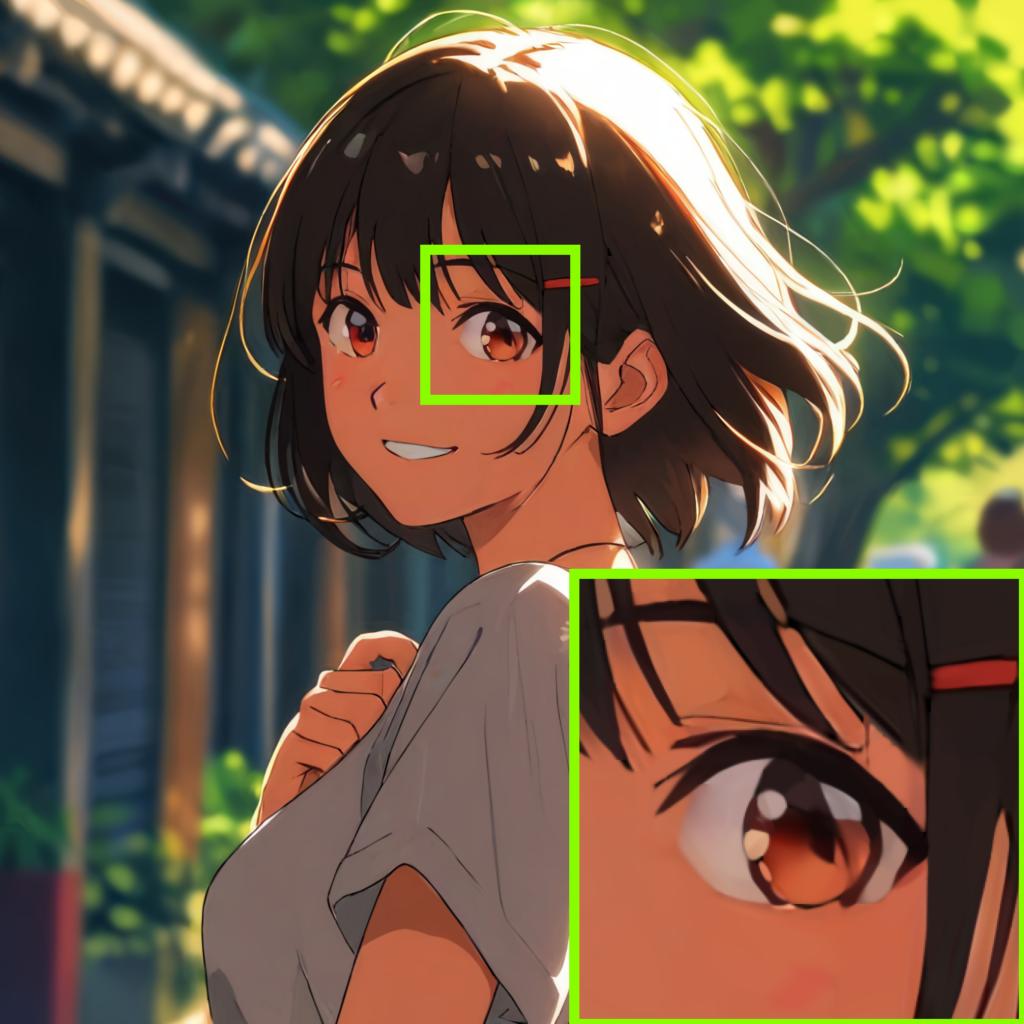}
  }
\subfloat{
    \includegraphics[width=\lodCompWidth,height=\lodCompWidth]{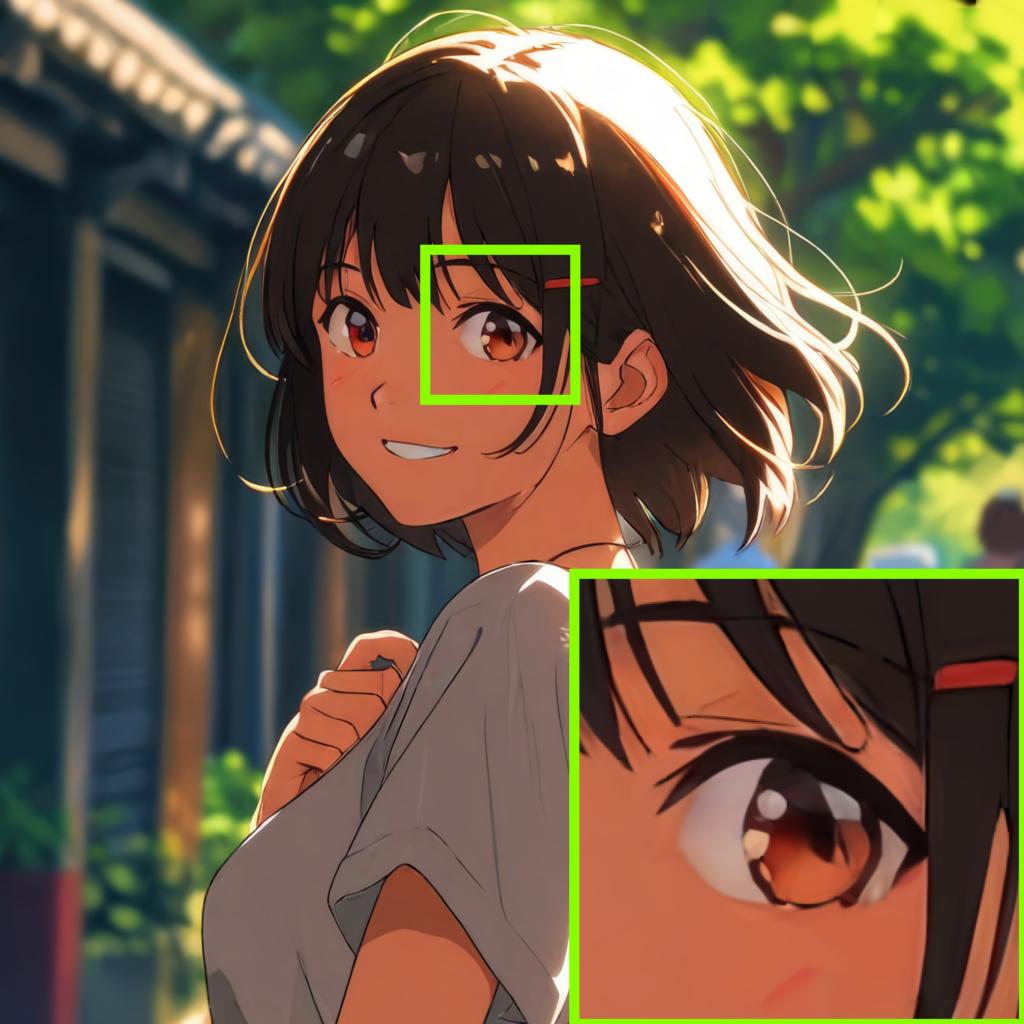}
  }
\subfloat{
    \includegraphics[width=\lodCompWidth,height=\lodCompWidth]{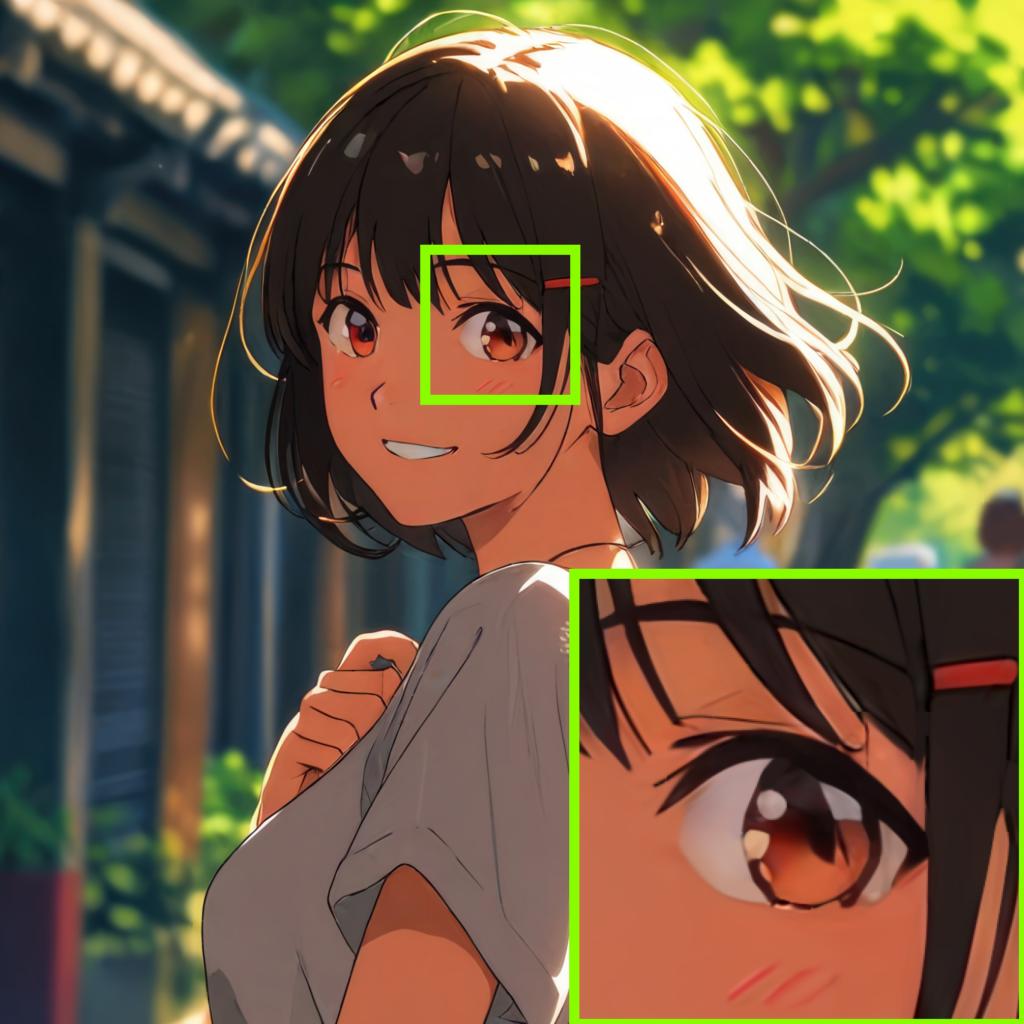}
  }
\subfloat{
    \includegraphics[width=\lodCompWidth,height=\lodCompWidth]{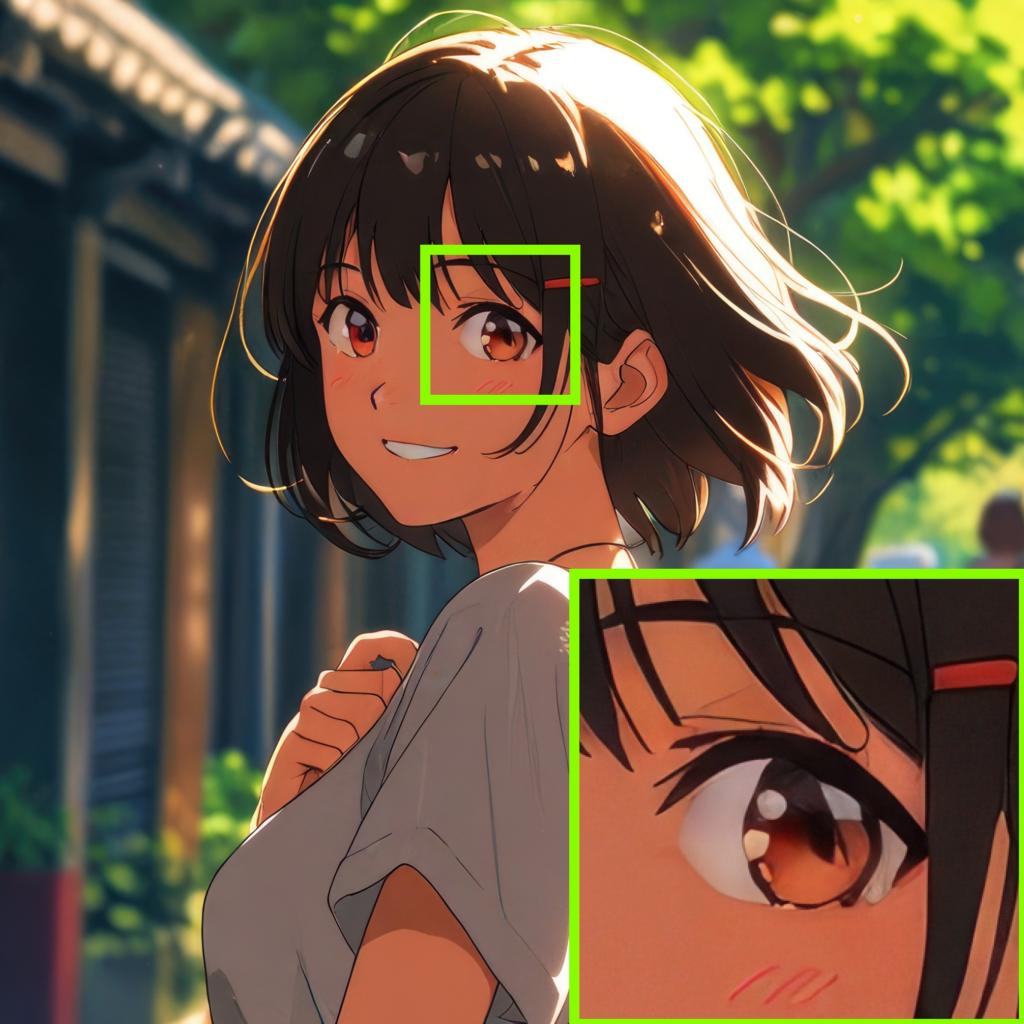}
  }
\vspace{2mm} \\
\subfloat{
    \includegraphics[width=\lodCompWidth,height=\lodCompWidth]{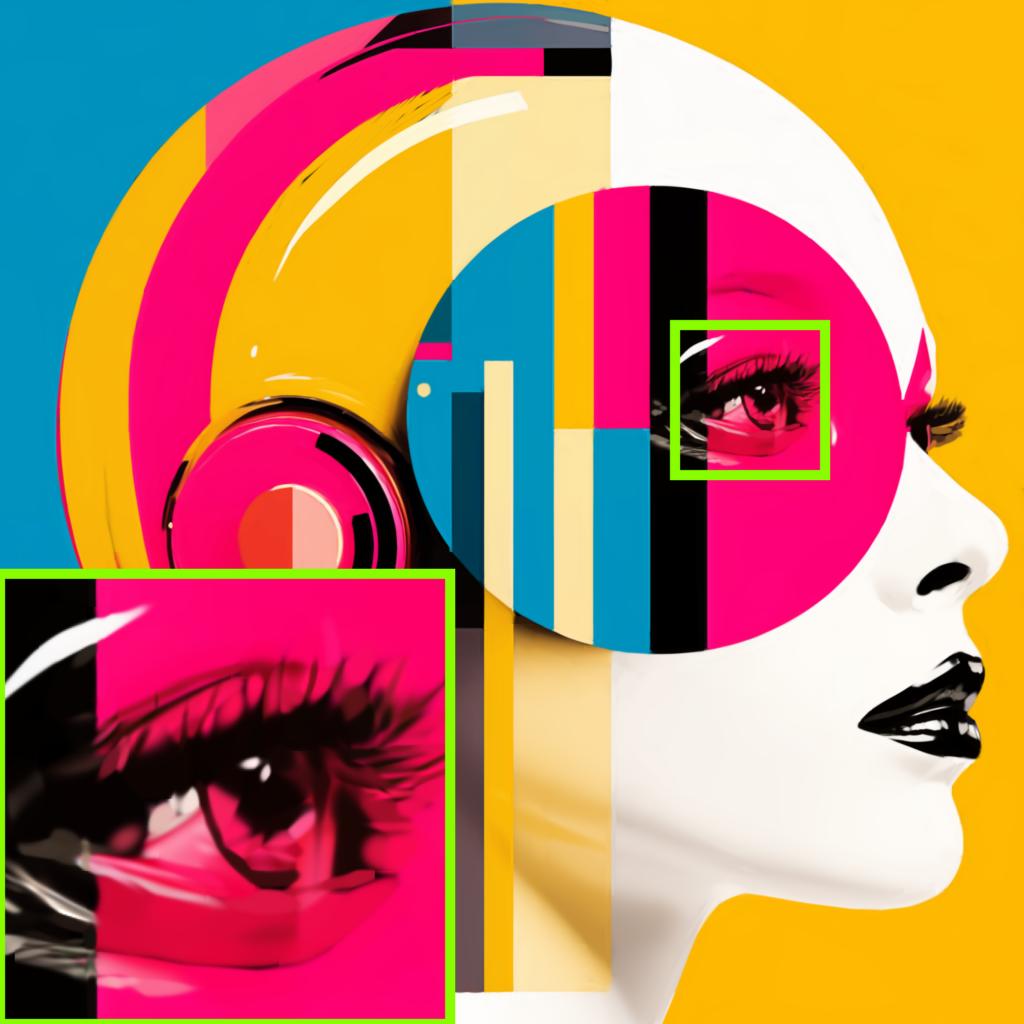}
  }
\subfloat{
    \includegraphics[width=\lodCompWidth,height=\lodCompWidth]{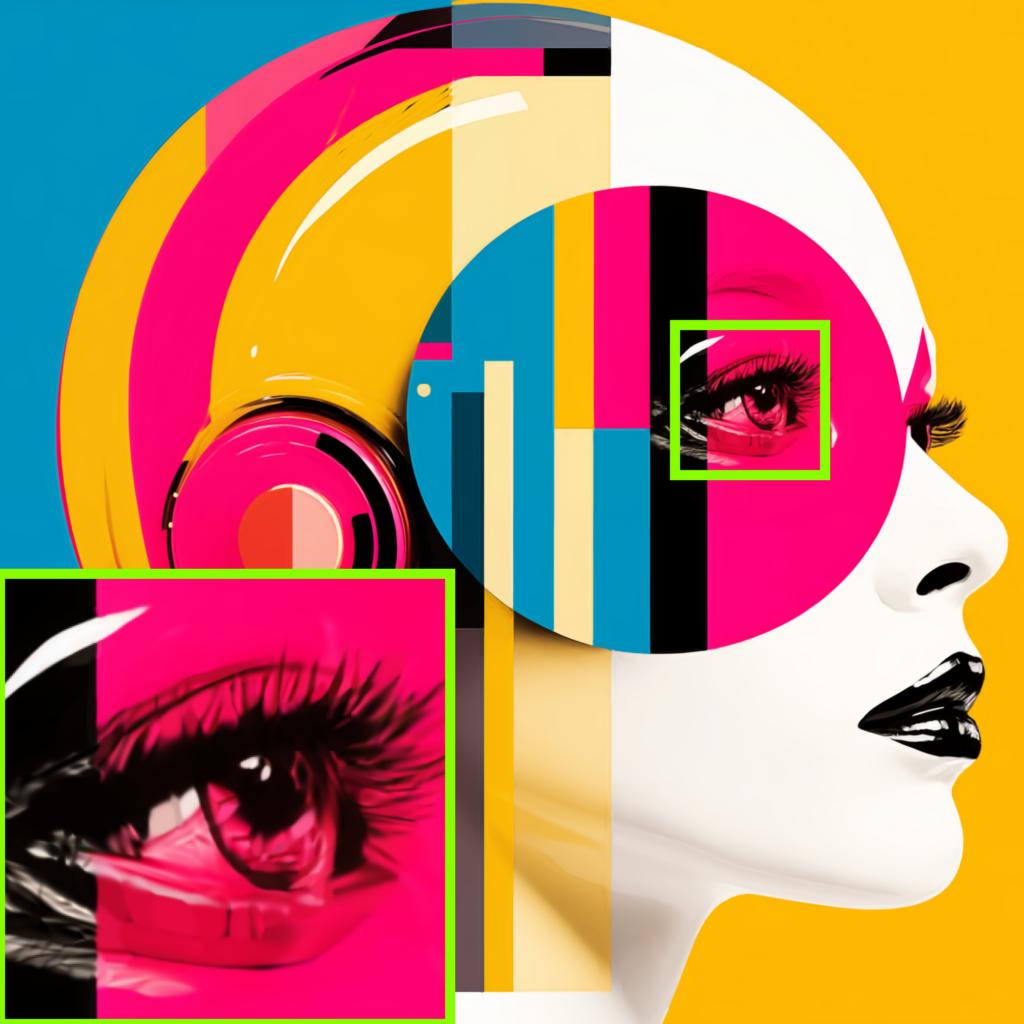}
  }
\subfloat{
    \includegraphics[width=\lodCompWidth,height=\lodCompWidth]{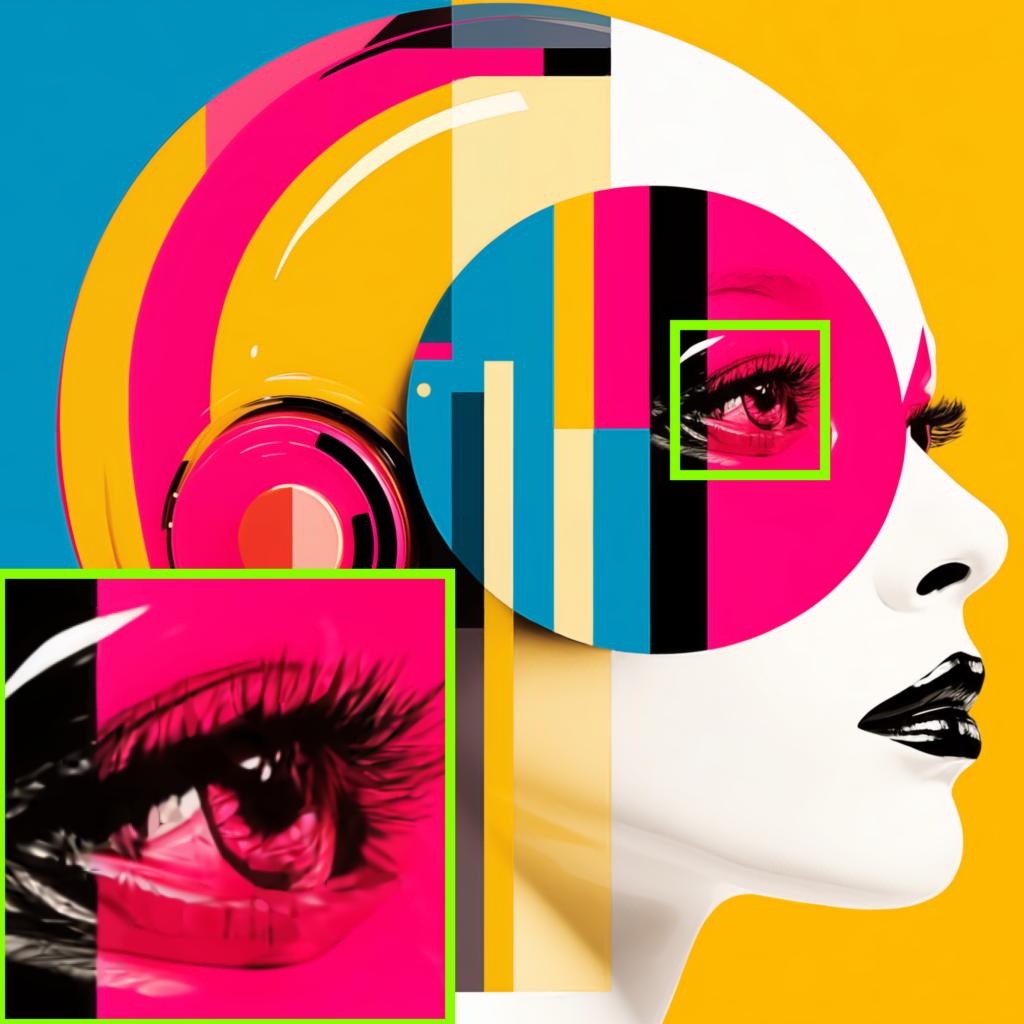}
  }
\subfloat{
    \includegraphics[width=\lodCompWidth,height=\lodCompWidth]{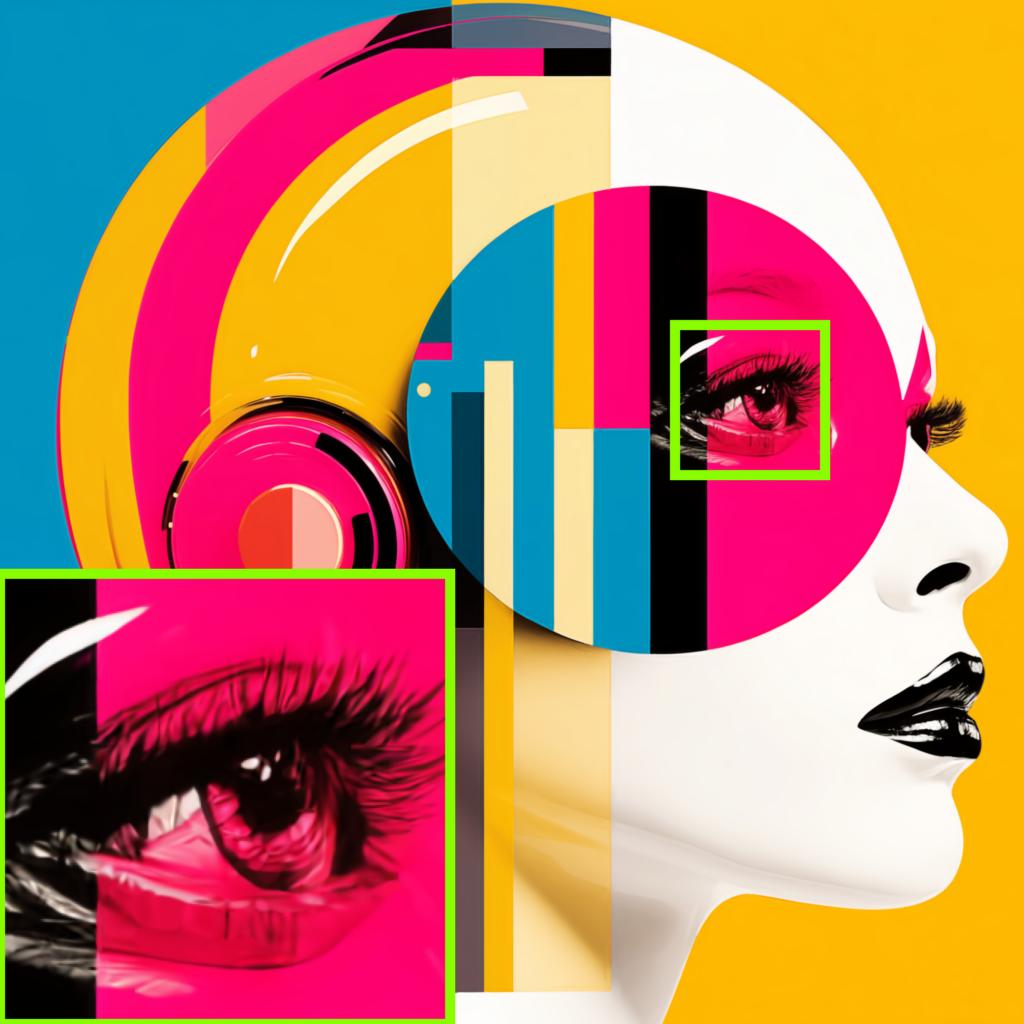}
  }
\subfloat{
    \includegraphics[width=\lodCompWidth,height=\lodCompWidth]{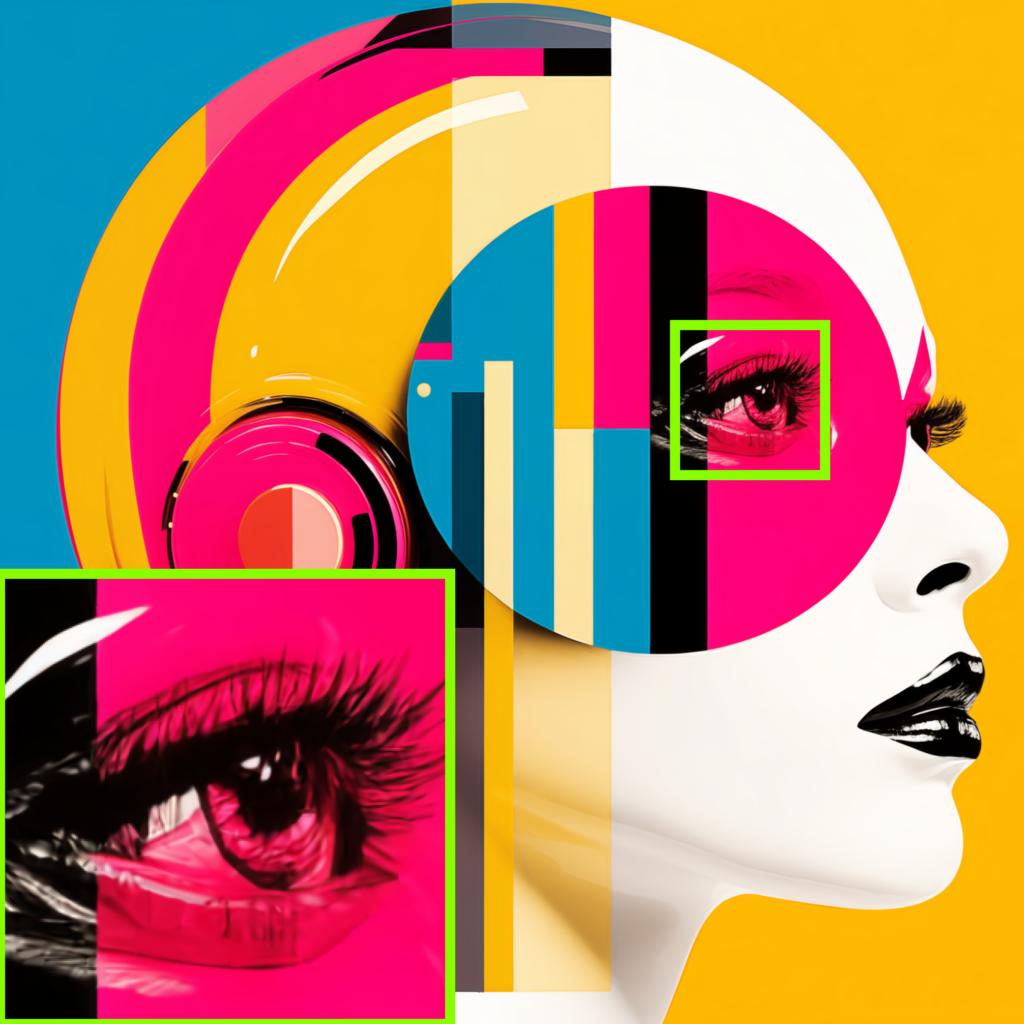}
  }
\subfloat{
    \includegraphics[width=\lodCompWidth,height=\lodCompWidth]{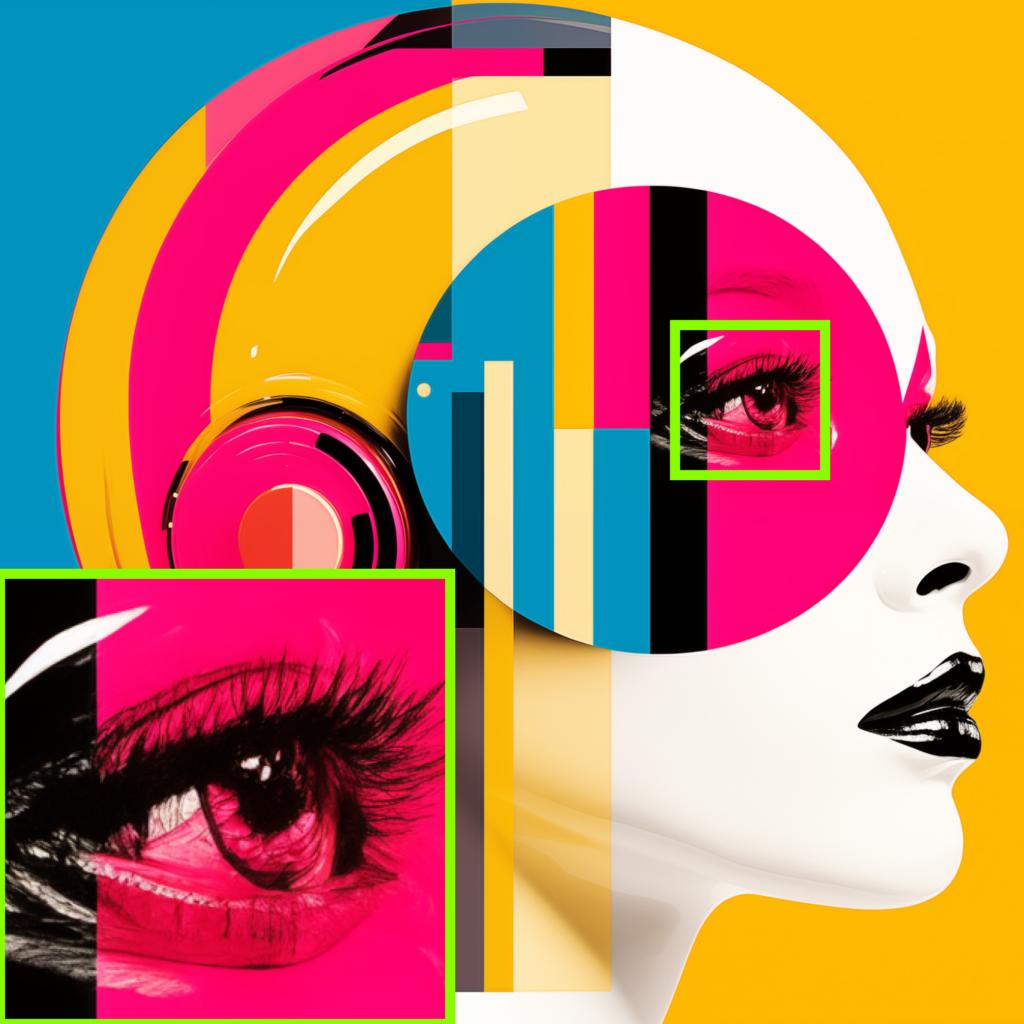}
  }
\vspace{2mm} \\
\subfloat{
    \includegraphics[width=\lodCompWidth,height=\lodCompWidth]{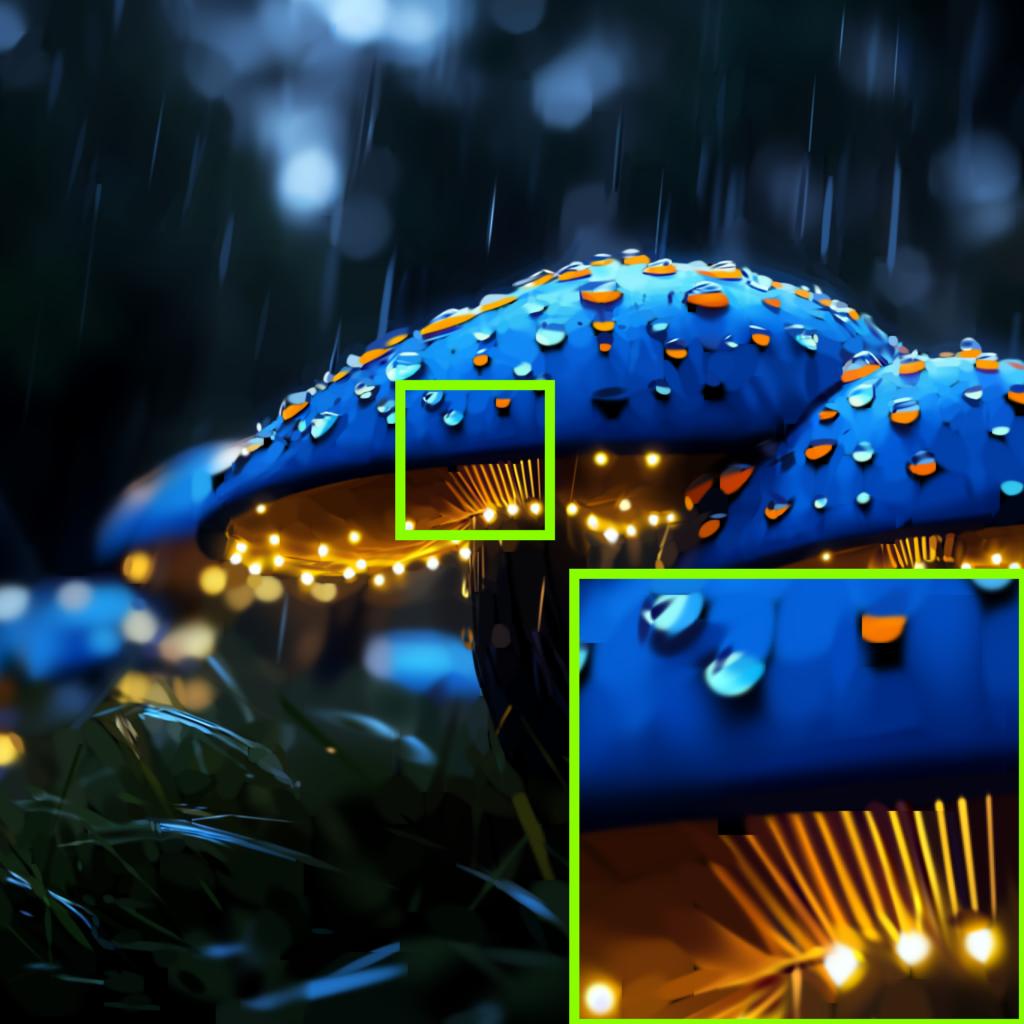}
  }
\subfloat{
    \includegraphics[width=\lodCompWidth,height=\lodCompWidth]{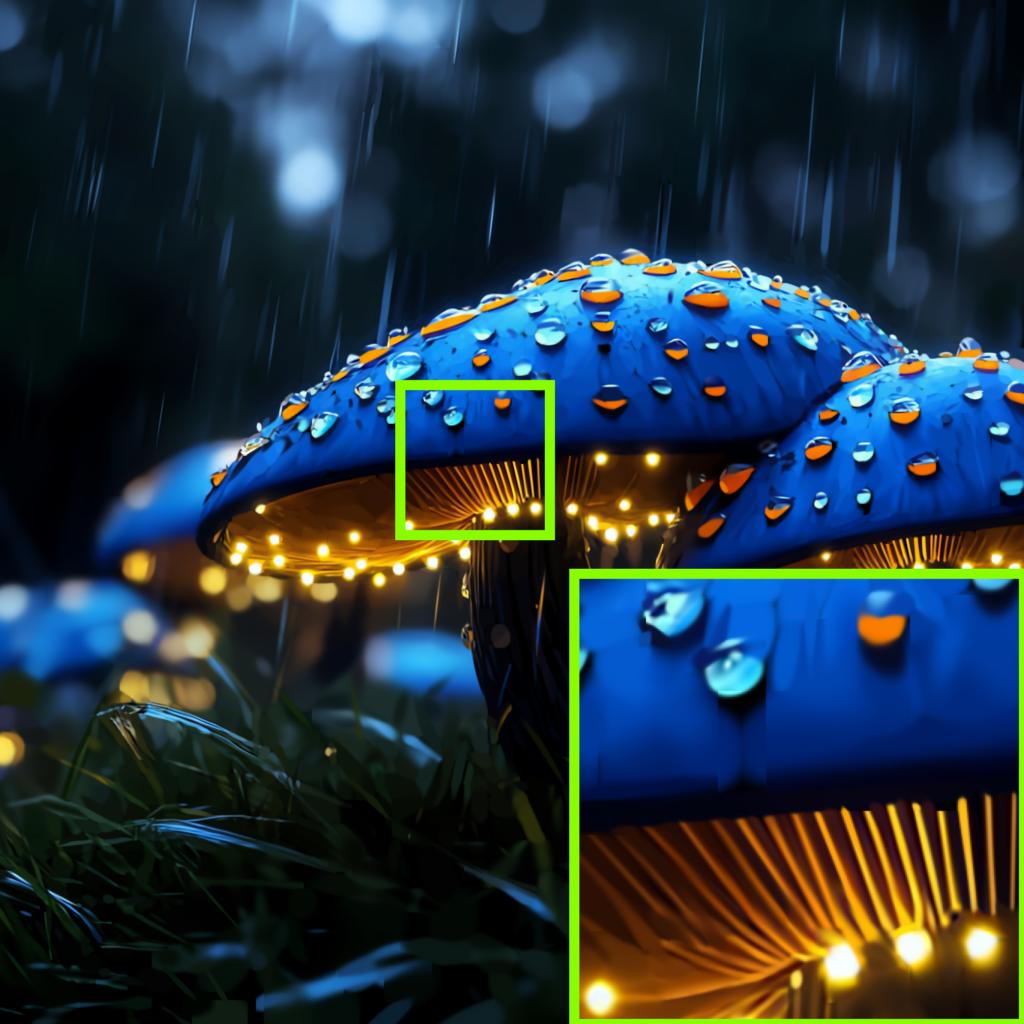}
  }
\subfloat{
    \includegraphics[width=\lodCompWidth,height=\lodCompWidth]{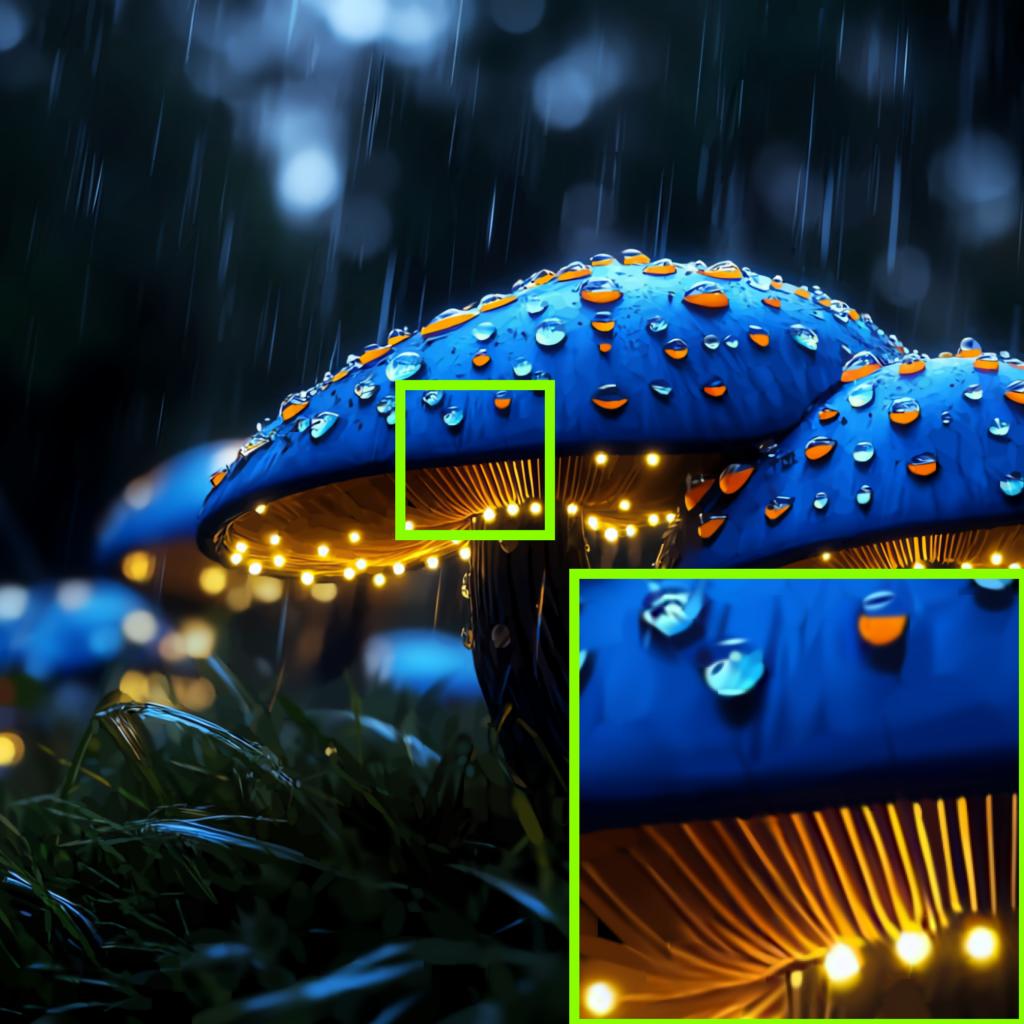}
  }
\subfloat{
    \includegraphics[width=\lodCompWidth,height=\lodCompWidth]{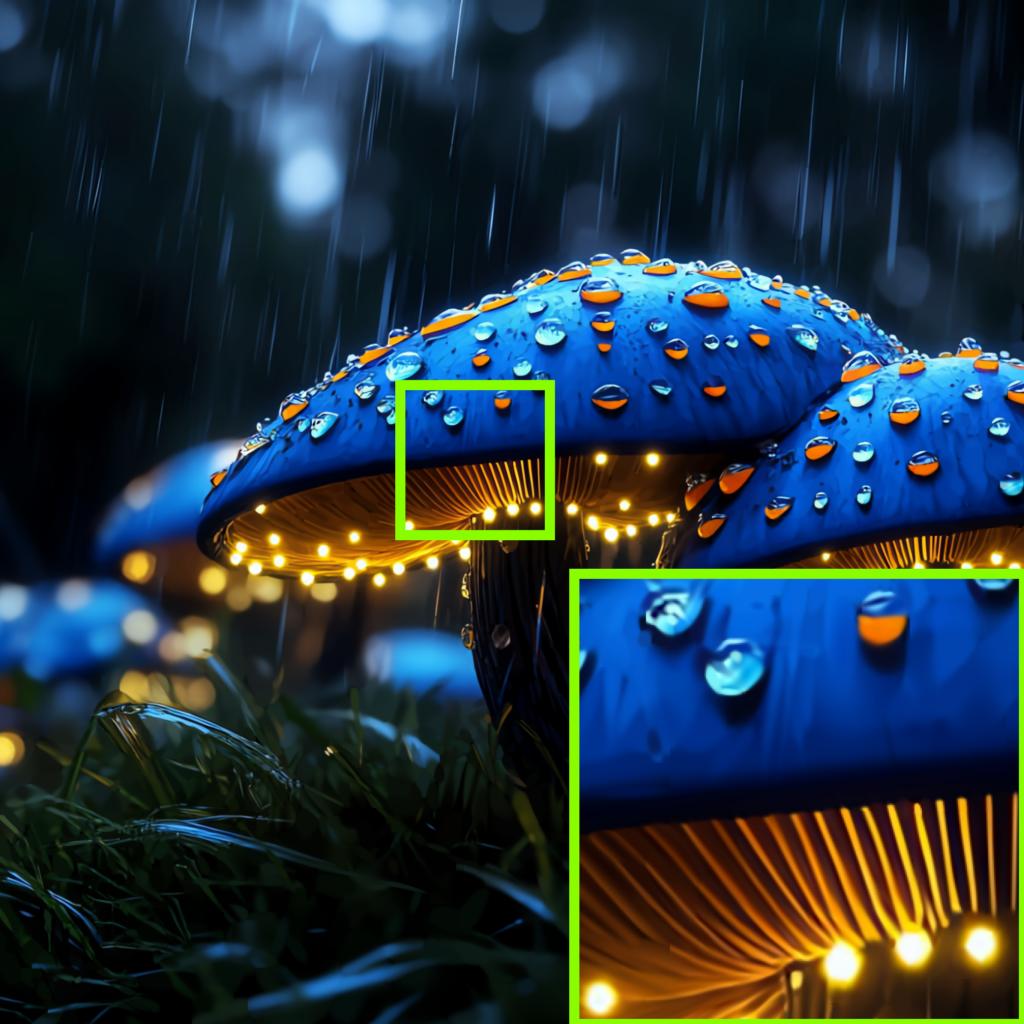}
  }
\subfloat{
    \includegraphics[width=\lodCompWidth,height=\lodCompWidth]{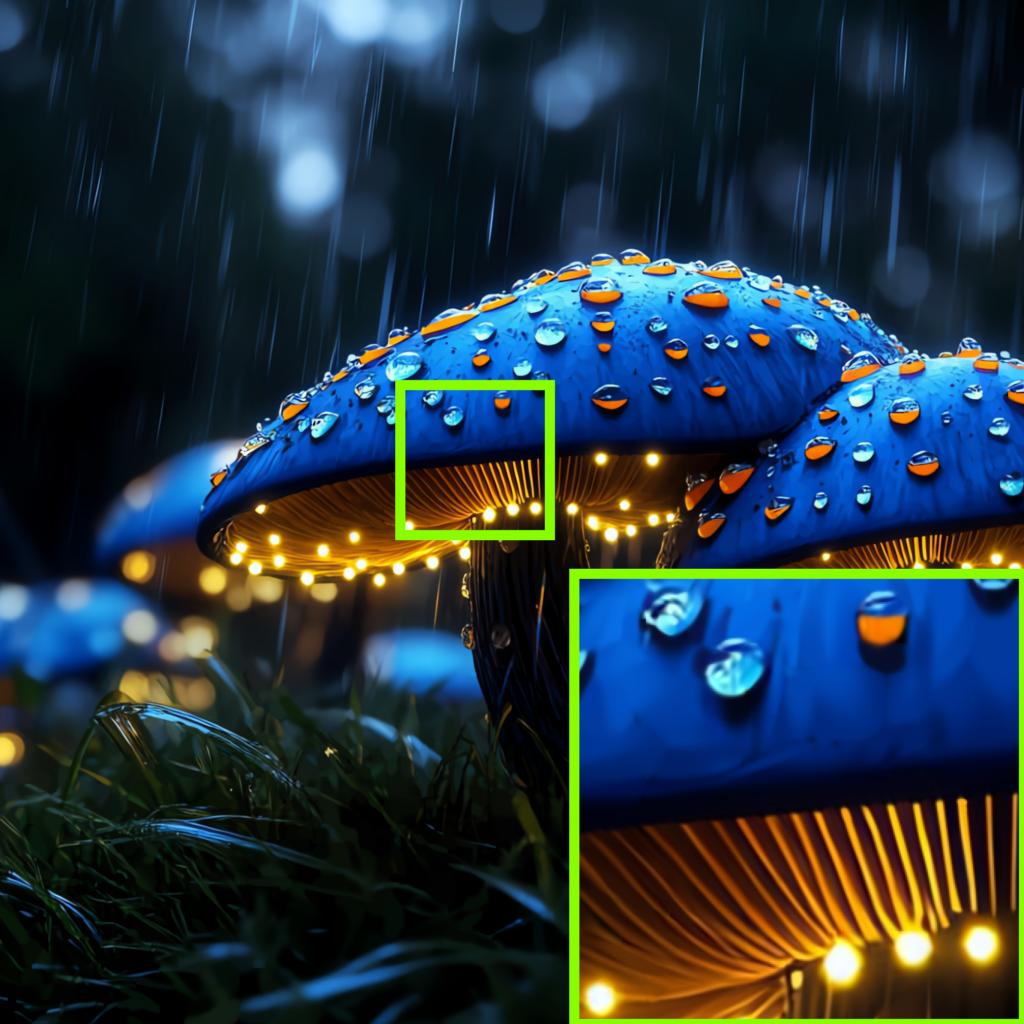}
  }
\subfloat{
    \includegraphics[width=\lodCompWidth,height=\lodCompWidth]{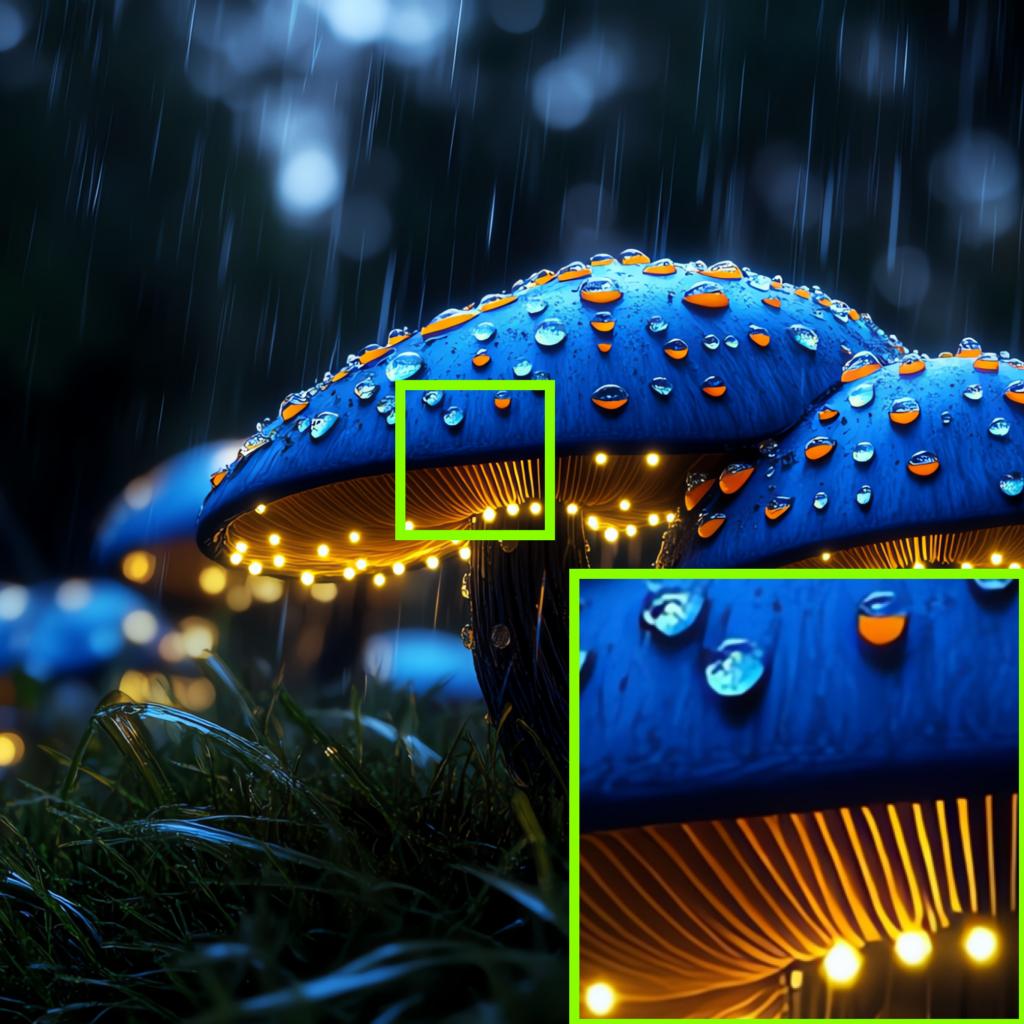}
  }
\vspace{2mm} \\
\subfloat{
    \includegraphics[width=\lodCompWidth,height=\lodCompWidth]{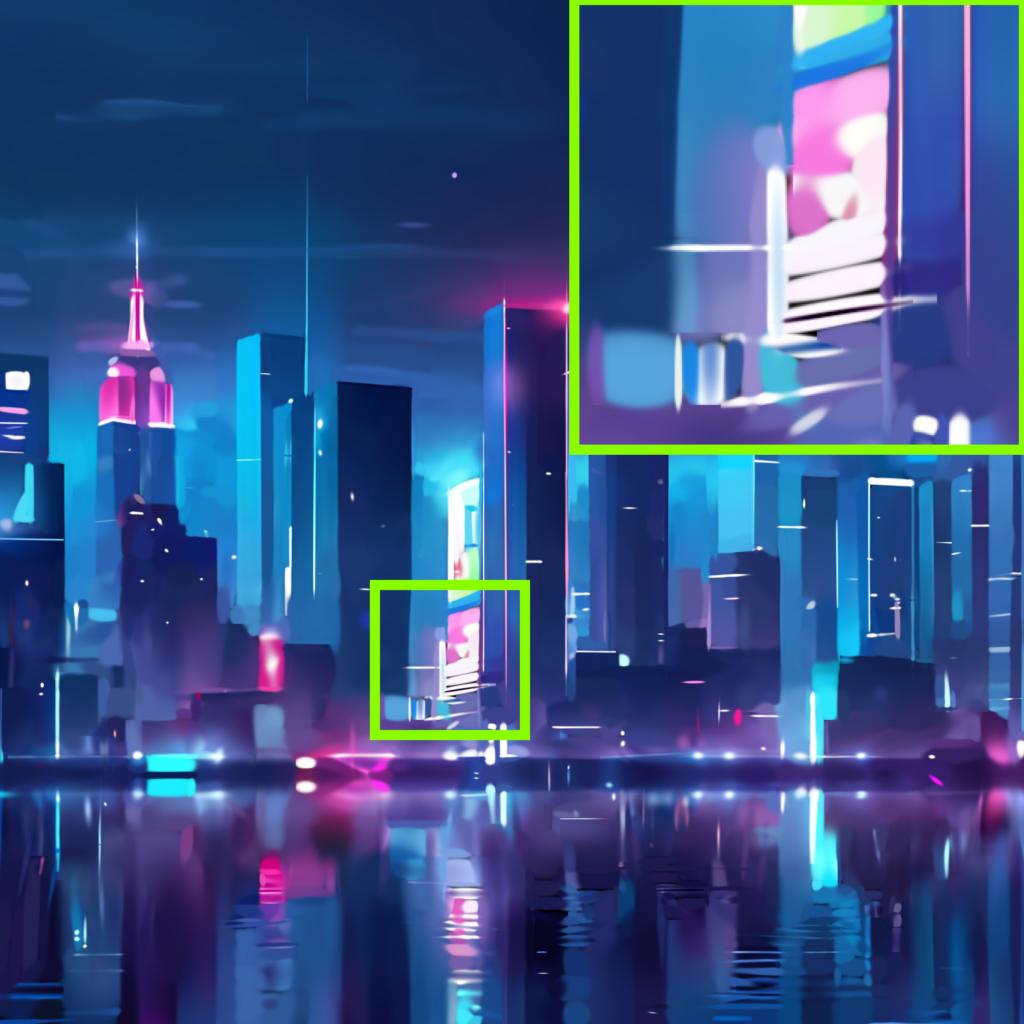}
  }
\subfloat{
    \includegraphics[width=\lodCompWidth,height=\lodCompWidth]{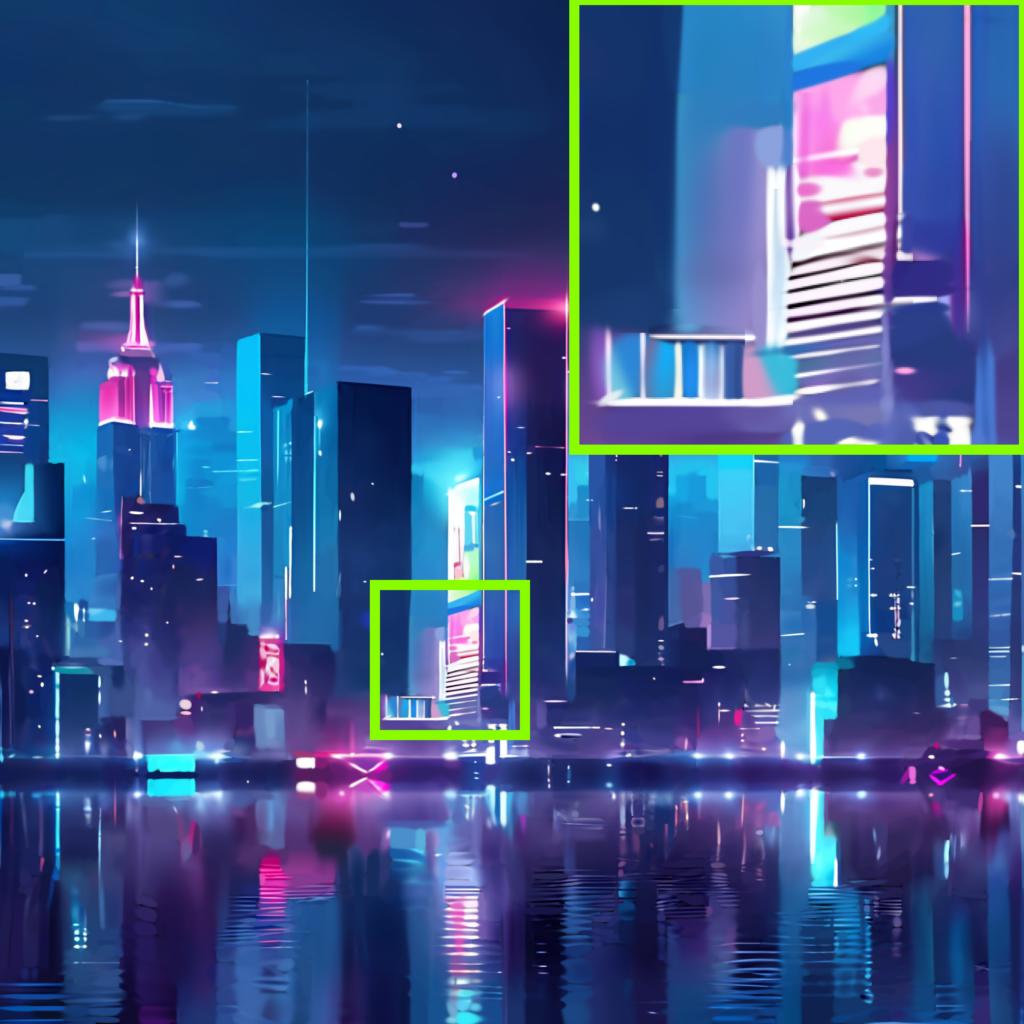}
  }
\subfloat{
    \includegraphics[width=\lodCompWidth,height=\lodCompWidth]{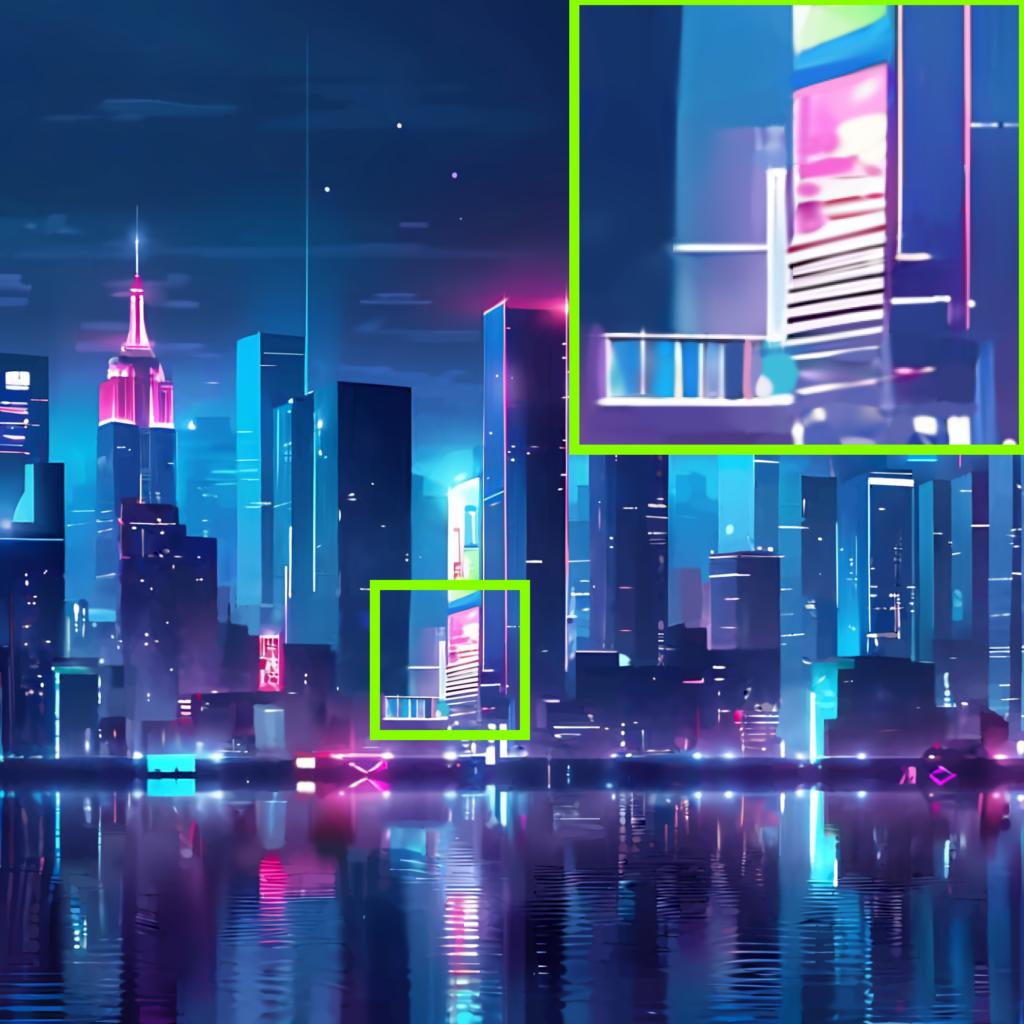}
  }
\subfloat{
    \includegraphics[width=\lodCompWidth,height=\lodCompWidth]{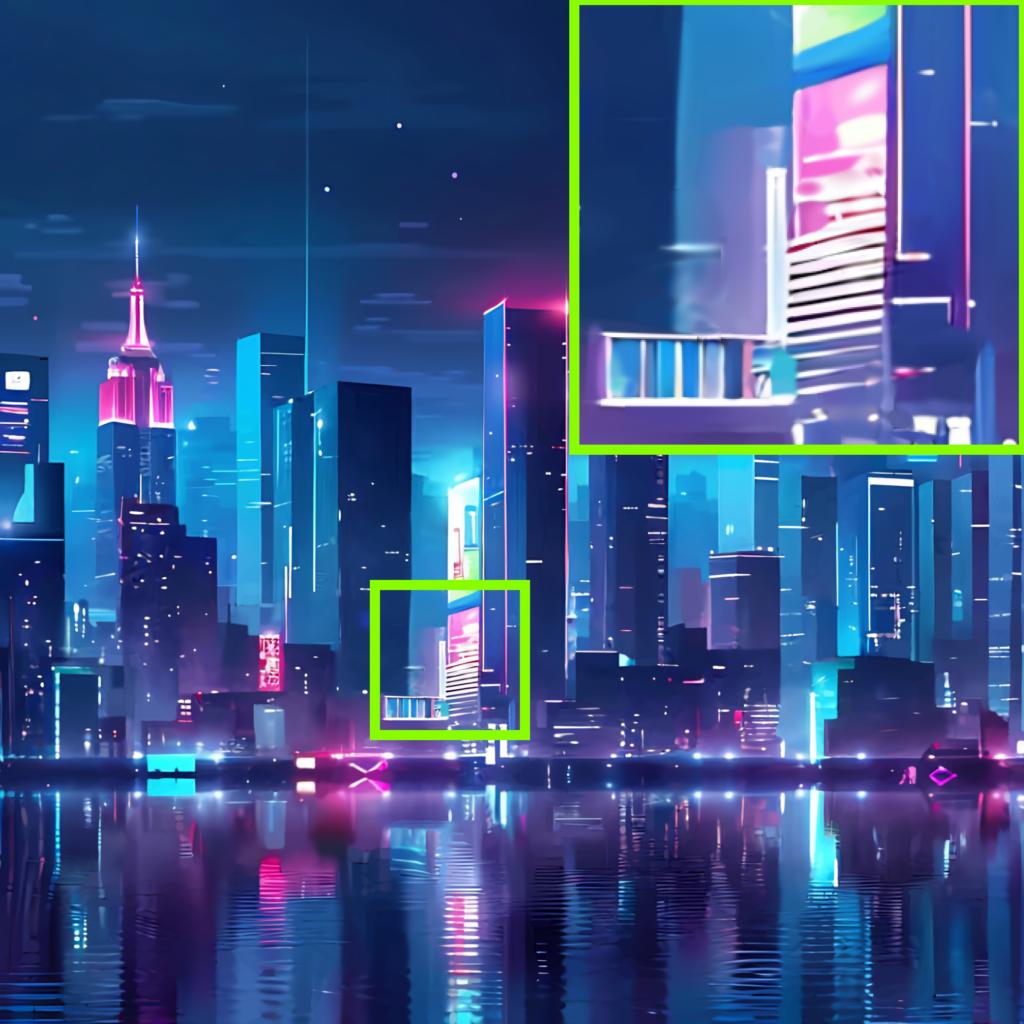}
  }
\subfloat{
    \includegraphics[width=\lodCompWidth,height=\lodCompWidth]{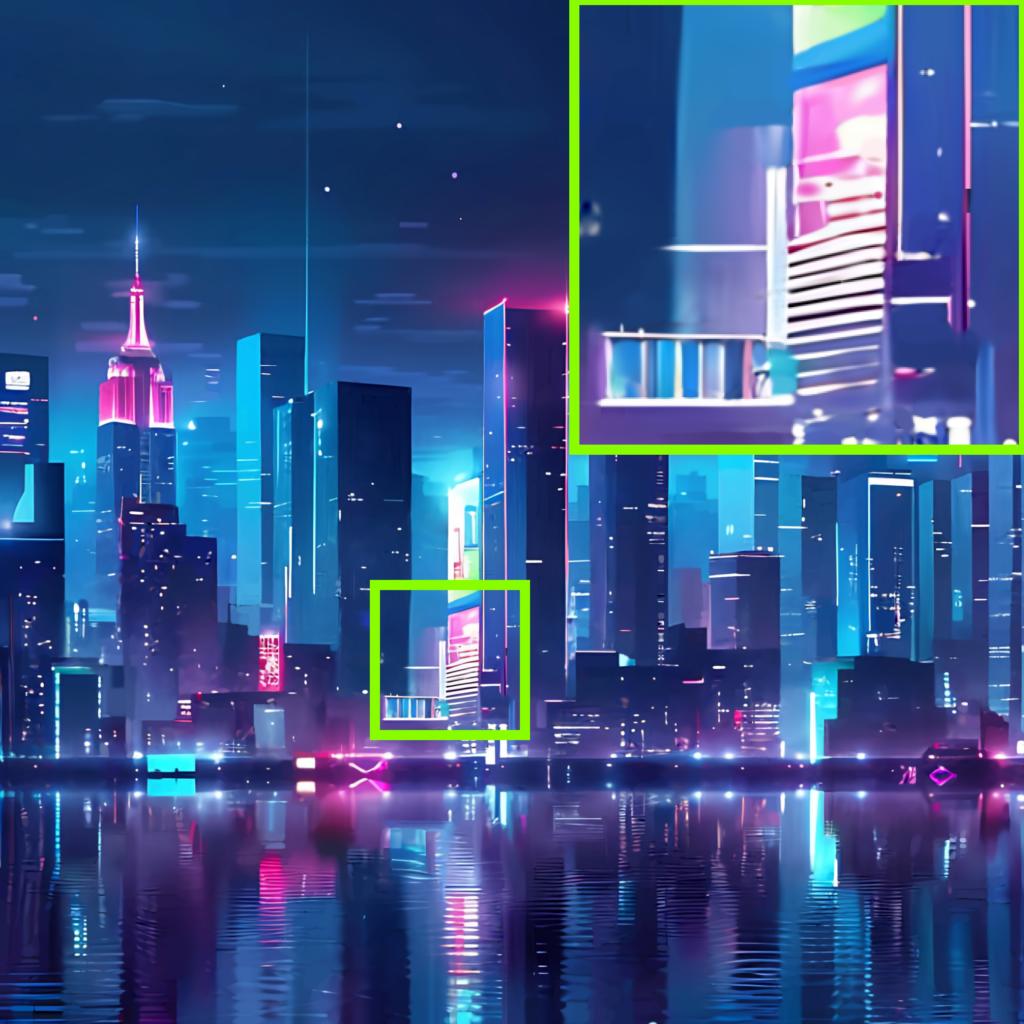}
  }
\subfloat{
    \includegraphics[width=\lodCompWidth,height=\lodCompWidth]{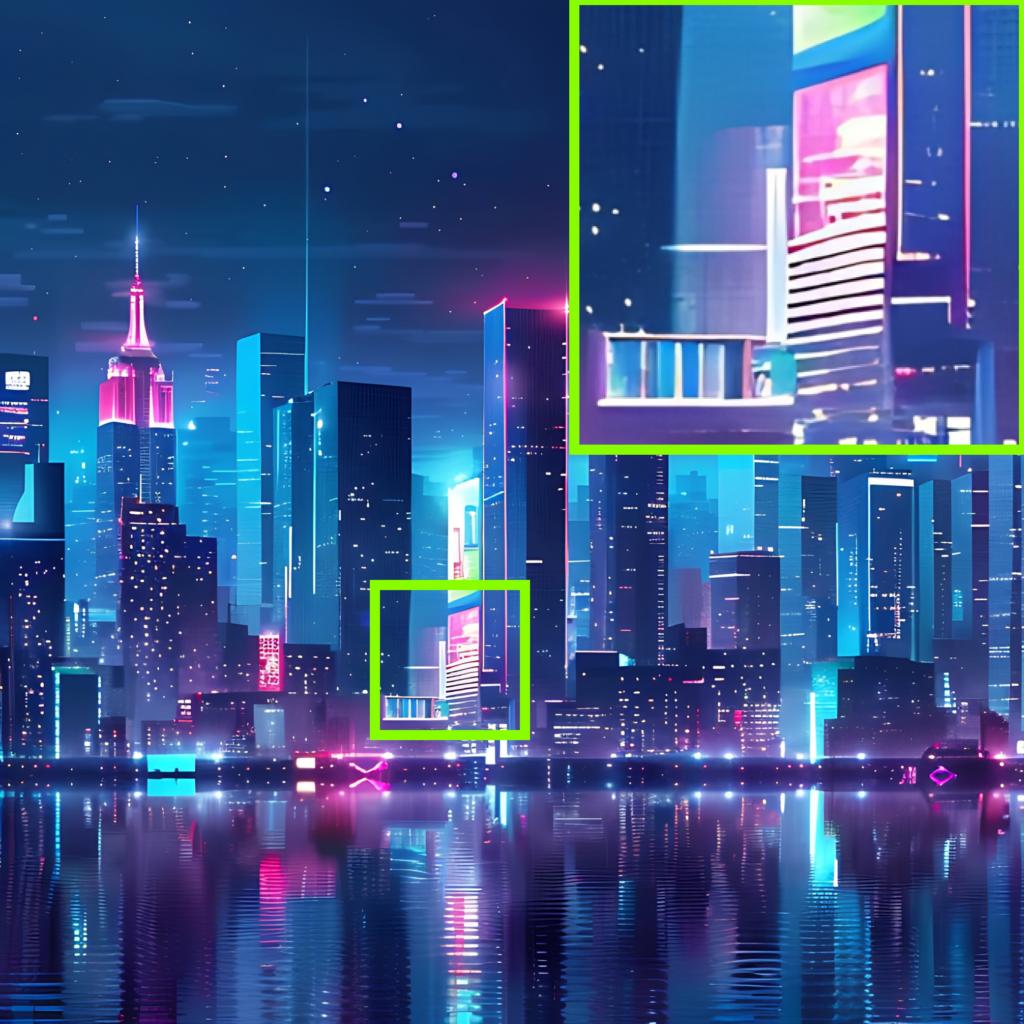}
  }
\vspace{2mm} \\
\subfloat{
    \includegraphics[width=\lodCompWidth,height=\lodCompWidth]{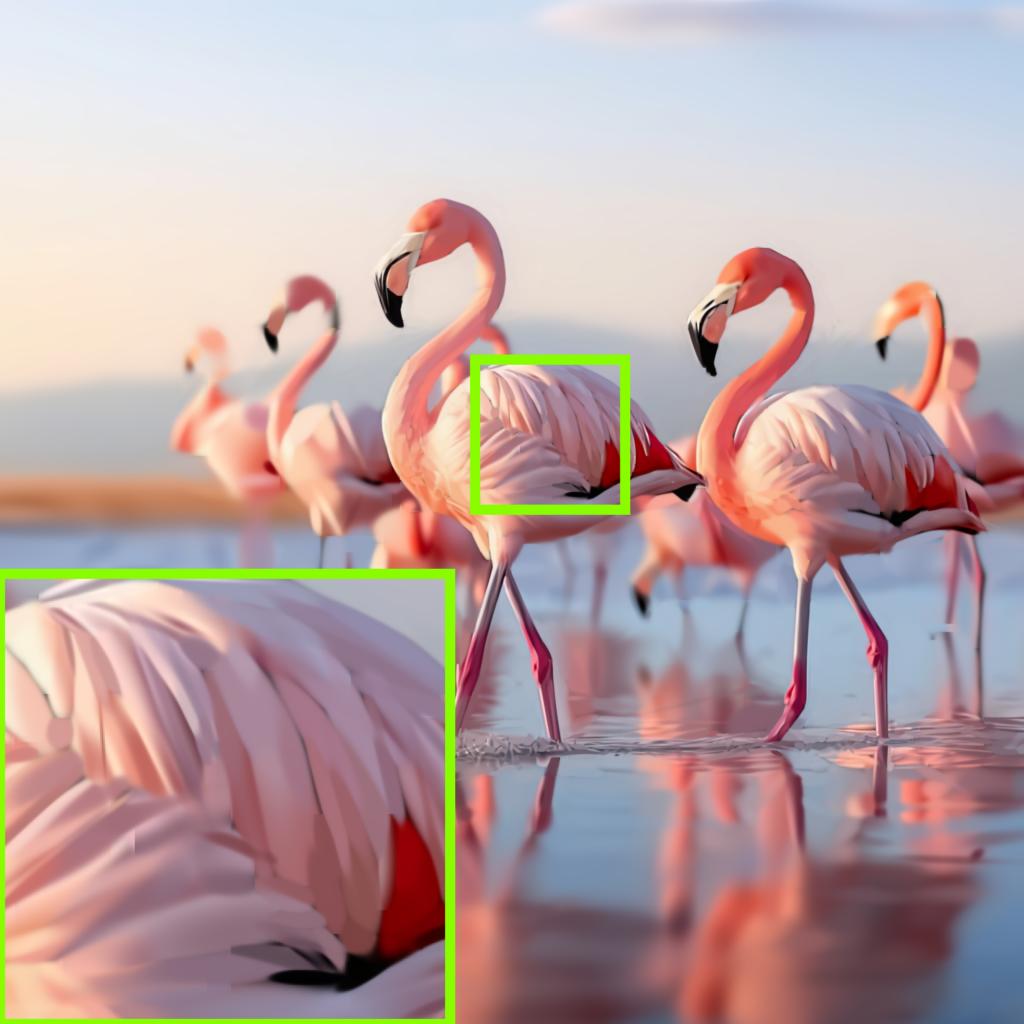}
  }
\subfloat{
    \includegraphics[width=\lodCompWidth,height=\lodCompWidth]{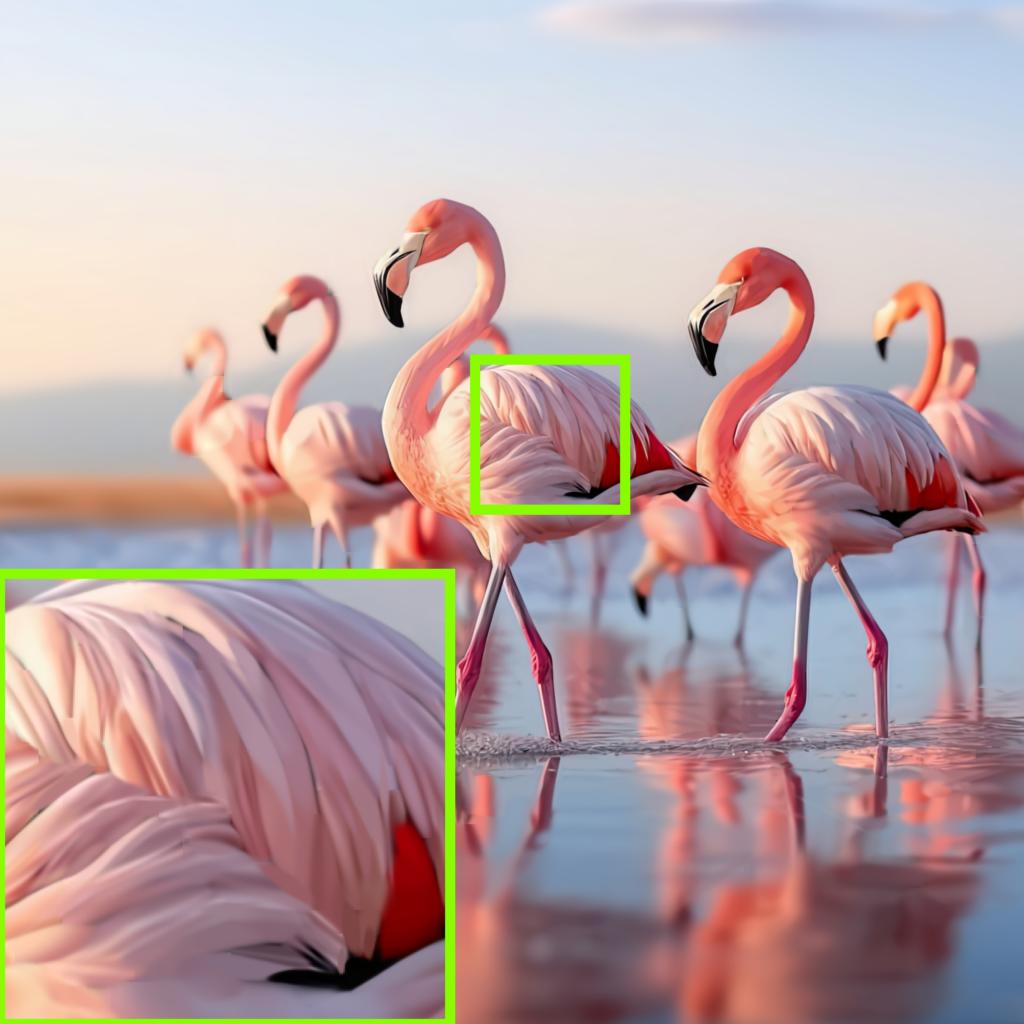}
  }
\subfloat{
    \includegraphics[width=\lodCompWidth,height=\lodCompWidth]{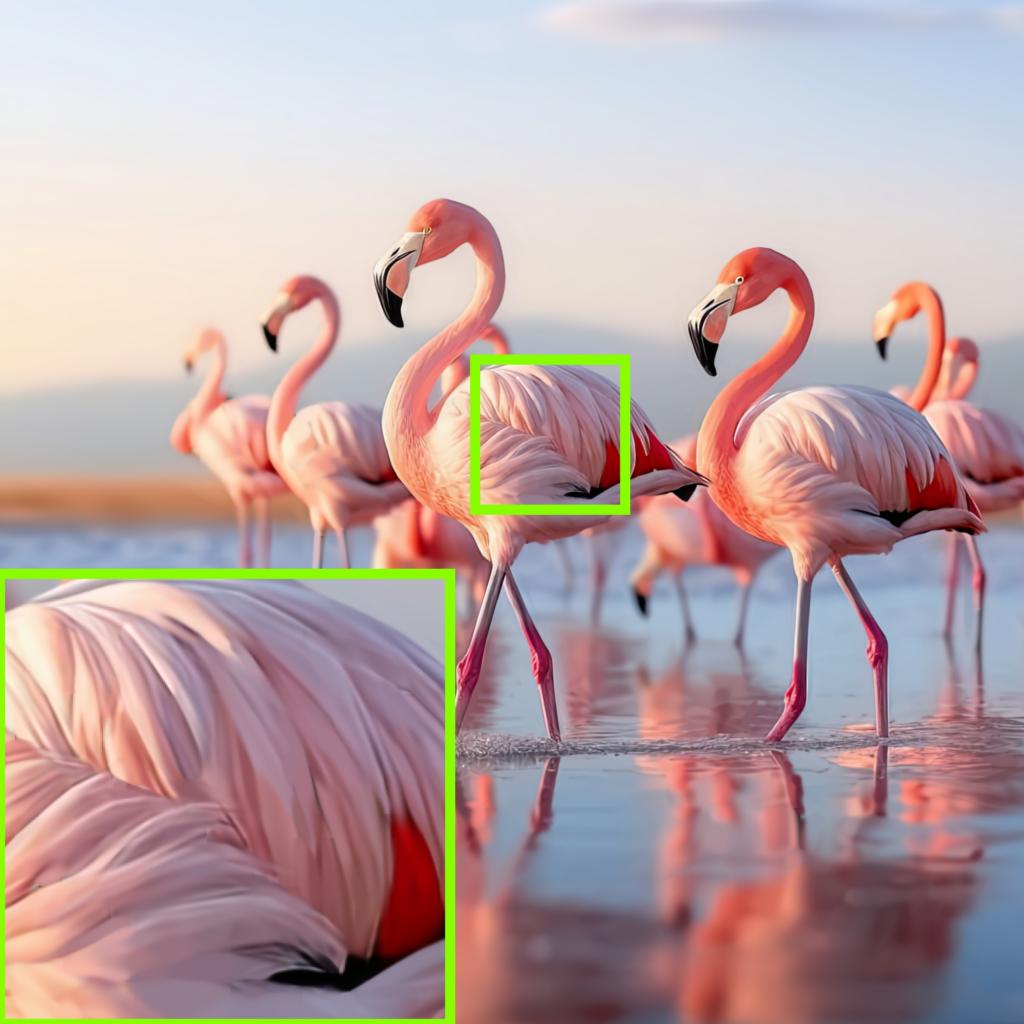}
  }
\subfloat{
    \includegraphics[width=\lodCompWidth,height=\lodCompWidth]{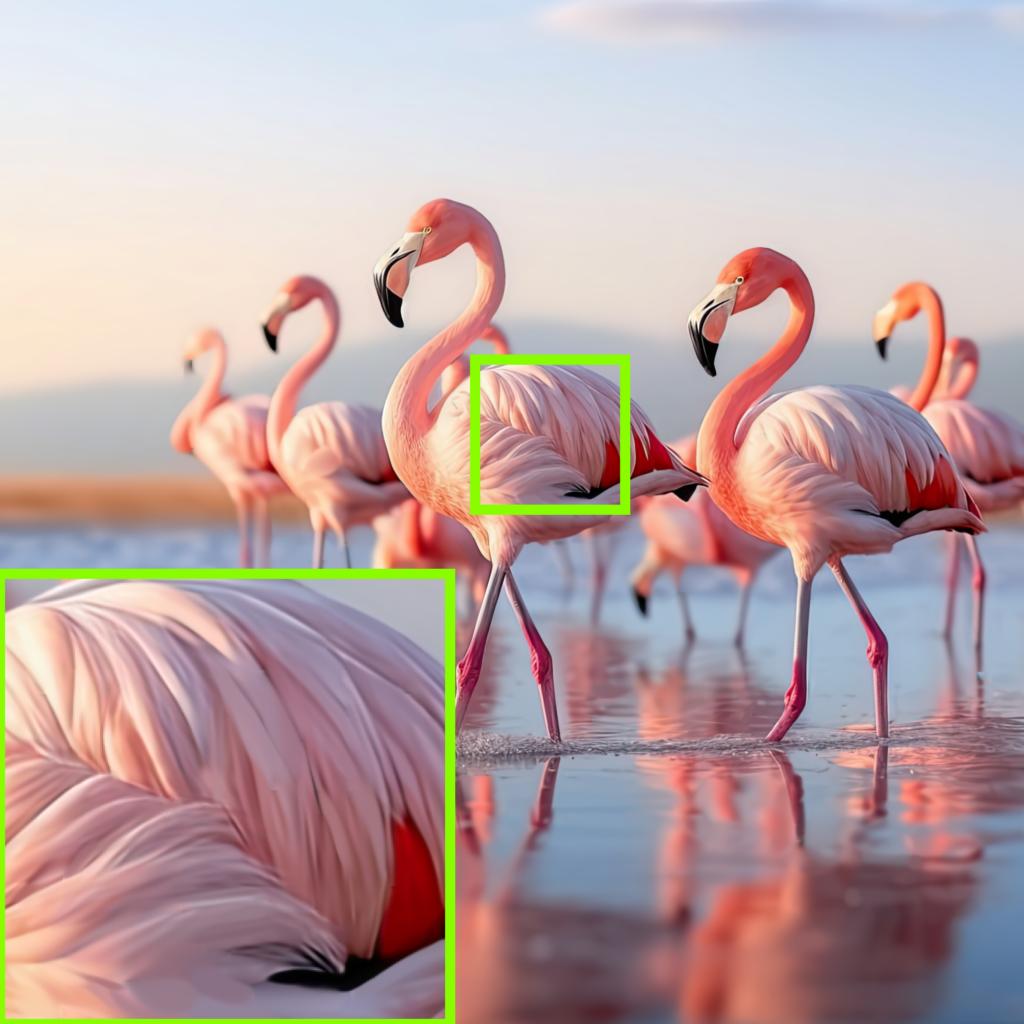}
  }
\subfloat{
    \includegraphics[width=\lodCompWidth,height=\lodCompWidth]{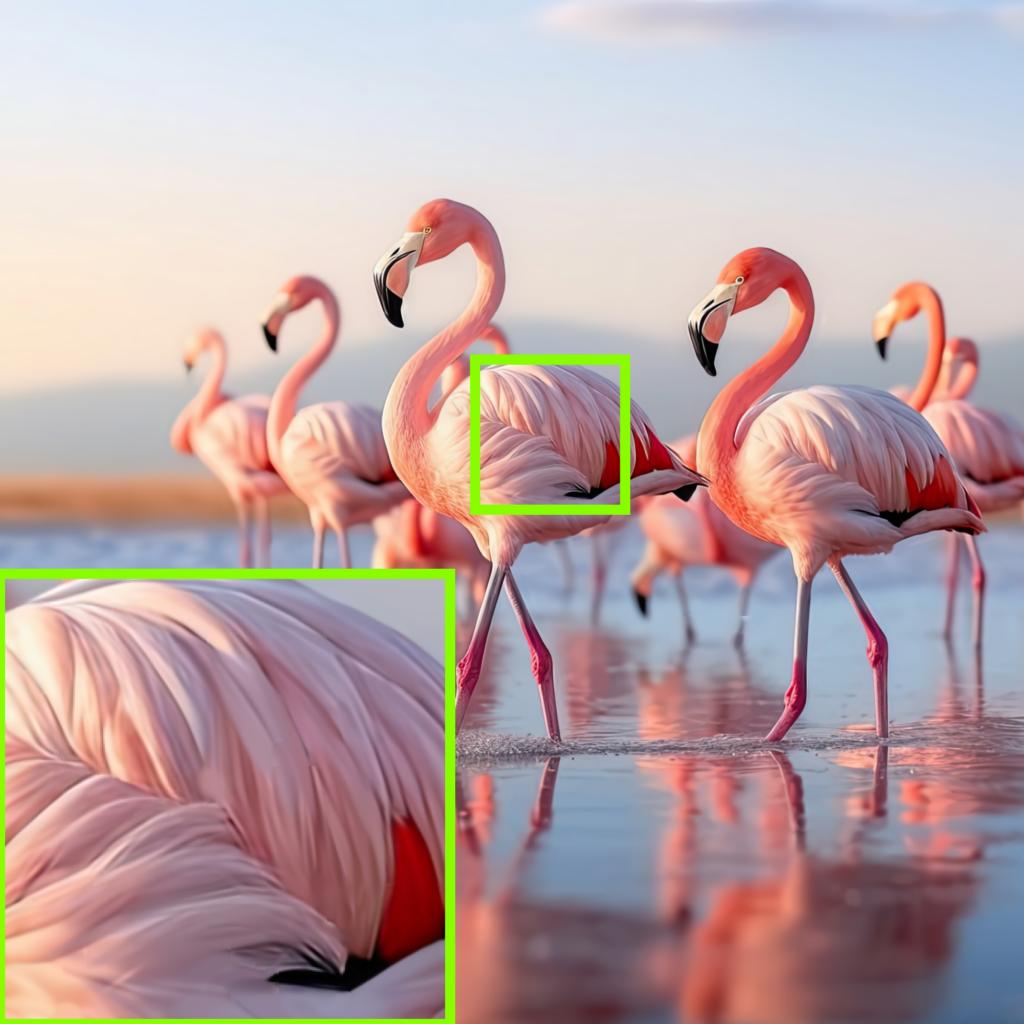}
  }
\subfloat{
    \includegraphics[width=\lodCompWidth,height=\lodCompWidth]{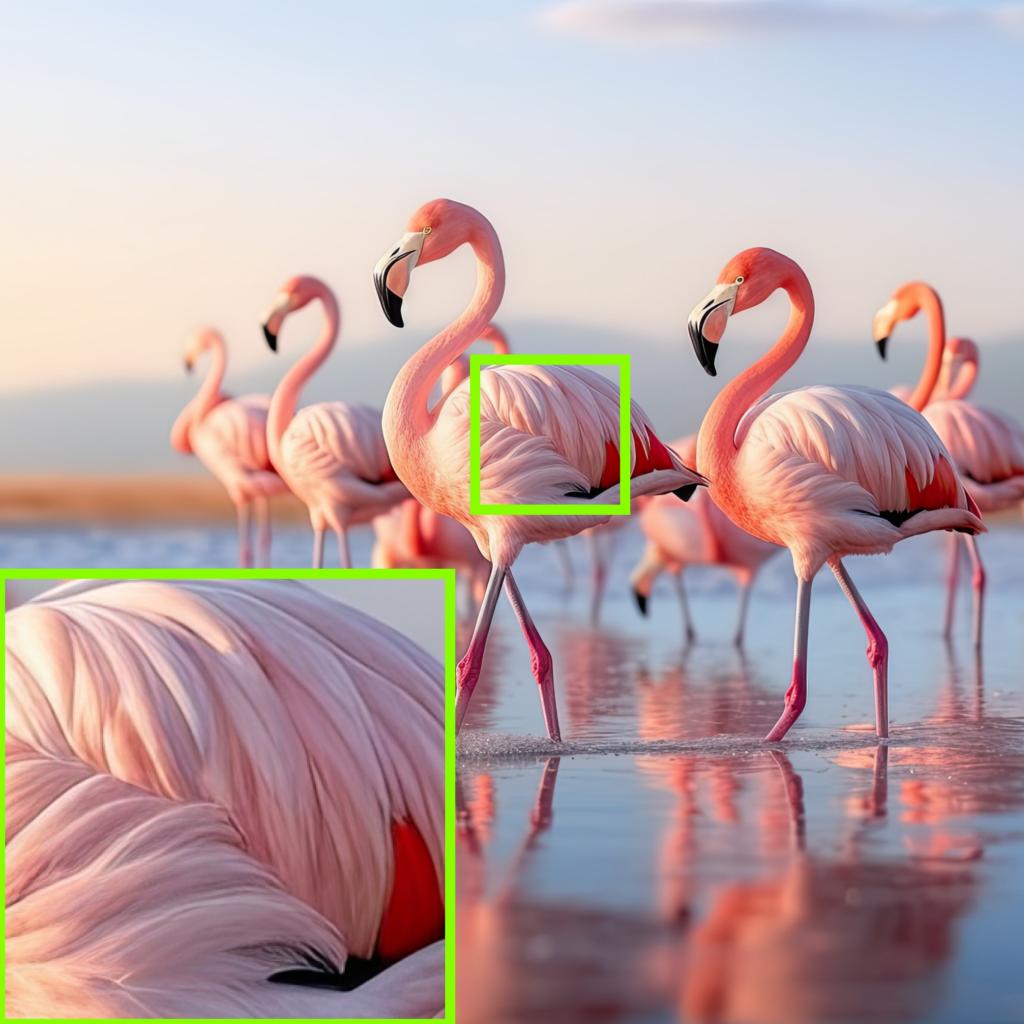}
  }
\vspace{2mm} \\
\subfloat{
    \includegraphics[width=\lodCompWidth,height=\lodCompWidth]{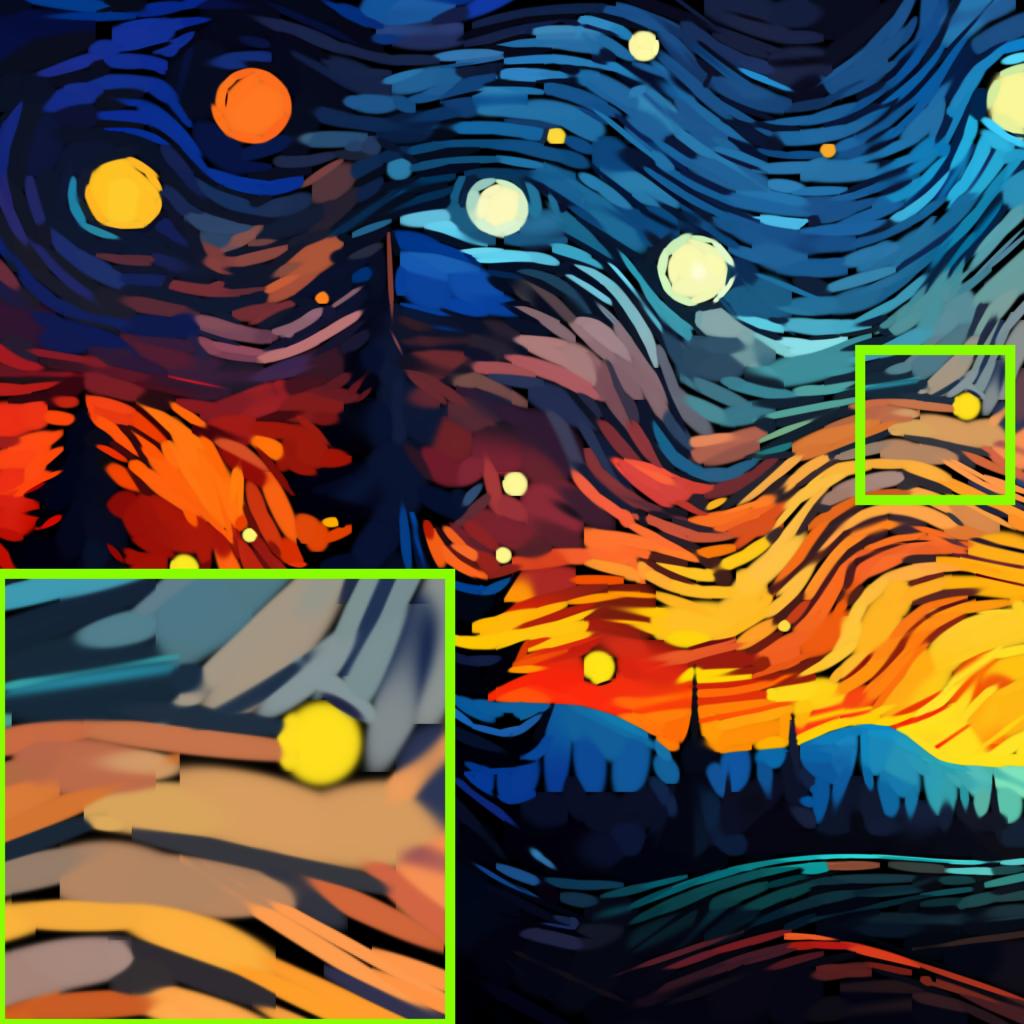}
  }
\subfloat{
    \includegraphics[width=\lodCompWidth,height=\lodCompWidth]{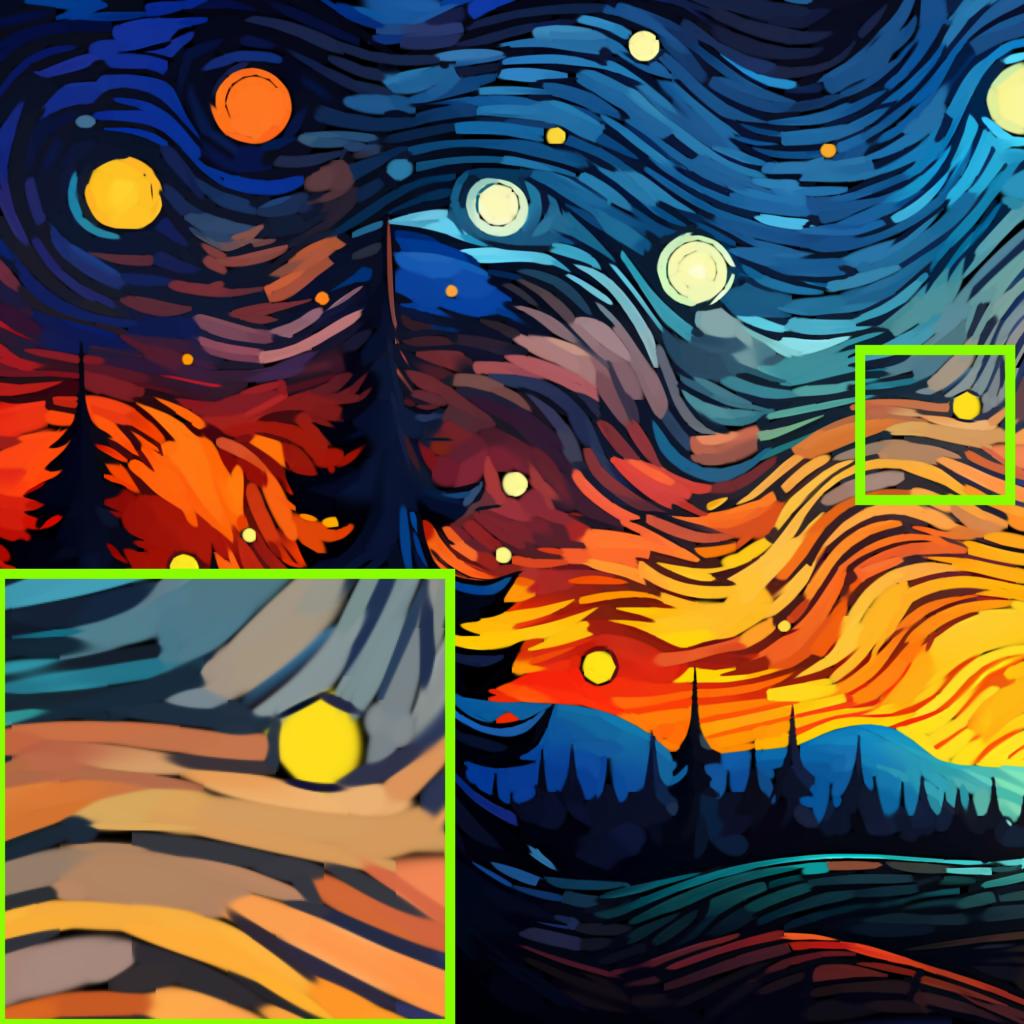}
  }
\subfloat{
    \includegraphics[width=\lodCompWidth,height=\lodCompWidth]{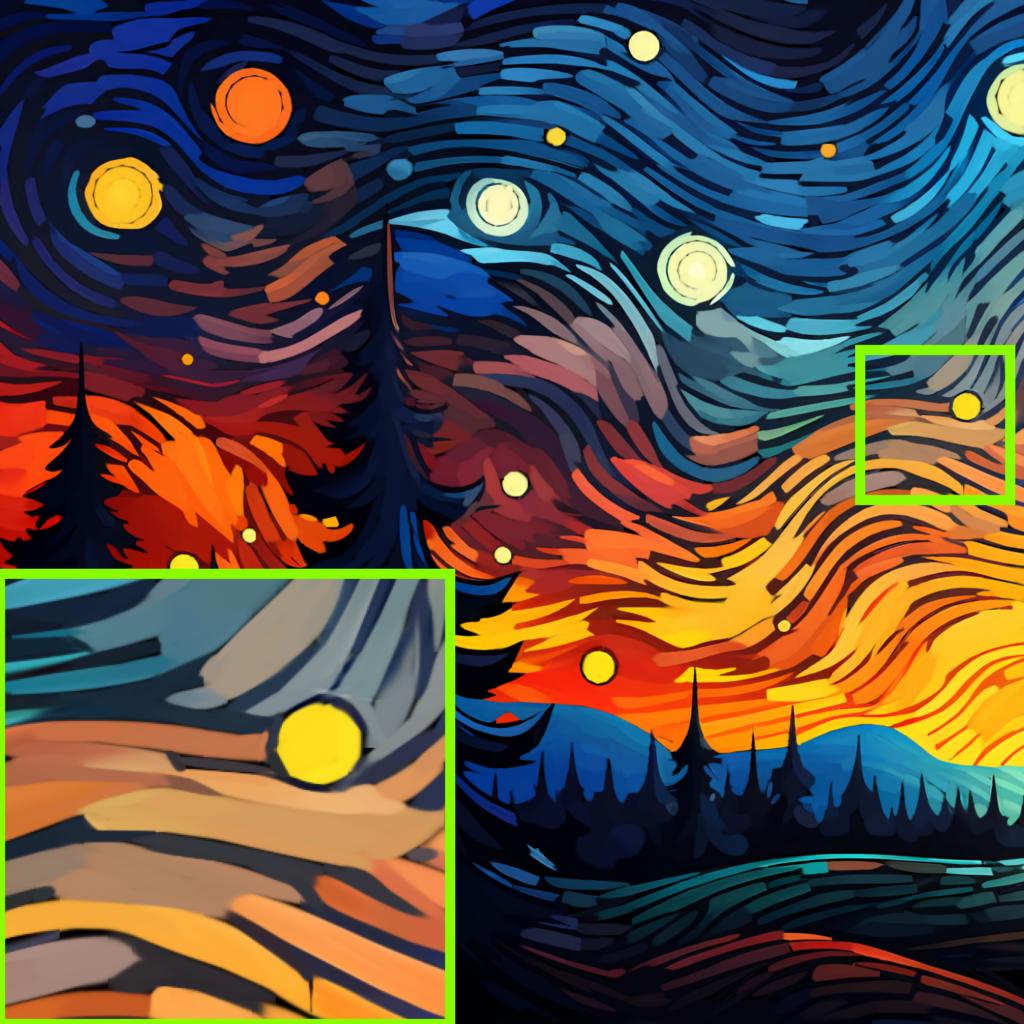}
  }
\subfloat{
    \includegraphics[width=\lodCompWidth,height=\lodCompWidth]{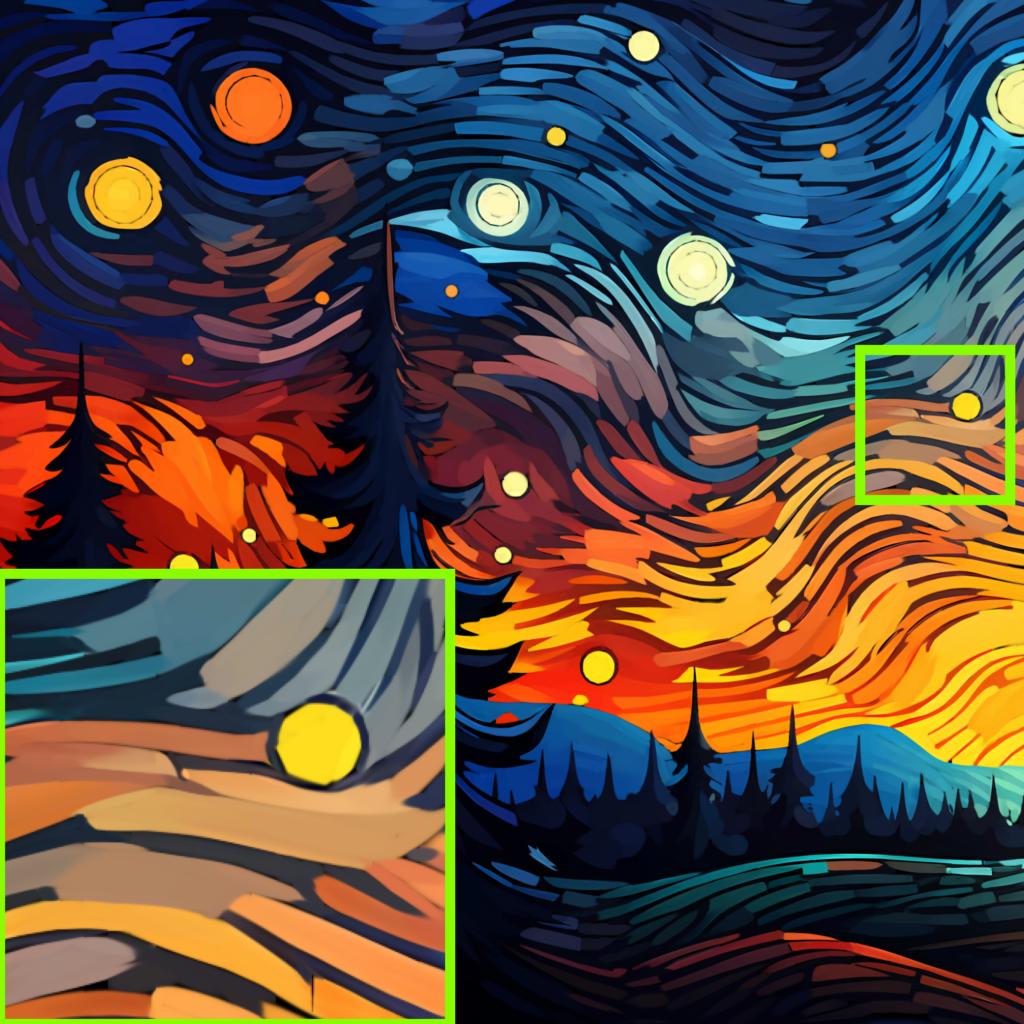}
  }
\subfloat{
    \includegraphics[width=\lodCompWidth,height=\lodCompWidth]{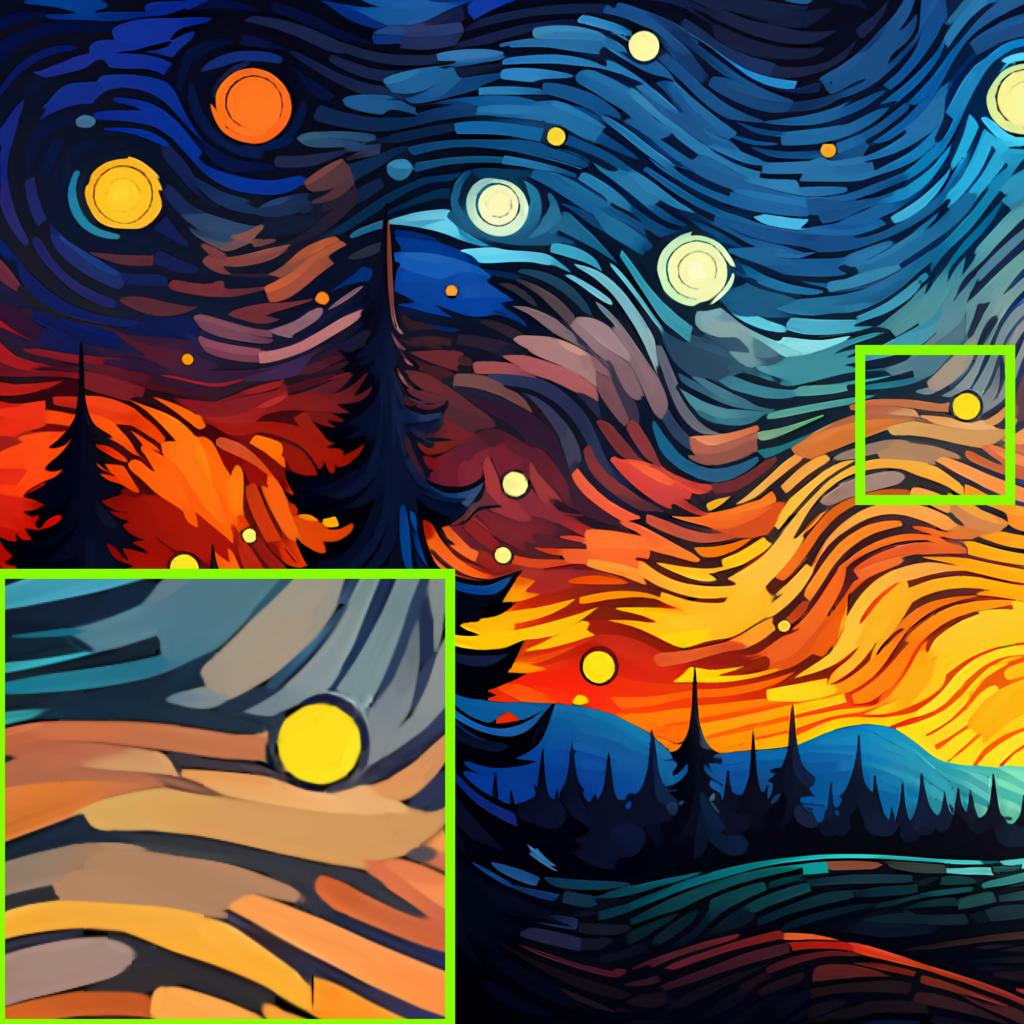}
  }
\subfloat{
    \includegraphics[width=\lodCompWidth,height=\lodCompWidth]{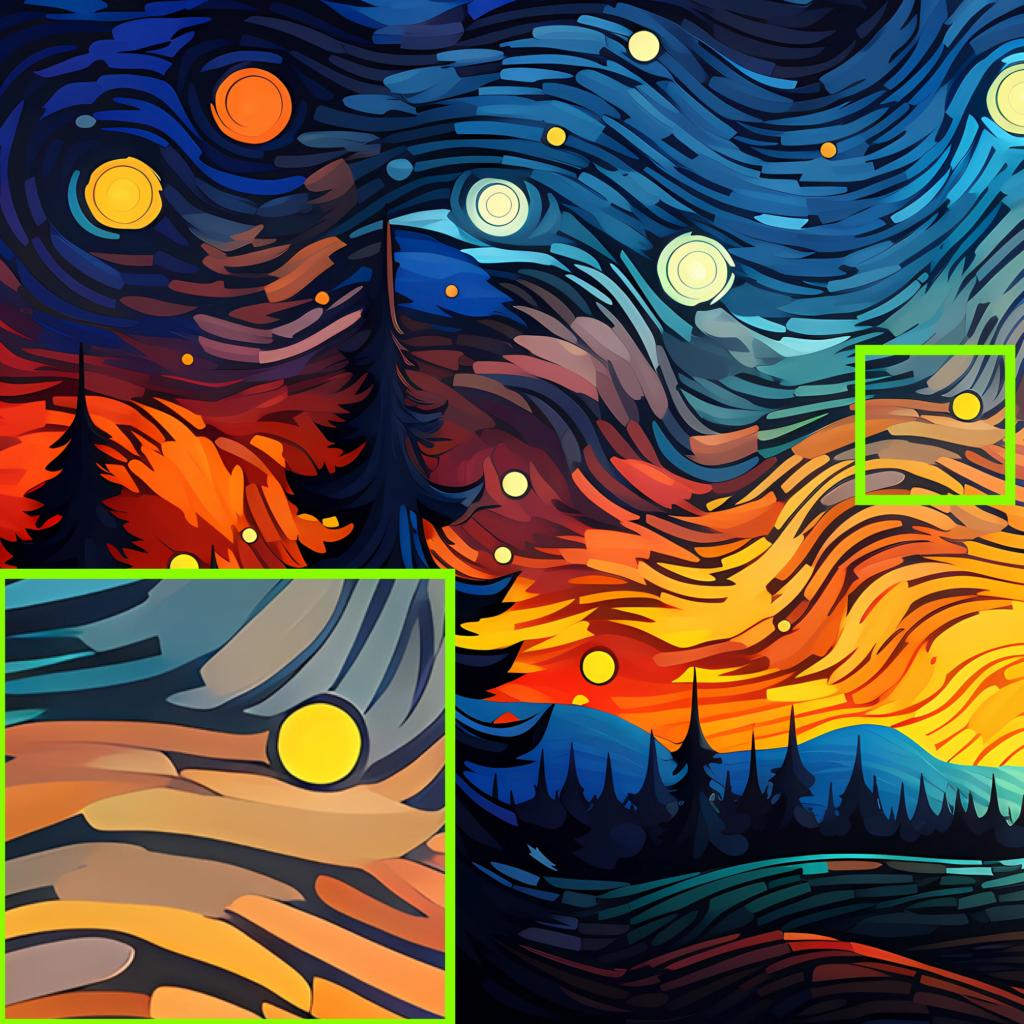}
  }
\vspace{2mm} \\
\setcounter{subfigure}{0}
\subfloat[Ours (0.061 bpp)]{
    \includegraphics[width=\lodCompWidth,height=\lodCompWidth]{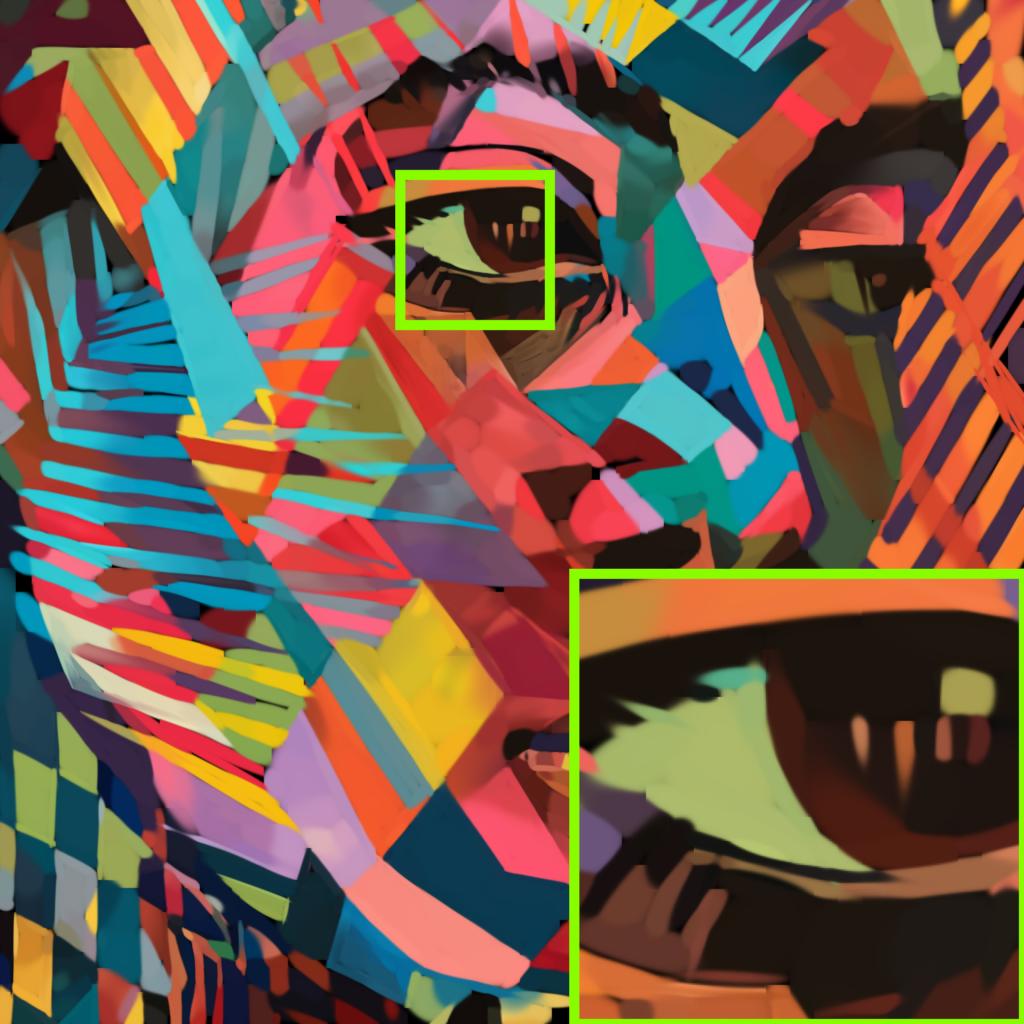}
  }
\subfloat[Ours (0.122 bpp)]{
    \includegraphics[width=\lodCompWidth,height=\lodCompWidth]{images/evaluation-lod/vector-9_2k/2000.jpg}
  }
\subfloat[Ours (0.183 bpp)]{
    \includegraphics[width=\lodCompWidth,height=\lodCompWidth]{images/evaluation-lod/vector-9_2k/2000.jpg}
  }
\subfloat[Ours (0.244 bpp)]{
    \includegraphics[width=\lodCompWidth,height=\lodCompWidth]{images/evaluation-lod/vector-9_2k/2000.jpg}
  }
\subfloat[Ours (0.305 bpp)]{
    \includegraphics[width=\lodCompWidth,height=\lodCompWidth]{images/evaluation-lod/vector-9_2k/2000.jpg}
  }
\subfloat[Reference]{
    \includegraphics[width=\lodCompWidth,height=\lodCompWidth]{images/evaluation-lod/vector-9_2k/2000.jpg}
  }
\Caption{\revise{\methodName's rate-distortion trade-off (\Cref{sec:evaluation-lod}).}}{\revise{Through error-guided progressive optimization (\Cref{sec:method-optimization}), \methodName naturally constructs a smooth level-of-detail hierarchy in a single optimization run without additional overhead, enabling flexible quality adaptation to device capabilities.}
}
\label{fig:evaluation-lod}
\end{figure*}


\newcommand{\TextureCompRes}{0.171\linewidth}
\begin{figure*}[t]
\centering
\subfloat{
    \includegraphics[width=\TextureCompRes]{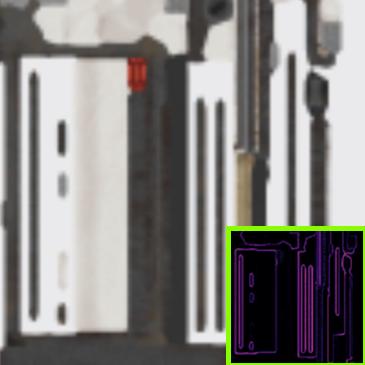}
  } \hspace{-0.18cm}
\subfloat{
    \includegraphics[width=\TextureCompRes]{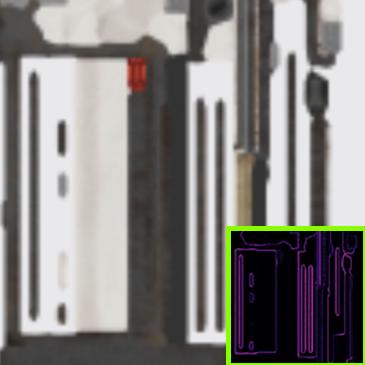}
  } \hspace{-0.18cm}
\subfloat{
    \includegraphics[width=\TextureCompRes]{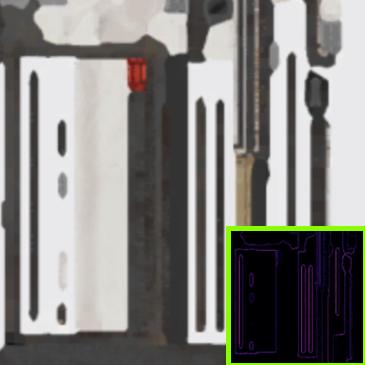}
  } \hspace{-0.18cm}
\subfloat{
    \includegraphics[width=\TextureCompRes]{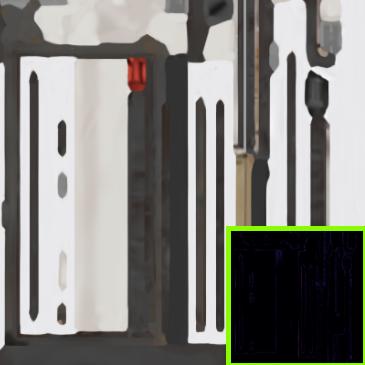}
  } \hspace{-0.18cm}
\subfloat{
    \includegraphics[width=\TextureCompRes]{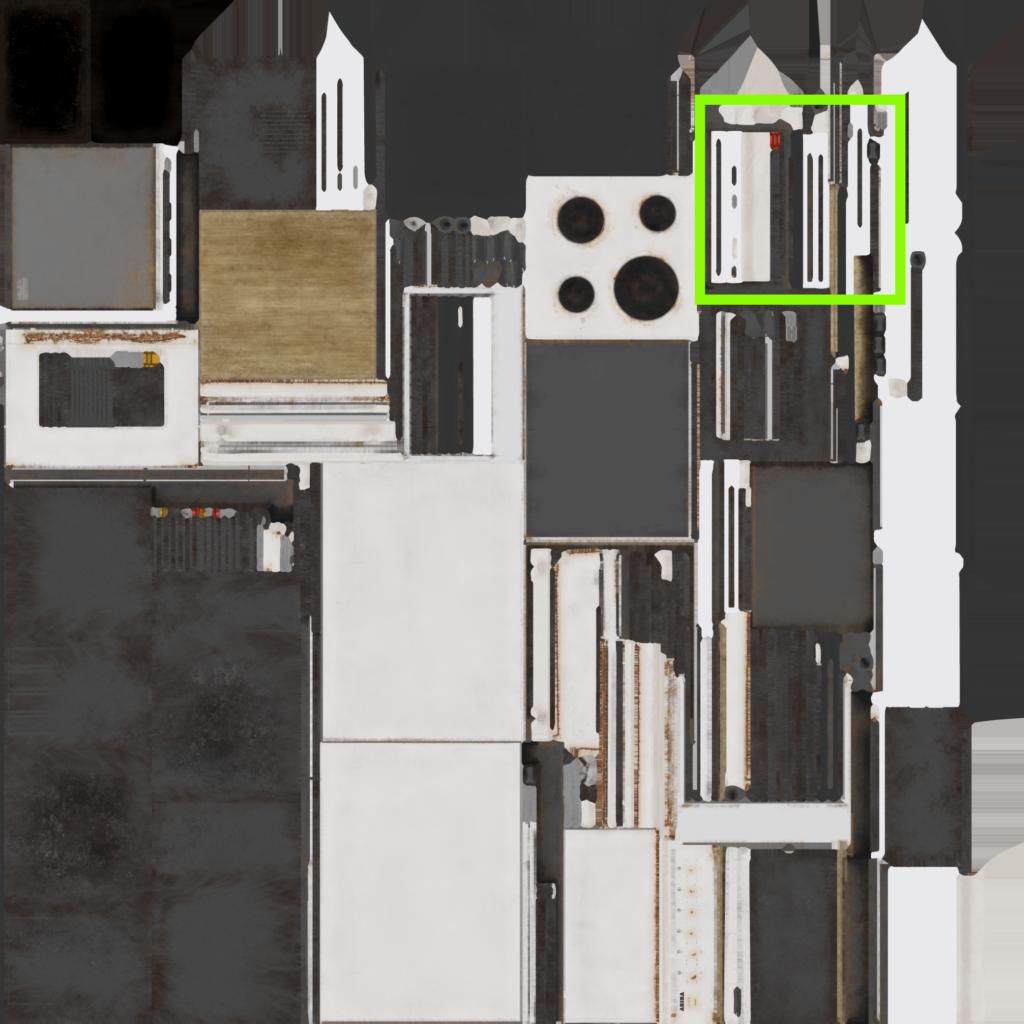}
  }
\vspace{0.2mm} \\
\subfloat{
    \includegraphics[width=\TextureCompRes]{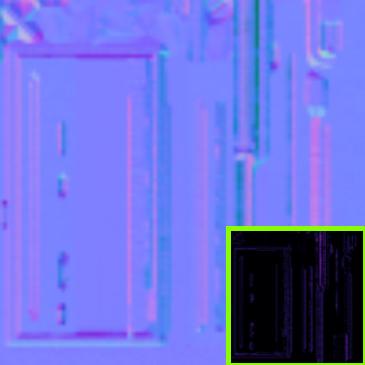}
  } \hspace{-0.18cm}
\subfloat{
    \includegraphics[width=\TextureCompRes]{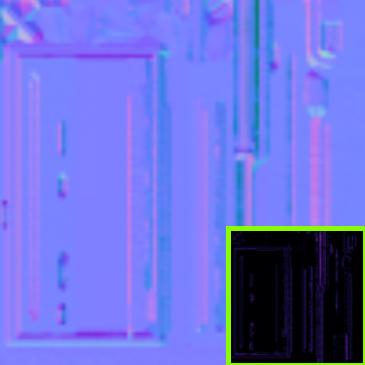}
  } \hspace{-0.18cm}
\subfloat{
    \includegraphics[width=\TextureCompRes]{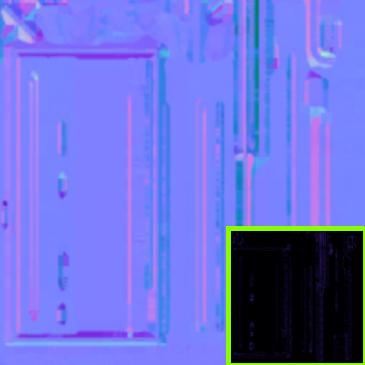}
  } \hspace{-0.18cm}
\subfloat{
    \includegraphics[width=\TextureCompRes]{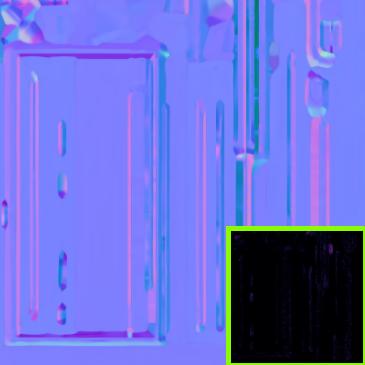}
  } \hspace{-0.18cm}
\subfloat{
    \includegraphics[width=\TextureCompRes]{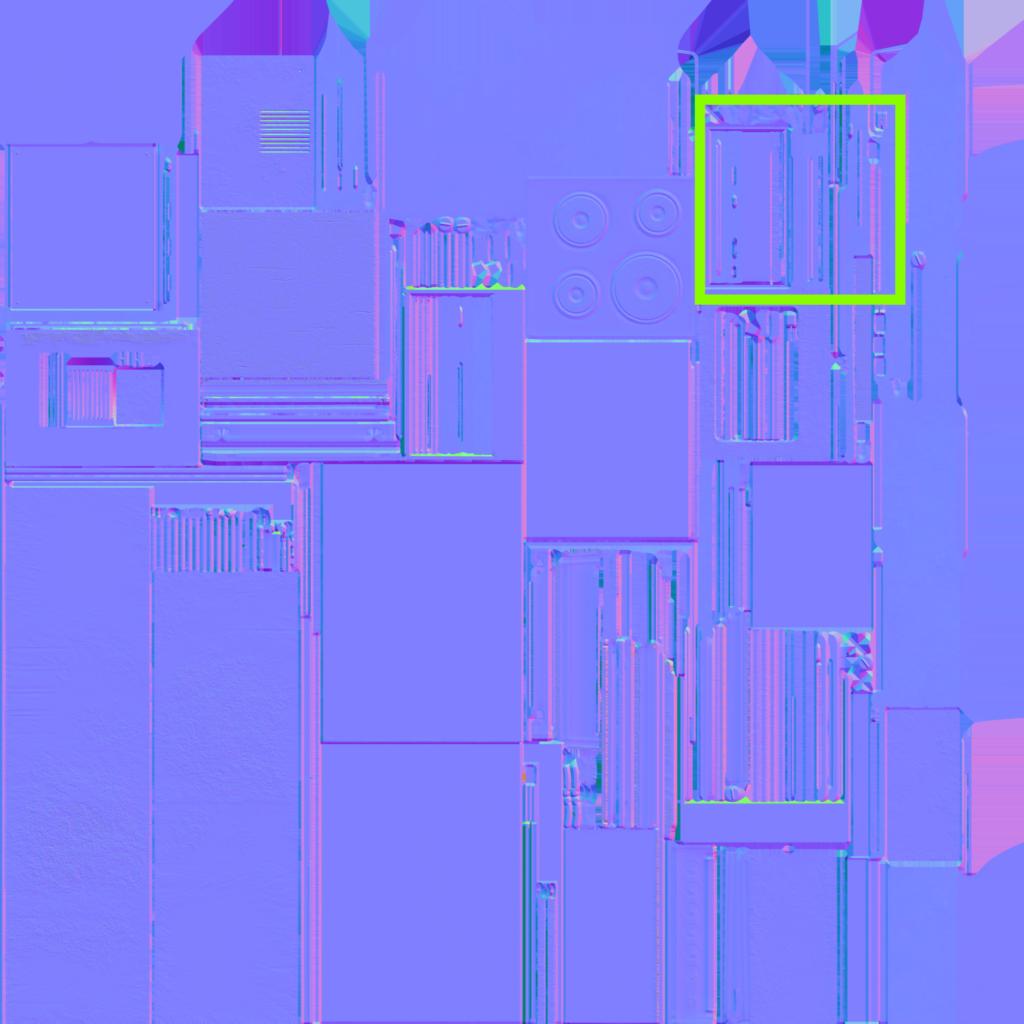}
  }
\vspace{0.2mm} \\
\setcounter{subfigure}{0}
\subfloat[BC1 (0.083 bppc)]{
    \includegraphics[width=\TextureCompRes]{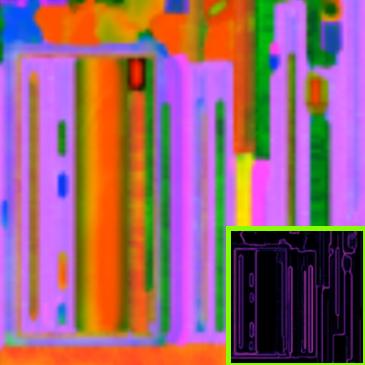}
  } \hspace{-0.18cm}
\subfloat[BC7 (0.167 bppc)]{
    \includegraphics[width=\TextureCompRes]{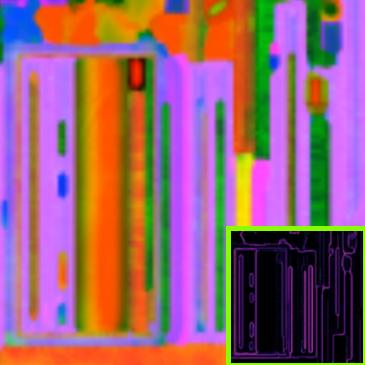}
  } \hspace{-0.18cm}
\subfloat[ASTC (0.074 bppc)]{
    \includegraphics[width=\TextureCompRes]{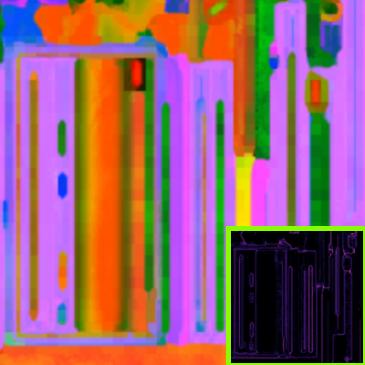}
  } \hspace{-0.18cm}
\subfloat[Ours (0.089 bppc)]{
    \includegraphics[width=\TextureCompRes]{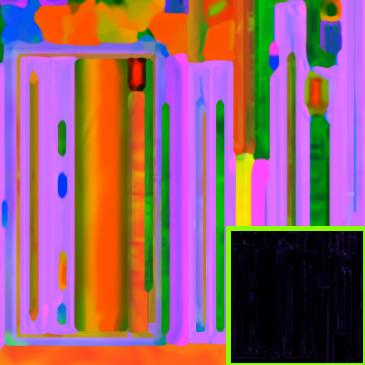}
  } \hspace{-0.18cm}
\subfloat[Reference]{
    \includegraphics[width=\TextureCompRes]{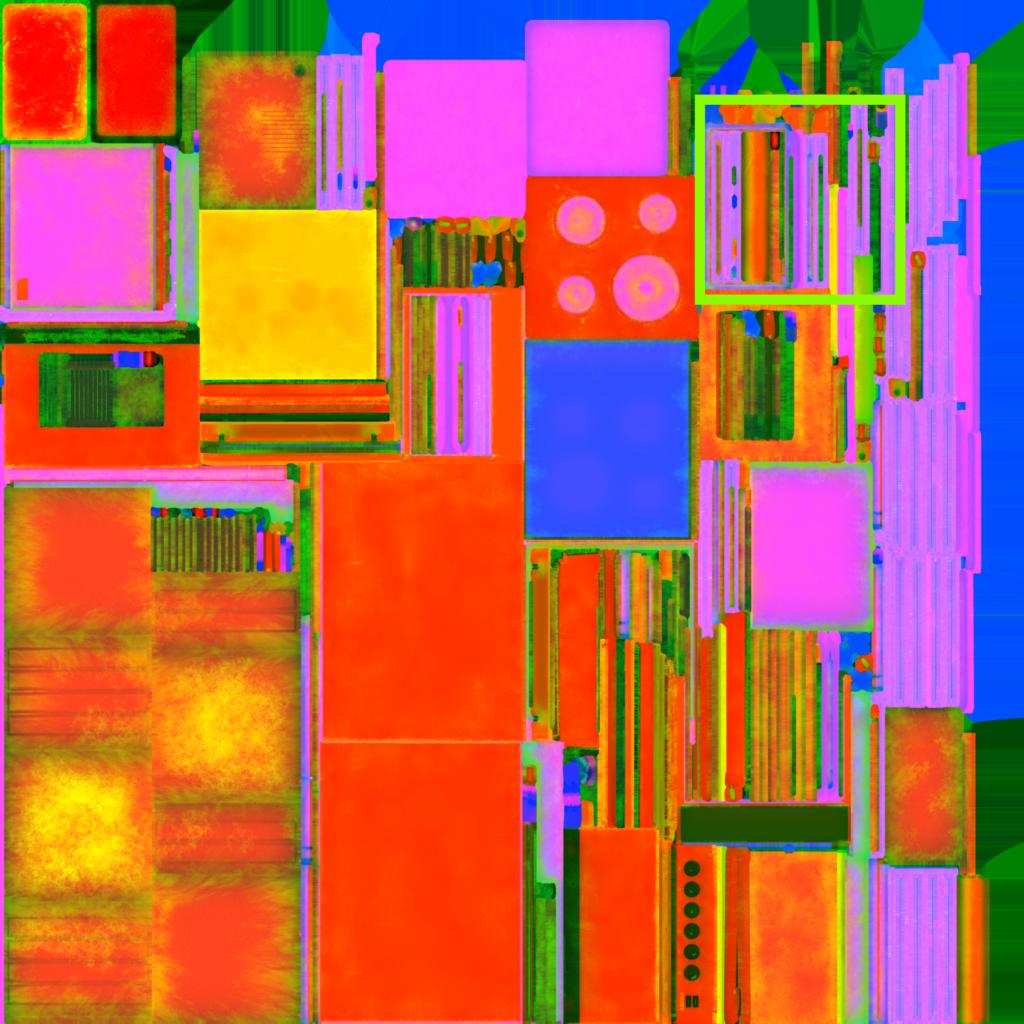}
  }
\Caption{\revise{Qualitative comparison against industry-standard GPU texture compression algorithms (\Cref{sec:evaluation-texture}).}}
{}
\label{fig:evaluation-texture}
\end{figure*}


As shown in \Cref{fig:evaluation-rate-distortion-image}, \methodName outperforms all neural baselines across the entire bitrate range we evaluate.
\revise{When the bitrate falls below 0.244, \methodName even surpasses JPEG by a significant margin. Notably, both JPEG and GI leverage entropy coding, which breaks data locality, to improve performance, whereas \methodName does not rely on such mechanisms.}
\Cref{fig:evaluation-image,fig:evaluation-image-supp} show several zoomed-in samples with error map overlays for visual comparison. Under this ultra-low bitrate regime, all baselines \revise{exhibit noticeable distortions and artifacts. For instance,} decreasing the resolution of feature grids (ReLU-F, I-NGP) leads to block artifacts due to feature vectors being interpolated at sparser locations. \revise{Implicit neural image representations (SIREN, FFN, WIRE) are prone to artifacts such as ringing, blurring, and ghosting, which become more pronounced} after reducing the base network parameters.

\revise{
We further ran \methodName on the professional validation split of the CLIC2020 dataset \cite{toderici2020workshop} to assess its compression performance on natural images. As illustrated in \Cref{fig:evaluation-rate-distortion-image-clic}, \methodName achieves lower scores compared to the results for stylized images in \Cref{sec:evaluation-image}. We attribute this performance drop to the more constrained Gaussian budgets on lower-resolution images at constant bitrates, as well as the prevalence of pixel-level camera sensor noise in natural images. As an explicit representation, \methodName requires a sufficient number of Gaussian primitives to accurately capture spatially distinct image features. \Cref{fig:evaluation-image-clic-supp-1,fig:evaluation-image-clic-supp-2} shows several samples with zoomed-in overlays for visual comparison.
}


\subsection{System Performance}
\label{sec:evaluation-system}

\revise{Our efficient implementation enables fast} training and rendering with \methodName. For instance, optimizing 10K Gaussians for 1K steps at 2K$\times$2K resolution takes an average of 18.74 seconds, \revise{while rendering (a single forward pass without gradient tracking) takes only 0.0037 seconds. This efficiency scales sub-linearly with the number of Gaussians: optimizing 50K Gaussians for 1K steps at 2K$\times$2K resolution takes 26.32 seconds, and rendering takes 0.0045 seconds.} All measurements were conducted on an Nvidia A6000 GPU.

As shown in \Cref{fig:evaluation-time}, \methodName's rendering speed is second only to I-NGP. Thanks to \revise{designs such as inverse scale training and top-$K$ normalization, \methodName converges rapidly, typically within 3–4K steps (indicated by less than 0.1 PSNR and 0.001 SSIM improvements). It reaches 95\% of its final performance in fewer than 400 steps and 99\% within 2K steps. This fast convergence significantly reduces the overall training time for \methodName.} See the supplementary video for a real-time training demonstration at 8K$\times$8K resolution.


\subsection{\revise{Texture Compression Performance}}
\label{sec:evaluation-texture}

\paragraph{Baselines}
We compare with 3 industry-standard texture compression algorithms, BC1, BC7, and ASTC, using their implementations from NVIDIA Texture Tools\footnote{https://developer.nvidia.com/texture-tools-exporter}. \revise{Since they only support bitrates down to 4.0, 8.0, and 0.89 bpp, respectively, for RGB(A) textures, we extracted the compressed textures from their higher mipmap levels to match our target bitrate range for iso-bitrate comparison. These lower-resolution mipmaps were upsampled to $2K\times2K$ resolution via bilinear interpolation before evaluation. We did not choose the alternative approach of running these baselines on downsampled versions of the target textures, as this would result in inconsistent reference across methods and make the comparison unfair.
}

As shown in \Cref{fig:evaluation-rate-distortion-texture}, \methodName consistently outperforms BC1 and BC7 while \revise{achieving comparable performance to ASTC. \methodName reaches quality metrics of $32.20\pm4.06$ (PSNR) and $0.869\pm0.107$ (SSIM) at an extreme bitrate of 0.059 bppc.} \Cref{fig:evaluation-texture,fig:evaluation-texture-supp-1,fig:evaluation-texture-supp-2,fig:evaluation-texture-supp-3} show several zoomed-in texture stacks with error map overlays.
\section{Applications}
\label{sec:application}


\subsection{Semantics-Aware Compression for Machine Vision}
\label{sec:application-machine-vision}

The exponential growth in the size of state-of-the-art machine vision models has been reshaping the AI deployment paradigm, with cloud provisioning increasingly supplanting local serving. In this context, the efficient storage and transfer of visual content, while preserving the information essential to the underlying applications, are critical to both the end users and cloud service providers \cite{hu2021learning}.

Thanks to the explicit nature of \methodName, we can readily factor in the distribution of such information through visual saliency analysis \cite{itti1998model} and accordingly distribute Gaussian primitives over the image domain to better preserve the important semantic content therein. Specifically, given a target image, we first take an off-the-shelf saliency predictor \cite{jia2020eml} to extract its saliency map $\smap \in \euclidOnePos^{H \times W}$, then perform saliency-guided Gaussian position initialization with sampling probability:
\begin{align}
    \probInit(\pixelLocation) = \frac{(1-\paramInit) \cdot \smap(\pixelLocation)}{\sum_{h=1}^{H}\sum_{w=1}^{W} \smap(\pixelLocation_{h,w})} + \frac{\paramInit}{H \cdot W} \quad \paramInit \in [0, 1]
\label{eq:saliency-init-sampling}
\end{align}
$\paramInit=0.1$ balances saliency guidance and uniform coverage.

\begin{figure}[t]
\centering
\includegraphics[width=0.99\linewidth]{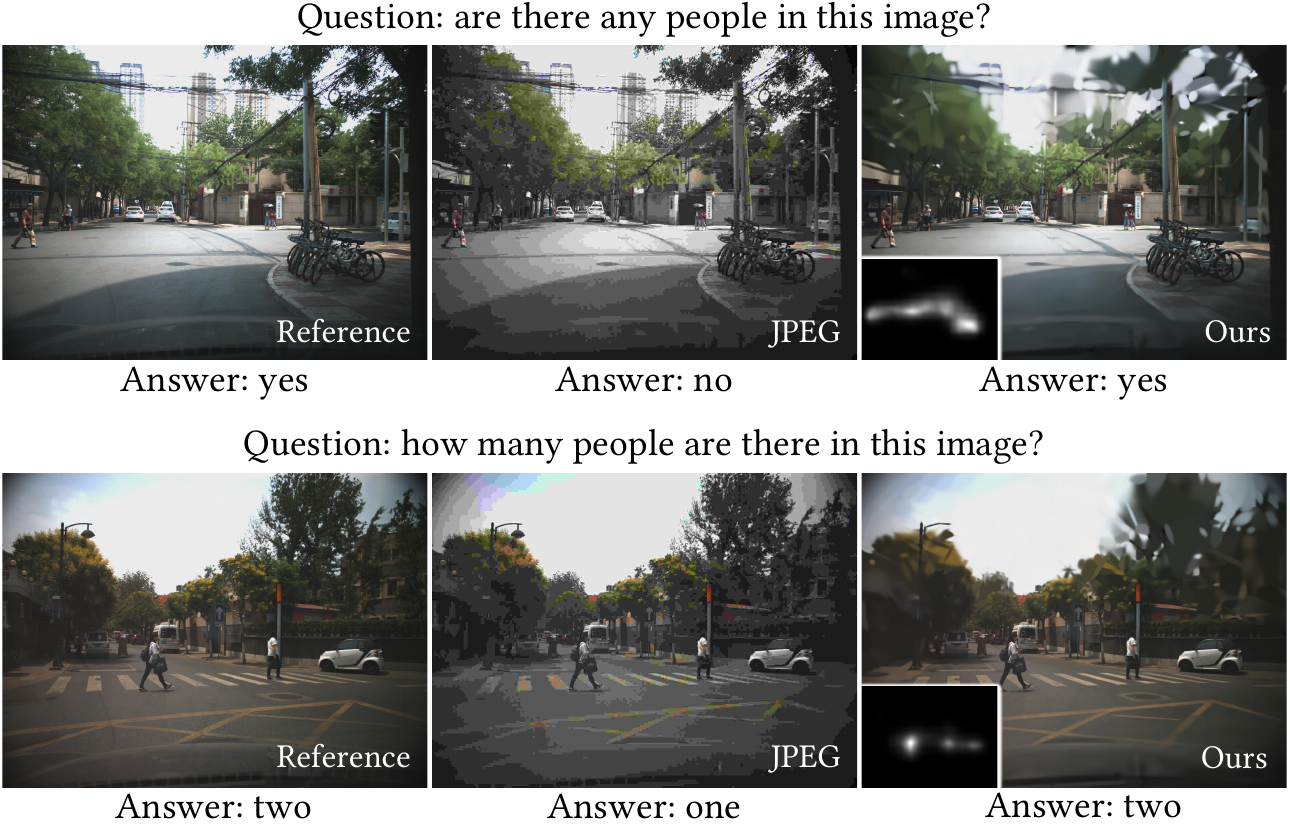}
\Caption{Semantics-aware compression (\Cref{sec:application-machine-vision}).}
{At an extreme rate of 0.2 bpp, \methodName enables more accurate VQA results with BLIP-2 than JPEG.}
\label{fig:application-machine-vision}
\end{figure}

We demonstrate the semantics-aware compression performance of \methodName on a vision-language task: visual question answering (VQA) \cite{antol2015vqa}. JPEG was used as the baseline and BLIP-2 \cite{li2023blip} was used to generate responses. We experimented with 20 randomly sampled images from the TJU-DHD dataset \cite{pang2020tju} and prepared custom questions for evaluation.

\Cref{fig:application-machine-vision} visualizes the results of 2 image-question pairs. At an extreme bitrate of 0.2 bpp, \methodName more effectively preserves task-relevant semantic image features during compression and enables outputs that better align with the uncompressed counterpart. Out of the 20 image-question pairs, \methodName achieved the same response as the reference 12 times, while JPEG only achieved 4 times. These results demonstrate \methodName's potential for serving as a compact yet robust encoding of visual inputs for machine vision applications.


\subsection{Joint Image Compression and Restoration}
\label{sec:application-image-restoration}

The low-pass nature of Gaussian functions endows \methodName with remarkable robustness against a range of common image distortions and artifacts at low bitrates, including the ringing and contouring patterns introduced by lossy compression \cite{vander1999jpeg}, the color banding and aliasing caused by quantization errors \cite{lorre1980artifacts}, and the various forms of noise that originate from transmission or imaging processes \cite{wei2020physics}.

We demonstrate that \methodName effectively achieves joint image compression and restoration at low bitrates. We experimented with 2 sets of 2K$\times$2K images, where 6 were JPEG-compressed (average size is 187 KB) and 8 were photos containing noticeable sensor noise. For quantitative evaluation, the corresponding uncompressed PNGs were used as ground truth for the 6 JPEG images, while AI-denoised outputs served as ground truth for the 8 photos.

\newcommand{\ImageRestorationRes}{0.32\linewidth}
\begin{figure}[t]
\centering
\subfloat{
    \includegraphics[width=\ImageRestorationRes]{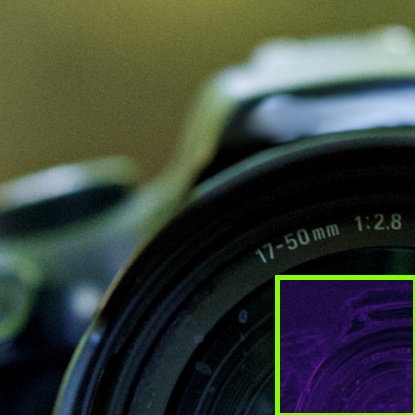}
  } \hspace{-0.18cm}
\subfloat{
    \includegraphics[width=\ImageRestorationRes]{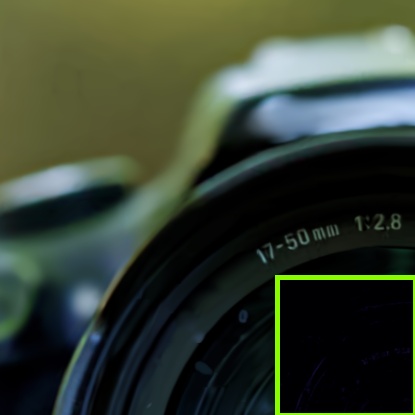}
  } \hspace{-0.18cm}
\subfloat{
    \includegraphics[width=\ImageRestorationRes]{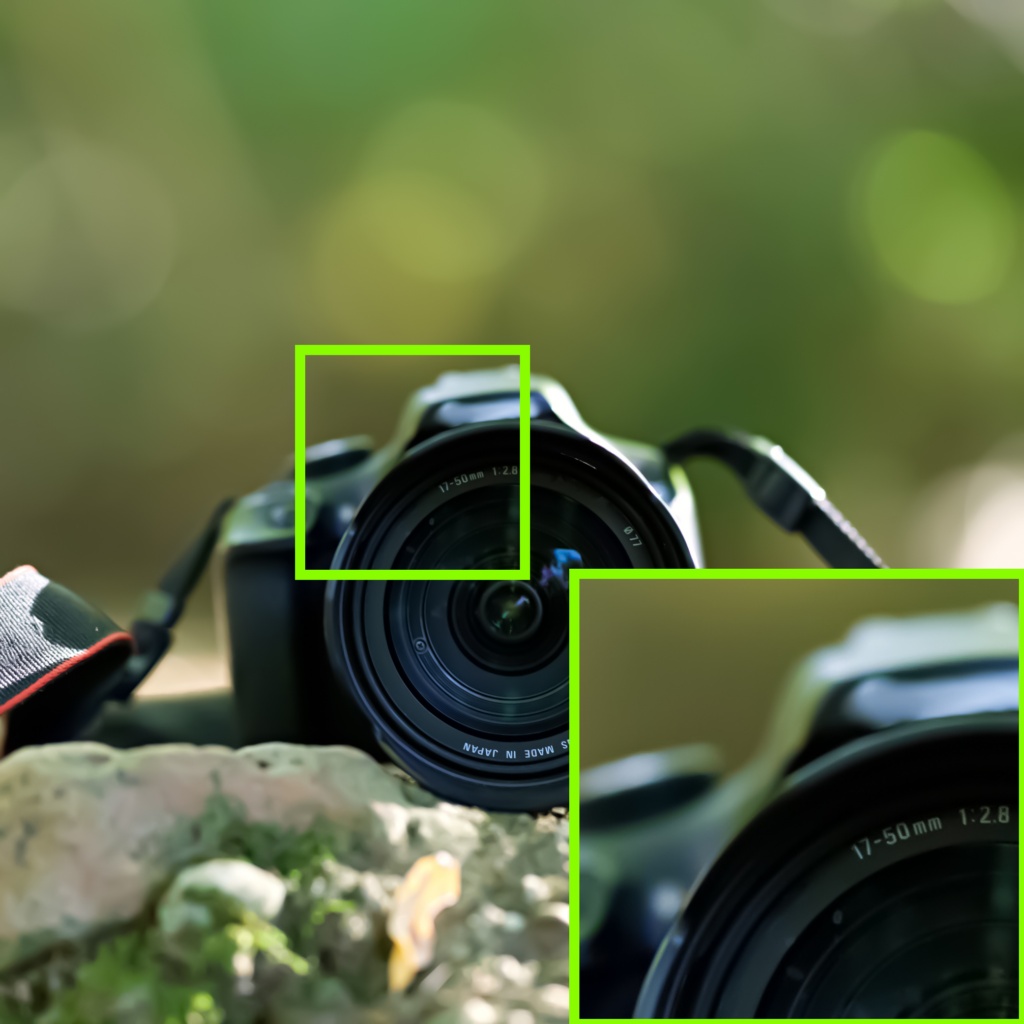}
  }
\vspace{0.2mm} \\
\setcounter{subfigure}{0}
\subfloat[Image with noise]{
    \includegraphics[width=\ImageRestorationRes]{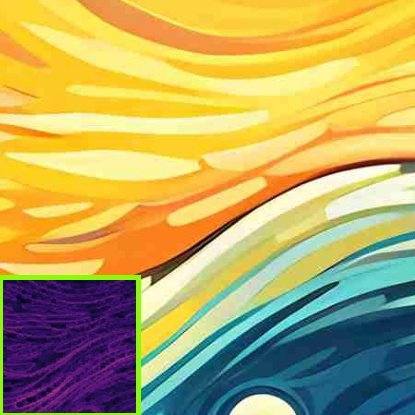}
  } \hspace{-0.18cm}
\subfloat[Restored by \methodName]{
    \includegraphics[width=\ImageRestorationRes]{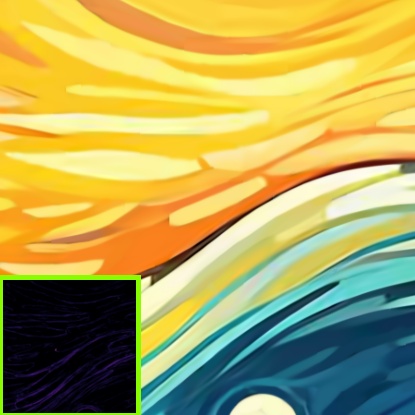}
  } \hspace{-0.18cm}
\subfloat[Reference]{
    \includegraphics[width=\ImageRestorationRes]{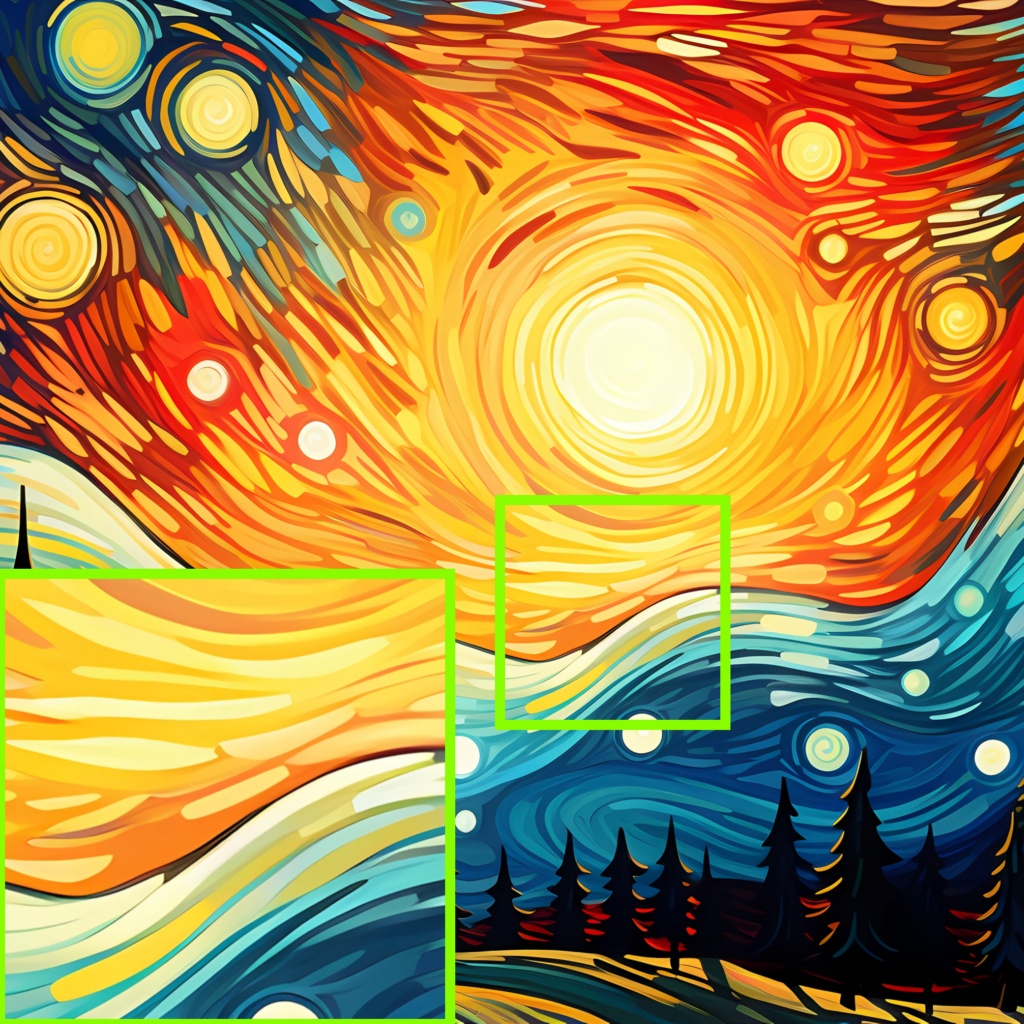}
  }
\Caption{Joint image compression and restoration (\Cref{sec:application-image-restoration}).}
{\revise{At low bitrates, \methodName effectively removes high-frequency artifacts from the input images while preserving detailed semantic content therein.}}
\label{fig:application-image-restoration}
\end{figure}

As shown in \Cref{fig:application-image-restoration}, \methodName eliminates most artifacts and noise from the input images while preserving the detailed content therein. This is because the limited representation budget (160 KB, $78.64\times$ compression) forces \methodName to prioritize bits on the more prominent image content instead of inconsistent pixel-level noise. Despite high compression ratios, images compressed by \methodName demonstrate consistently improved fidelity to the ground truth. On the 8 noise-corrupted photos, it achieves average gains of 1.782 in PSNR and 0.011 in MS-SSIM. On the 6 JPEG-compressed images, it achieves average gains of 0.354 in PSNR and 0.012 in MS-SSIM. Notably, these restoration effects naturally emerge from \methodName's compression process, requiring no additional post-processing. More results can be found in \Cref{fig:application-image-restoration-supp-1,fig:application-image-restoration-supp-2}.
\section{Limitations and Discussion}
\label{sec:discussion}

\paragraph{Spatially adaptive optimization}
Although \methodName is designed to be content-adaptive, its current optimization pipeline prioritizes large image features and struggles with reconstructing images that are rich in pixel-level details, such as natural images (see examples in \Cref{fig:evaluation-image-clic-supp-1,fig:evaluation-image-clic-supp-2}). Inspired by hybrid representations with spatially adaptive optimization \cite{martel2021acorn}, we plan to incorporate a dynamic binary space partitioning tree to guide the spatial distribution of Gaussians and adaptively scale the gradients they receive, with the tree structure and Gaussian attributes jointly updated during optimization. This approach encourages comparable importance across image features of varying scales.

\paragraph{Dynamic content}
We demonstrated in this research that images can be efficiently represented using an explicit basis of colored 2D Gaussians. Following recent works that incorporate dynamics into Gaussian-based scene representations \cite{luiten2024dynamic,diolatzis2024n}, we plan to extend \methodName to efficient video representations by modeling the motion of 2D Gaussians in the image plane. We envision this extension benefiting graphics applications such as panoramic video streaming in extended reality.

\section{Conclusion}

In this work, we proposed \methodName, an explicit image representation based on anisotropic, colored 2D Gaussians. \methodName supports favorable rate-distortion trade-offs, hardware-friendly fast random access, and flexible quality controls through a smooth level-of-detail stack. Its content-adaptive design effectively captures non-uniform image features and preserves fine details under constrained memory budgets. We hope this research inspires future advances in developing novel representations of visual data.
\begin{acks}
This research is partially supported by the NSF grants \#2232817 and \#2225861, and an Intel-sponsored research program.
\end{acks}


\bibliographystyle{ACM-Reference-Format}
\bibliography{paper.bib}

\appendixpageoff
\appendixtitleoff
\renewcommand{\appendixtocname}{Supplementary material}
\begin{appendices}
\crefalias{section}{supp}
\normalsize

\clearpage
\pagenumbering{roman}
\clearpage
\onecolumn
\clearpage
\newcommand{\imageCompWidthSupp}{0.114\linewidth}
\section{\revise{Additional Image Compression Results}}
\label{fig:evaluation-image-supp}
\begin{figure*}[h]
\centering
\subfloat{
    \includegraphics[width=\imageCompWidthSupp,height=\imageCompWidthSupp]{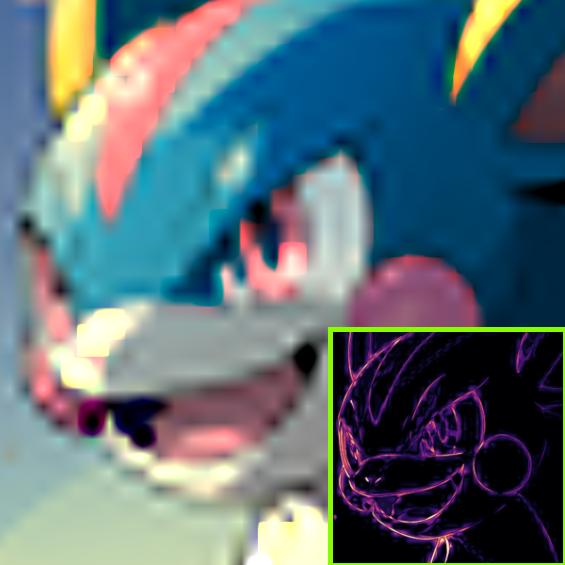}
  } \hspace{\reduceWidth}
\subfloat{
    \includegraphics[width=\imageCompWidthSupp,height=\imageCompWidthSupp]{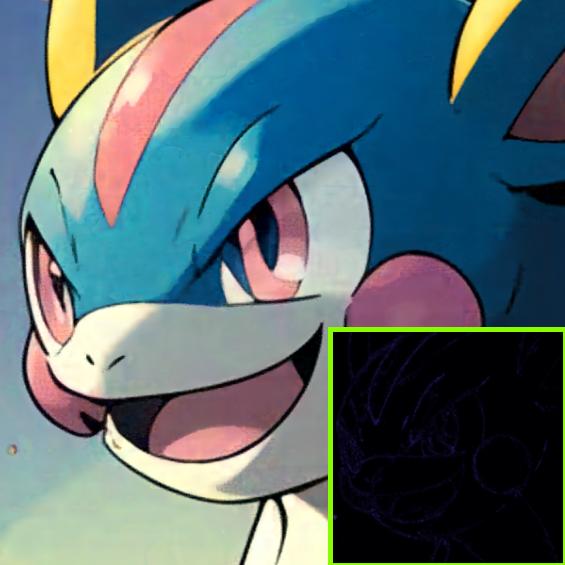}
  } \hspace{\reduceWidth}
\subfloat{
    \includegraphics[width=\imageCompWidthSupp,height=\imageCompWidthSupp]{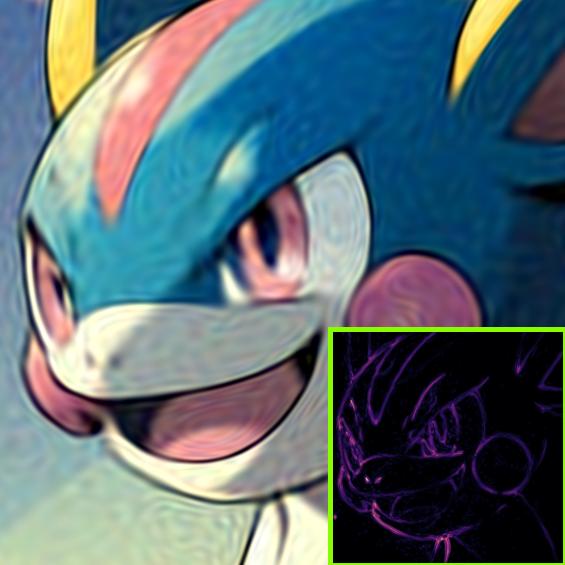}
  } \hspace{\reduceWidth}
\subfloat{
    \includegraphics[width=\imageCompWidthSupp,height=\imageCompWidthSupp]{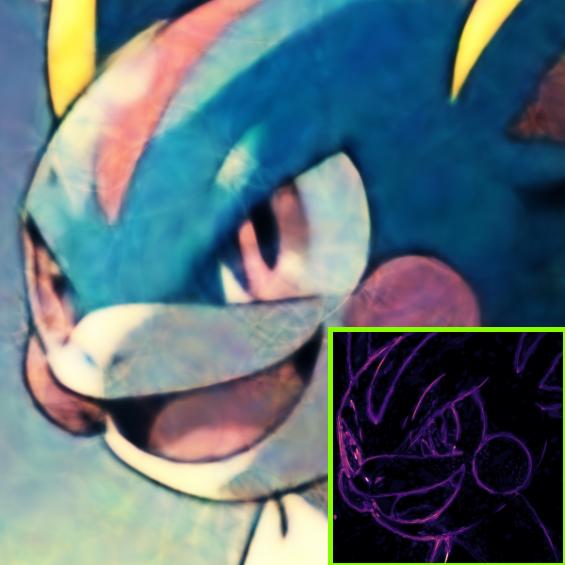}
  } \hspace{\reduceWidth}
\subfloat{
    \includegraphics[width=\imageCompWidthSupp,height=\imageCompWidthSupp]{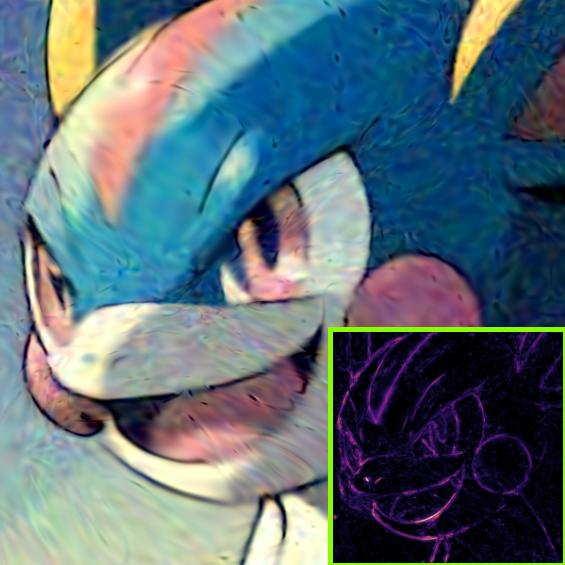}
  } \hspace{\reduceWidth}
\subfloat{
    \includegraphics[width=\imageCompWidthSupp,height=\imageCompWidthSupp]{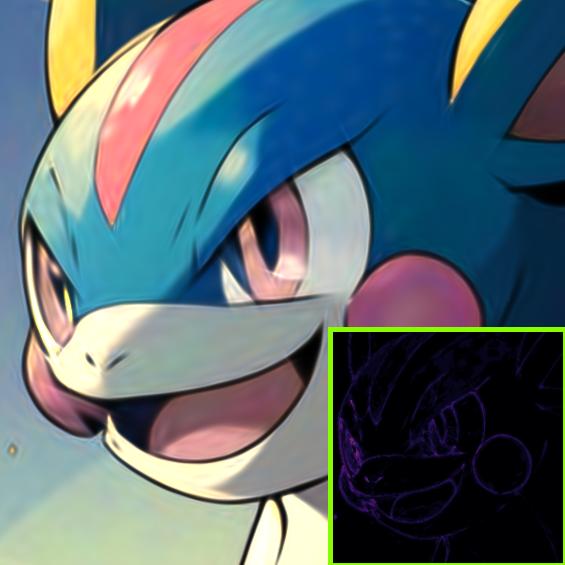}
  } \hspace{\reduceWidth}
\subfloat{
    \includegraphics[width=\imageCompWidthSupp,height=\imageCompWidthSupp]{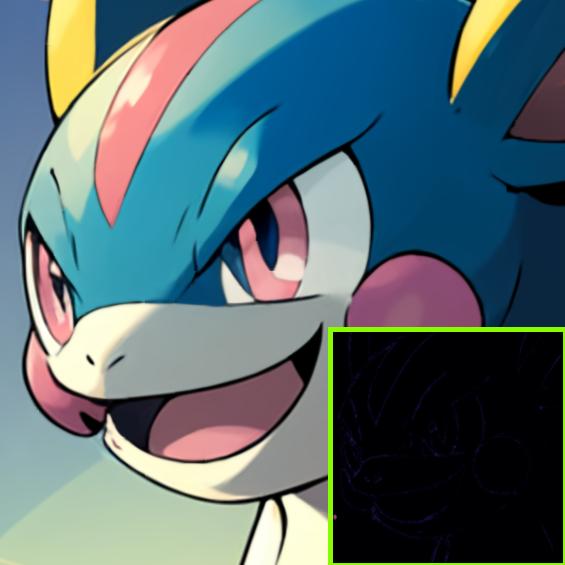}
  } \hspace{\reduceWidth}
\subfloat{
    \includegraphics[width=\imageCompWidthSupp,height=\imageCompWidthSupp]{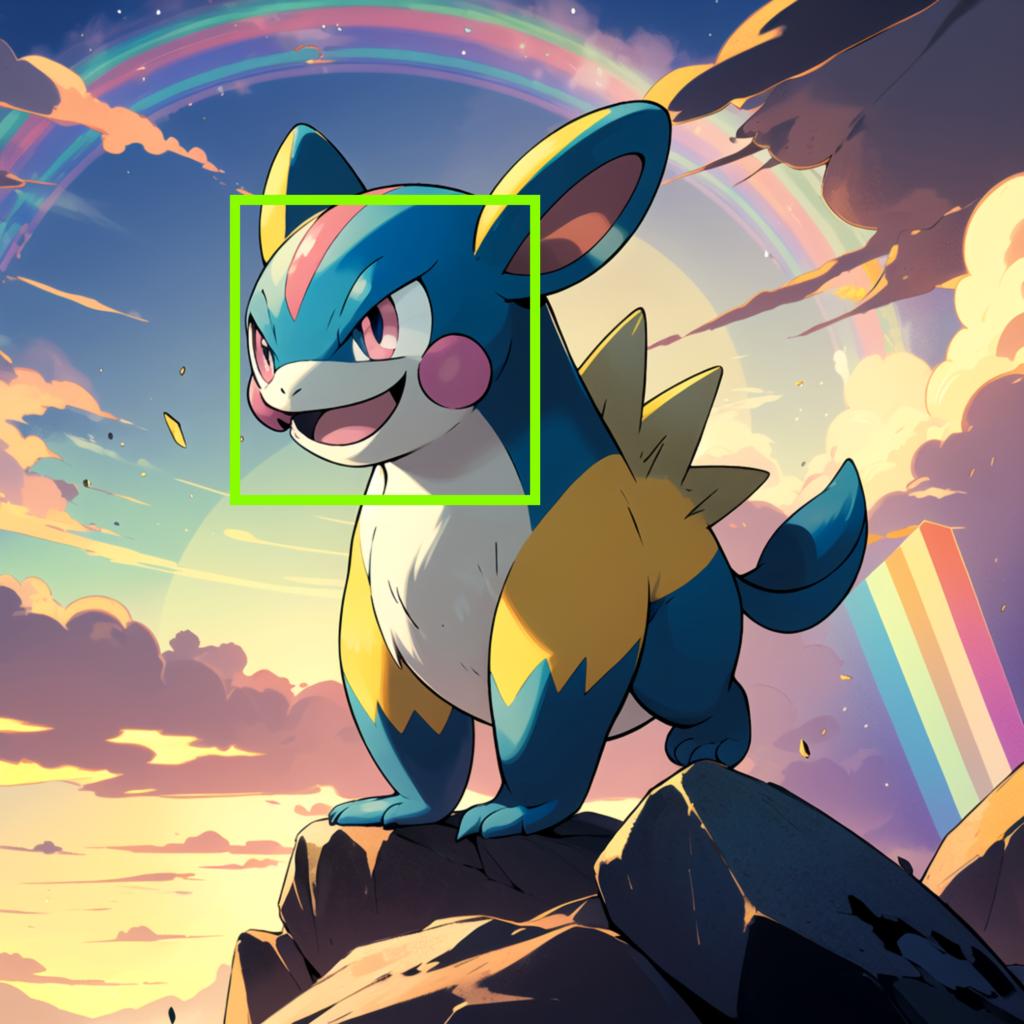}
  }
\vspace{0.2mm} \\
\subfloat{
    \includegraphics[width=\imageCompWidthSupp,height=\imageCompWidthSupp]{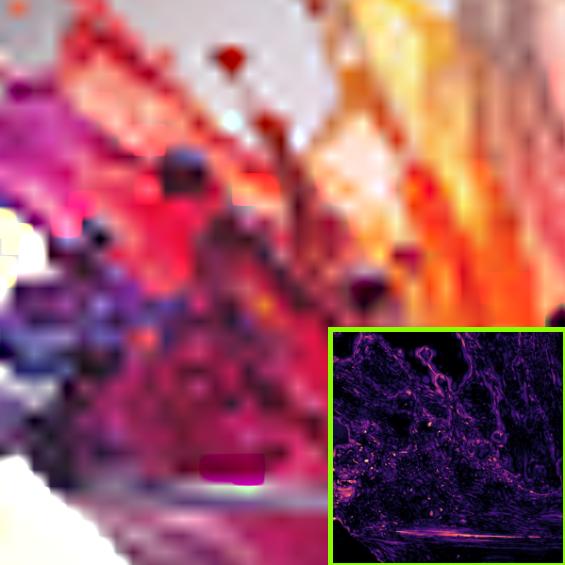}
  } \hspace{\reduceWidth}
\subfloat{
    \includegraphics[width=\imageCompWidthSupp,height=\imageCompWidthSupp]{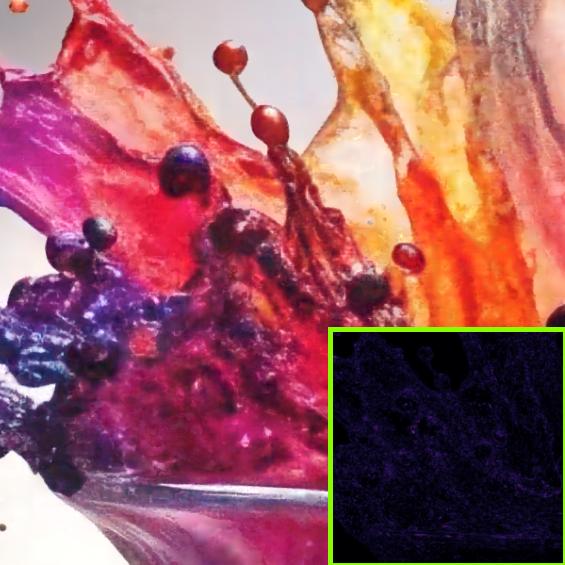}
  } \hspace{\reduceWidth}
\subfloat{
    \includegraphics[width=\imageCompWidthSupp,height=\imageCompWidthSupp]{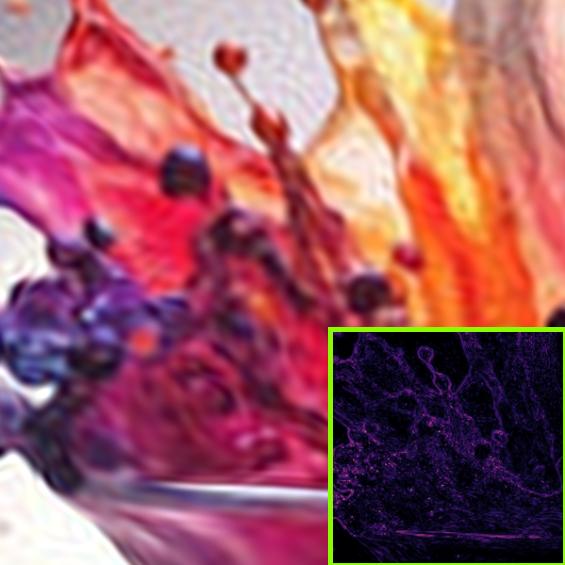}
  } \hspace{\reduceWidth}
\subfloat{
    \includegraphics[width=\imageCompWidthSupp,height=\imageCompWidthSupp]{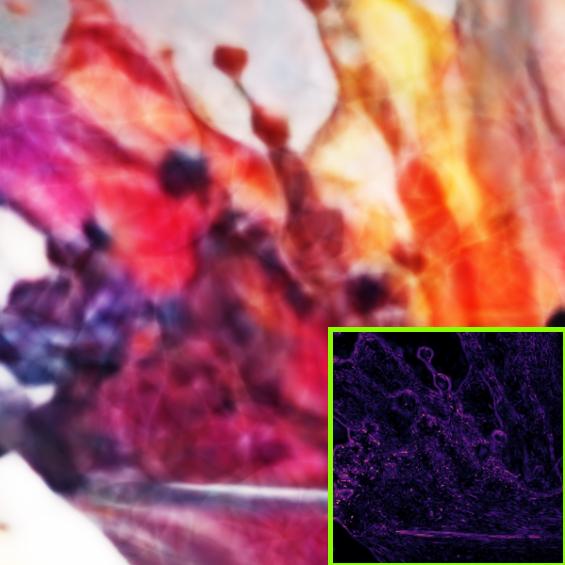}
  } \hspace{\reduceWidth}
\subfloat{
    \includegraphics[width=\imageCompWidthSupp,height=\imageCompWidthSupp]{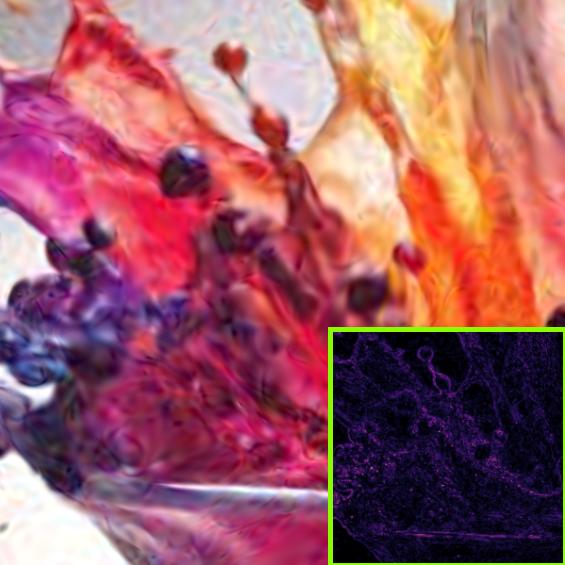}
  } \hspace{\reduceWidth}
\subfloat{
    \includegraphics[width=\imageCompWidthSupp,height=\imageCompWidthSupp]{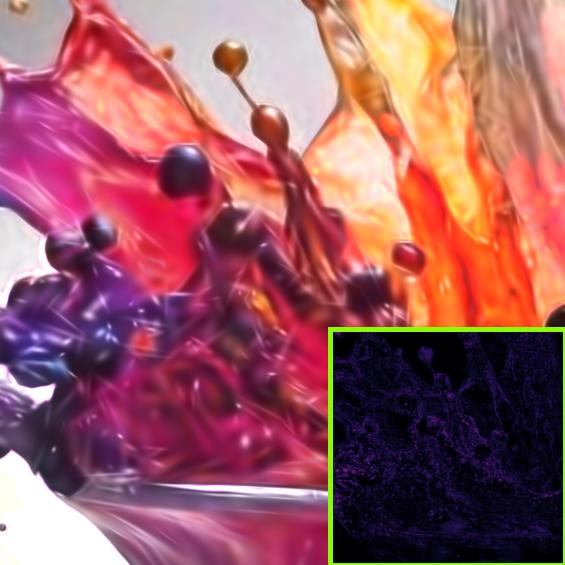}
  } \hspace{\reduceWidth}
\subfloat{
    \includegraphics[width=\imageCompWidthSupp,height=\imageCompWidthSupp]{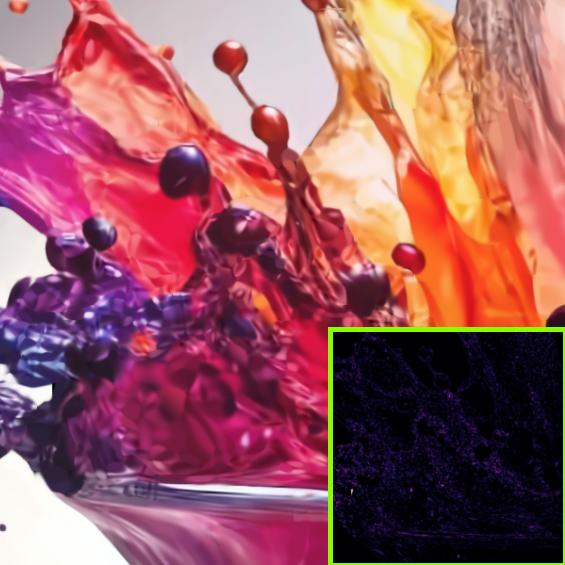}
  } \hspace{\reduceWidth}
\subfloat{
    \includegraphics[width=\imageCompWidthSupp,height=\imageCompWidthSupp]{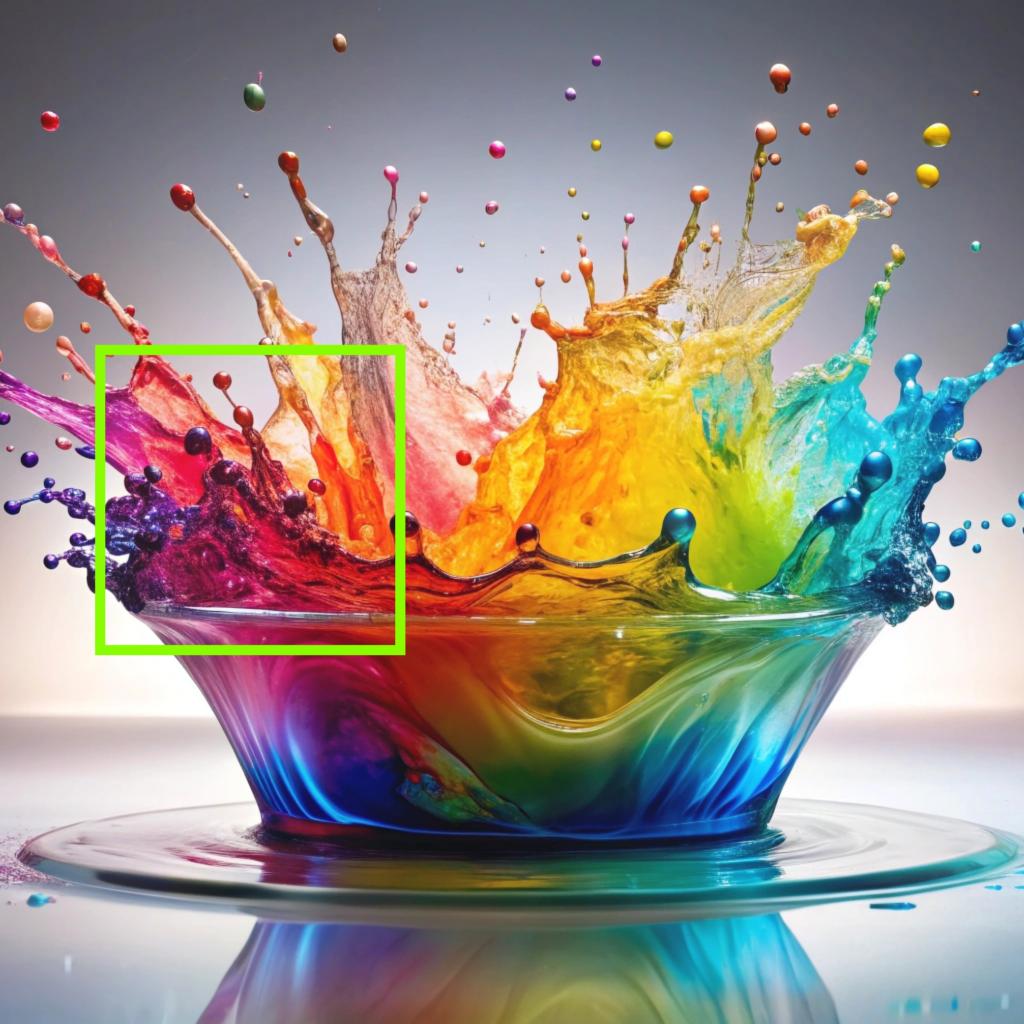}
  }
\vspace{0.2mm} \\
\subfloat{
    \includegraphics[width=\imageCompWidthSupp,height=\imageCompWidthSupp]{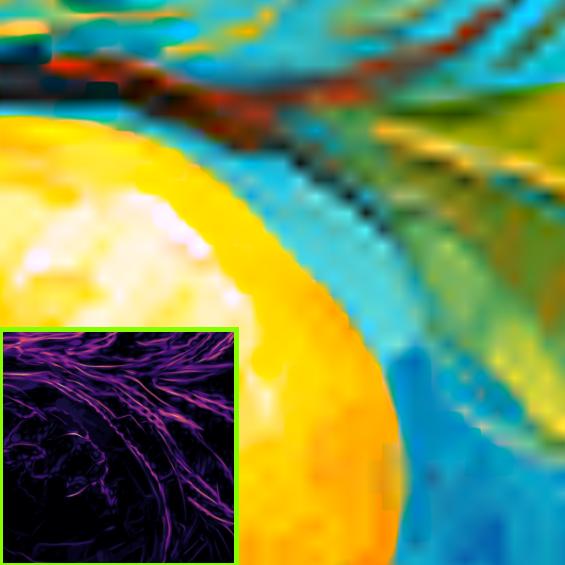}
  } \hspace{\reduceWidth}
\subfloat{
    \includegraphics[width=\imageCompWidthSupp,height=\imageCompWidthSupp]{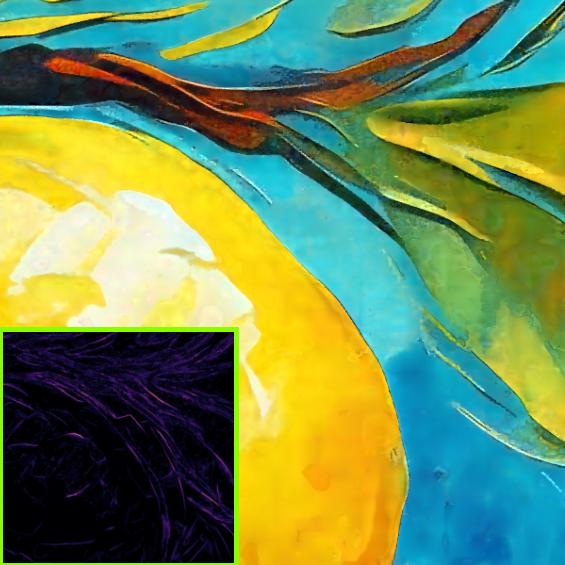}
  } \hspace{\reduceWidth}
\subfloat{
    \includegraphics[width=\imageCompWidthSupp,height=\imageCompWidthSupp]{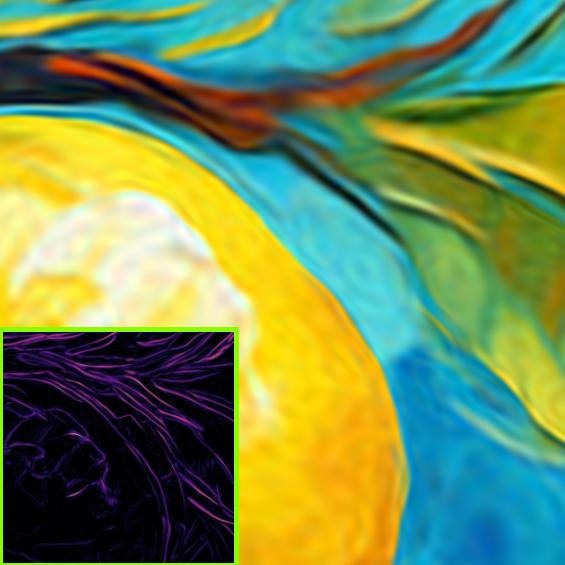}
  } \hspace{\reduceWidth}
\subfloat{
    \includegraphics[width=\imageCompWidthSupp,height=\imageCompWidthSupp]{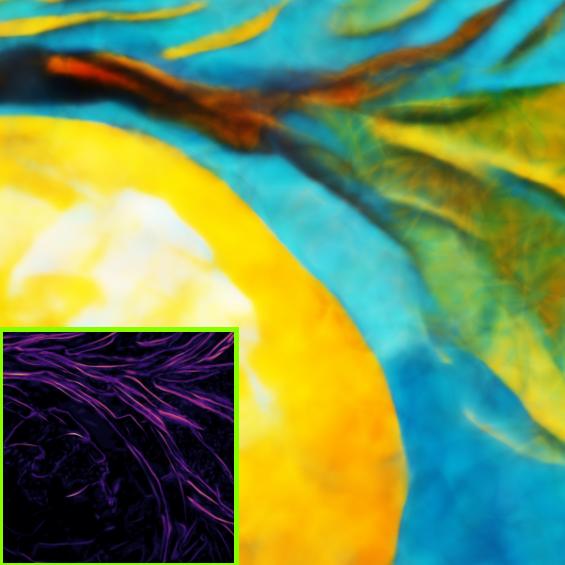}
  } \hspace{\reduceWidth}
\subfloat{
    \includegraphics[width=\imageCompWidthSupp,height=\imageCompWidthSupp]{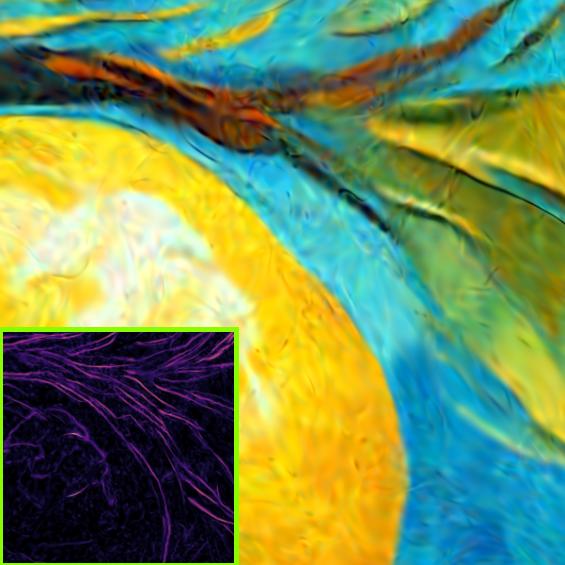}
  } \hspace{\reduceWidth}
\subfloat{
    \includegraphics[width=\imageCompWidthSupp,height=\imageCompWidthSupp]{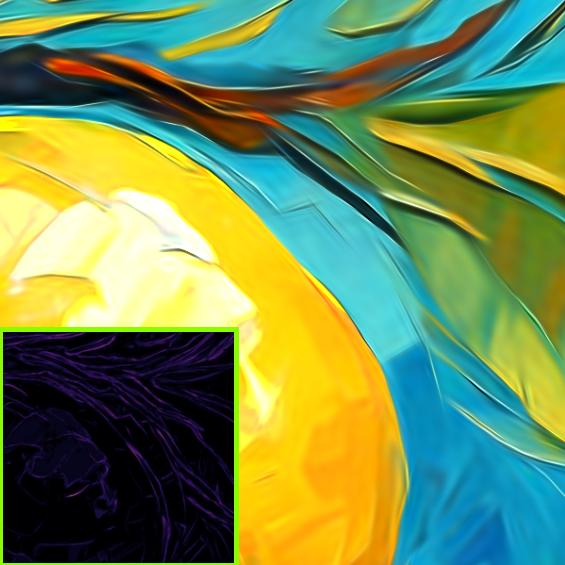}
  } \hspace{\reduceWidth}
\subfloat{
    \includegraphics[width=\imageCompWidthSupp,height=\imageCompWidthSupp]{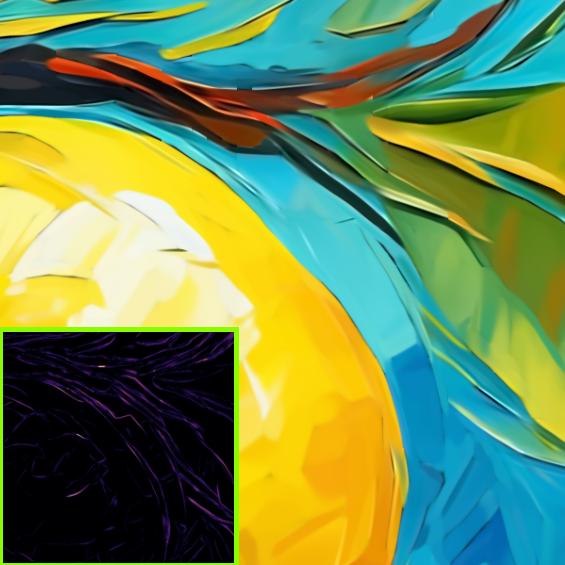}
  } \hspace{\reduceWidth}
\subfloat{
    \includegraphics[width=\imageCompWidthSupp,height=\imageCompWidthSupp]{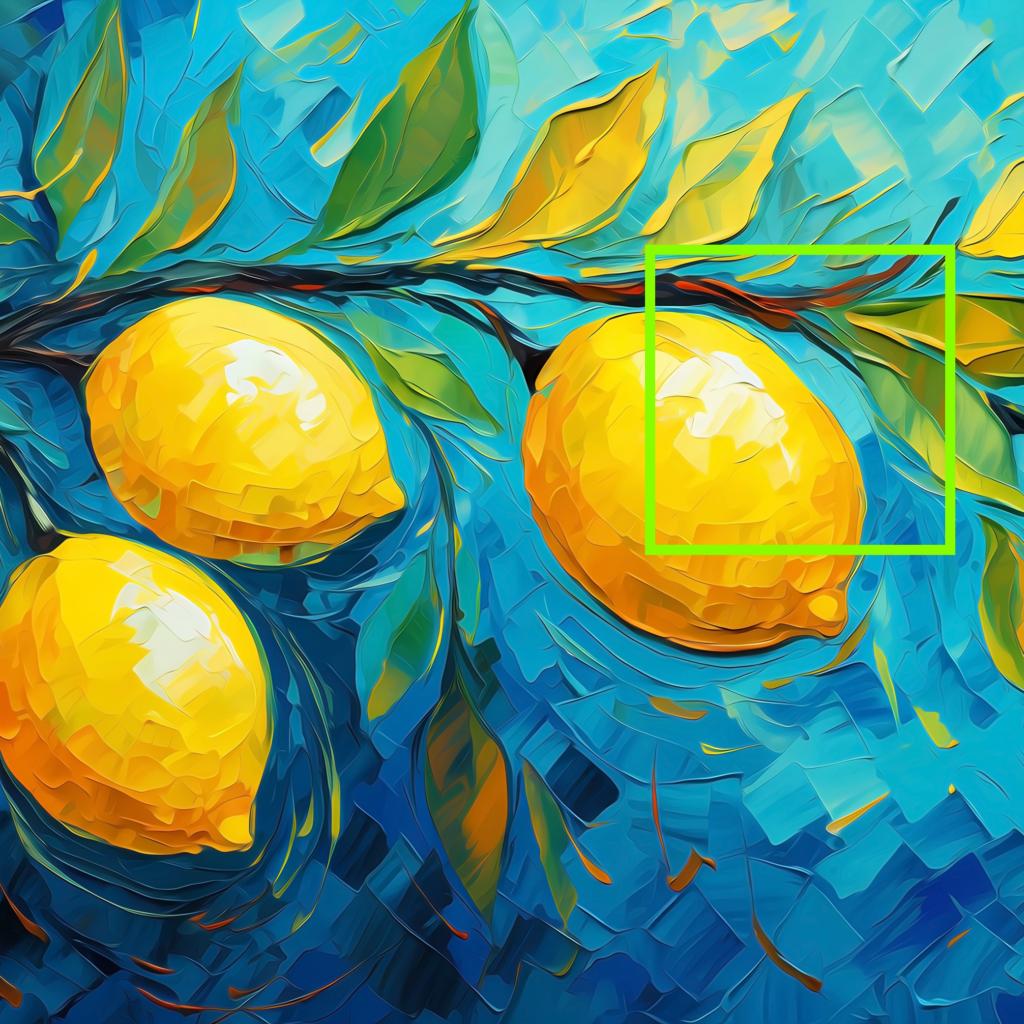}
  }
\vspace{0.2mm} \\
\subfloat{
    \includegraphics[width=\imageCompWidthSupp,height=\imageCompWidthSupp]{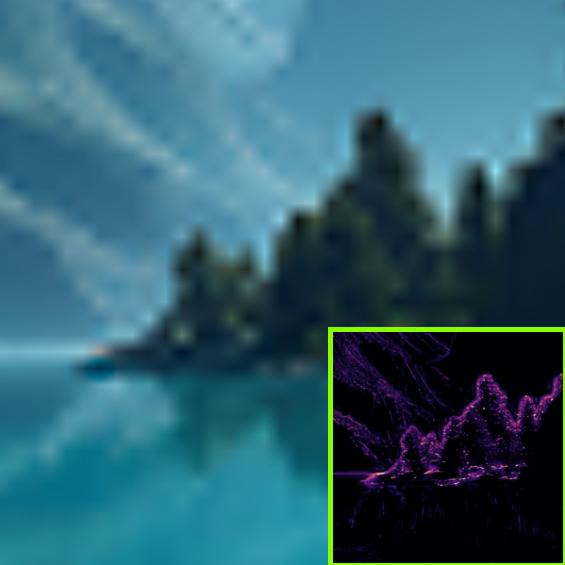}
  } \hspace{\reduceWidth}
\subfloat{
    \includegraphics[width=\imageCompWidthSupp,height=\imageCompWidthSupp]{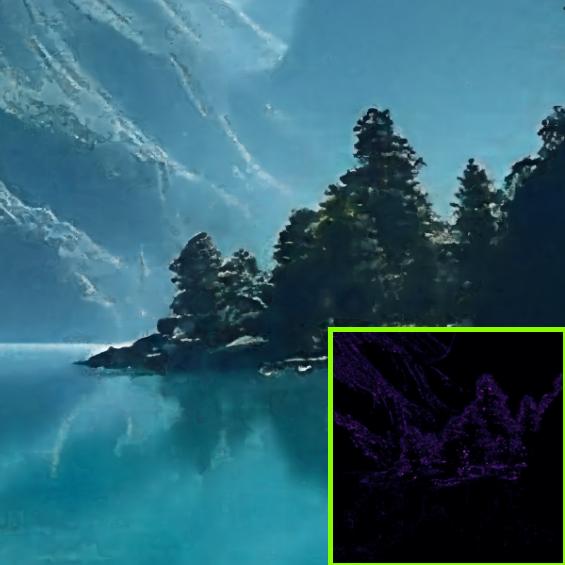}
  } \hspace{\reduceWidth}
\subfloat{
    \includegraphics[width=\imageCompWidthSupp,height=\imageCompWidthSupp]{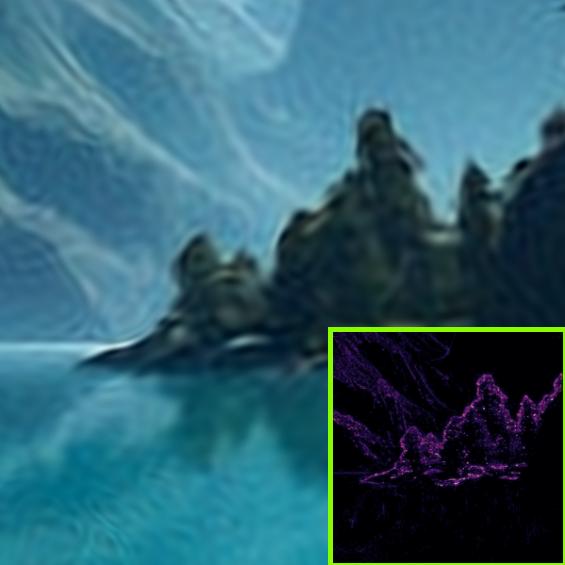}
  } \hspace{\reduceWidth}
\subfloat{
    \includegraphics[width=\imageCompWidthSupp,height=\imageCompWidthSupp]{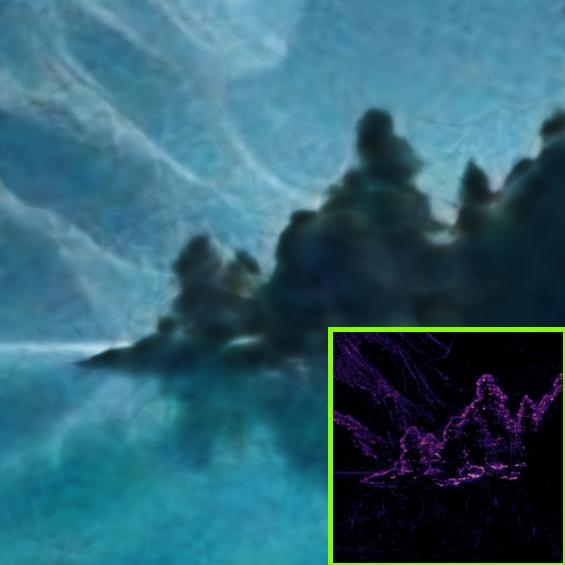}
  } \hspace{\reduceWidth}
\subfloat{
    \includegraphics[width=\imageCompWidthSupp,height=\imageCompWidthSupp]{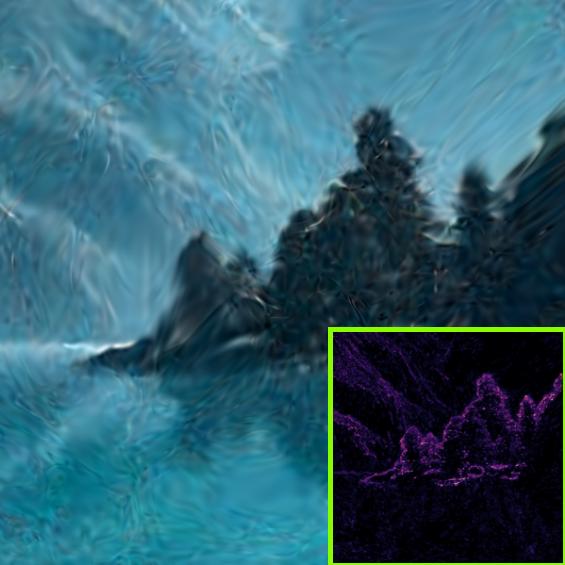}
  } \hspace{\reduceWidth}
\subfloat{
    \includegraphics[width=\imageCompWidthSupp,height=\imageCompWidthSupp]{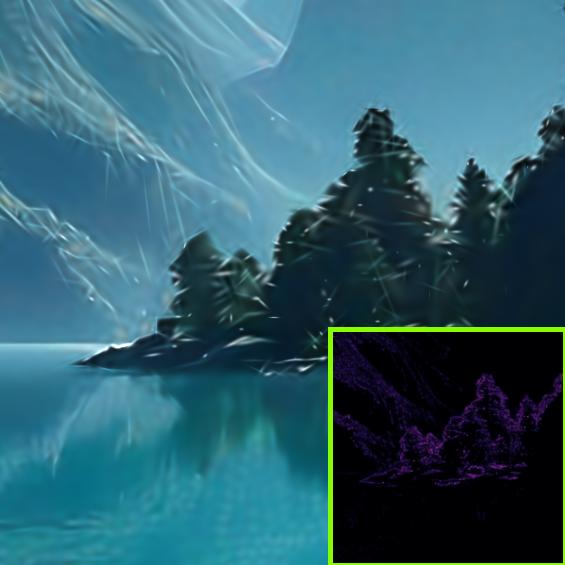}
  } \hspace{\reduceWidth}
\subfloat{
    \includegraphics[width=\imageCompWidthSupp,height=\imageCompWidthSupp]{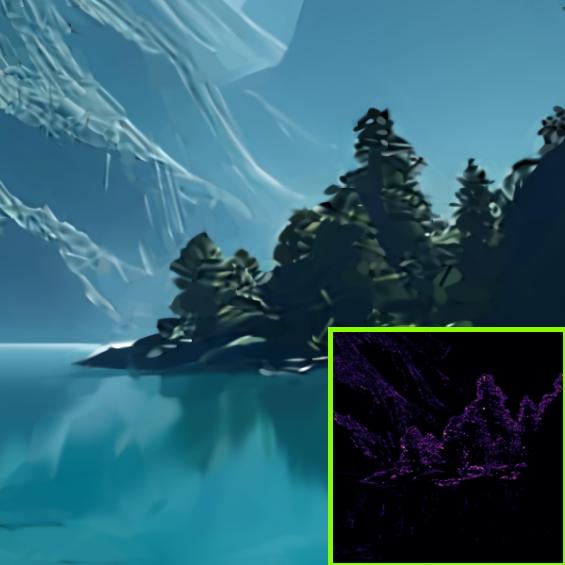}
  } \hspace{\reduceWidth}
\subfloat{
    \includegraphics[width=\imageCompWidthSupp,height=\imageCompWidthSupp]{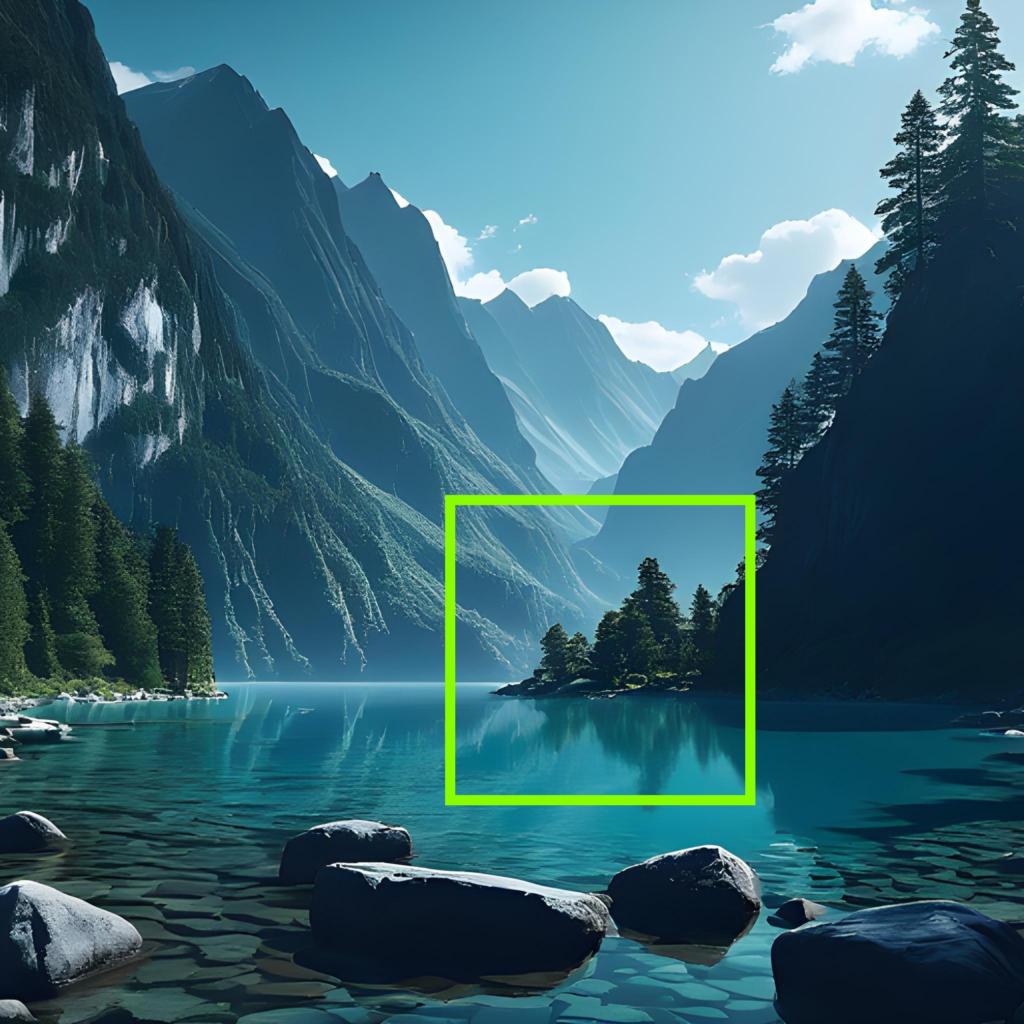}
  }
\vspace{0.2mm} \\
\subfloat{
    \includegraphics[width=\imageCompWidthSupp,height=\imageCompWidthSupp]{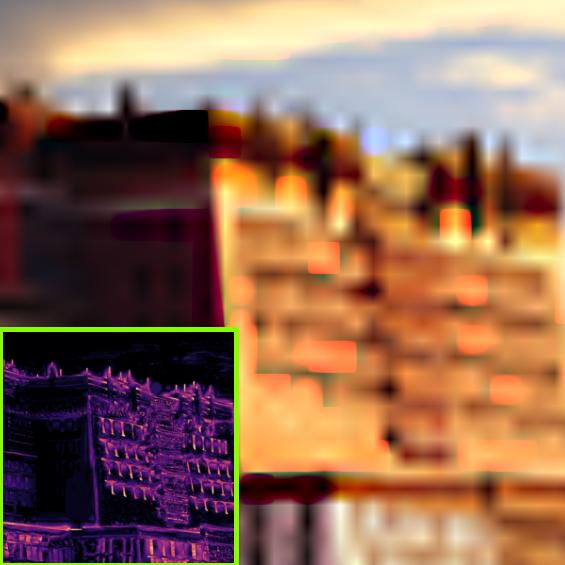}
  } \hspace{\reduceWidth}
\subfloat{
    \includegraphics[width=\imageCompWidthSupp,height=\imageCompWidthSupp]{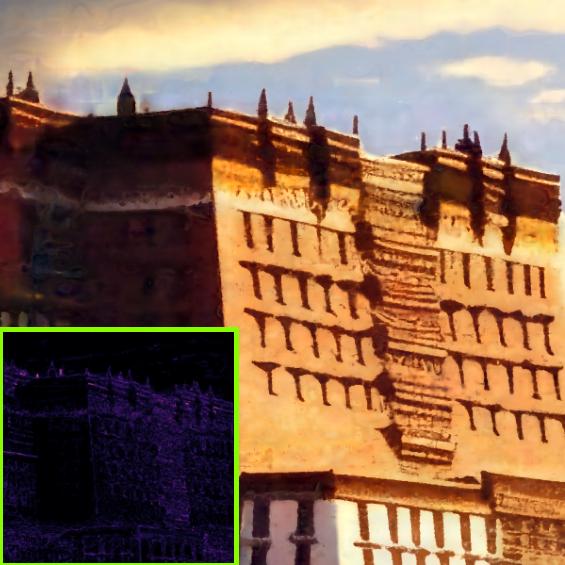}
  } \hspace{\reduceWidth}
\subfloat{
    \includegraphics[width=\imageCompWidthSupp,height=\imageCompWidthSupp]{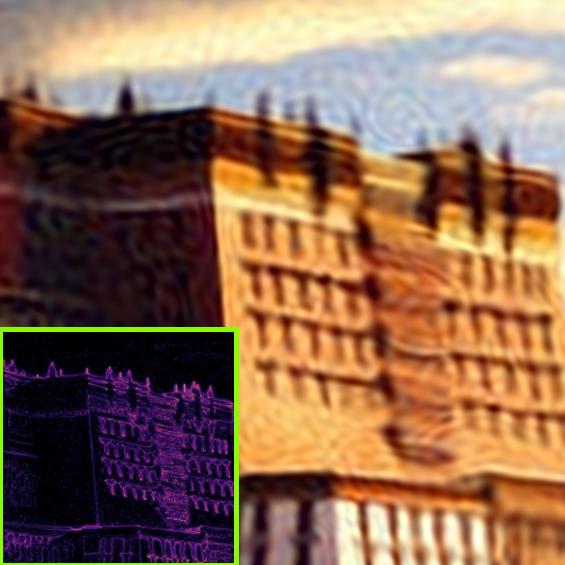}
  } \hspace{\reduceWidth}
\subfloat{
    \includegraphics[width=\imageCompWidthSupp,height=\imageCompWidthSupp]{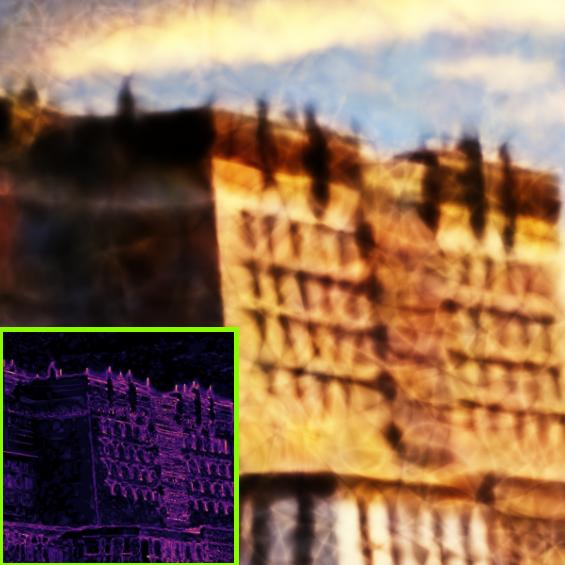}
  } \hspace{\reduceWidth}
\subfloat{
    \includegraphics[width=\imageCompWidthSupp,height=\imageCompWidthSupp]{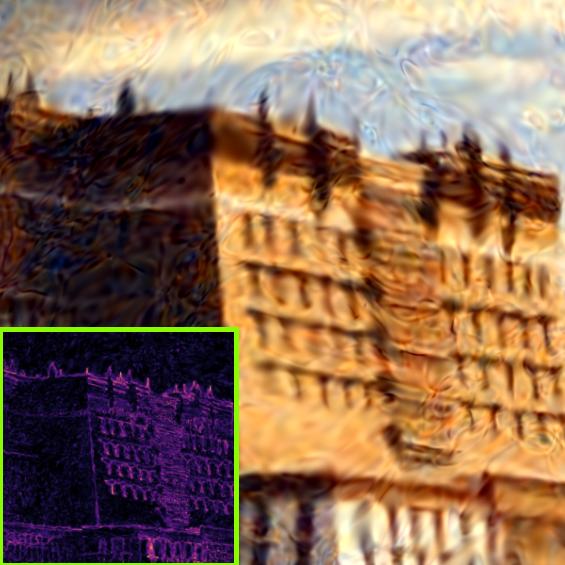}
  } \hspace{\reduceWidth}
\subfloat{
    \includegraphics[width=\imageCompWidthSupp,height=\imageCompWidthSupp]{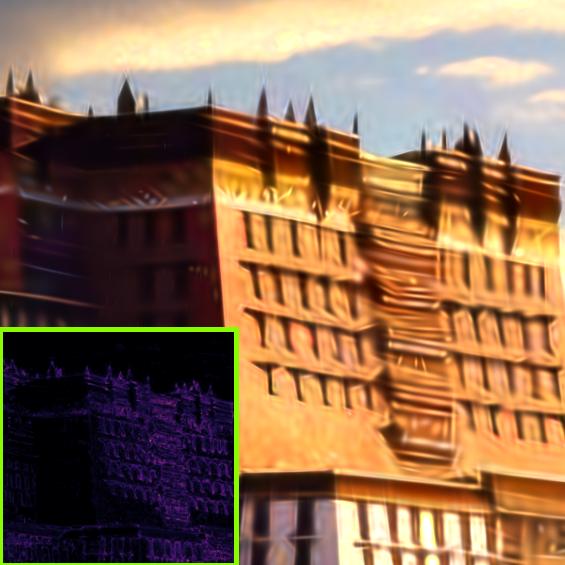}
  } \hspace{\reduceWidth}
\subfloat{
    \includegraphics[width=\imageCompWidthSupp,height=\imageCompWidthSupp]{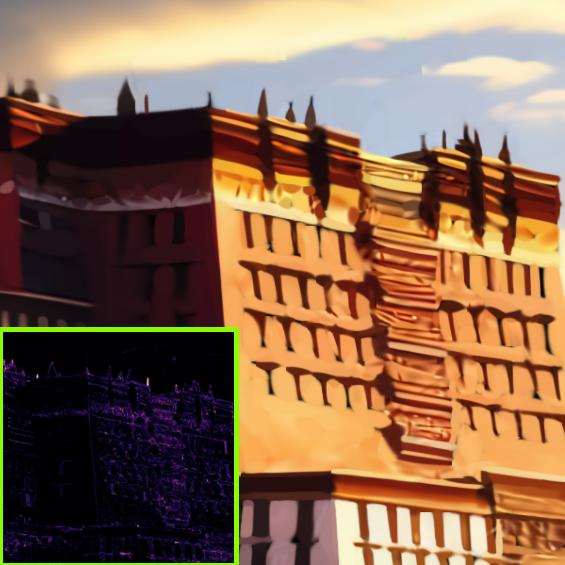}
  } \hspace{\reduceWidth}
\subfloat{
    \includegraphics[width=\imageCompWidthSupp,height=\imageCompWidthSupp]{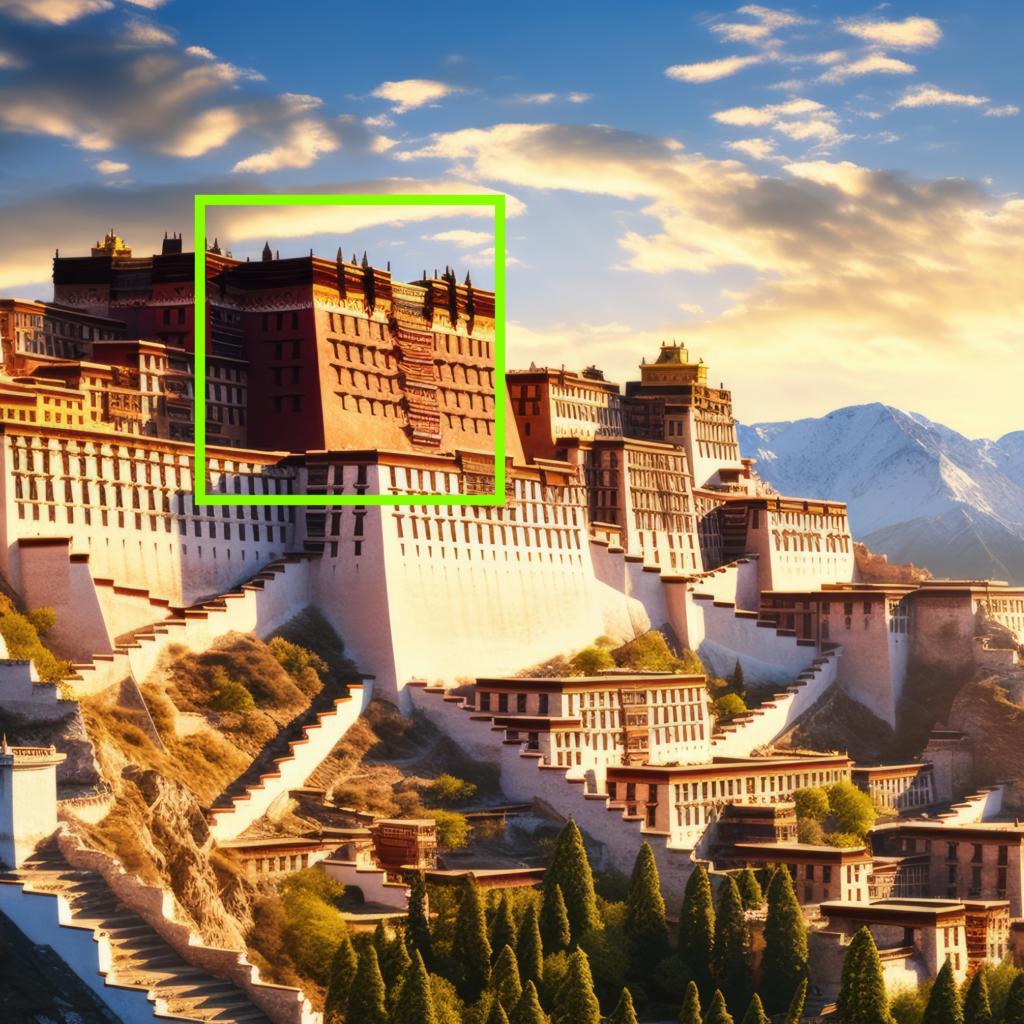}
  }
\vspace{0.2mm} \\
\subfloat{
    \includegraphics[width=\imageCompWidthSupp,height=\imageCompWidthSupp]{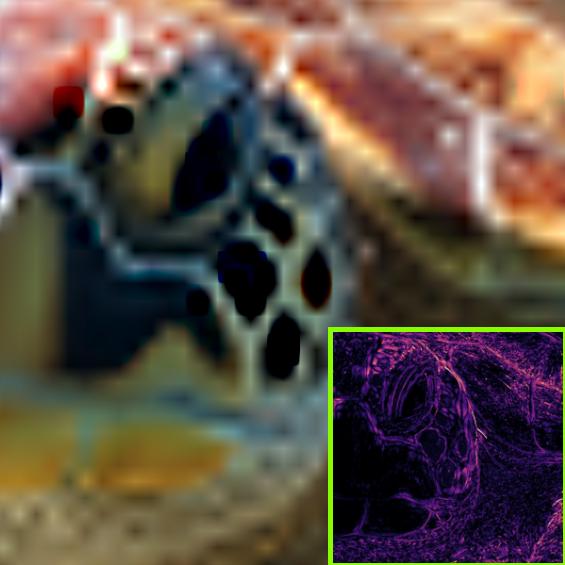}
  } \hspace{\reduceWidth}
\subfloat{
    \includegraphics[width=\imageCompWidthSupp,height=\imageCompWidthSupp]{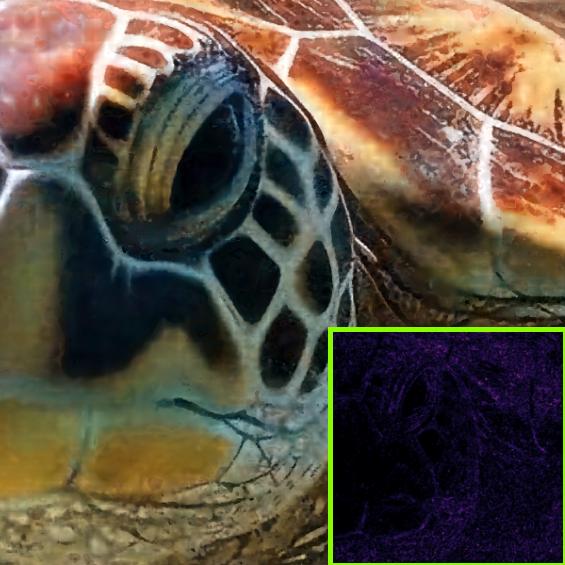}
  } \hspace{\reduceWidth}
\subfloat{
    \includegraphics[width=\imageCompWidthSupp,height=\imageCompWidthSupp]{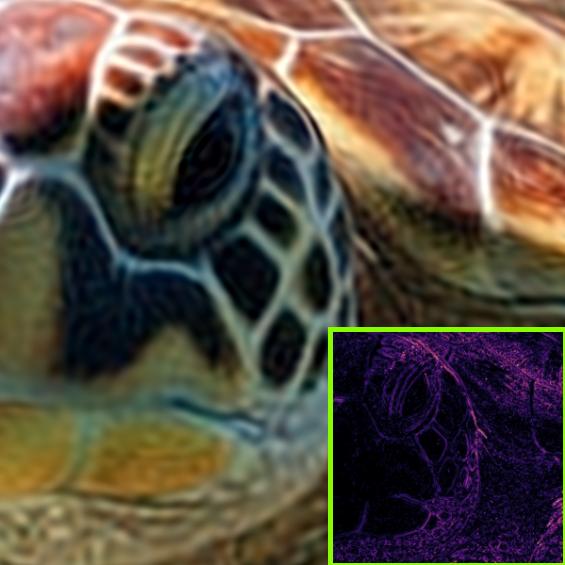}
  } \hspace{\reduceWidth}
\subfloat{
    \includegraphics[width=\imageCompWidthSupp,height=\imageCompWidthSupp]{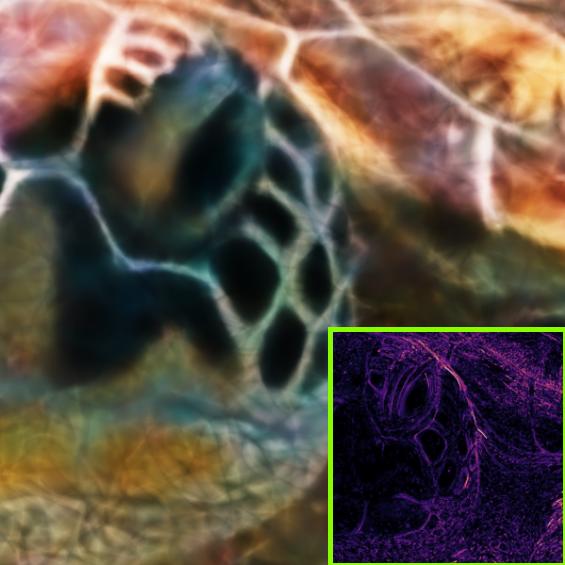}
  } \hspace{\reduceWidth}
\subfloat{
    \includegraphics[width=\imageCompWidthSupp,height=\imageCompWidthSupp]{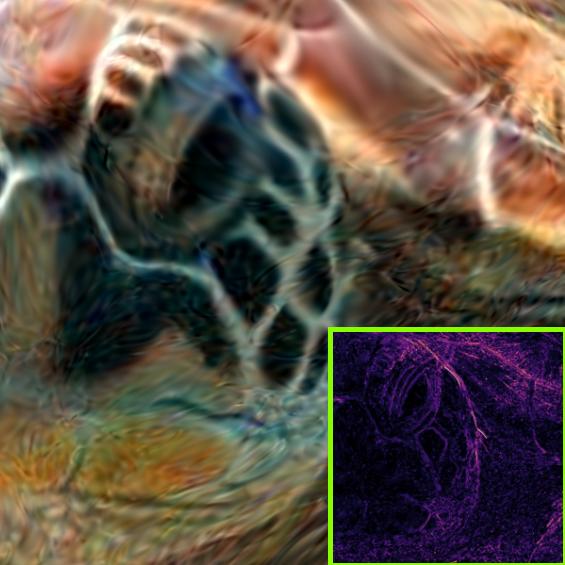}
  } \hspace{\reduceWidth}
\subfloat{
    \includegraphics[width=\imageCompWidthSupp,height=\imageCompWidthSupp]{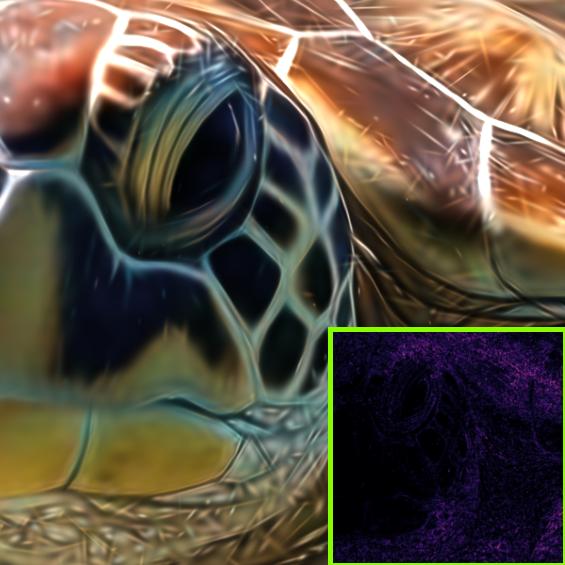}
  } \hspace{\reduceWidth}
\subfloat{
    \includegraphics[width=\imageCompWidthSupp,height=\imageCompWidthSupp]{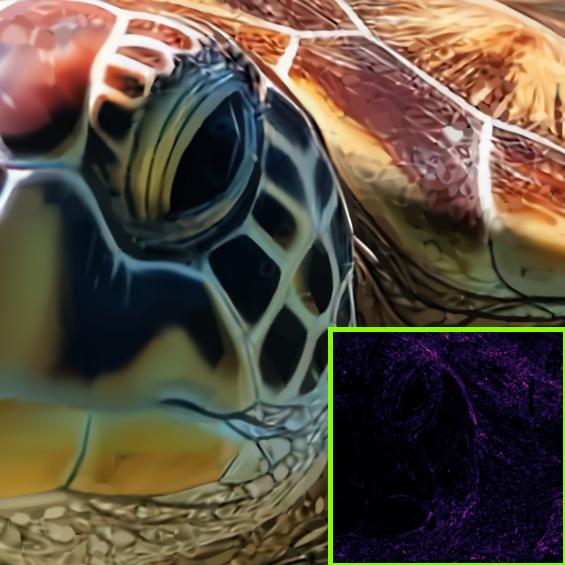}
  } \hspace{\reduceWidth}
\subfloat{
    \includegraphics[width=\imageCompWidthSupp,height=\imageCompWidthSupp]{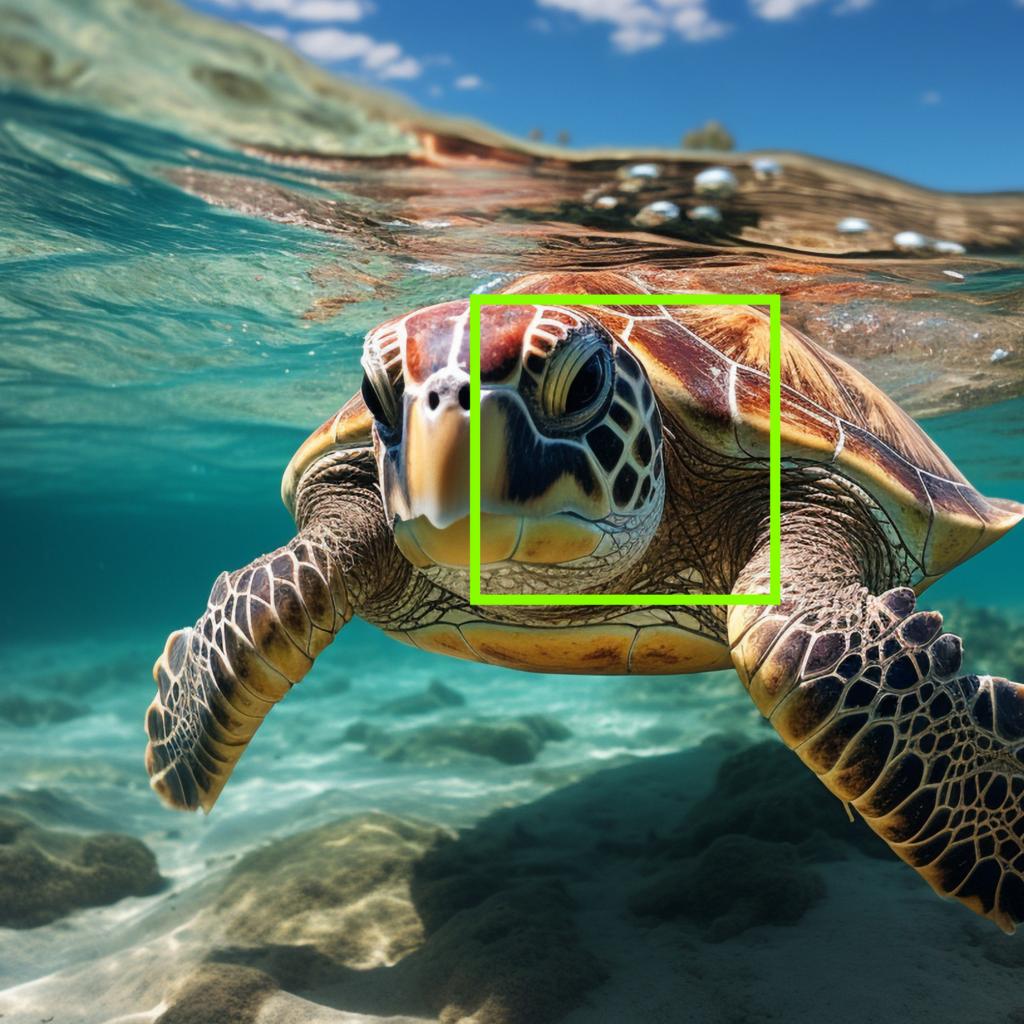}
  }
\vspace{0.2mm} \\
\subfloat{
    \includegraphics[width=\imageCompWidthSupp,height=\imageCompWidthSupp]{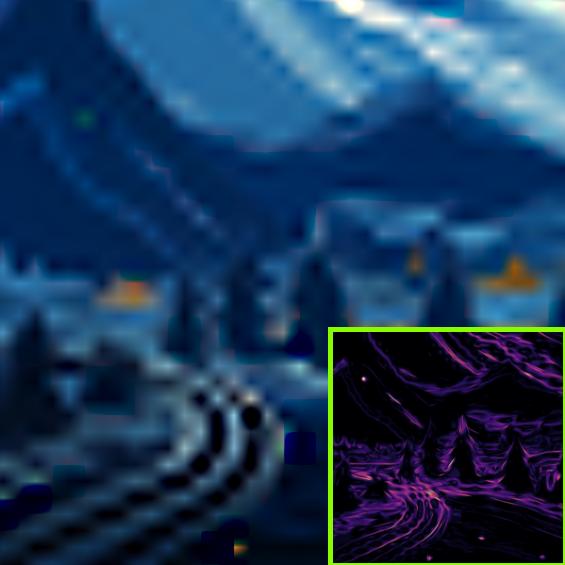}
  } \hspace{\reduceWidth}
\subfloat{
    \includegraphics[width=\imageCompWidthSupp,height=\imageCompWidthSupp]{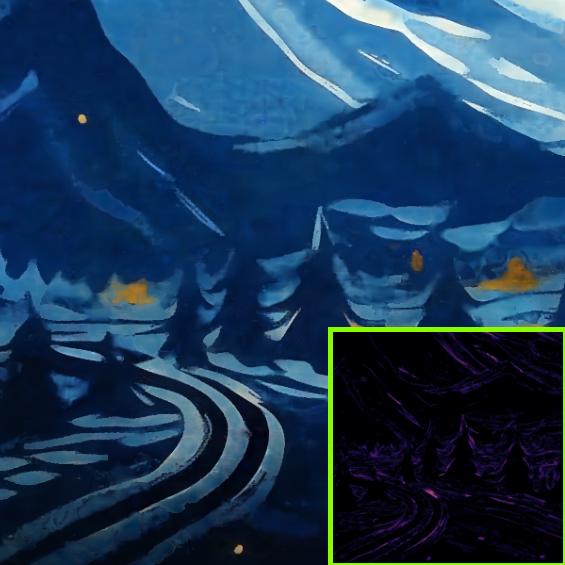}
  } \hspace{\reduceWidth}
\subfloat{
    \includegraphics[width=\imageCompWidthSupp,height=\imageCompWidthSupp]{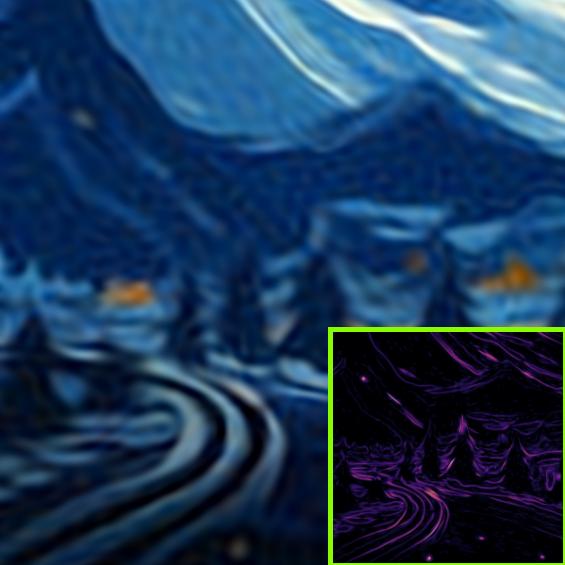}
  } \hspace{\reduceWidth}
\subfloat{
    \includegraphics[width=\imageCompWidthSupp,height=\imageCompWidthSupp]{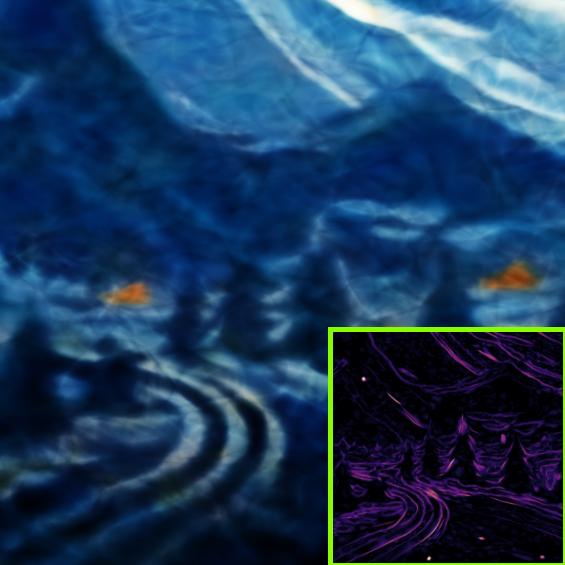}
  } \hspace{\reduceWidth}
\subfloat{
    \includegraphics[width=\imageCompWidthSupp,height=\imageCompWidthSupp]{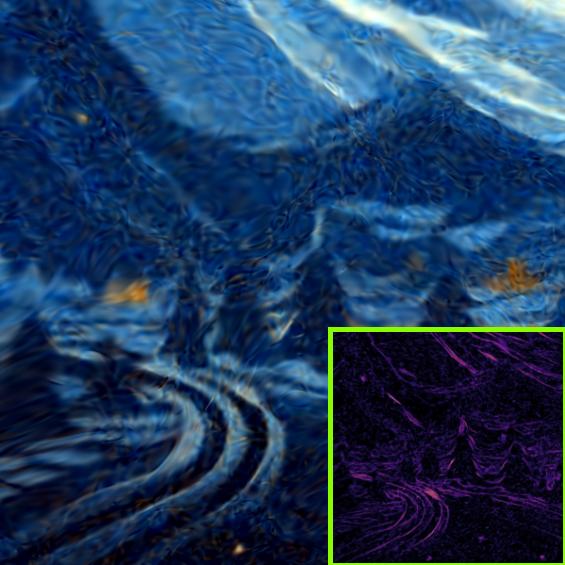}
  } \hspace{\reduceWidth}
\subfloat{
    \includegraphics[width=\imageCompWidthSupp,height=\imageCompWidthSupp]{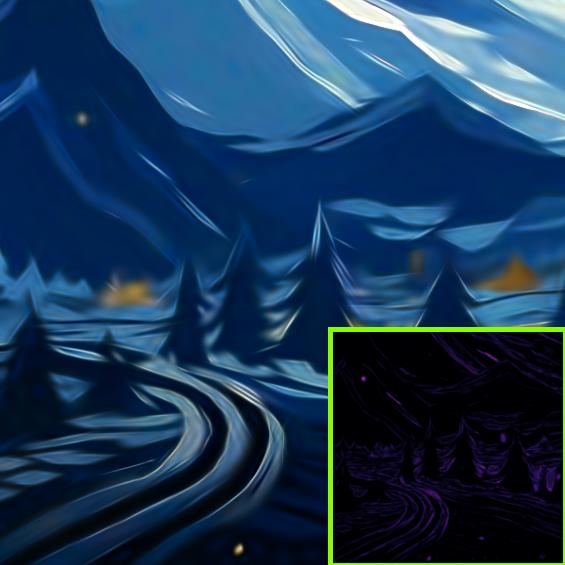}
  } \hspace{\reduceWidth}
\subfloat{
    \includegraphics[width=\imageCompWidthSupp,height=\imageCompWidthSupp]{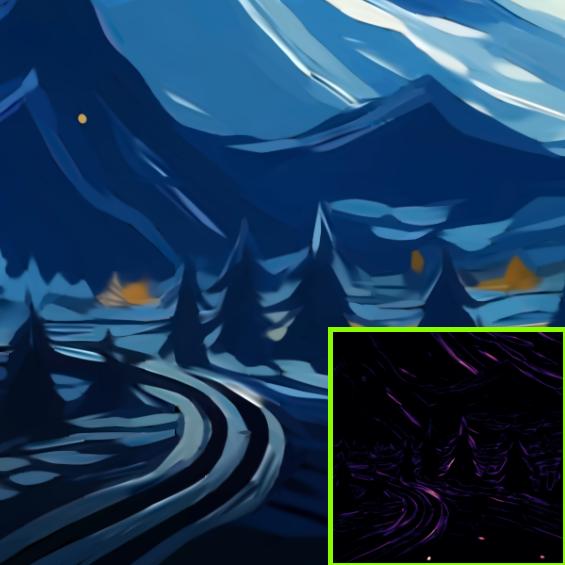}
  } \hspace{\reduceWidth}
\subfloat{
    \includegraphics[width=\imageCompWidthSupp,height=\imageCompWidthSupp]{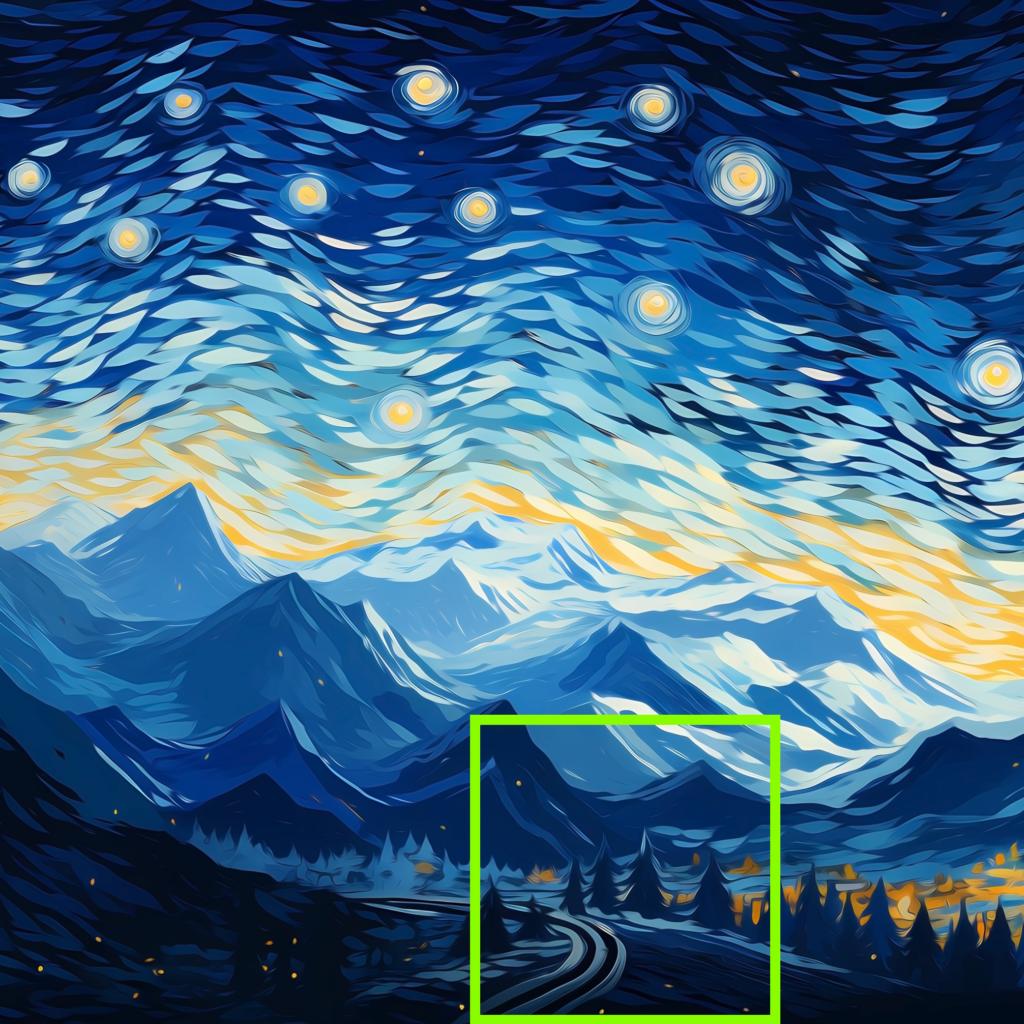}
  }
\vspace{0.2mm} \\
\subfloat{
    \includegraphics[width=\imageCompWidthSupp,height=\imageCompWidthSupp]{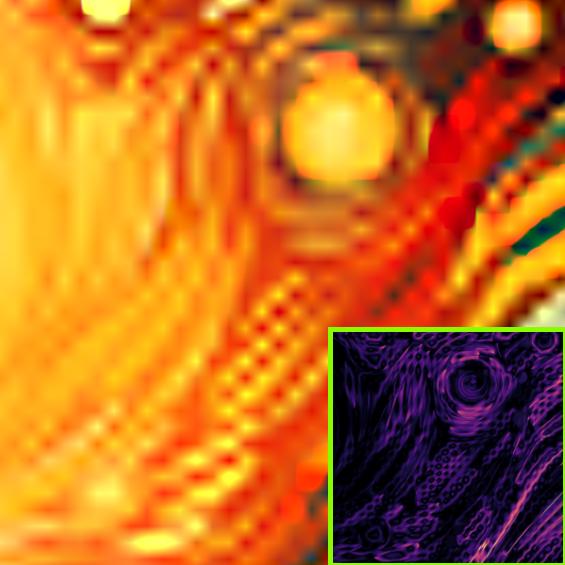}
  } \hspace{\reduceWidth}
\subfloat{
    \includegraphics[width=\imageCompWidthSupp,height=\imageCompWidthSupp]{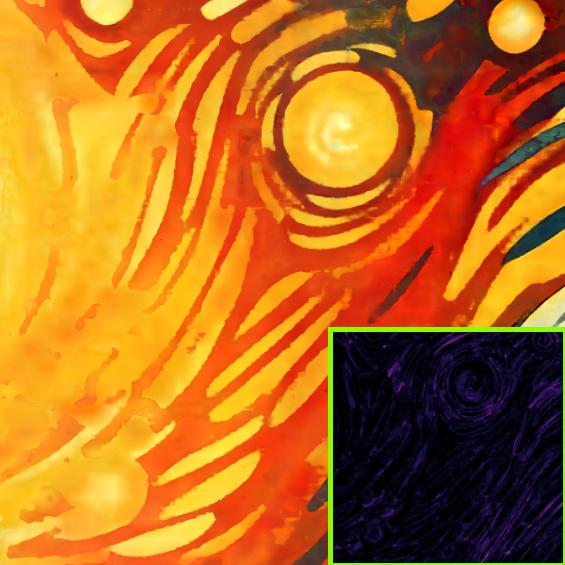}
  } \hspace{\reduceWidth}
\subfloat{
    \includegraphics[width=\imageCompWidthSupp,height=\imageCompWidthSupp]{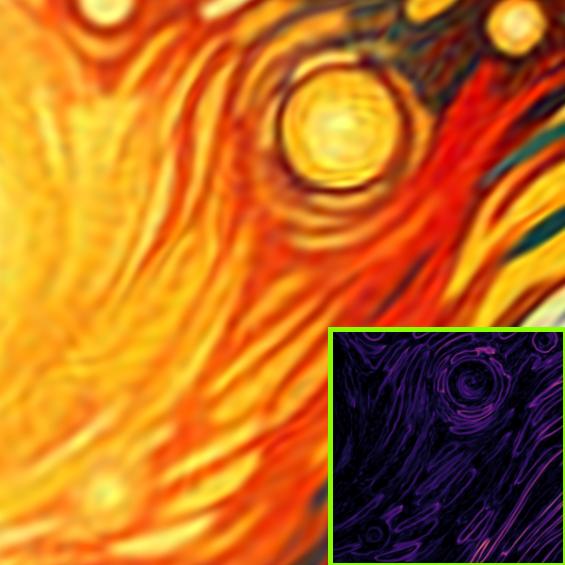}
  } \hspace{\reduceWidth}
\subfloat{
    \includegraphics[width=\imageCompWidthSupp,height=\imageCompWidthSupp]{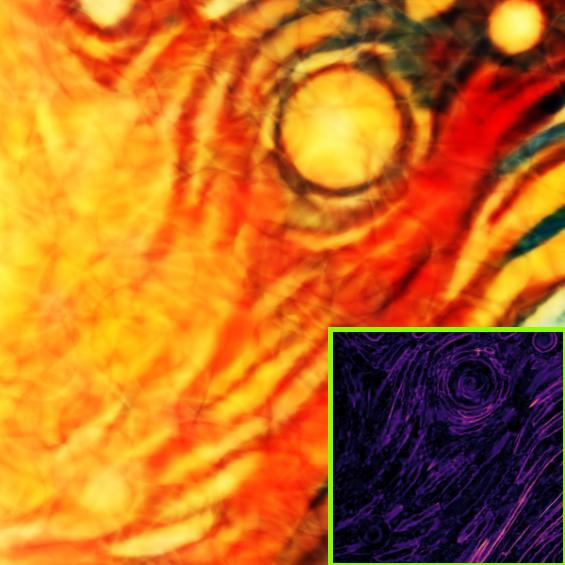}
  } \hspace{\reduceWidth}
\subfloat{
    \includegraphics[width=\imageCompWidthSupp,height=\imageCompWidthSupp]{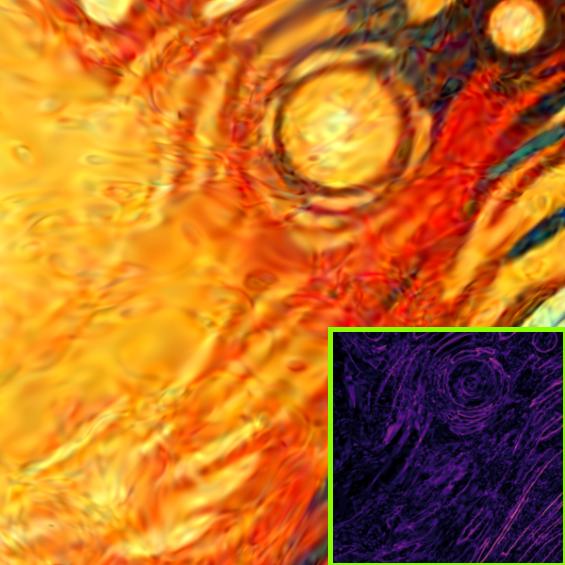}
  } \hspace{\reduceWidth}
\subfloat{
    \includegraphics[width=\imageCompWidthSupp,height=\imageCompWidthSupp]{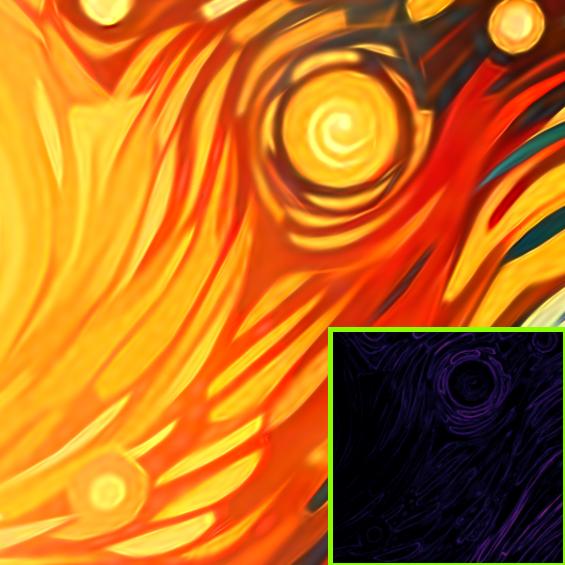}
  } \hspace{\reduceWidth}
\subfloat{
    \includegraphics[width=\imageCompWidthSupp,height=\imageCompWidthSupp]{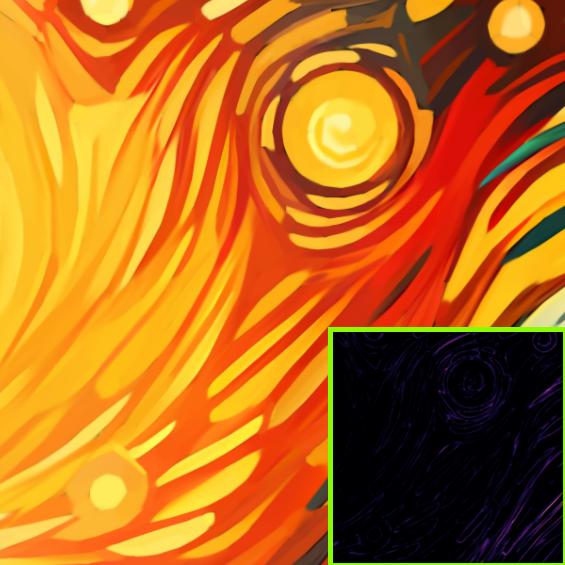}
  } \hspace{\reduceWidth}
\subfloat{
    \includegraphics[width=\imageCompWidthSupp,height=\imageCompWidthSupp]{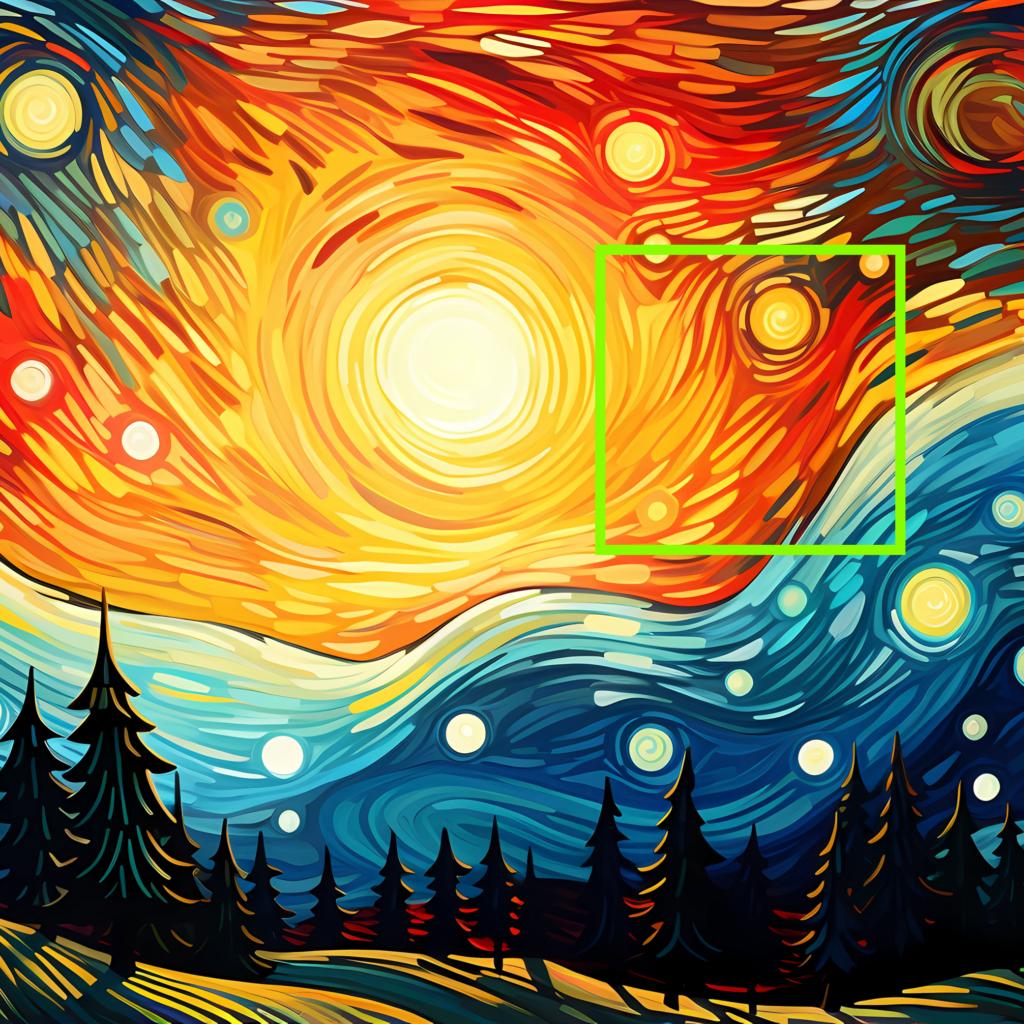}
  }
\vspace{0.2mm} \\
\setcounter{subfigure}{0}
\subfloat[ReLU-F]{
    \includegraphics[width=\imageCompWidthSupp,height=\imageCompWidthSupp]{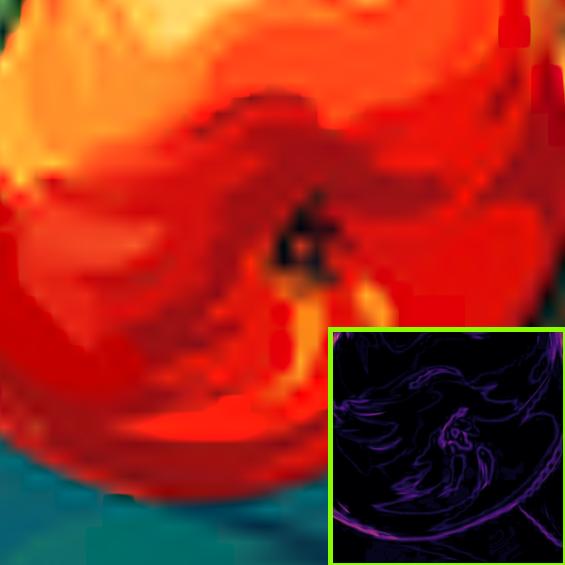}
  } \hspace{\reduceWidth}
\subfloat[I-NGP]{
    \includegraphics[width=\imageCompWidthSupp,height=\imageCompWidthSupp]{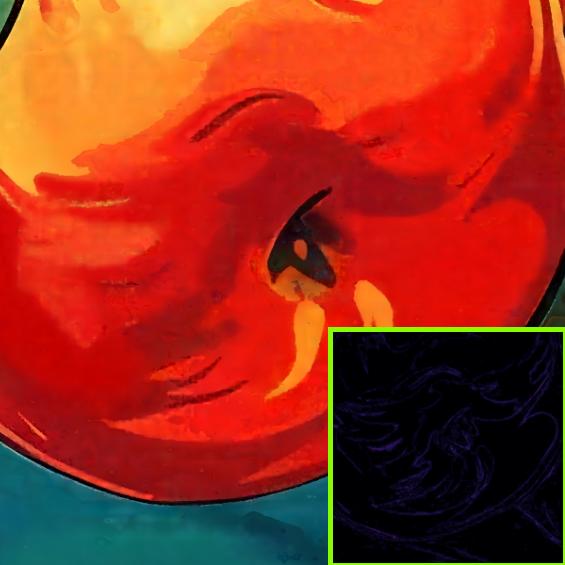}
  } \hspace{\reduceWidth}
\subfloat[SIREN]{
    \includegraphics[width=\imageCompWidthSupp,height=\imageCompWidthSupp]{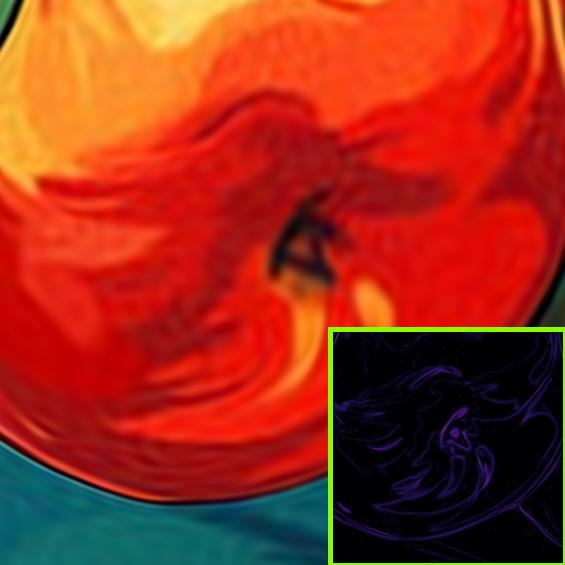}
  } \hspace{\reduceWidth}
\subfloat[FFN]{
    \includegraphics[width=\imageCompWidthSupp,height=\imageCompWidthSupp]{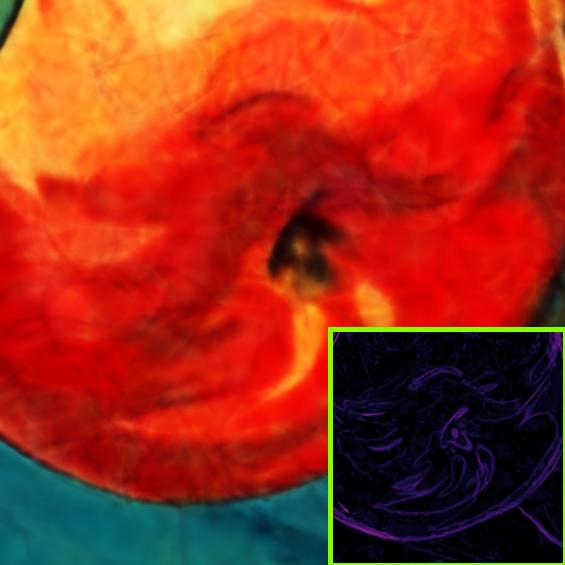}
  } \hspace{\reduceWidth}
\subfloat[WIRE]{
    \includegraphics[width=\imageCompWidthSupp,height=\imageCompWidthSupp]{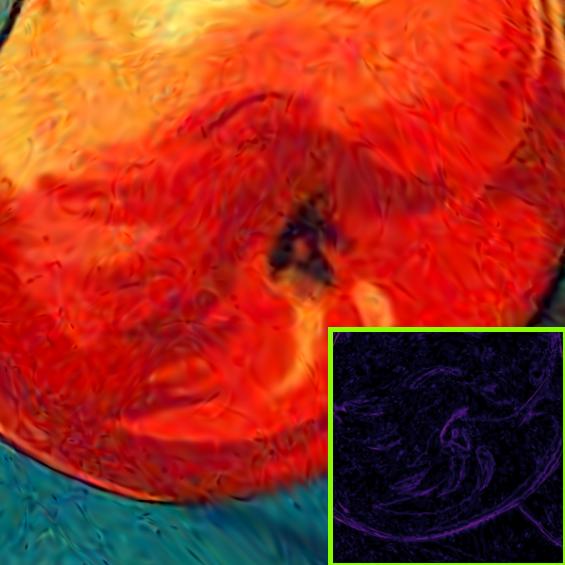}
  } \hspace{\reduceWidth}
\subfloat[GI]{
    \includegraphics[width=\imageCompWidthSupp,height=\imageCompWidthSupp]{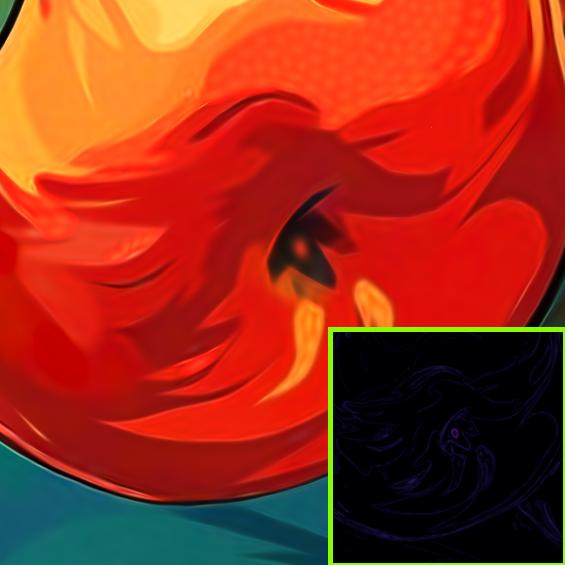}
  } \hspace{\reduceWidth}
\subfloat[Ours]{
    \includegraphics[width=\imageCompWidthSupp,height=\imageCompWidthSupp]{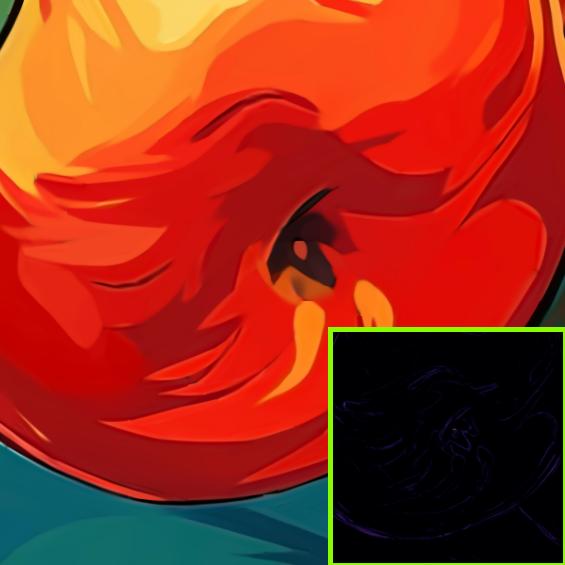}
  } \hspace{\reduceWidth}
\subfloat[Reference]{
    \includegraphics[width=\imageCompWidthSupp,height=\imageCompWidthSupp]{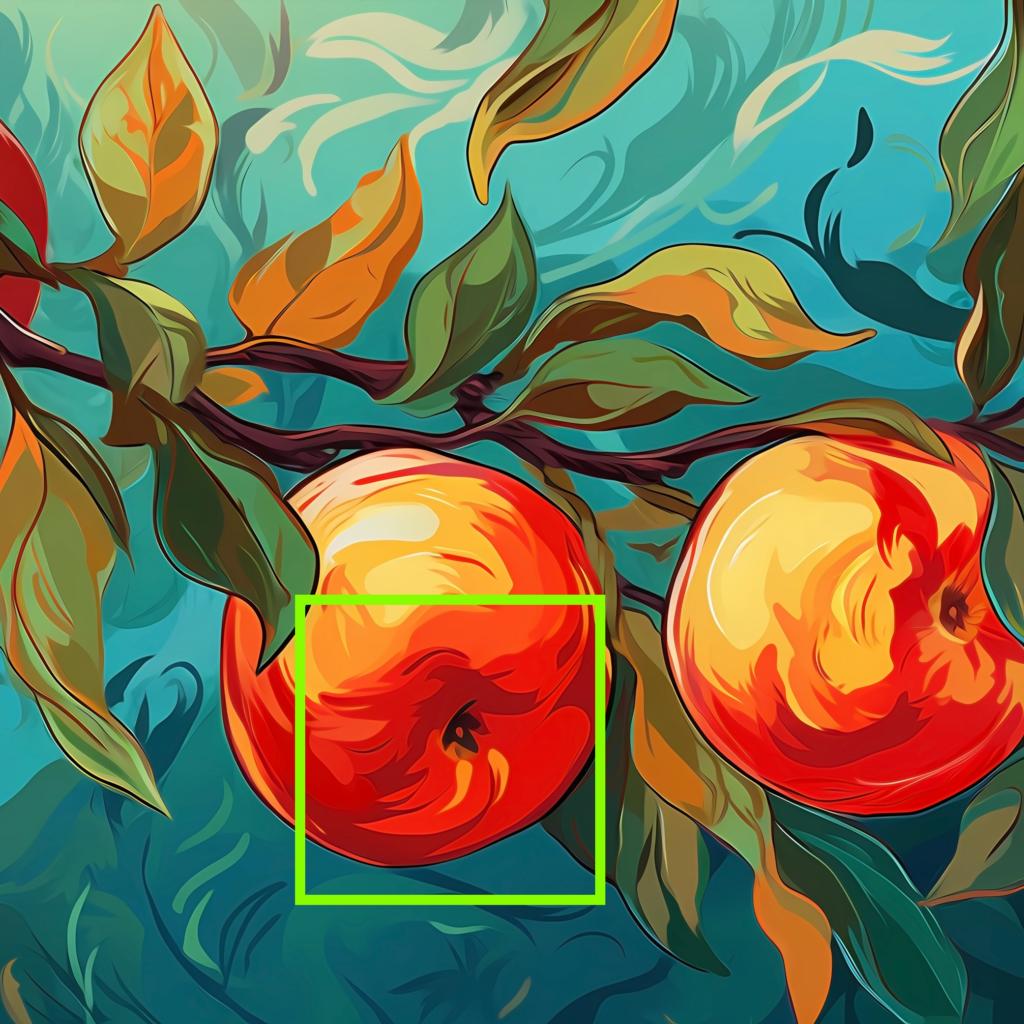}
  }
\Caption{\revise{Qualitative comparison against conventional and neural image representations (\Cref{sec:evaluation-image}).}}{\revise{For the 2K$\times$2K-resolution results shown here, the model sizes (in KB) of ReLU-F, I-NGP, SIREN, FFN, WIRE, GI, and \methodName are 164, 166, 161, 154, 159, 164, and 160, respectively.}}
\end{figure*}
\clearpage
\newcommand{\lodCompWidthSupp}{0.139\linewidth}
\section{\revise{Additional Rate-Distortion Trade-off Results}}
\label{fig:evaluation-lod-supp}
\begin{figure*}[h]
\centering
\subfloat{
    \includegraphics[width=\lodCompWidthSupp,height=\lodCompWidthSupp]{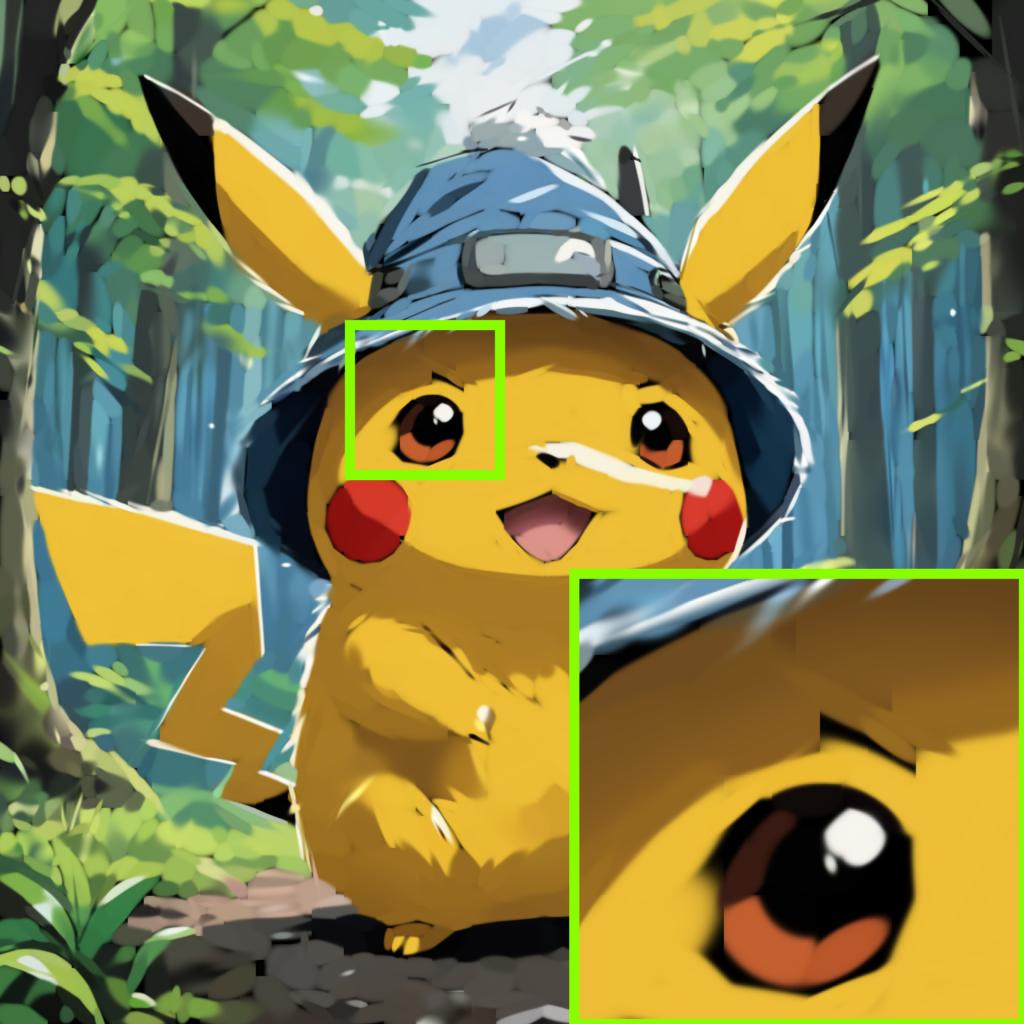}
  }
\subfloat{
    \includegraphics[width=\lodCompWidthSupp,height=\lodCompWidthSupp]{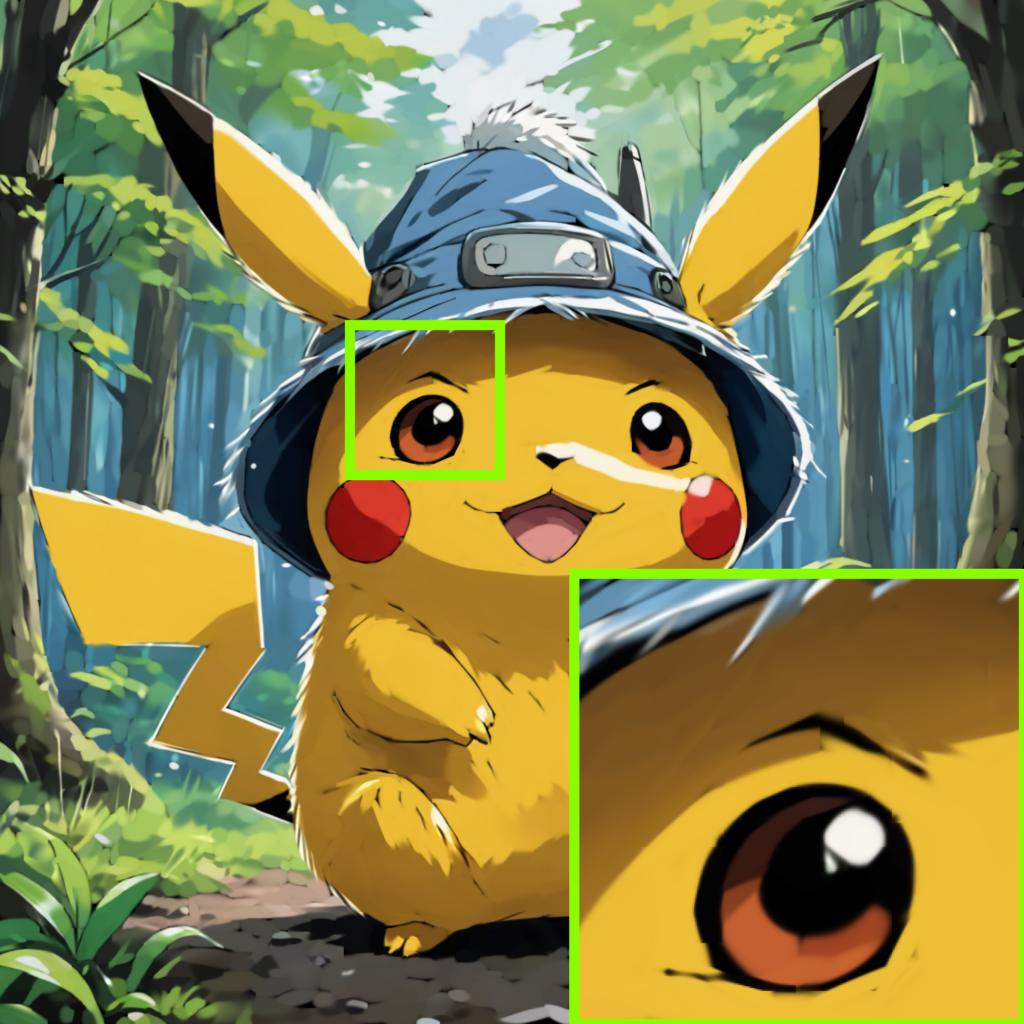}
  }
\subfloat{
    \includegraphics[width=\lodCompWidthSupp,height=\lodCompWidthSupp]{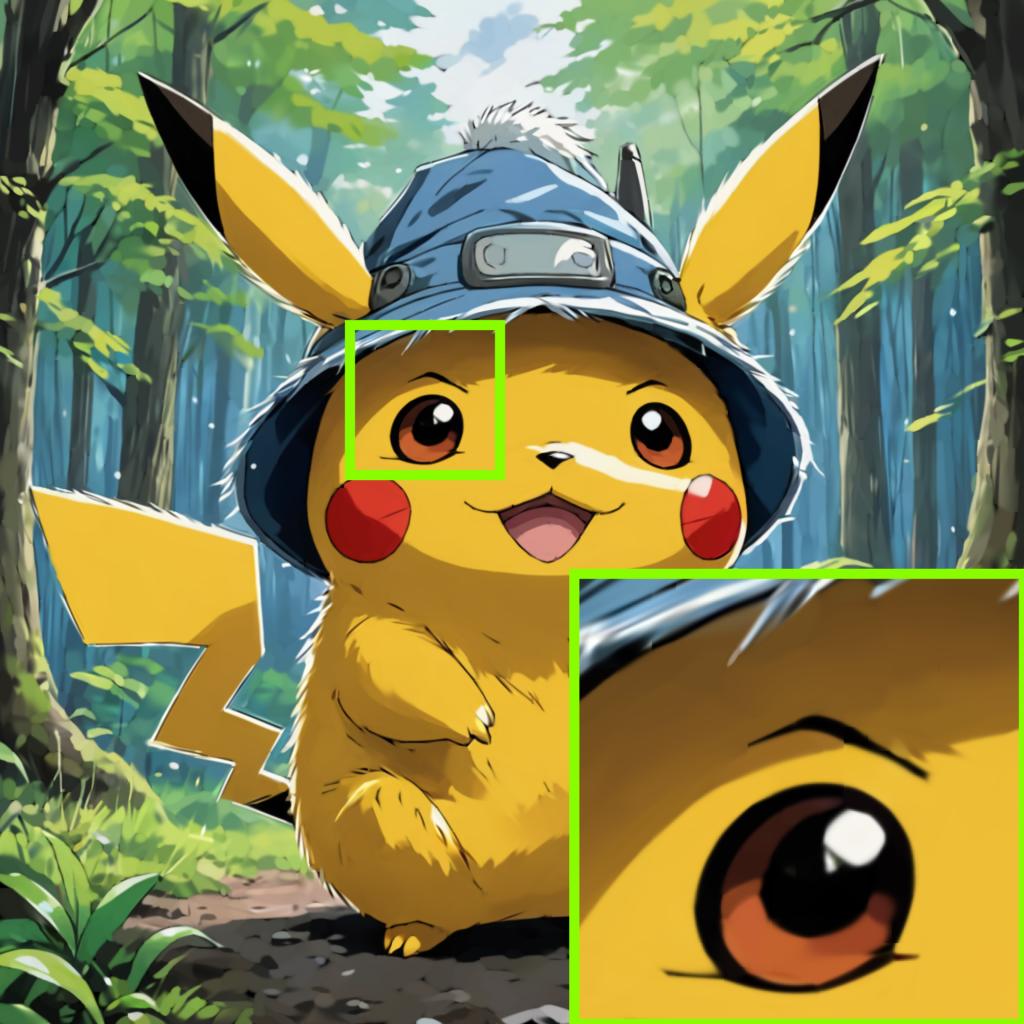}
  }
\subfloat{
    \includegraphics[width=\lodCompWidthSupp,height=\lodCompWidthSupp]{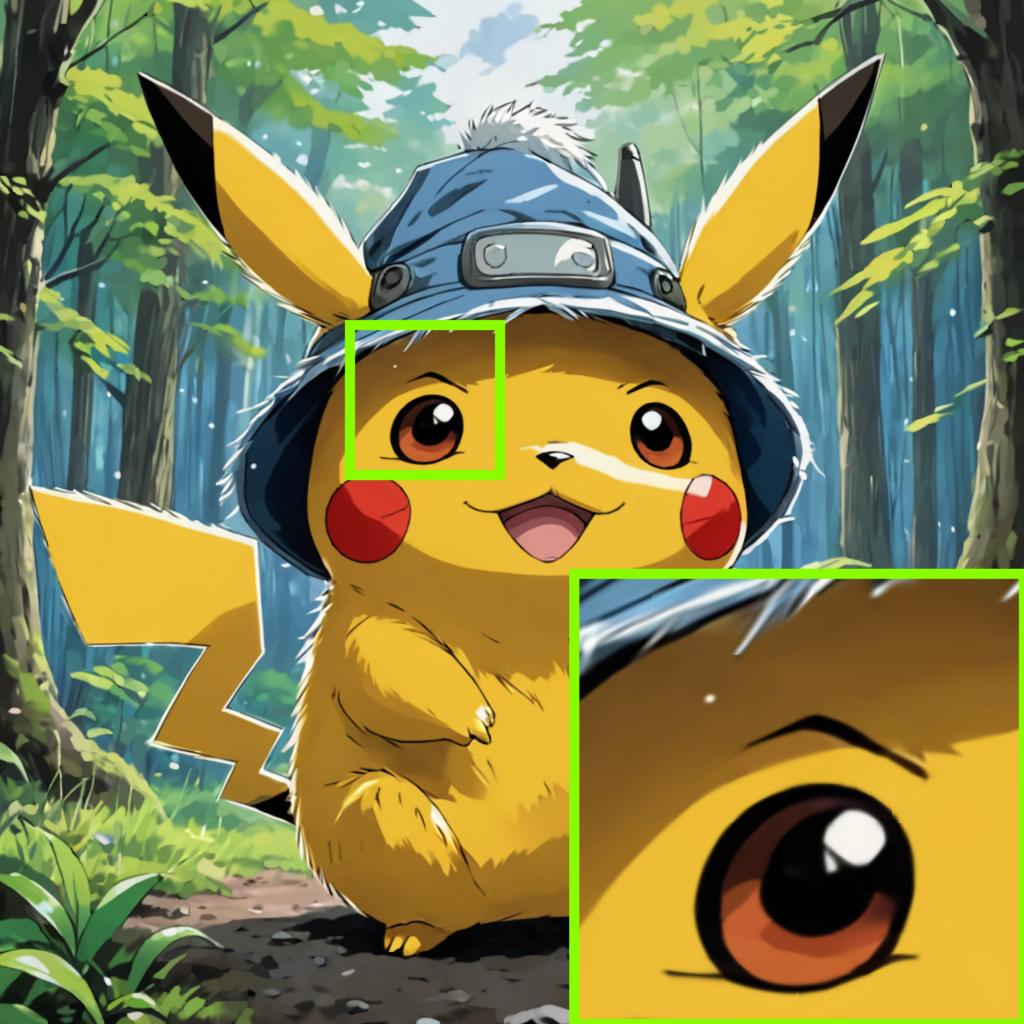}
  }
\subfloat{
    \includegraphics[width=\lodCompWidthSupp,height=\lodCompWidthSupp]{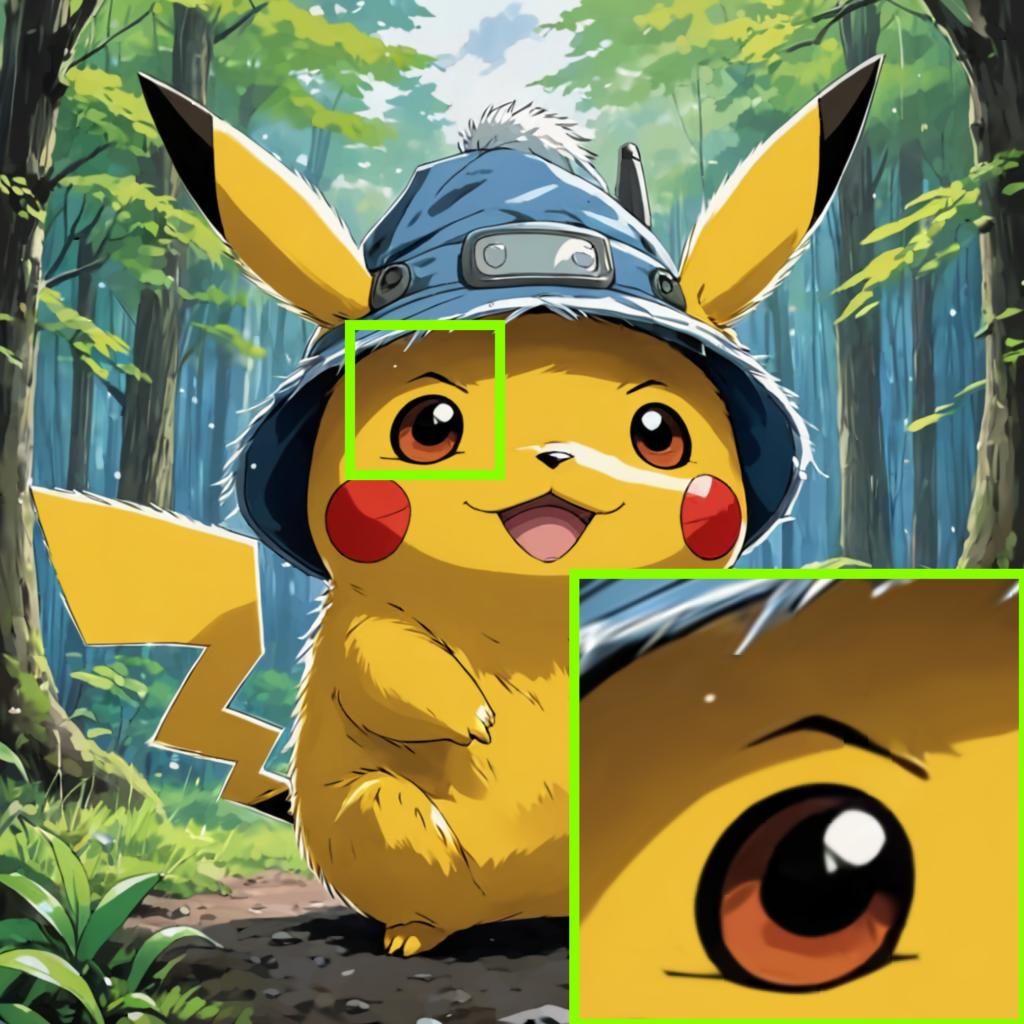}
  }
\subfloat{
    \includegraphics[width=\lodCompWidthSupp,height=\lodCompWidthSupp]{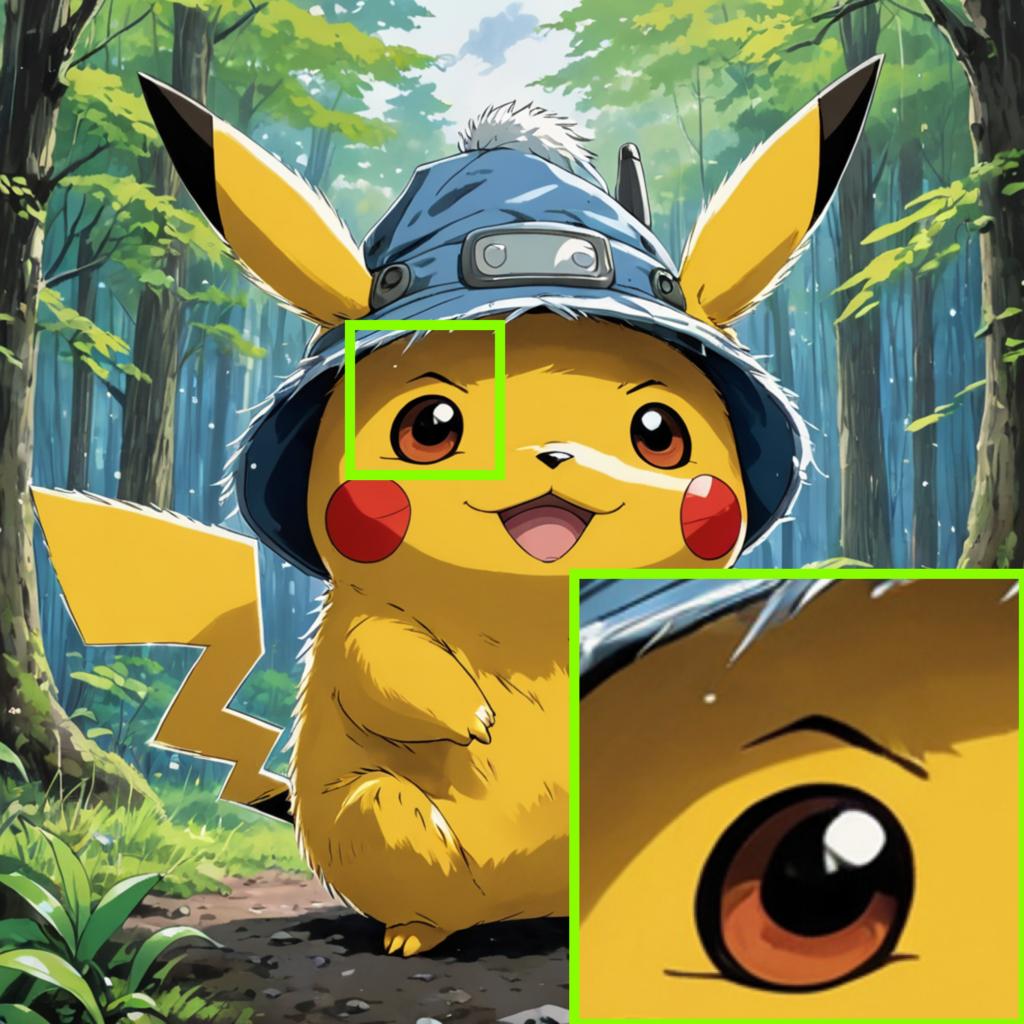}
  }
\vspace{2mm} \\
\subfloat{
    \includegraphics[width=\lodCompWidthSupp,height=\lodCompWidthSupp]{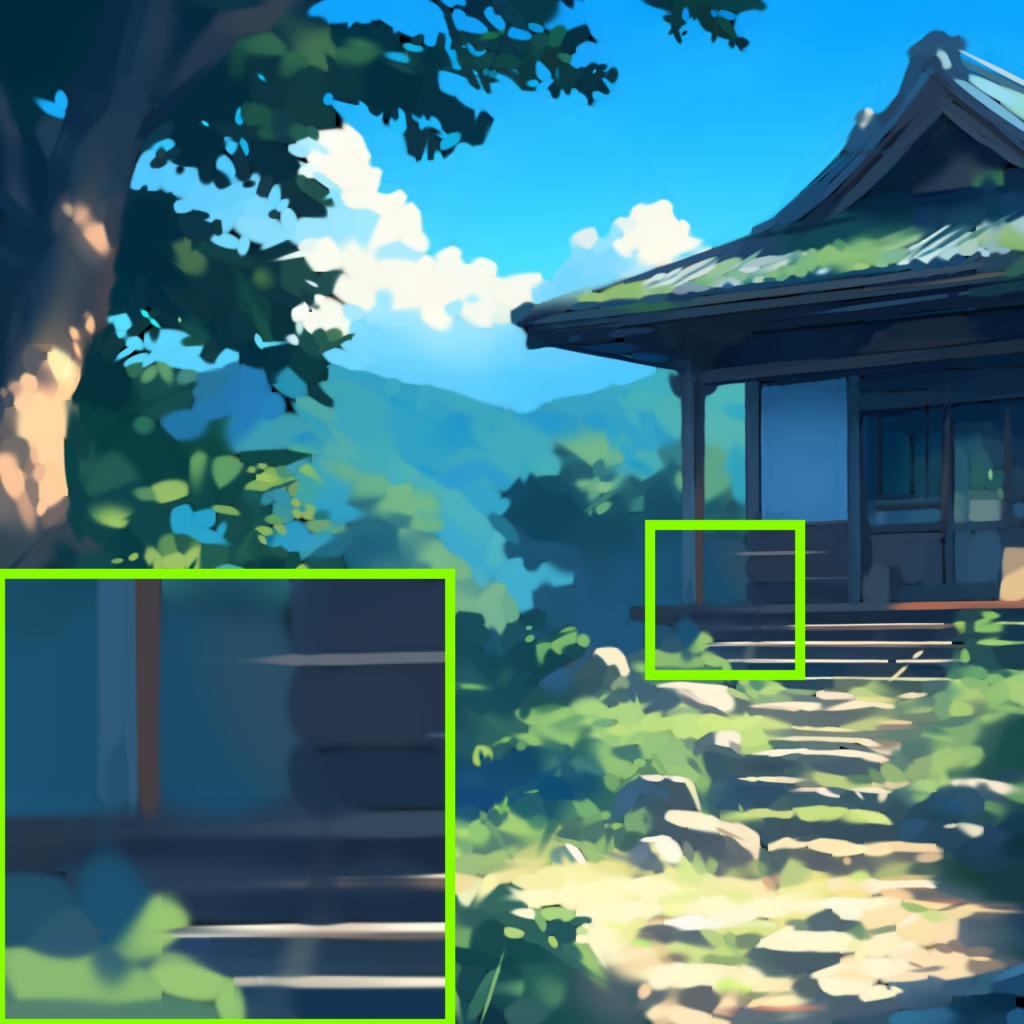}
  }
\subfloat{
    \includegraphics[width=\lodCompWidthSupp,height=\lodCompWidthSupp]{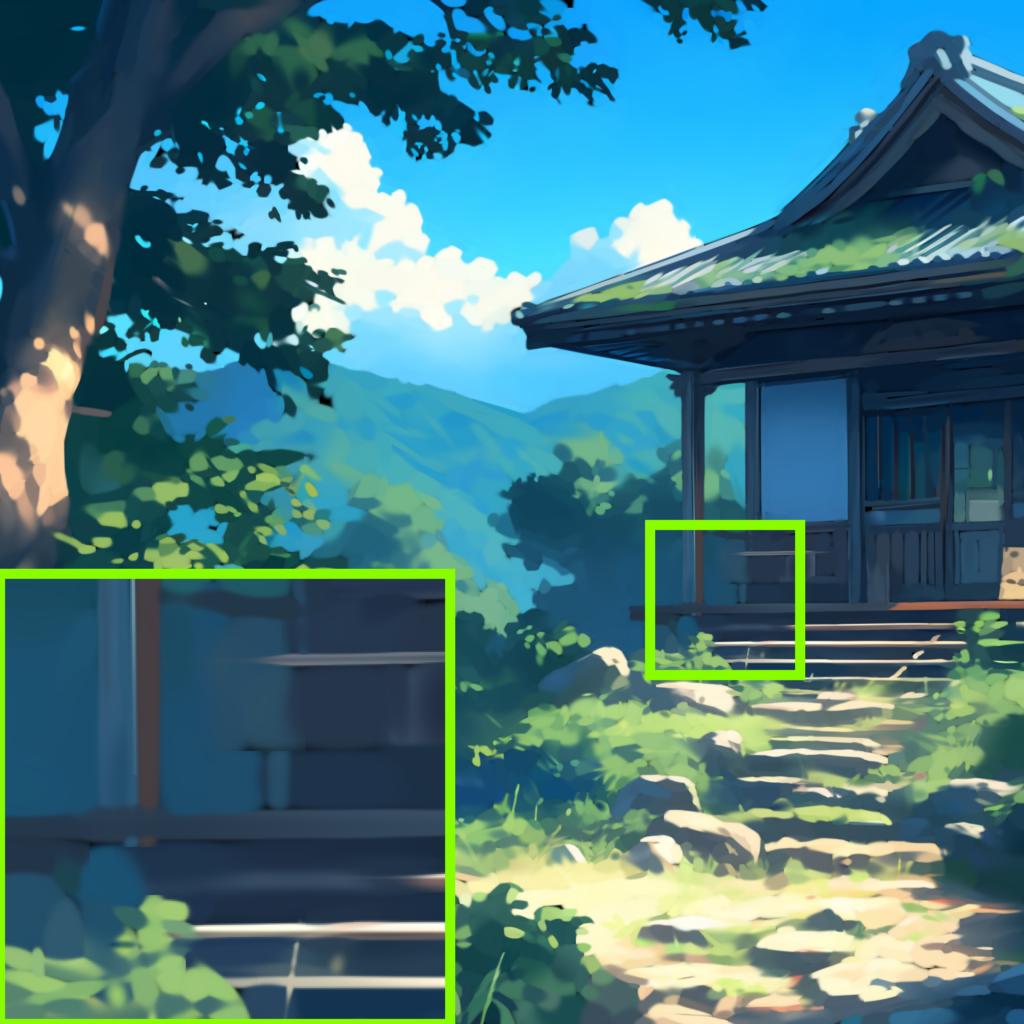}
  }
\subfloat{
    \includegraphics[width=\lodCompWidthSupp,height=\lodCompWidthSupp]{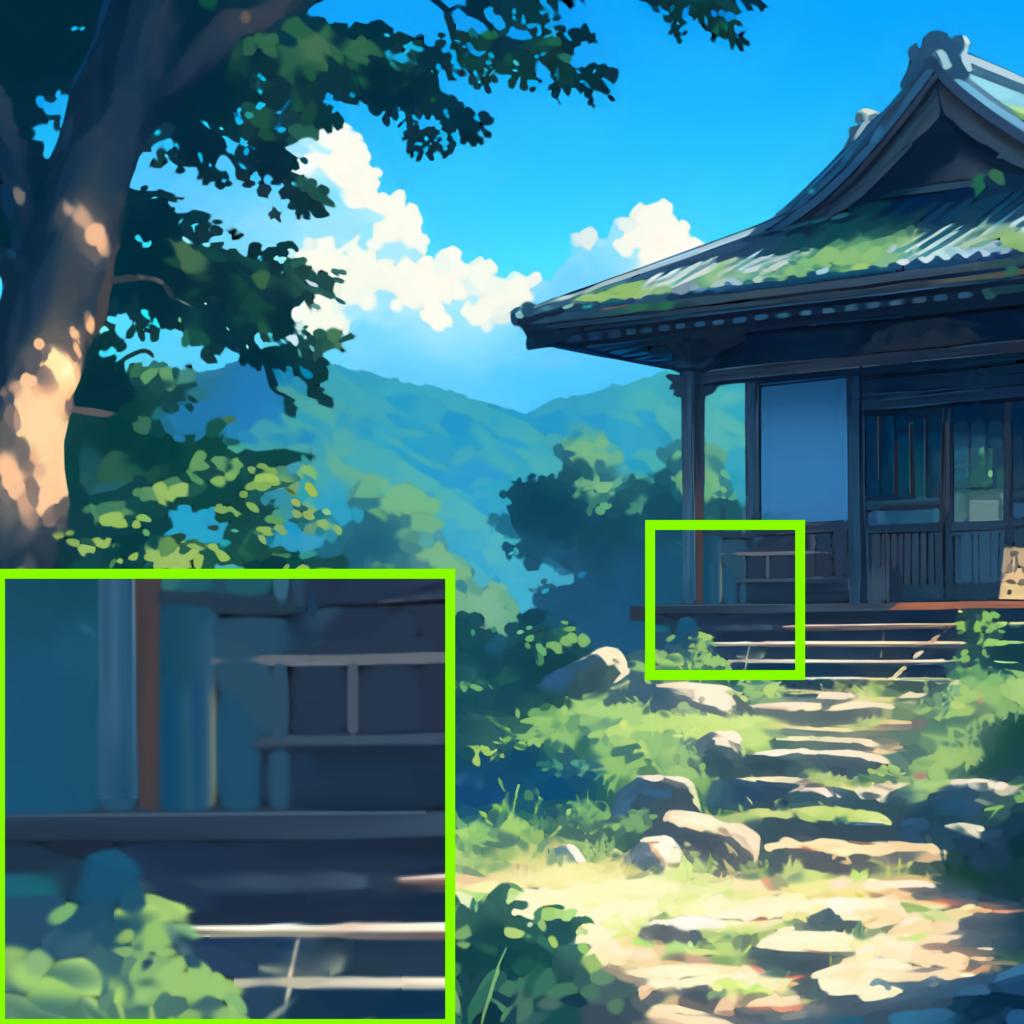}
  }
\subfloat{
    \includegraphics[width=\lodCompWidthSupp,height=\lodCompWidthSupp]{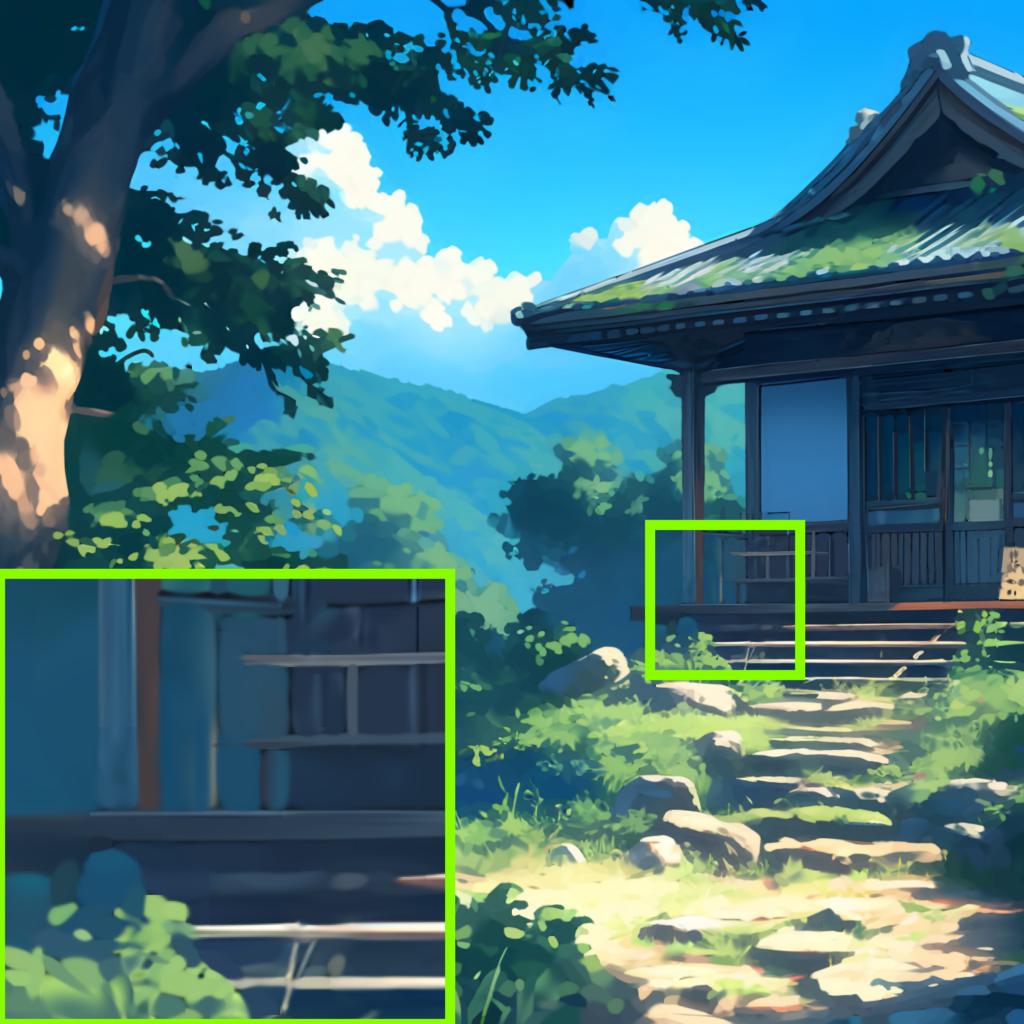}
  }
\subfloat{
    \includegraphics[width=\lodCompWidthSupp,height=\lodCompWidthSupp]{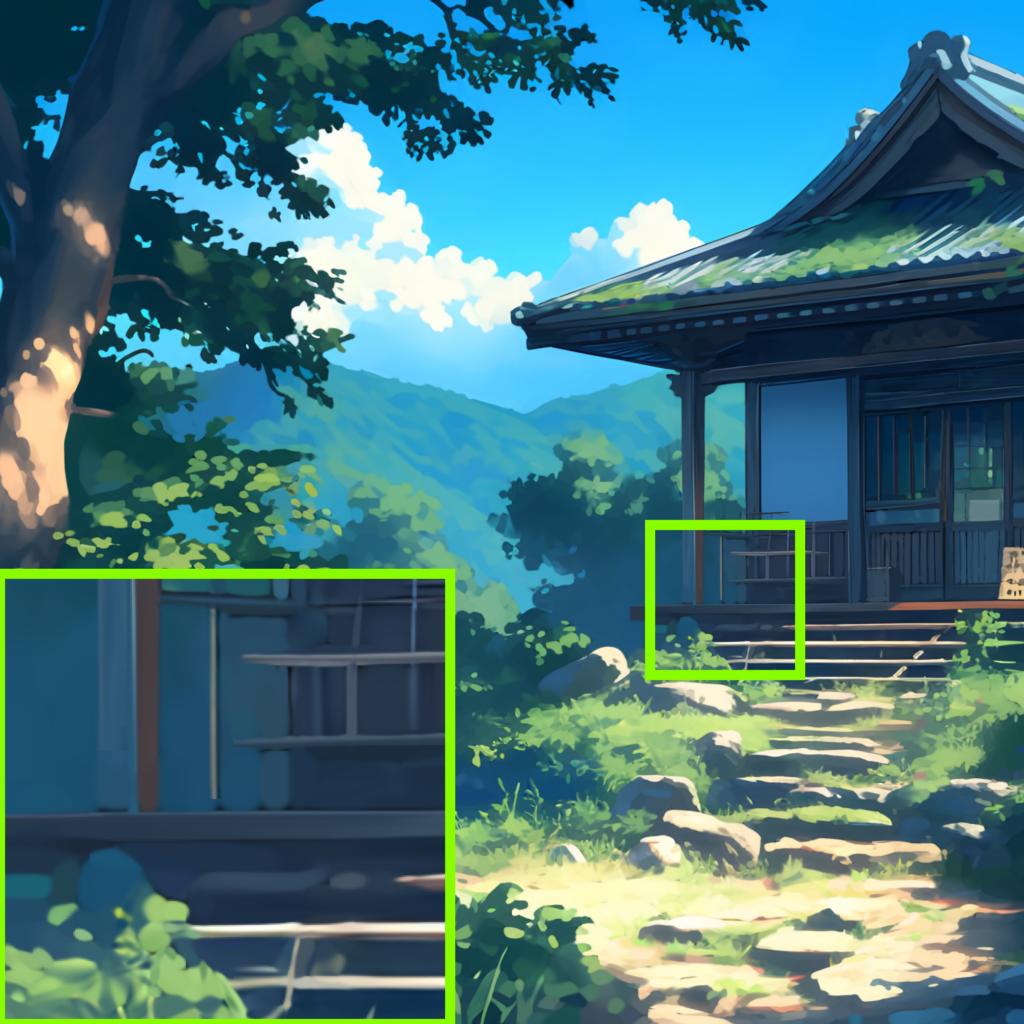}
  }
\subfloat{
    \includegraphics[width=\lodCompWidthSupp,height=\lodCompWidthSupp]{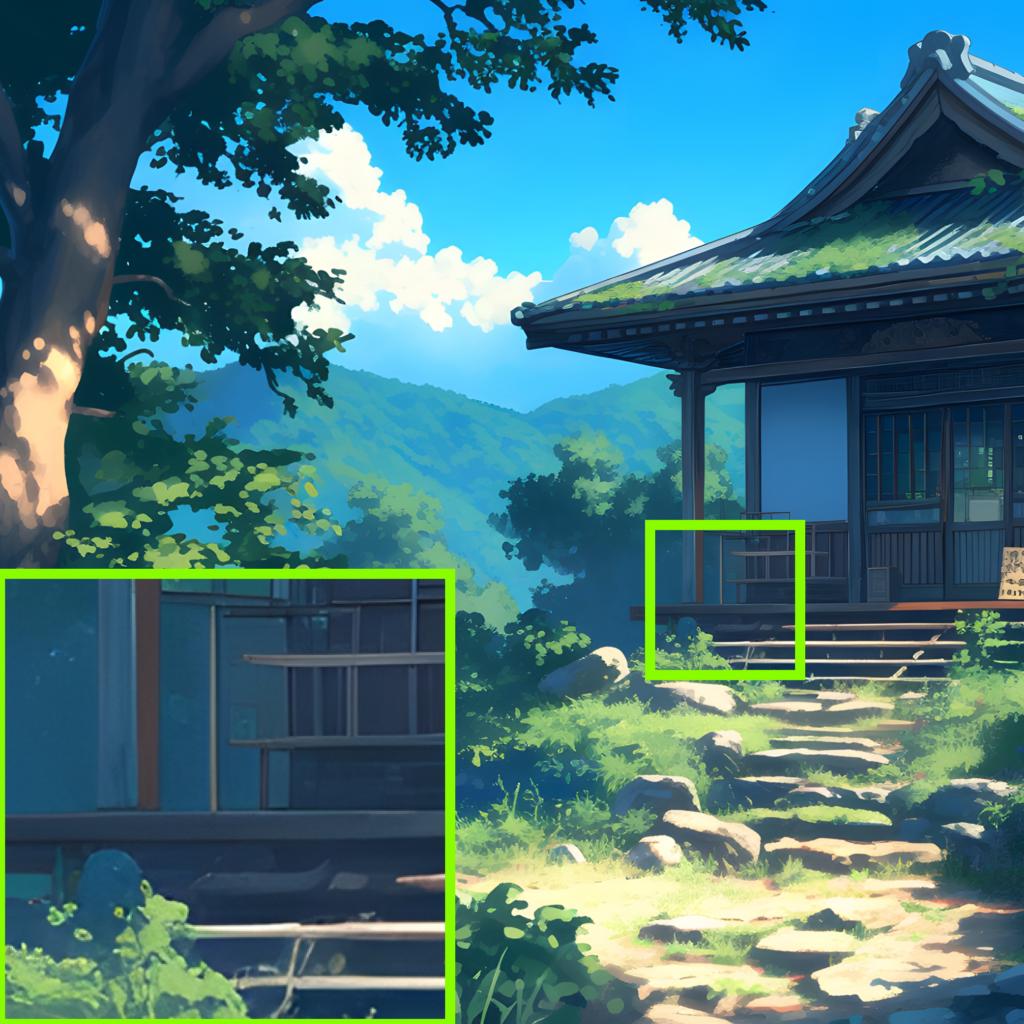}
  }
\vspace{2mm} \\
\subfloat{
    \includegraphics[width=\lodCompWidthSupp,height=\lodCompWidthSupp]{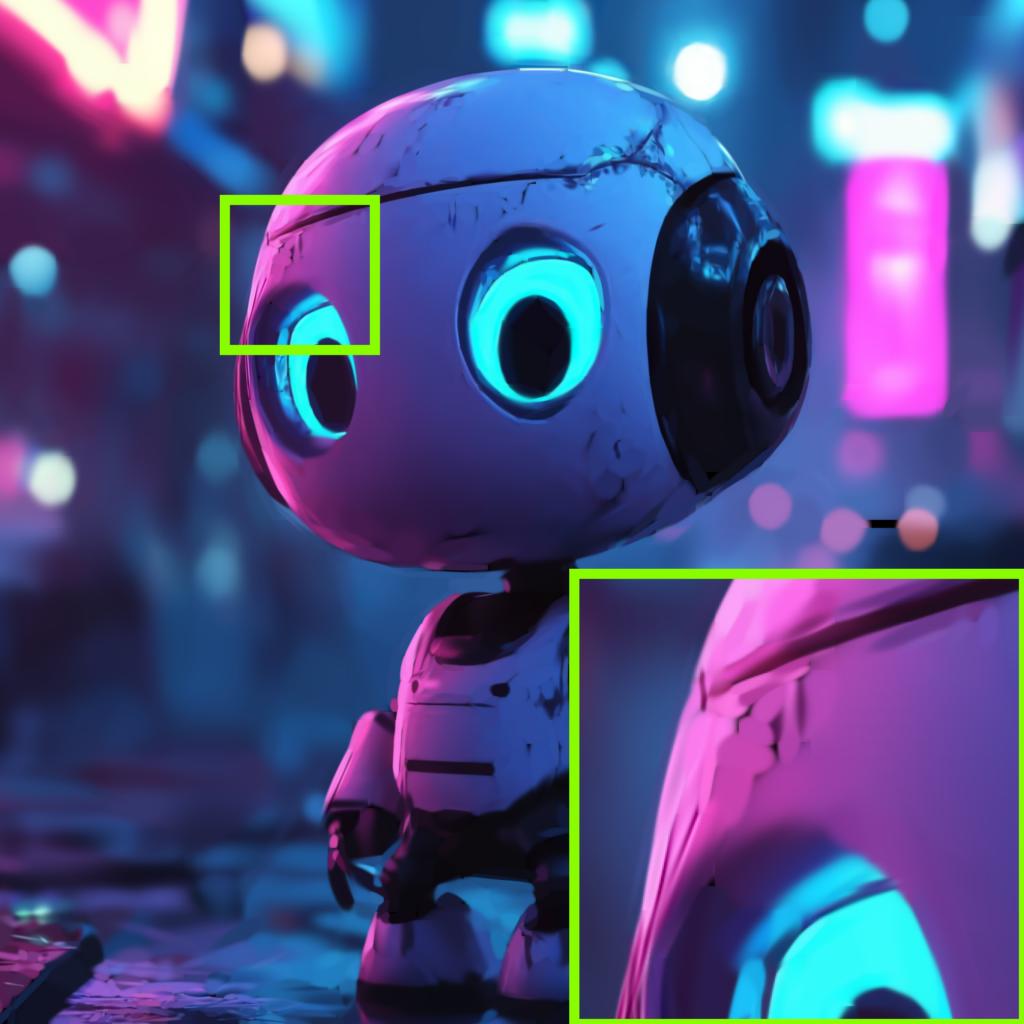}
  }
\subfloat{
    \includegraphics[width=\lodCompWidthSupp,height=\lodCompWidthSupp]{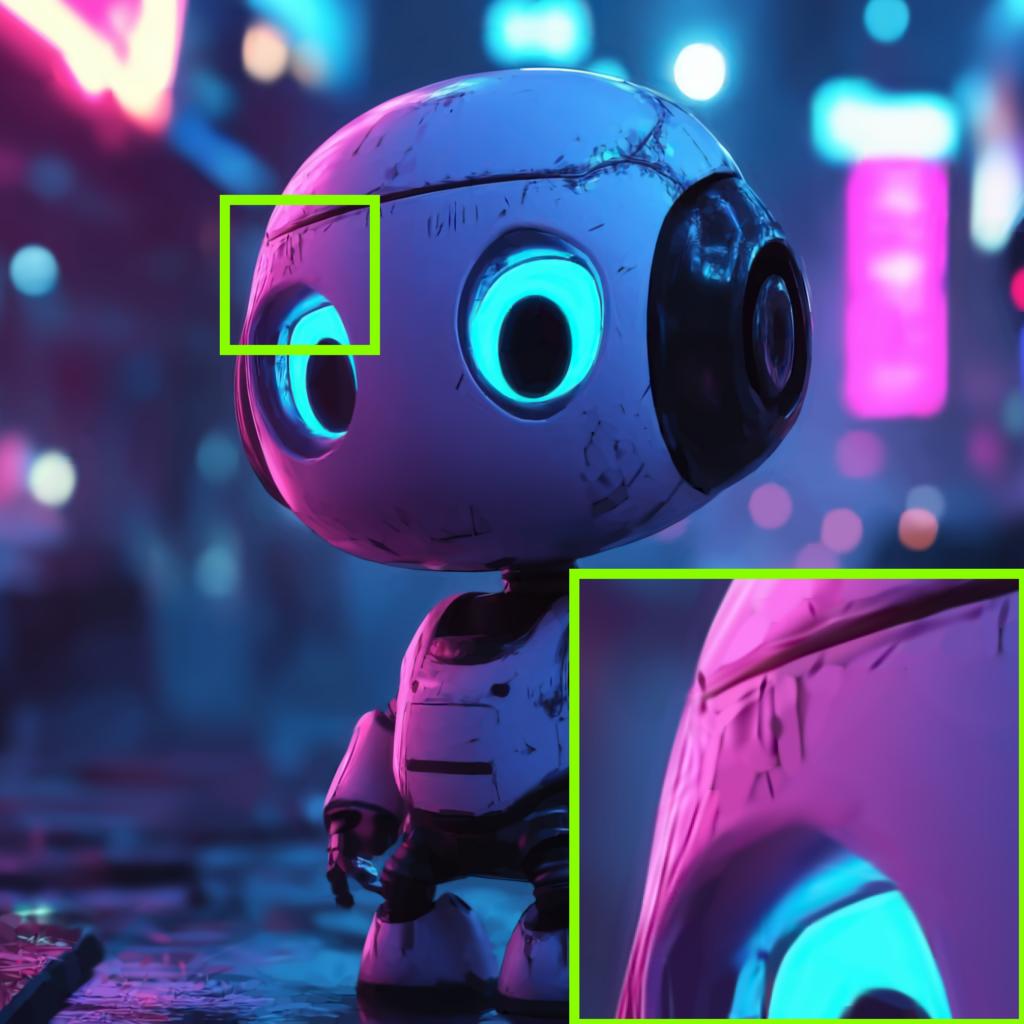}
  }
\subfloat{
    \includegraphics[width=\lodCompWidthSupp,height=\lodCompWidthSupp]{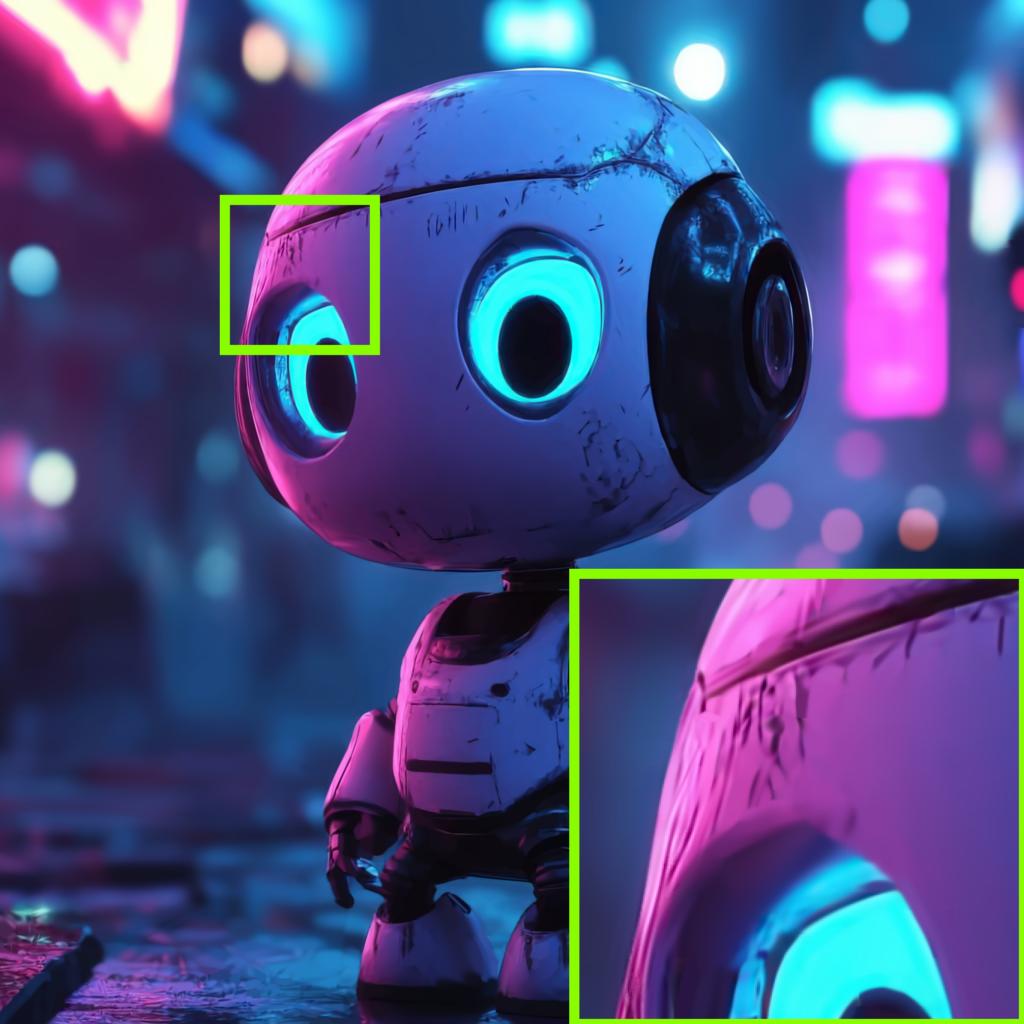}
  }
\subfloat{
    \includegraphics[width=\lodCompWidthSupp,height=\lodCompWidthSupp]{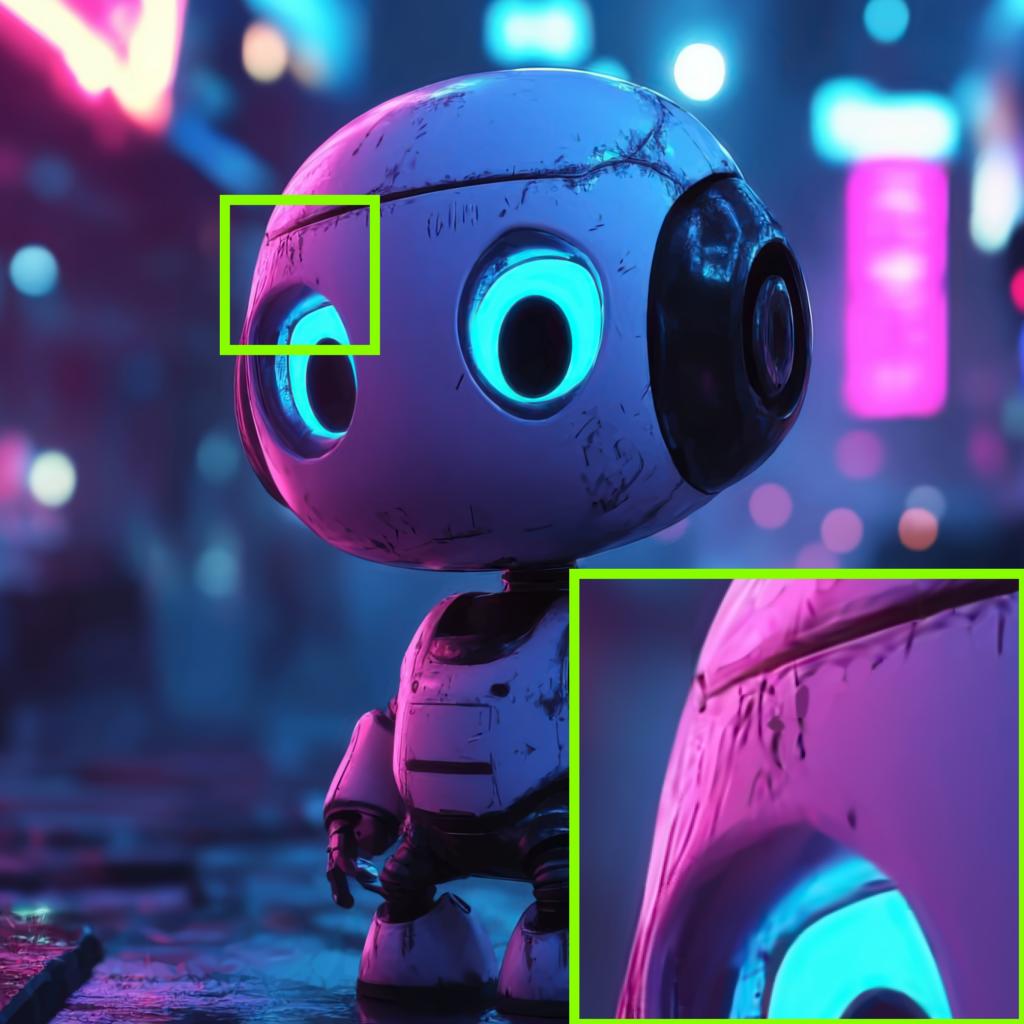}
  }
\subfloat{
    \includegraphics[width=\lodCompWidthSupp,height=\lodCompWidthSupp]{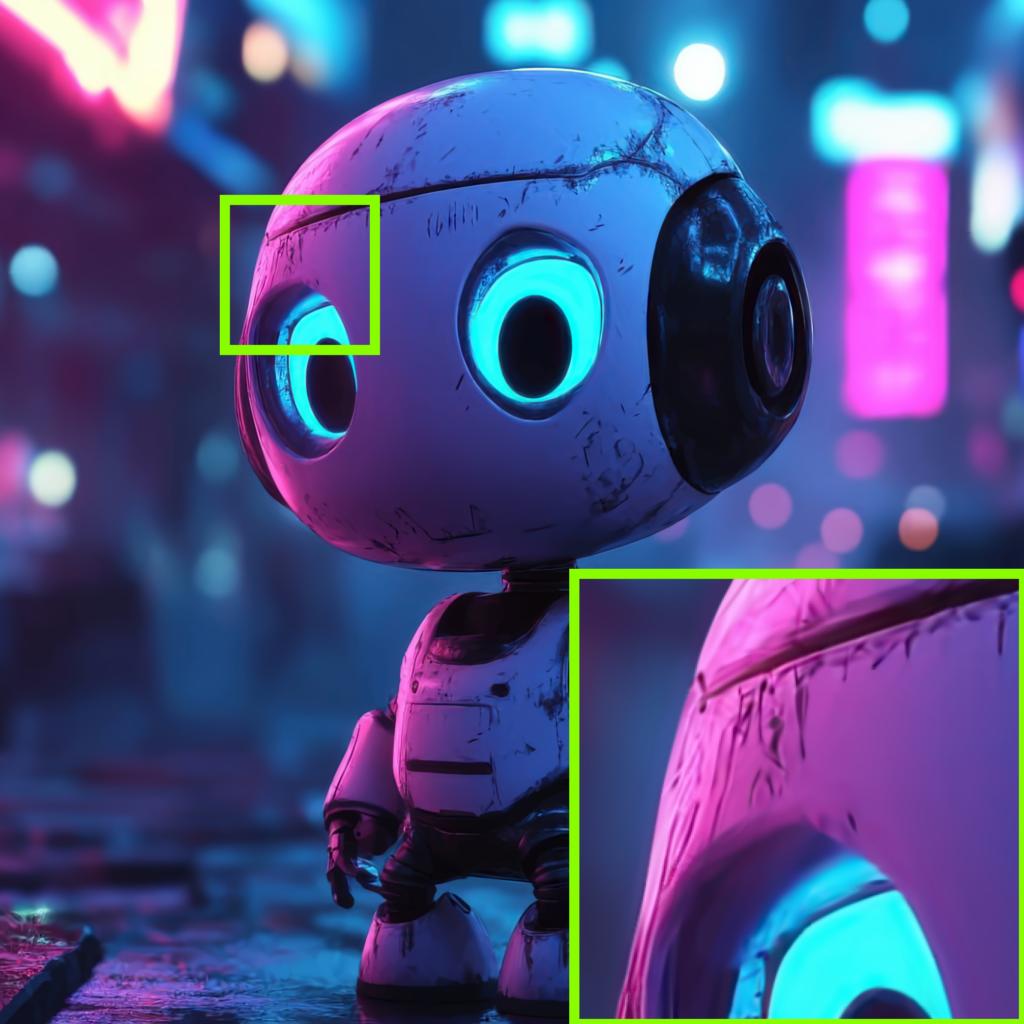}
  }
\subfloat{
    \includegraphics[width=\lodCompWidthSupp,height=\lodCompWidthSupp]{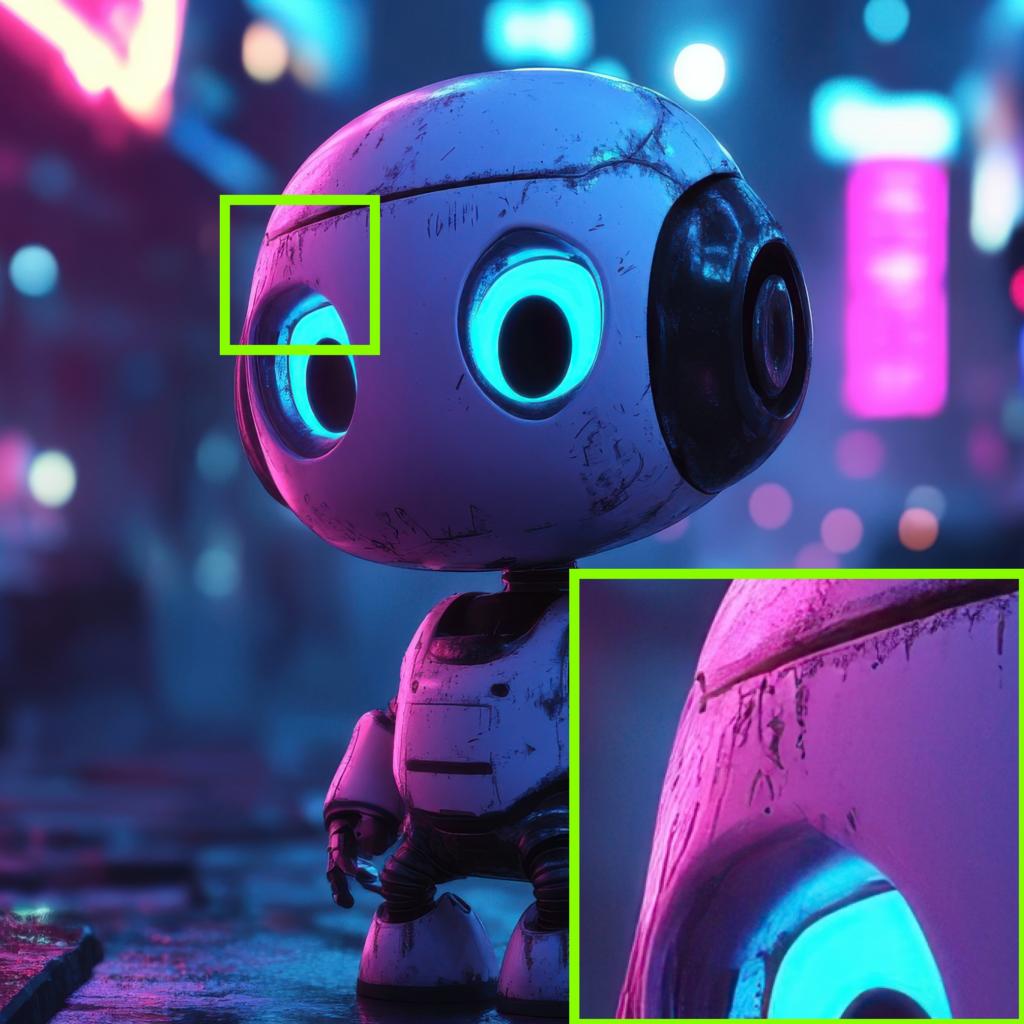}
  }
\vspace{2mm} \\
\subfloat{
    \includegraphics[width=\lodCompWidthSupp,height=\lodCompWidthSupp]{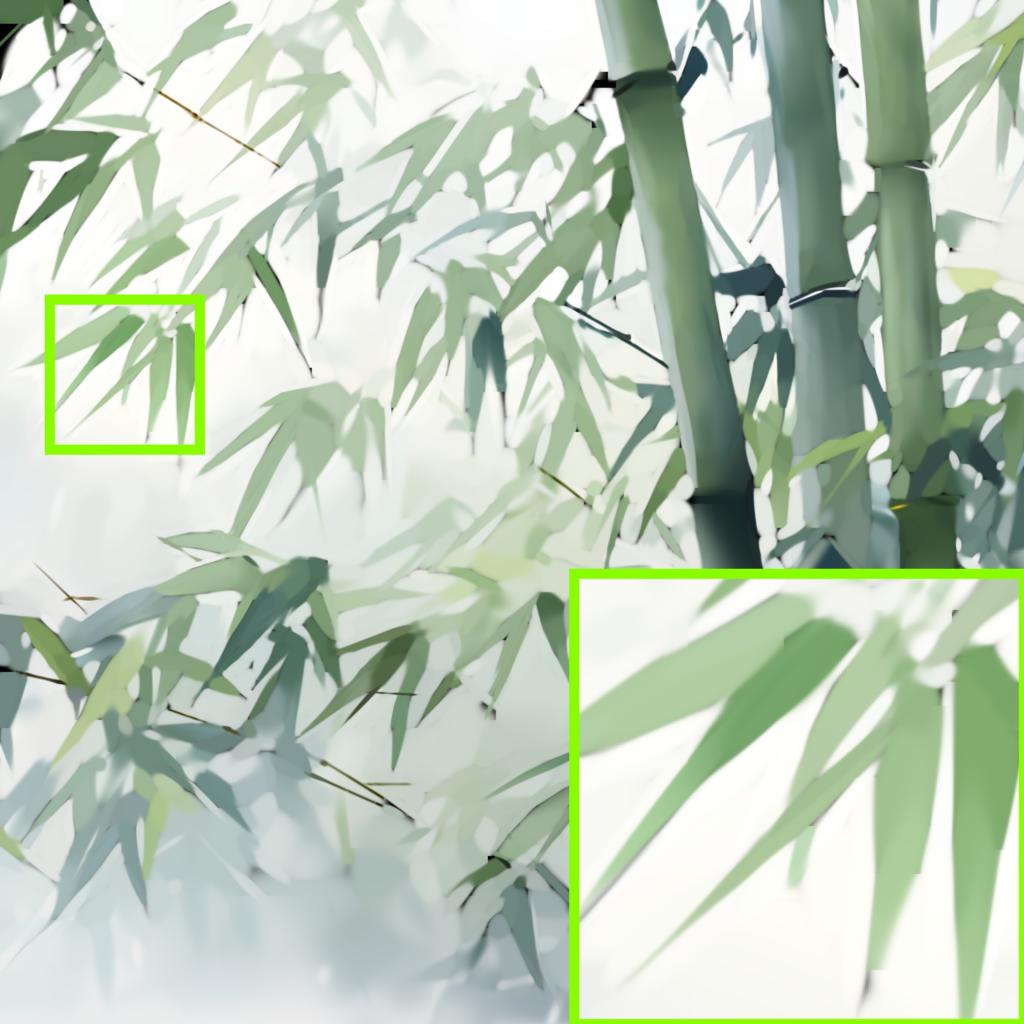}
  }
\subfloat{
    \includegraphics[width=\lodCompWidthSupp,height=\lodCompWidthSupp]{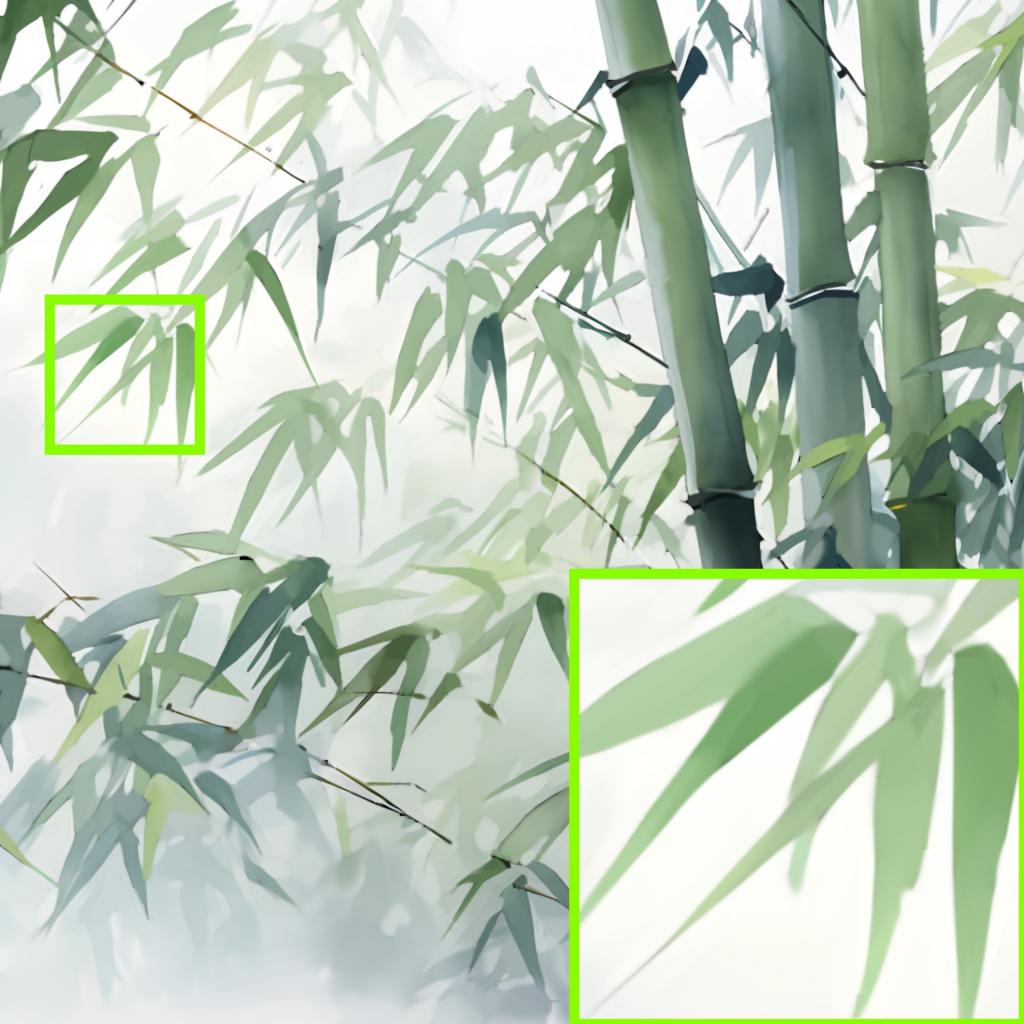}
  }
\subfloat{
    \includegraphics[width=\lodCompWidthSupp,height=\lodCompWidthSupp]{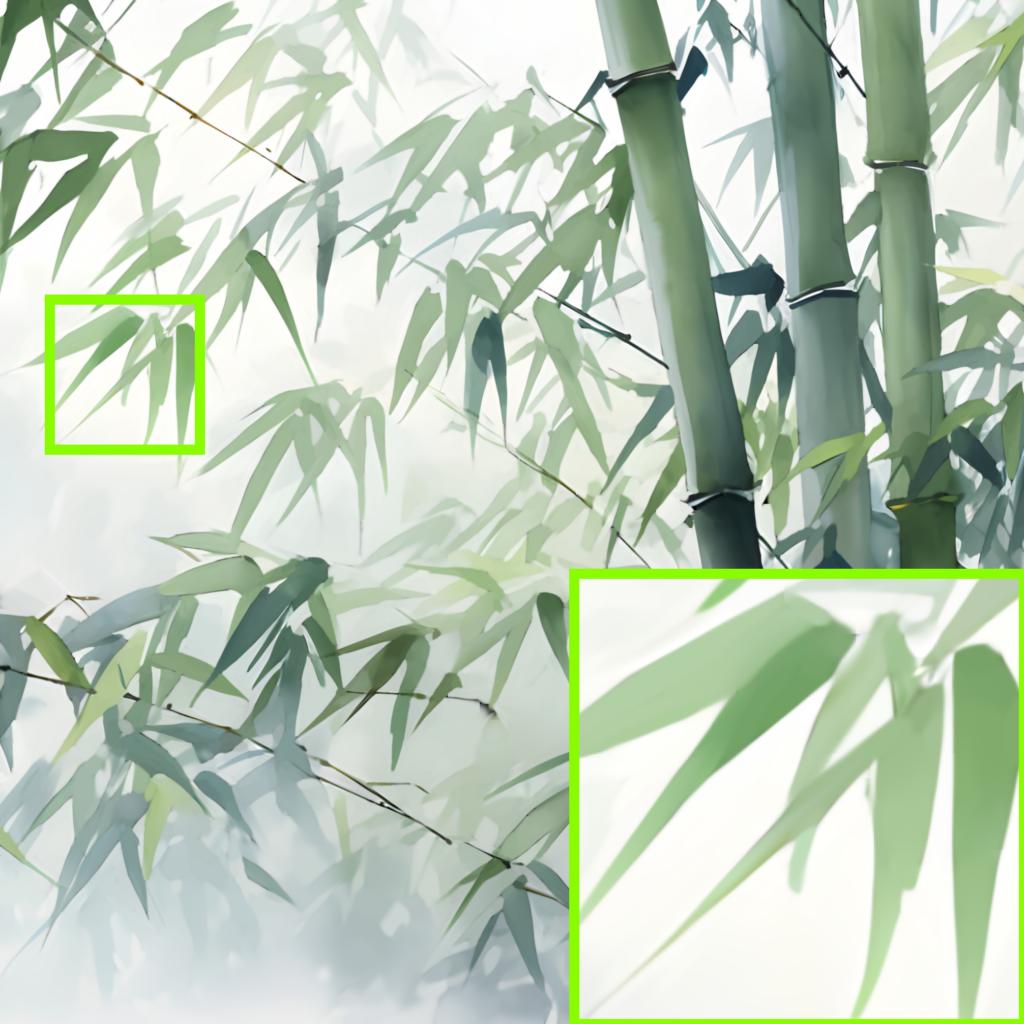}
  }
\subfloat{
    \includegraphics[width=\lodCompWidthSupp,height=\lodCompWidthSupp]{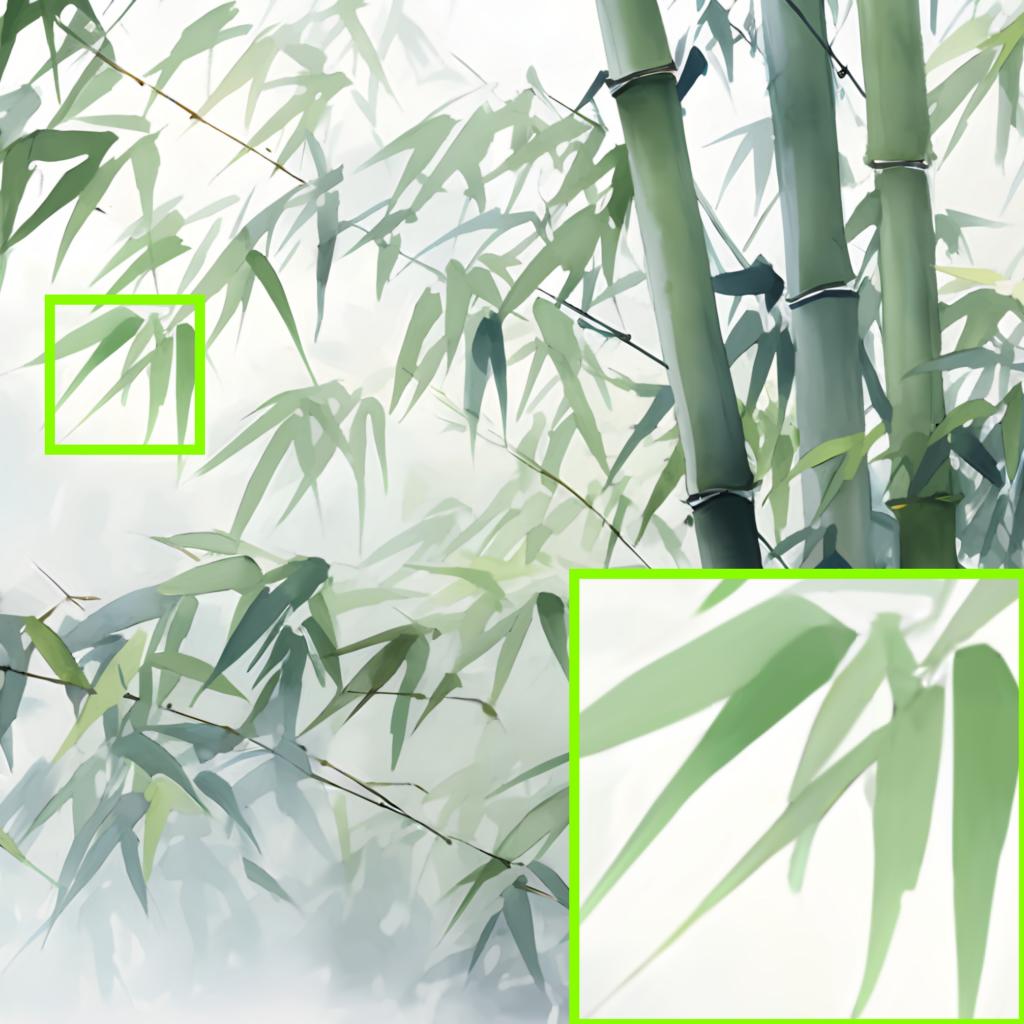}
  }
\subfloat{
    \includegraphics[width=\lodCompWidthSupp,height=\lodCompWidthSupp]{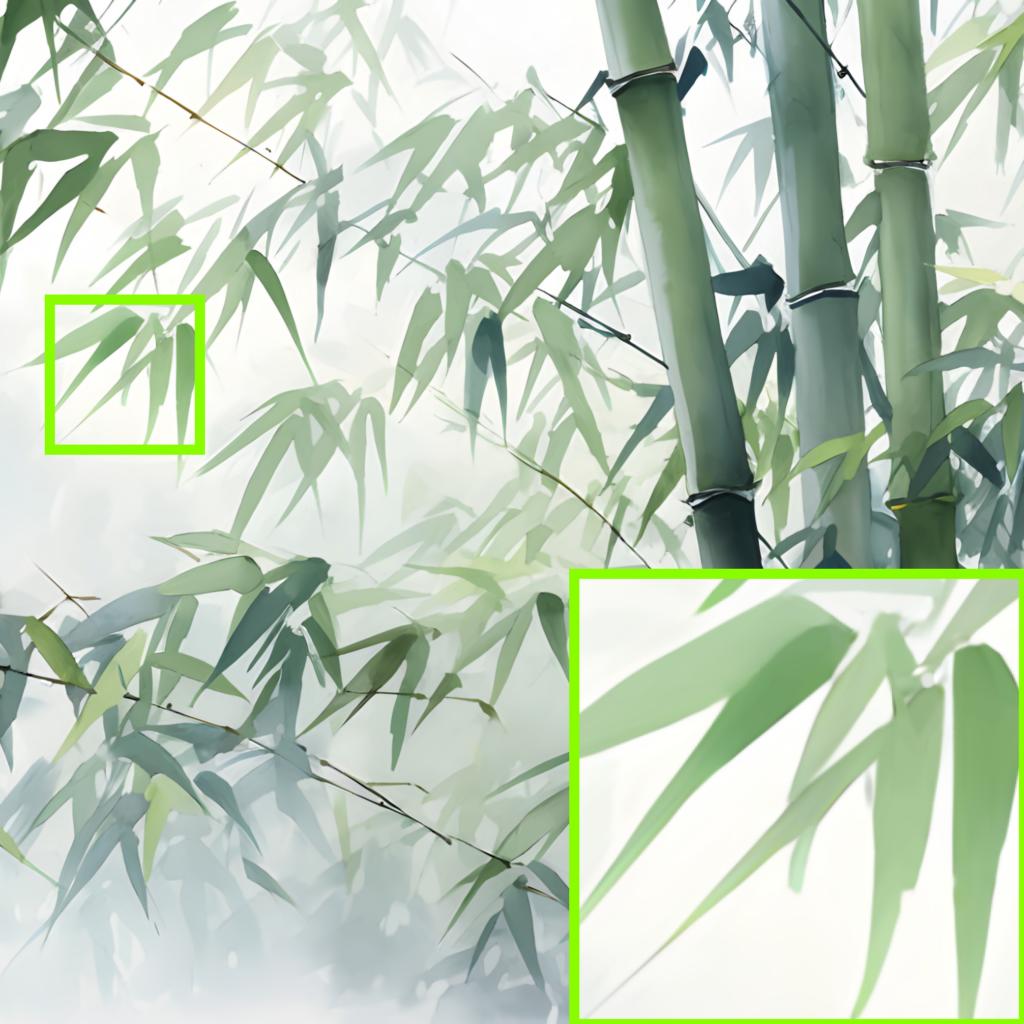}
  }
\subfloat{
    \includegraphics[width=\lodCompWidthSupp,height=\lodCompWidthSupp]{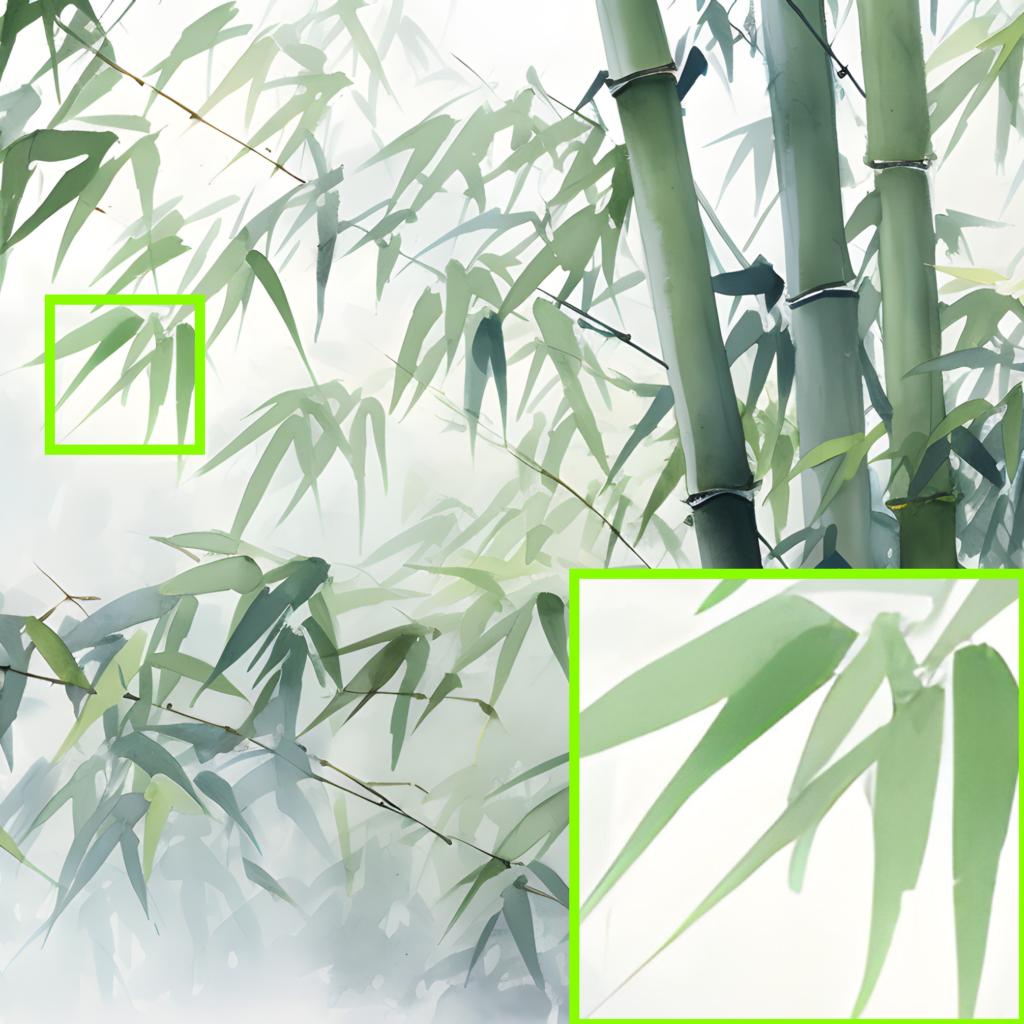}
  }
\vspace{2mm} \\
\subfloat{
    \includegraphics[width=\lodCompWidthSupp,height=\lodCompWidthSupp]{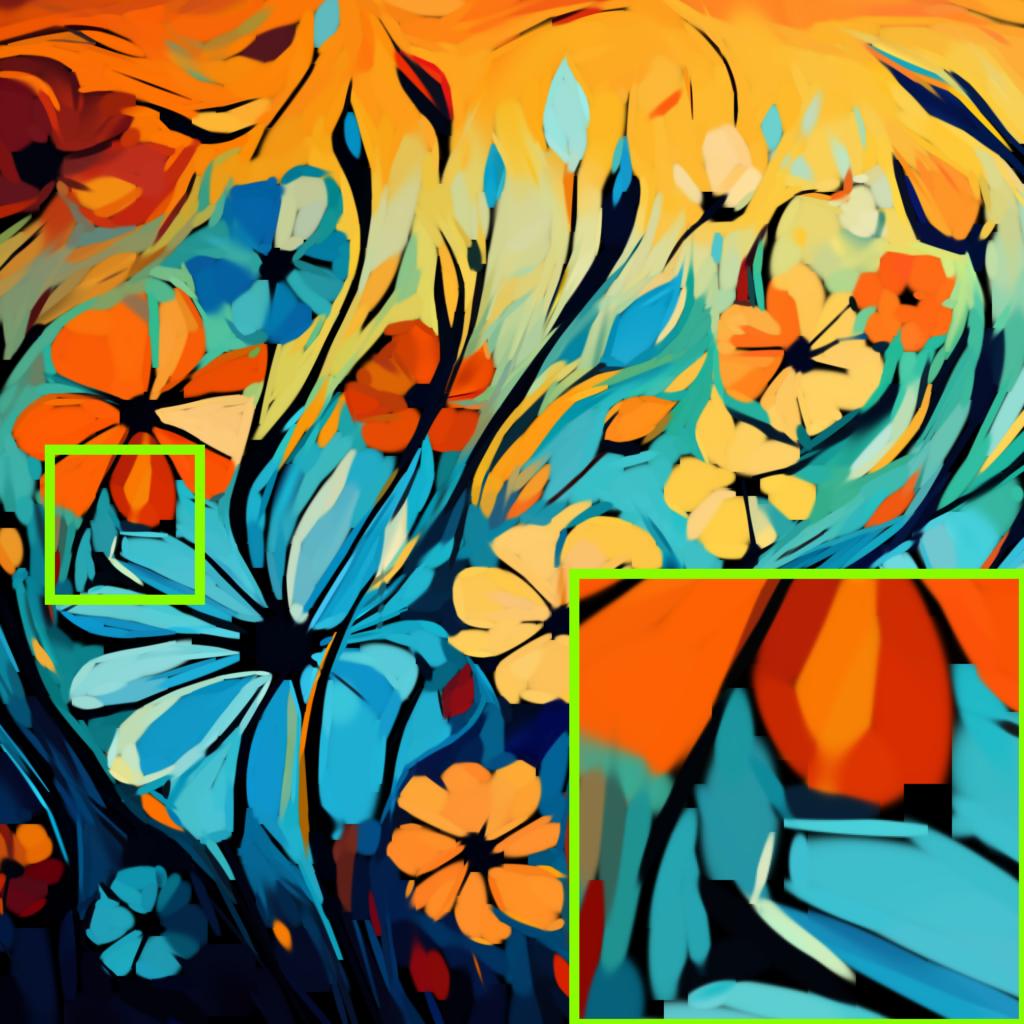}
  }
\subfloat{
    \includegraphics[width=\lodCompWidthSupp,height=\lodCompWidthSupp]{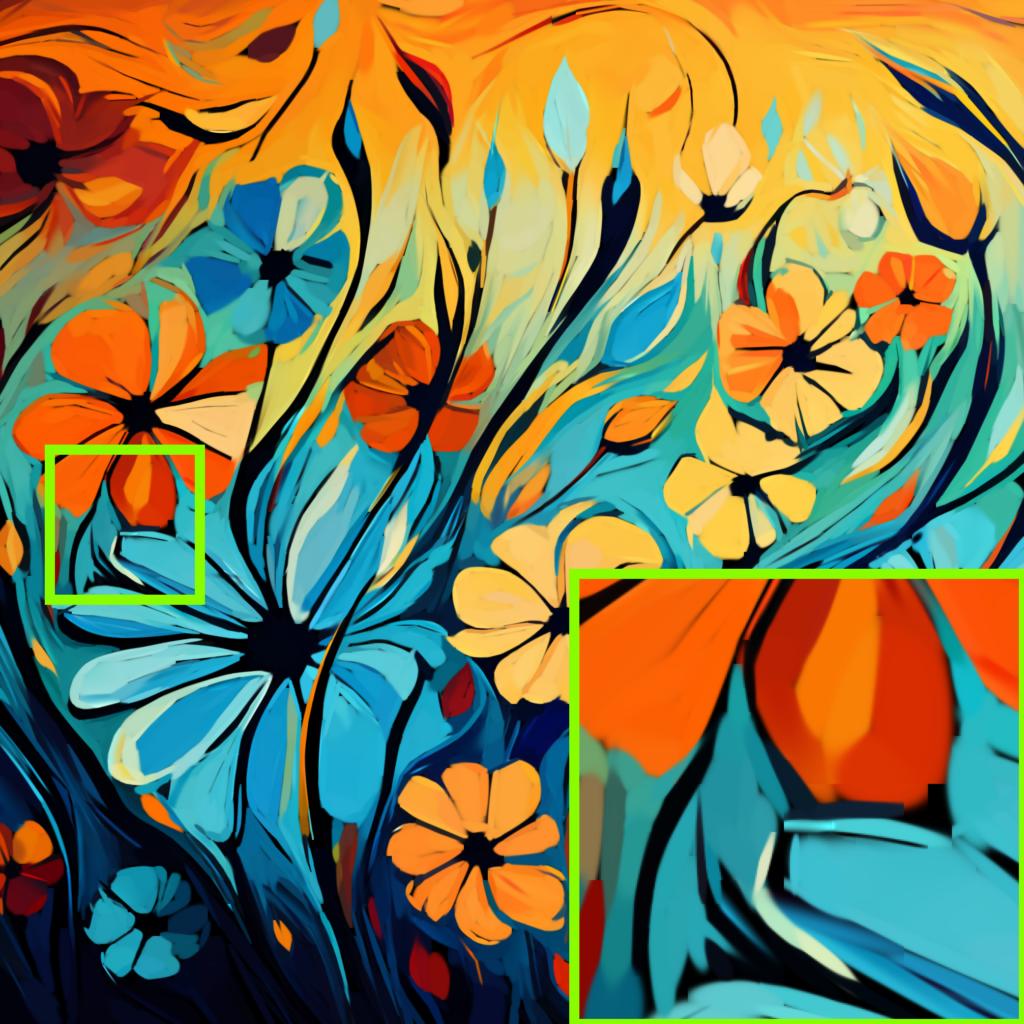}
  }
\subfloat{
    \includegraphics[width=\lodCompWidthSupp,height=\lodCompWidthSupp]{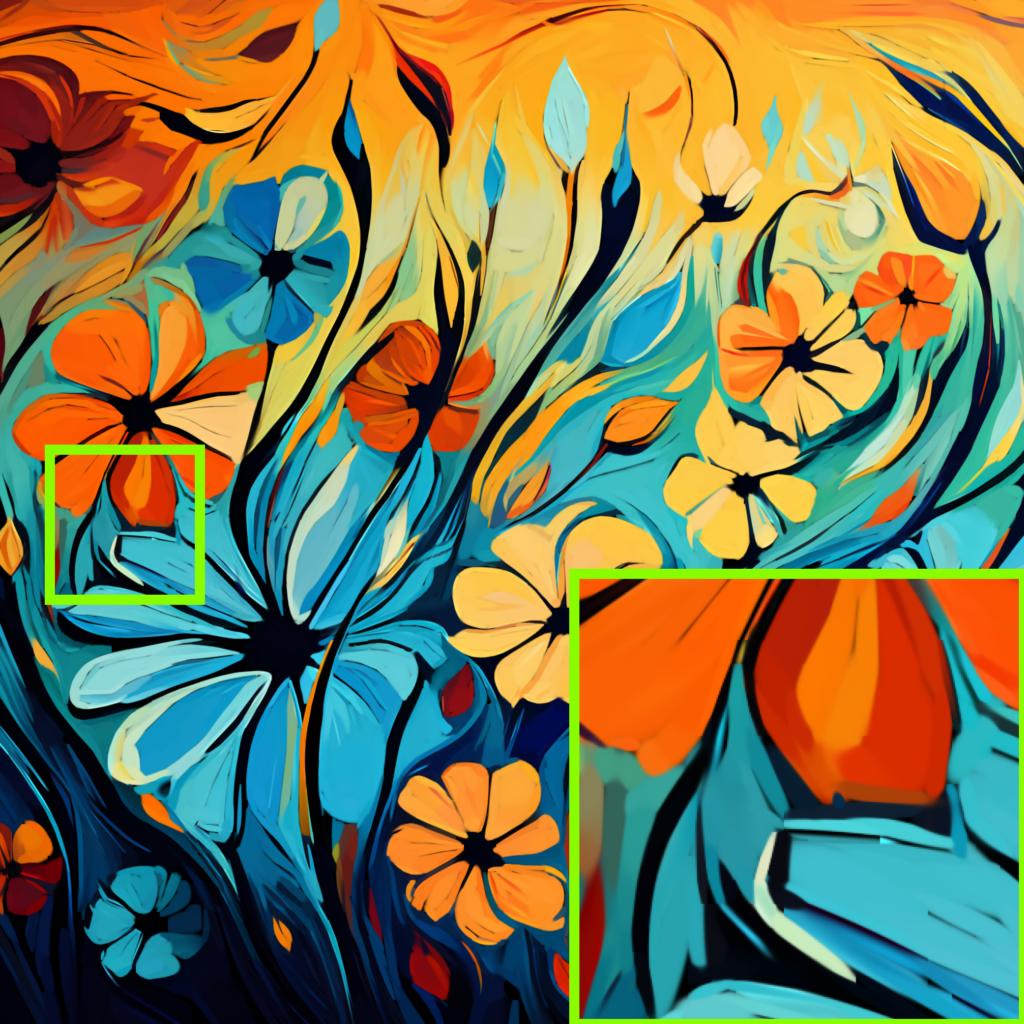}
  }
\subfloat{
    \includegraphics[width=\lodCompWidthSupp,height=\lodCompWidthSupp]{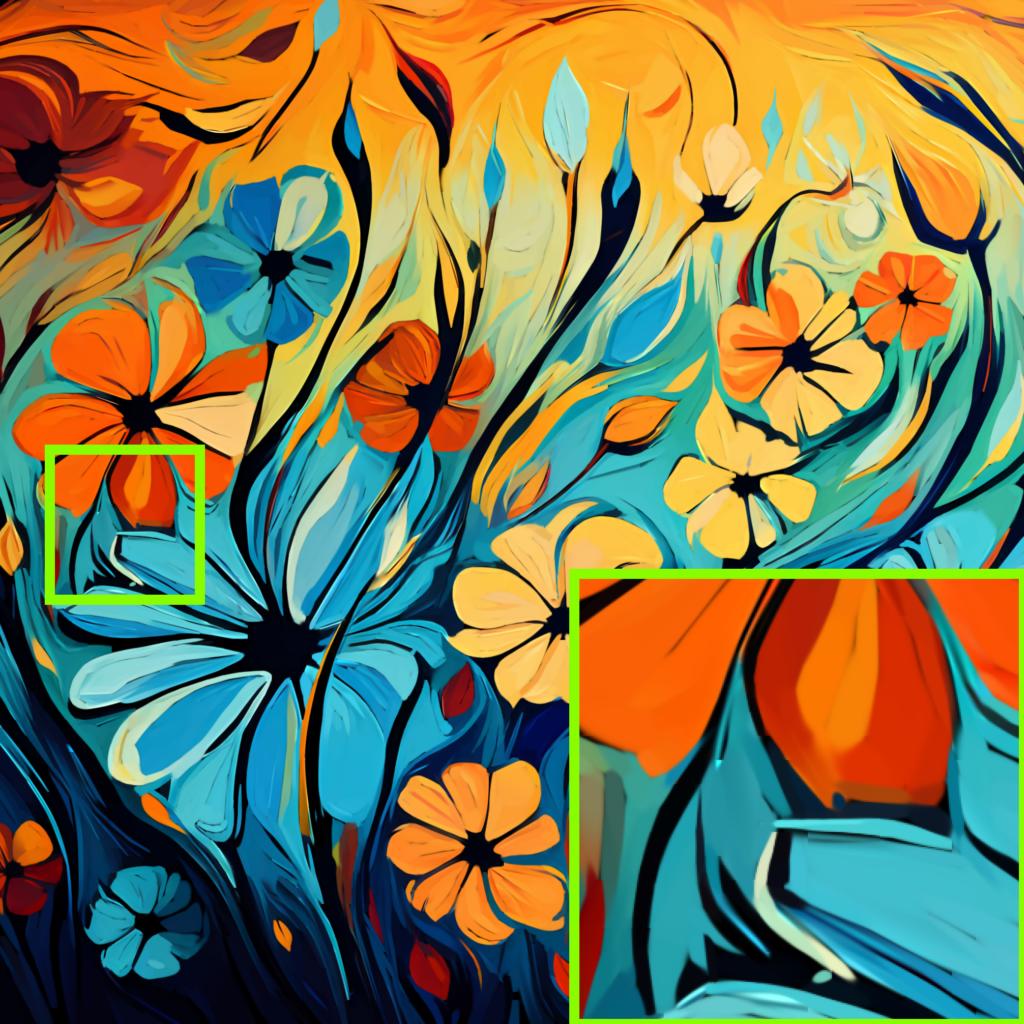}
  }
\subfloat{
    \includegraphics[width=\lodCompWidthSupp,height=\lodCompWidthSupp]{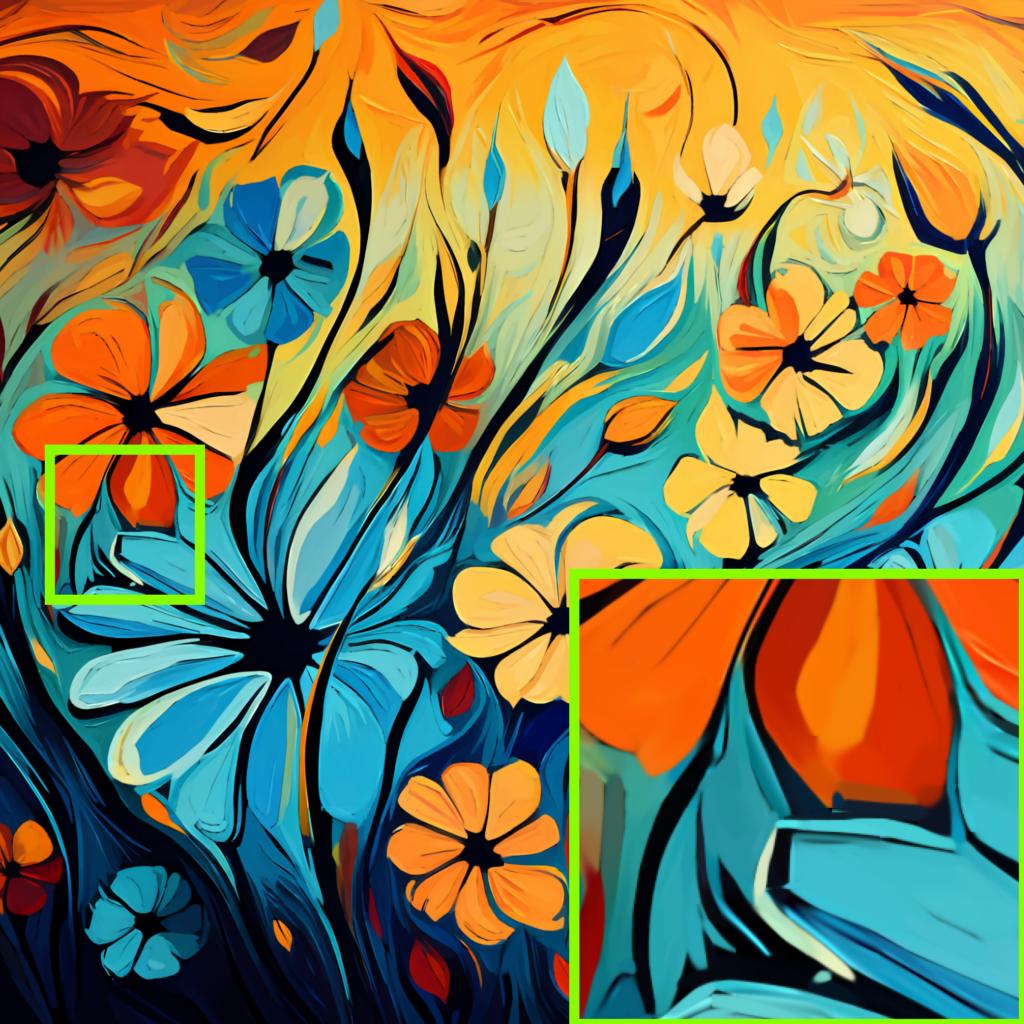}
  }
\subfloat{
    \includegraphics[width=\lodCompWidthSupp,height=\lodCompWidthSupp]{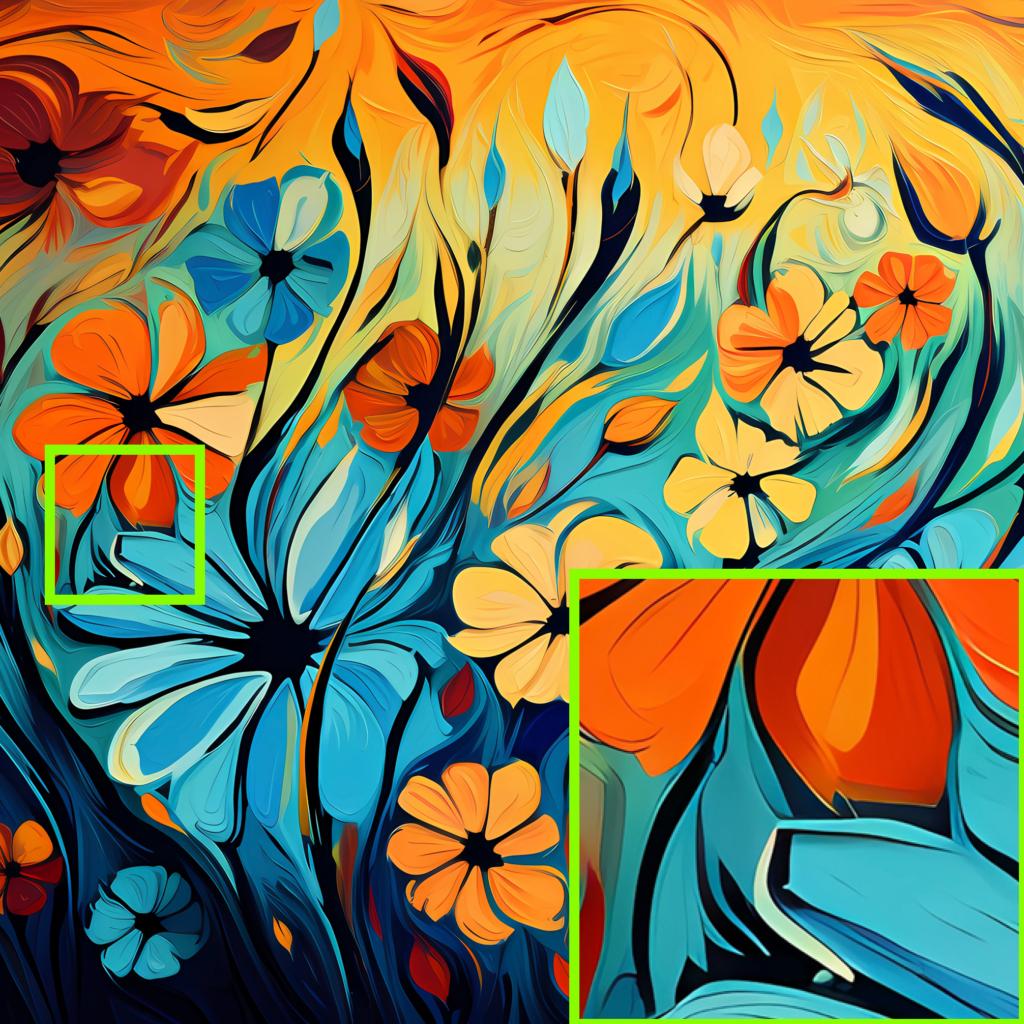}
  }
\vspace{2mm} \\
\subfloat{
    \includegraphics[width=\lodCompWidthSupp,height=\lodCompWidthSupp]{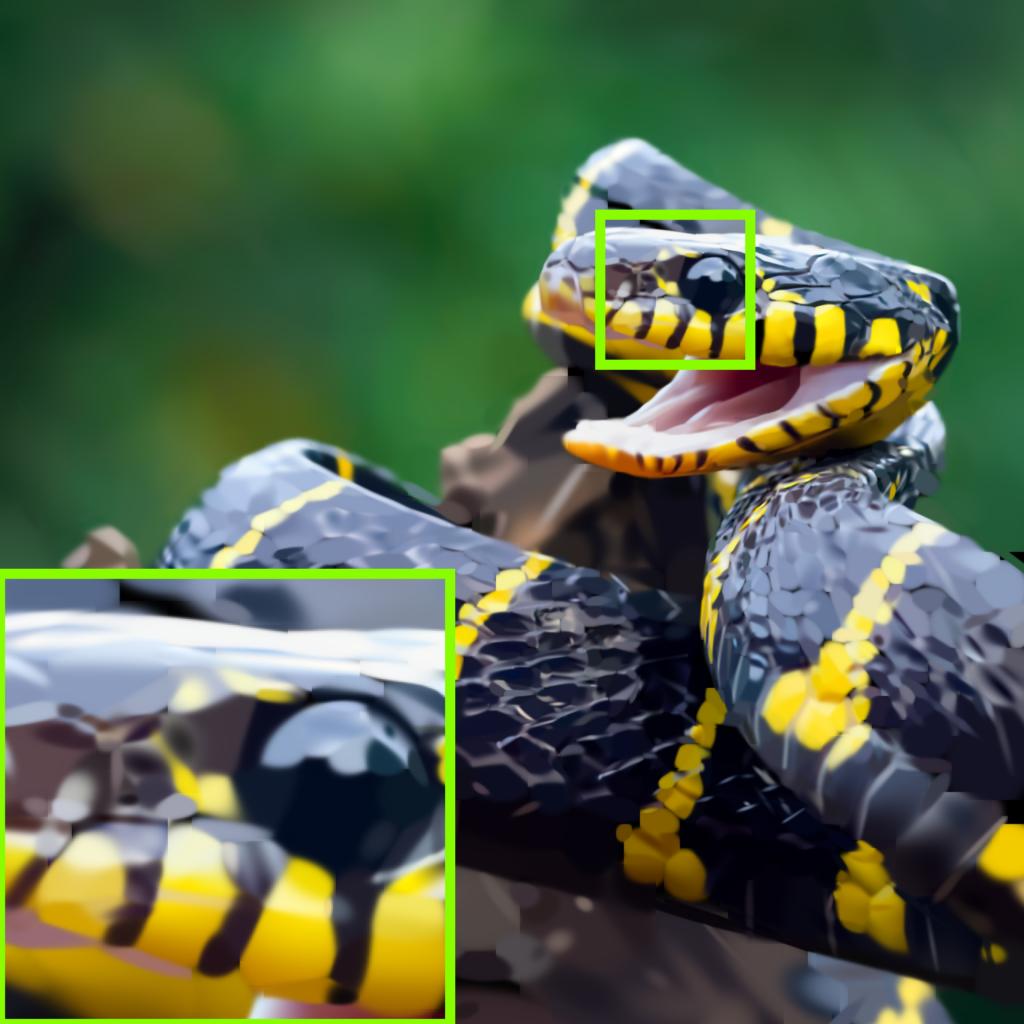}
  }
\subfloat{
    \includegraphics[width=\lodCompWidthSupp,height=\lodCompWidthSupp]{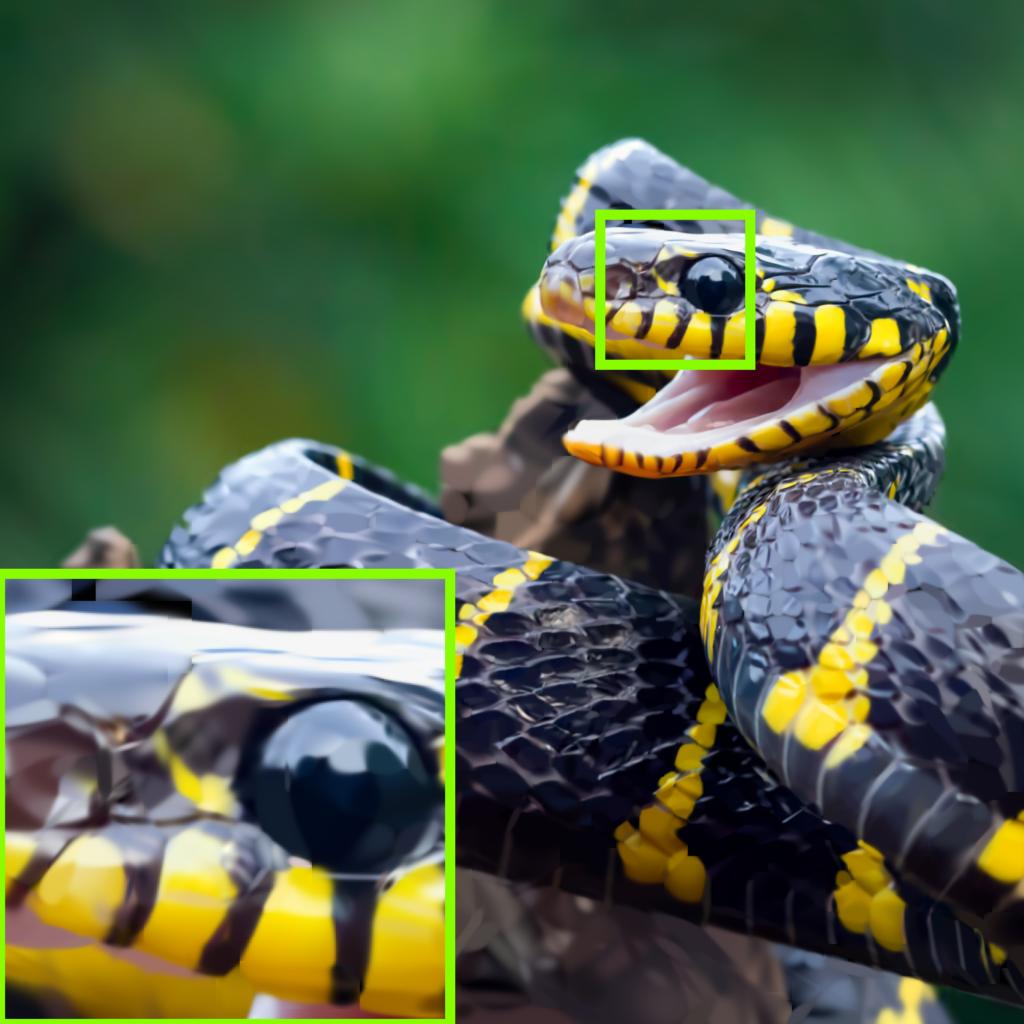}
  }
\subfloat{
    \includegraphics[width=\lodCompWidthSupp,height=\lodCompWidthSupp]{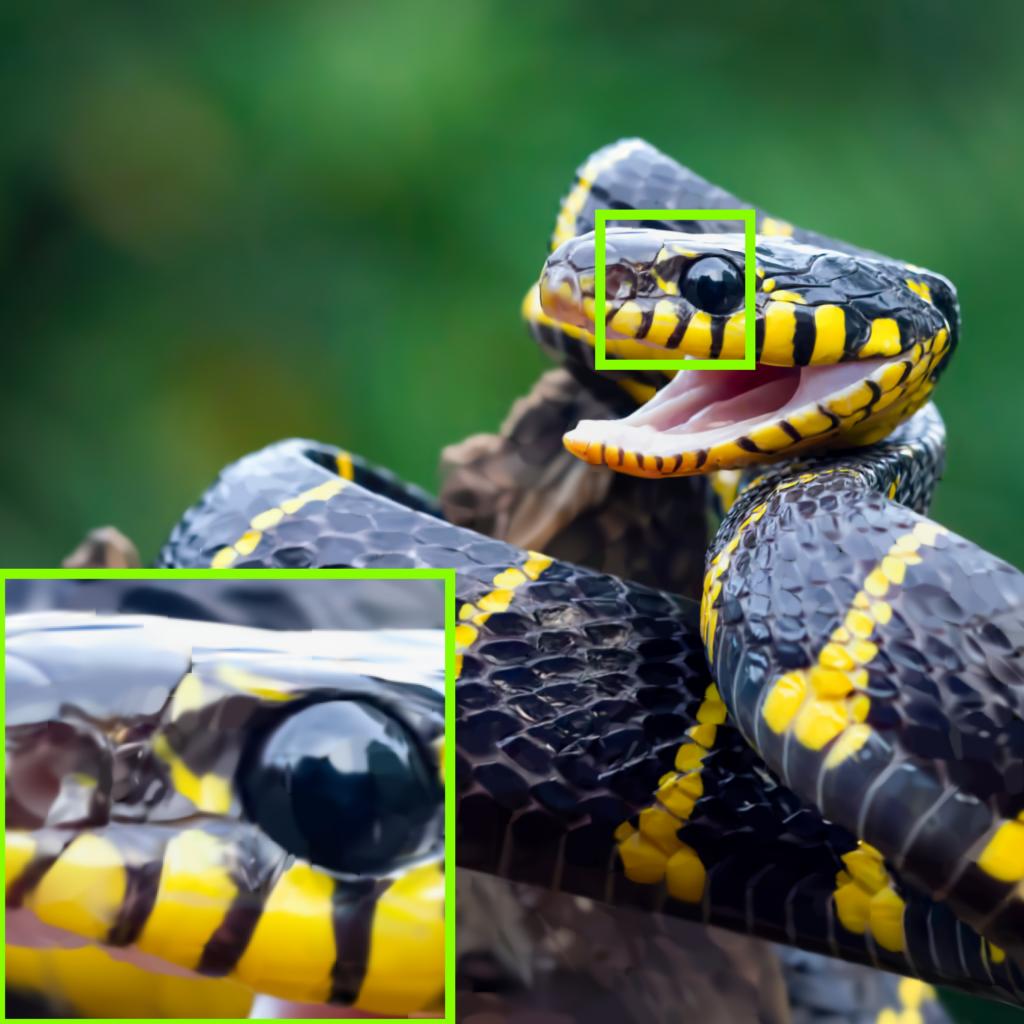}
  }
\subfloat{
    \includegraphics[width=\lodCompWidthSupp,height=\lodCompWidthSupp]{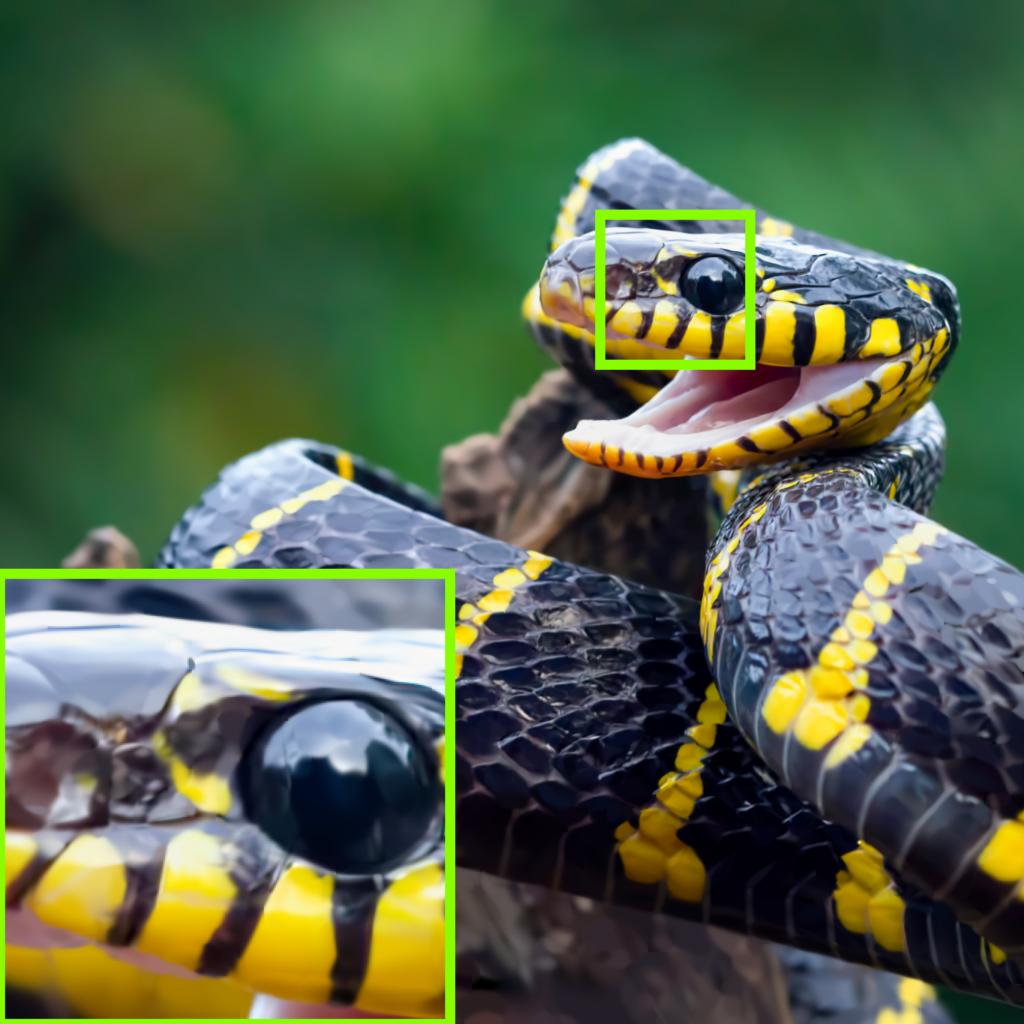}
  }
\subfloat{
    \includegraphics[width=\lodCompWidthSupp,height=\lodCompWidthSupp]{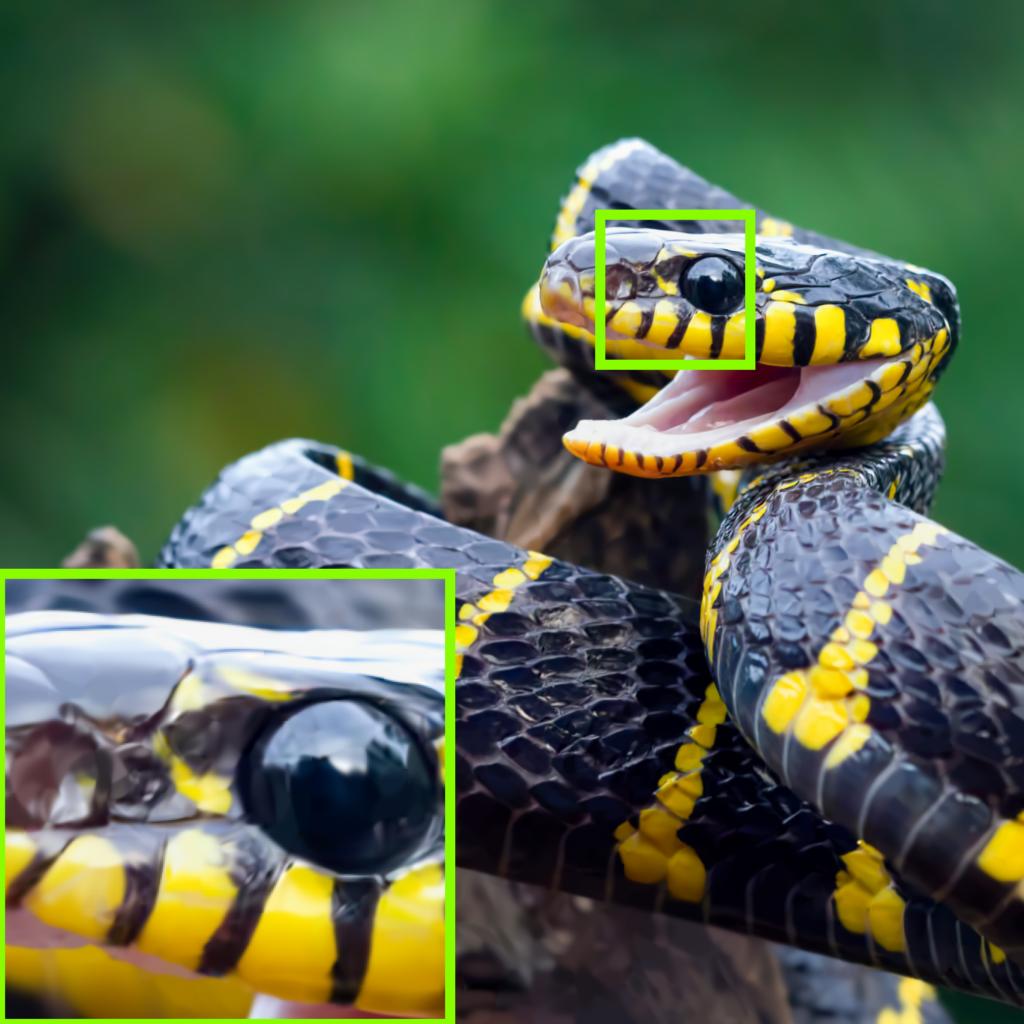}
  }
\subfloat{
    \includegraphics[width=\lodCompWidthSupp,height=\lodCompWidthSupp]{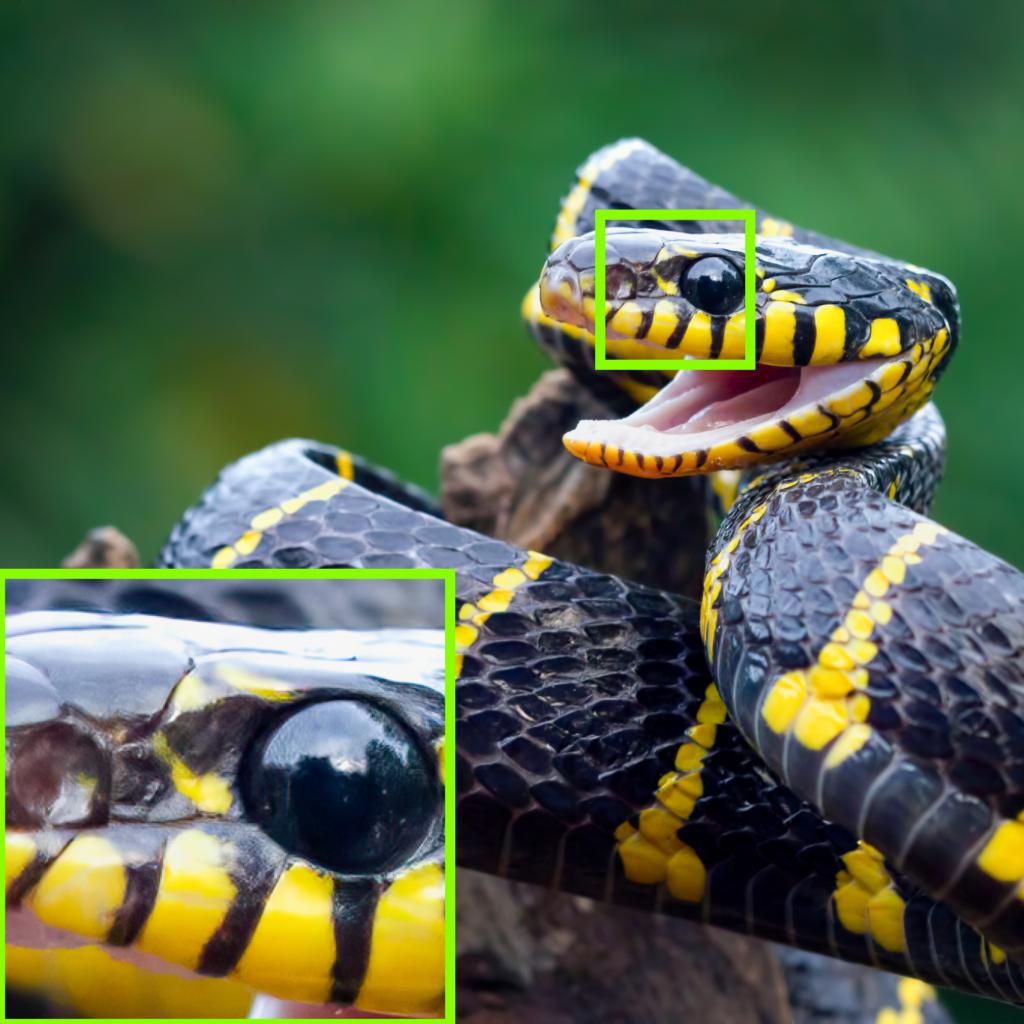}
  }
\vspace{2mm} \\
\setcounter{subfigure}{0}
\subfloat[Ours (0.061 bpp)]{
    \includegraphics[width=\lodCompWidthSupp,height=\lodCompWidthSupp]{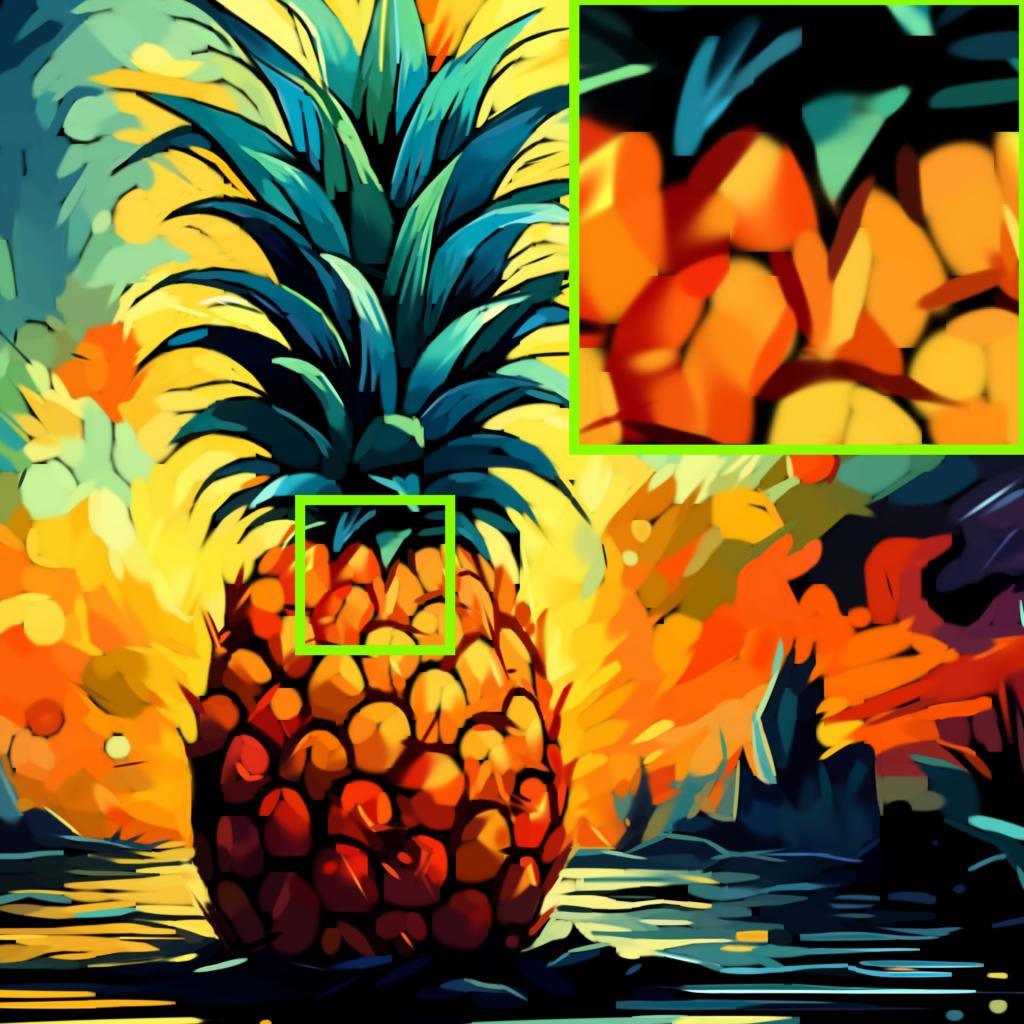}
  }
\subfloat[Ours (0.122 bpp)]{
    \includegraphics[width=\lodCompWidthSupp,height=\lodCompWidthSupp]{images/evaluation-lod/vector-1_2k/2000.jpg}
  }
\subfloat[Ours (0.183 bpp)]{
    \includegraphics[width=\lodCompWidthSupp,height=\lodCompWidthSupp]{images/evaluation-lod/vector-1_2k/2000.jpg}
  }
\subfloat[Ours (0.244 bpp)]{
    \includegraphics[width=\lodCompWidthSupp,height=\lodCompWidthSupp]{images/evaluation-lod/vector-1_2k/2000.jpg}
  }
\subfloat[Ours (0.305 bpp)]{
    \includegraphics[width=\lodCompWidthSupp,height=\lodCompWidthSupp]{images/evaluation-lod/vector-1_2k/2000.jpg}
  }
\subfloat[Reference]{
    \includegraphics[width=\lodCompWidthSupp,height=\lodCompWidthSupp]{images/evaluation-lod/vector-1_2k/2000.jpg}
  }
\Caption{\revise{\methodName's rate-distortion trade-off (\Cref{sec:evaluation-lod}).}}{\revise{Through error-guided progressive optimization (\Cref{sec:method-optimization}), \methodName naturally constructs a smooth level-of-detail hierarchy in a single optimization run without additional overhead, enabling flexible quality adaptation to device capabilities.}
}
\end{figure*}
\clearpage
\section{\revise{Additional Image Compression Results on The CLIC2020 Benchmark}}
\label{fig:evaluation-image-clic-supp-1}
\label{fig:evaluation-image-clic-supp-2}
\newcommand{\ImageCompCLICSuppRes}{0.41\linewidth}
\begin{figure*}[h]
\centering
\subfloat{
    \includegraphics[width=\ImageCompCLICSuppRes]{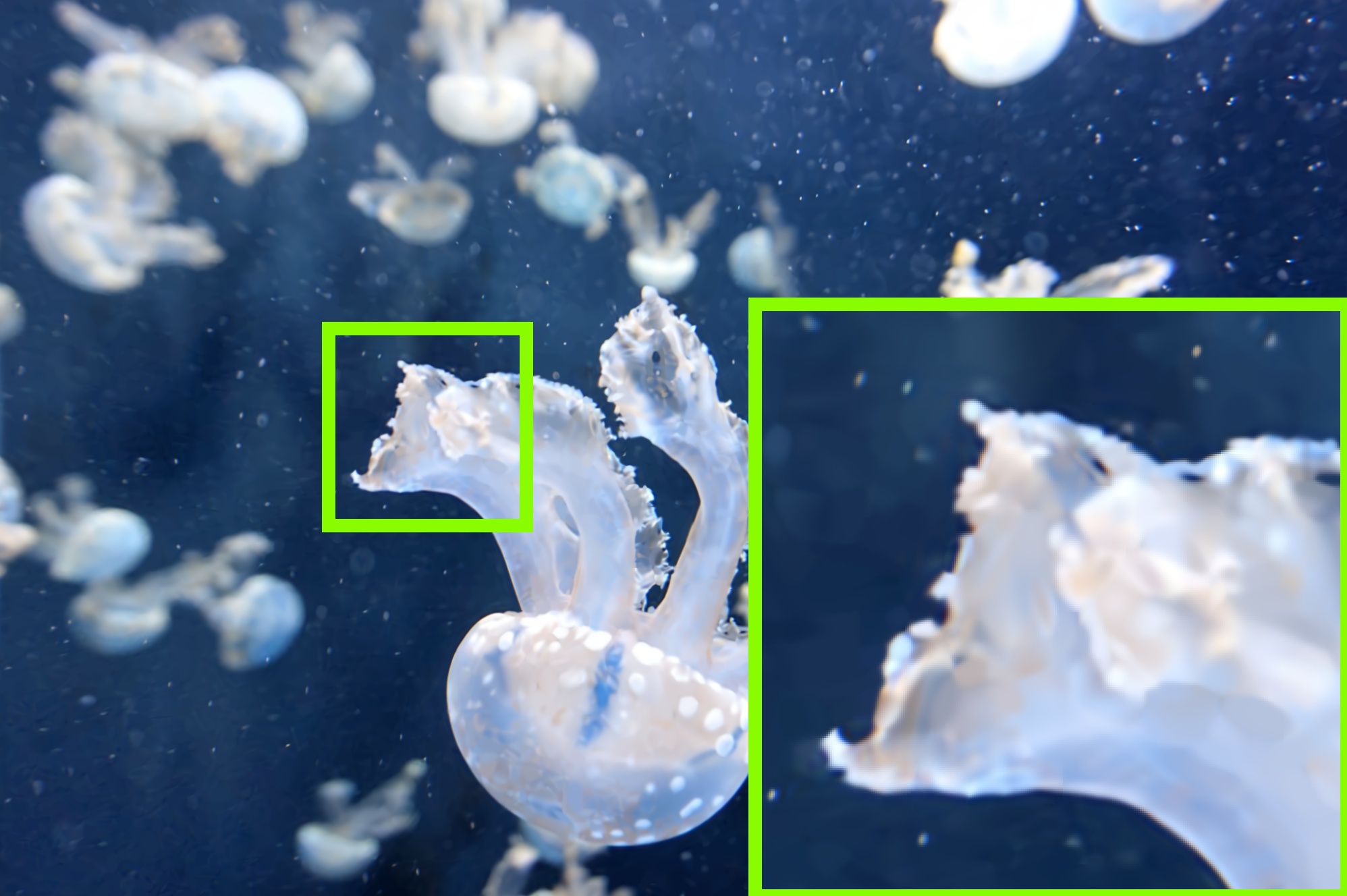}
  } \hspace{-0.18cm}
\subfloat{
    \includegraphics[width=\ImageCompCLICSuppRes]{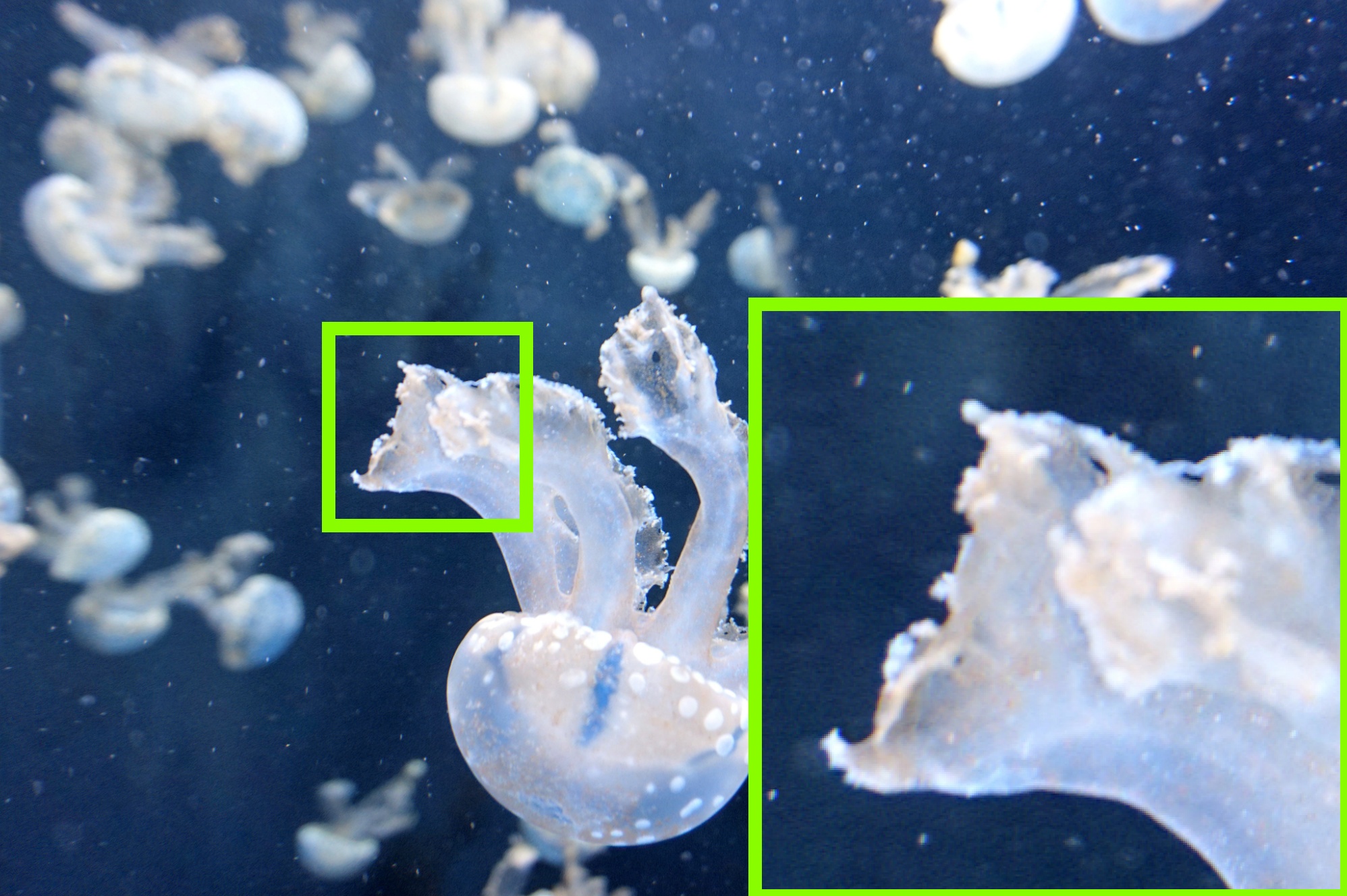}
  }
\vspace{0.2mm} \\
\subfloat{
    \includegraphics[width=\ImageCompCLICSuppRes]{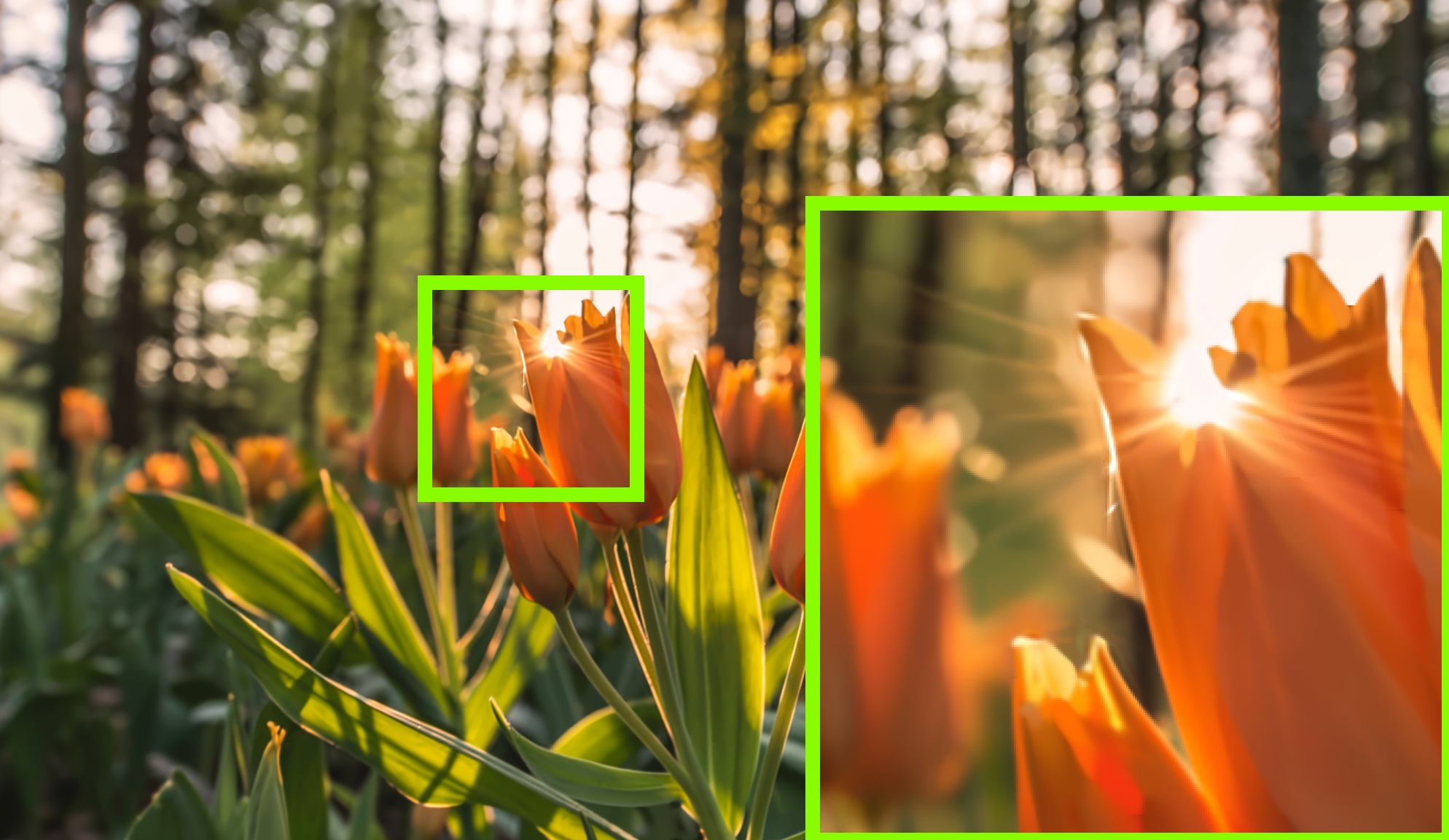}
  } \hspace{-0.18cm}
\subfloat{
    \includegraphics[width=\ImageCompCLICSuppRes]{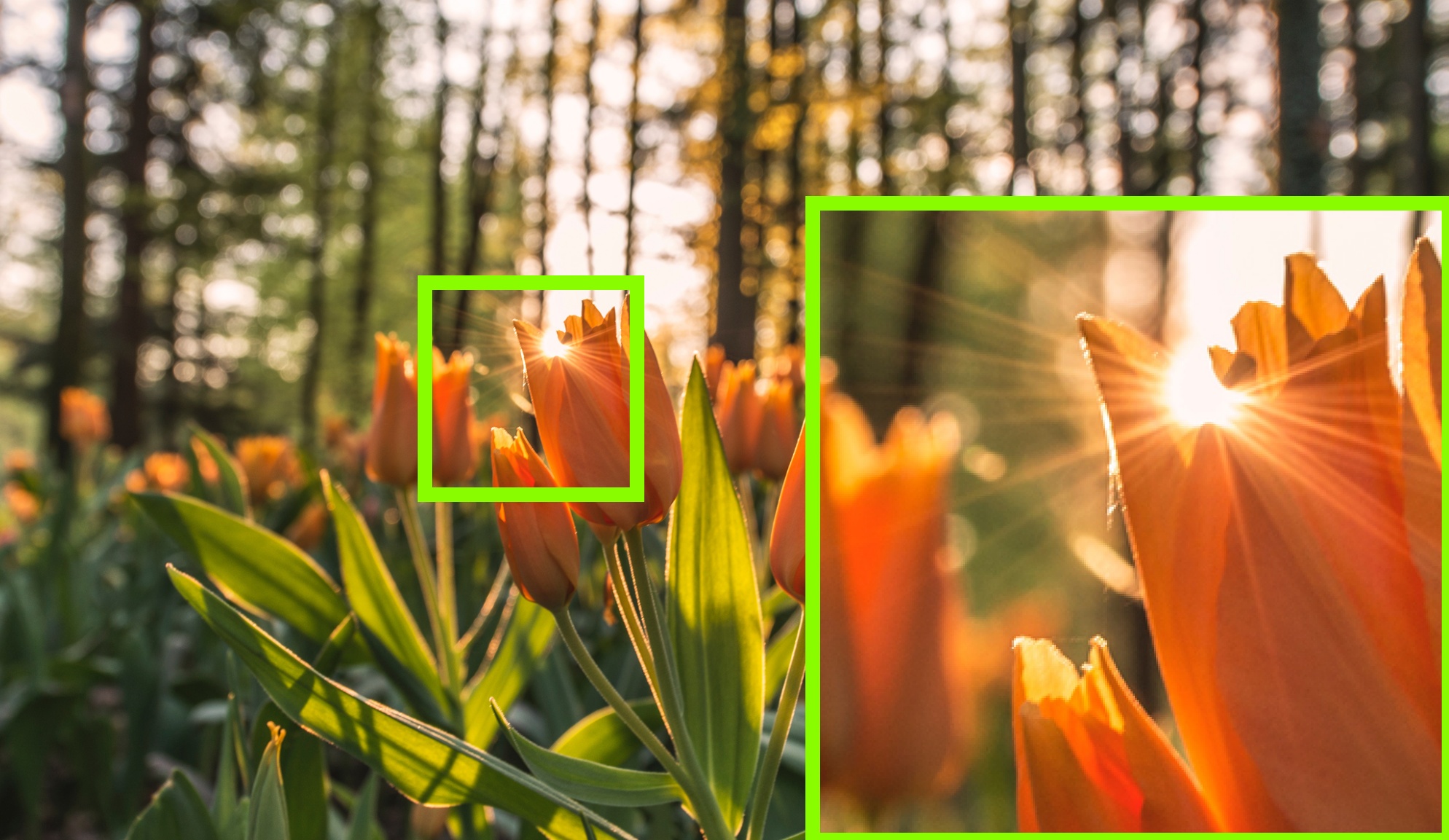}
  }
\vspace{0.2mm} \\
\subfloat{
    \includegraphics[width=\ImageCompCLICSuppRes]{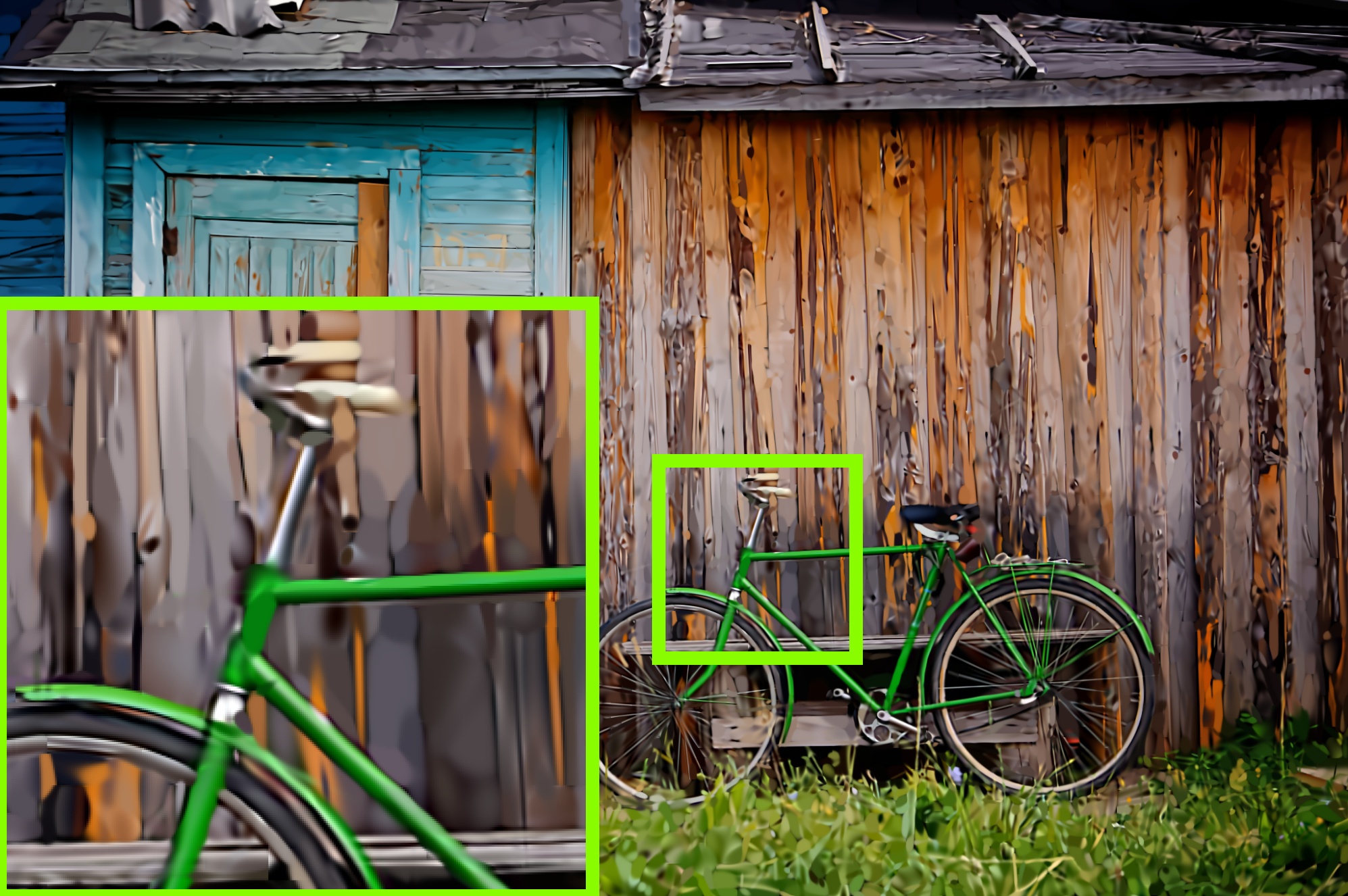}
  } \hspace{-0.18cm}
\subfloat{
    \includegraphics[width=\ImageCompCLICSuppRes]{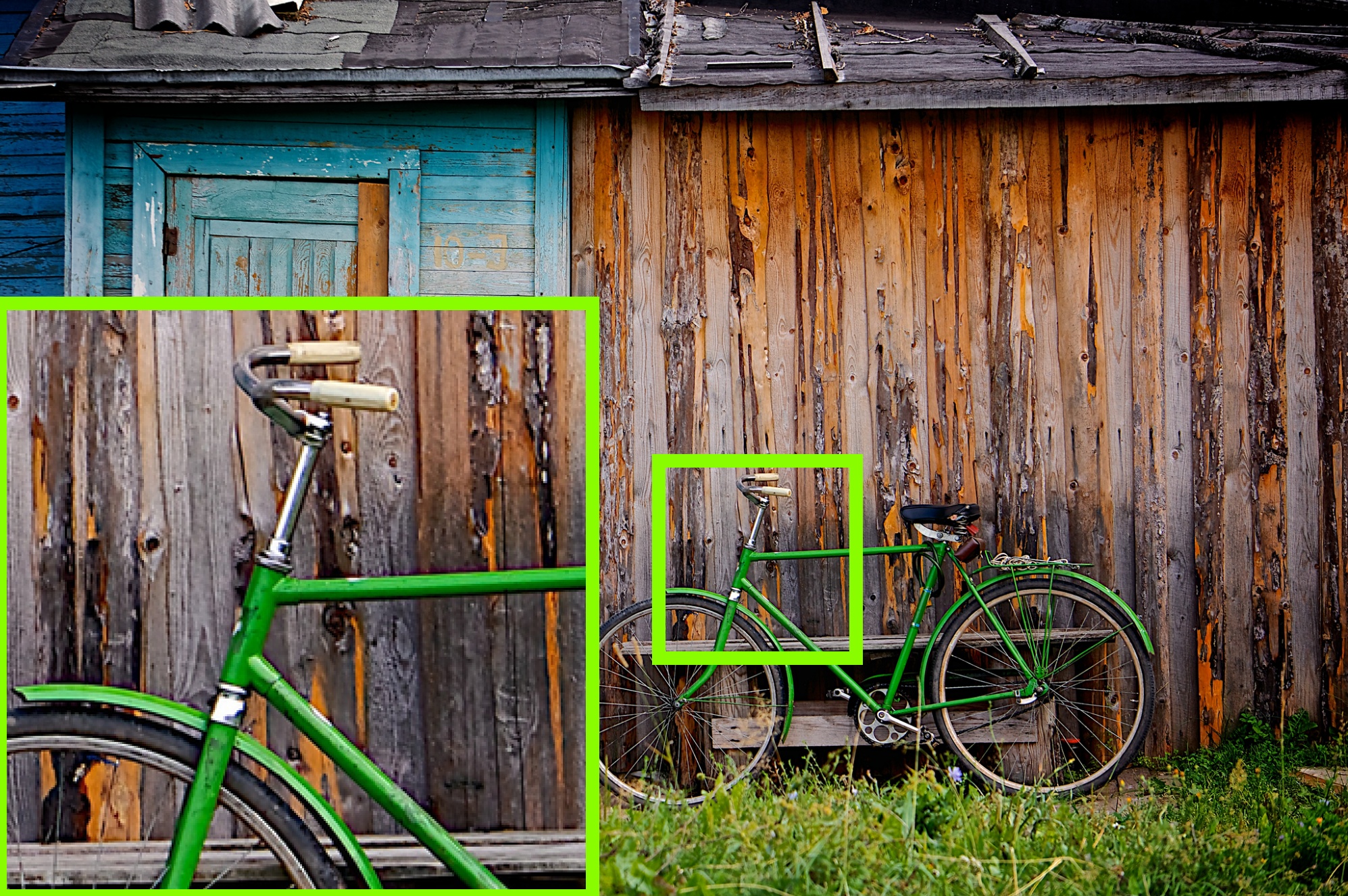}
  }
\vspace{0.2mm} \\
\setcounter{subfigure}{0}
\subfloat[Ours (0.50 bpp)]{
    \includegraphics[width=\ImageCompCLICSuppRes]{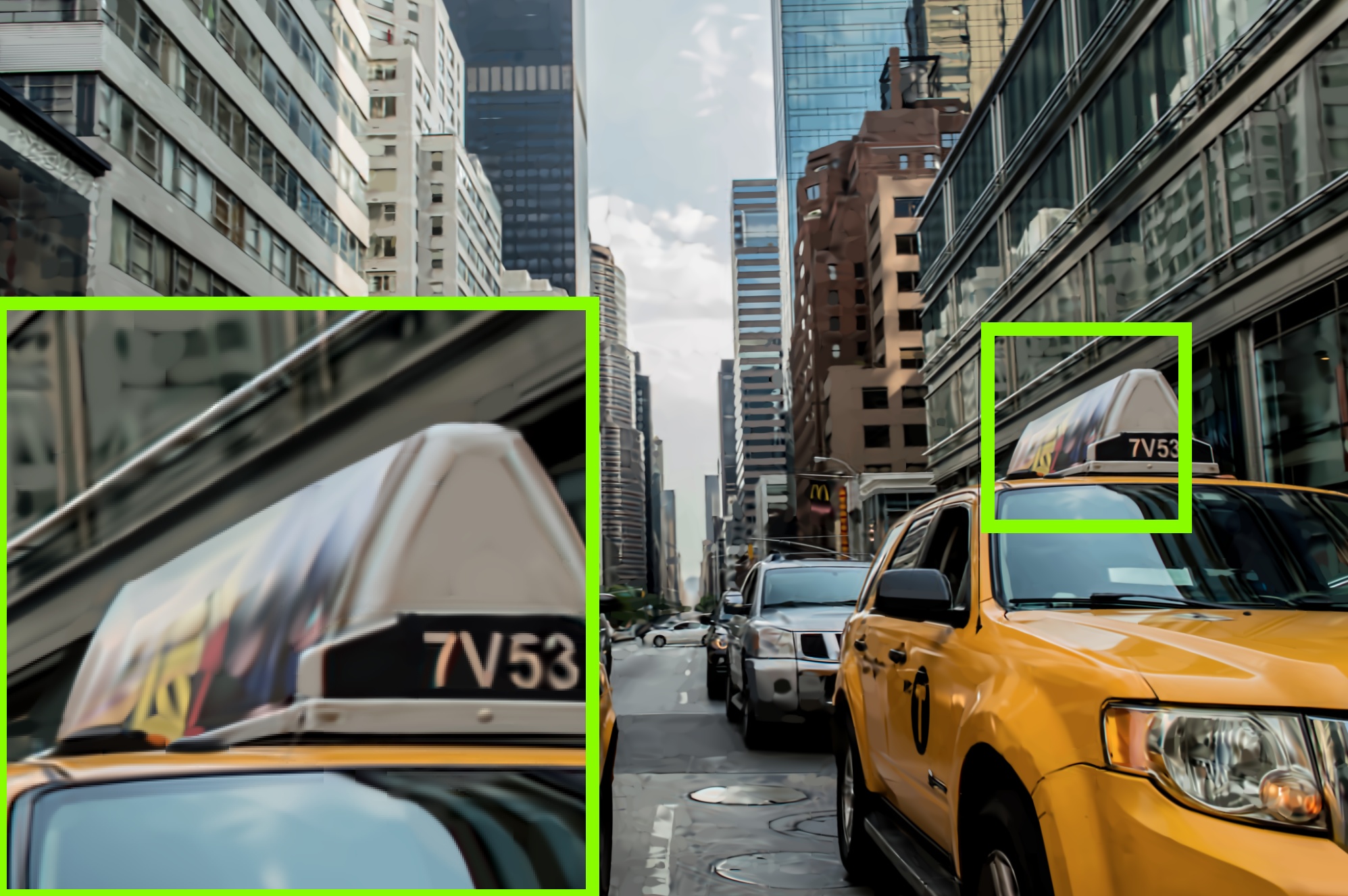}
  } \hspace{-0.18cm}
\subfloat[Reference]{
    \includegraphics[width=\ImageCompCLICSuppRes]{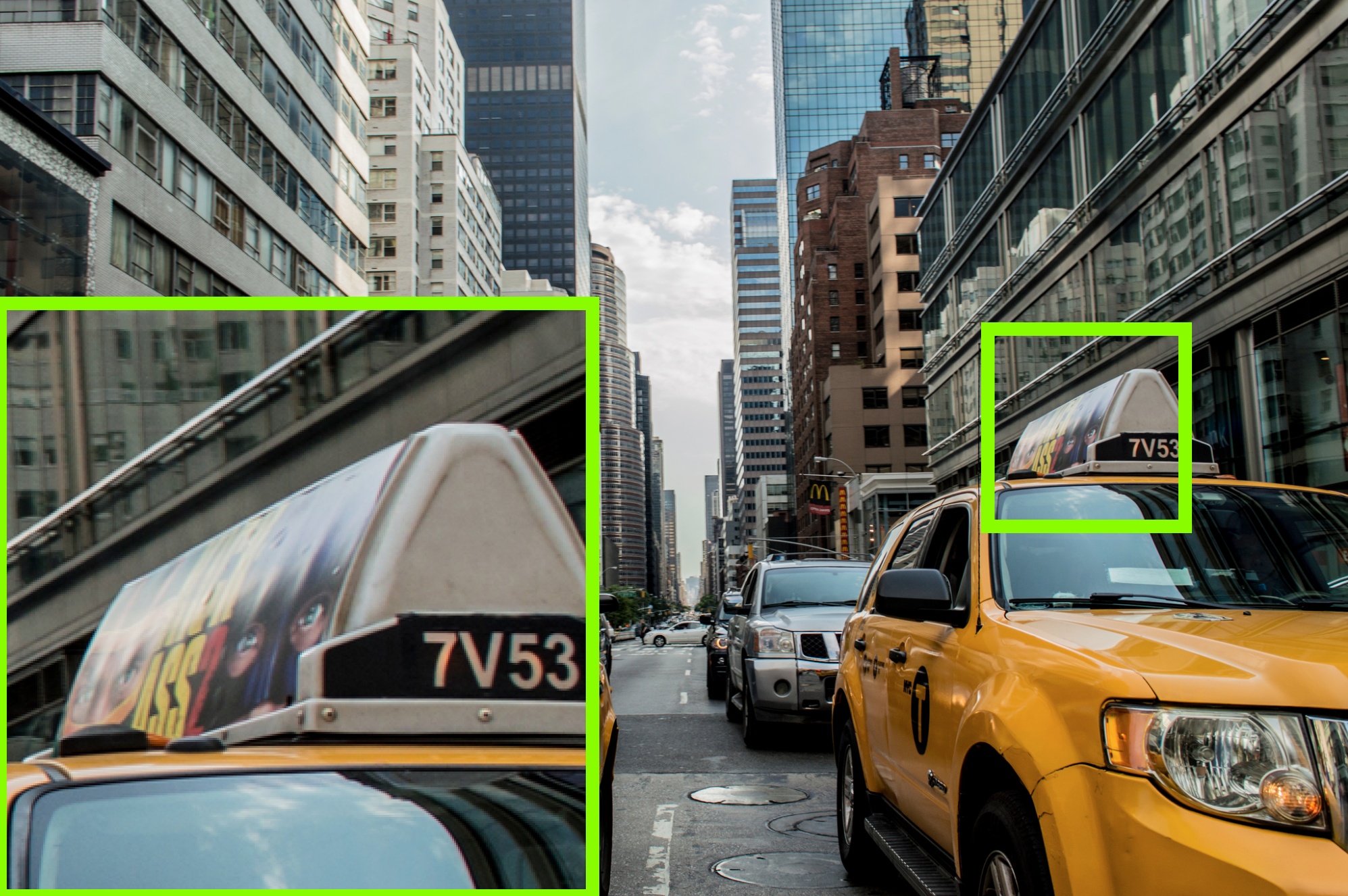}
  }
\Caption{\revise{Qualitative results on the professional validation split of the CLIC2020 dataset \cite{toderici2020workshop} (\Cref{sec:evaluation-image}).}}
{}
\end{figure*}
\begin{figure*}[h]
\centering
\subfloat{
    \includegraphics[width=\ImageCompCLICSuppRes]{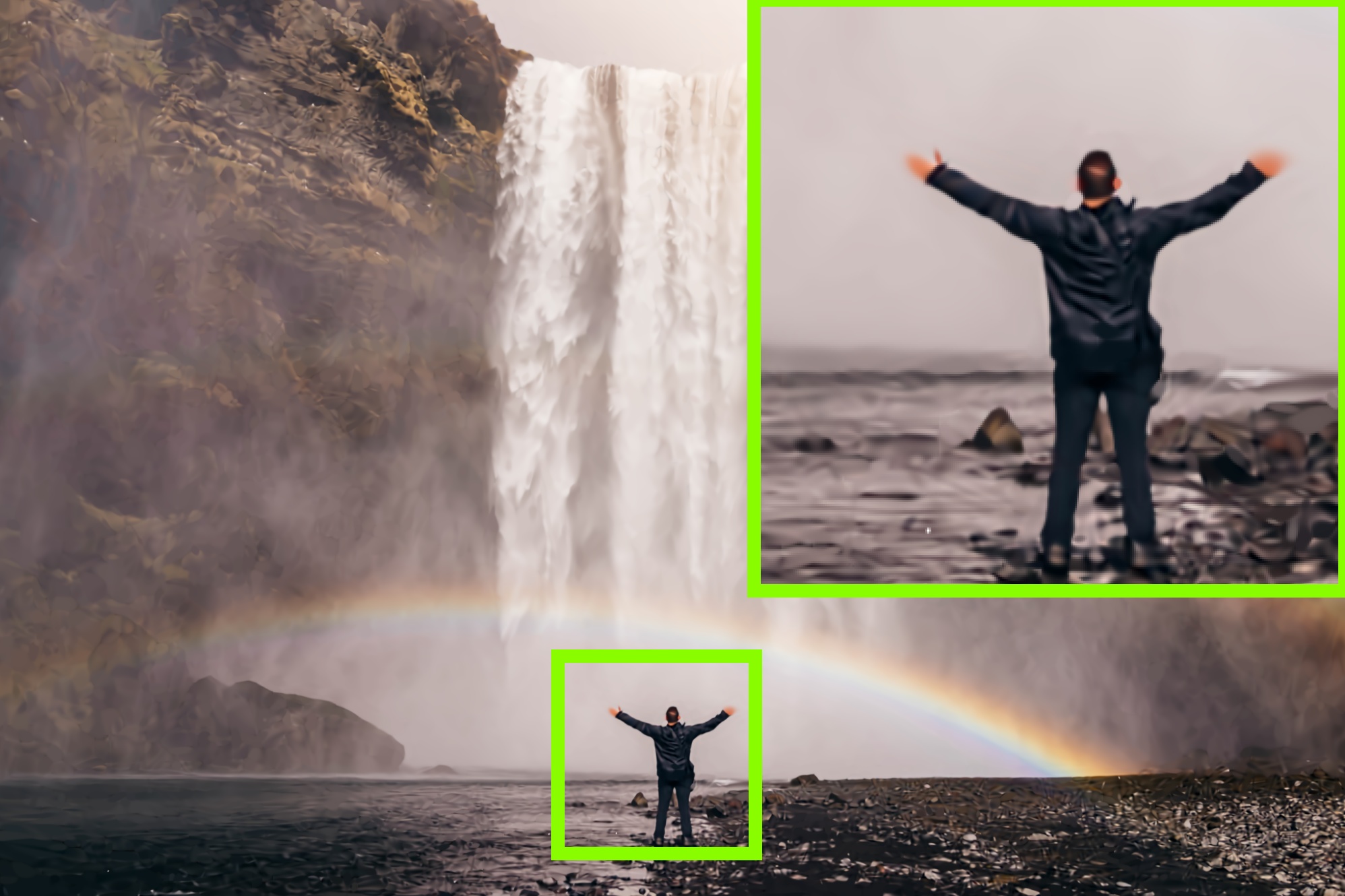}
  } \hspace{-0.18cm}
\subfloat{
    \includegraphics[width=\ImageCompCLICSuppRes]{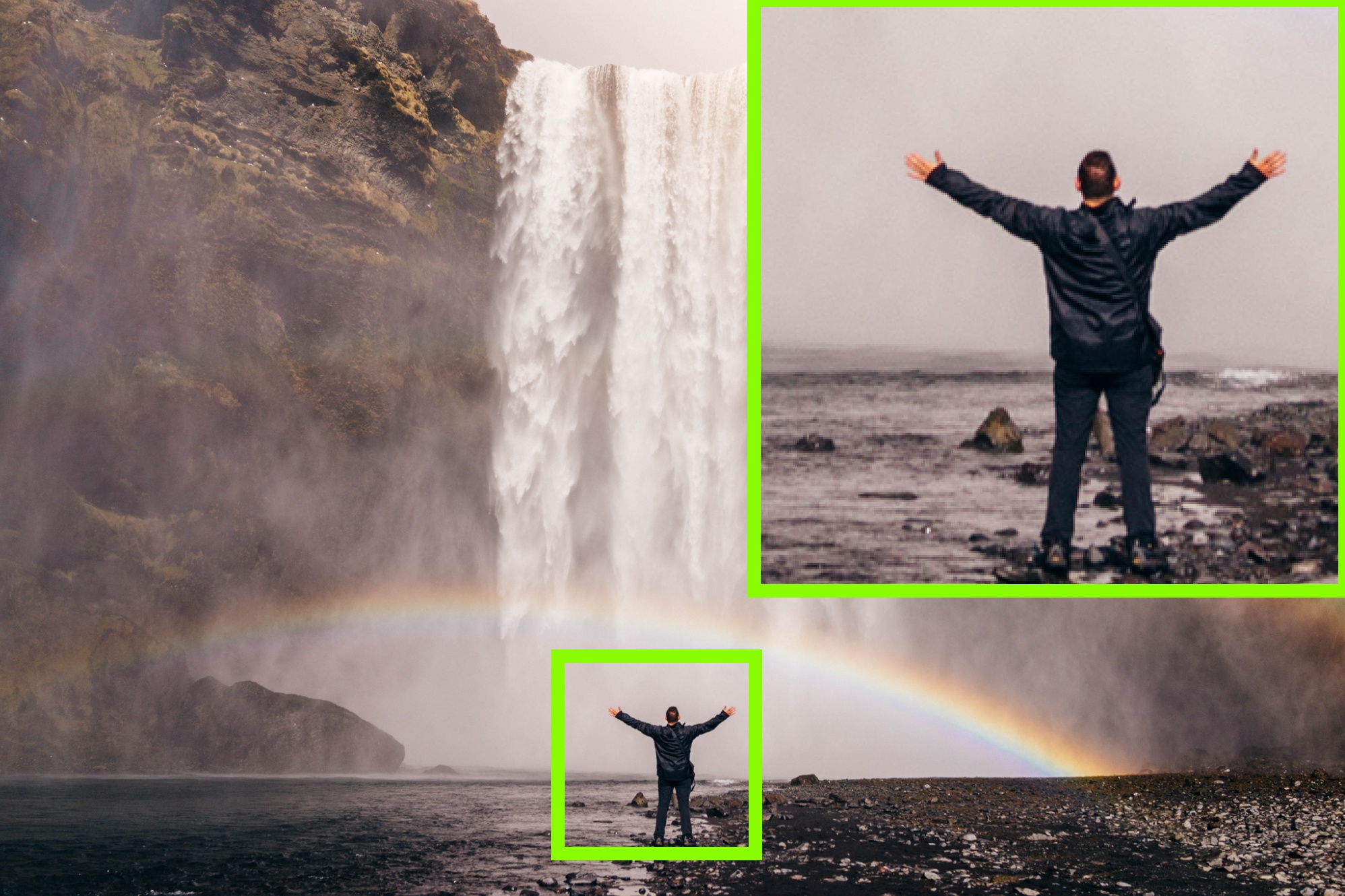}
  }
\vspace{0.2mm} \\
\subfloat{
    \includegraphics[width=\ImageCompCLICSuppRes]{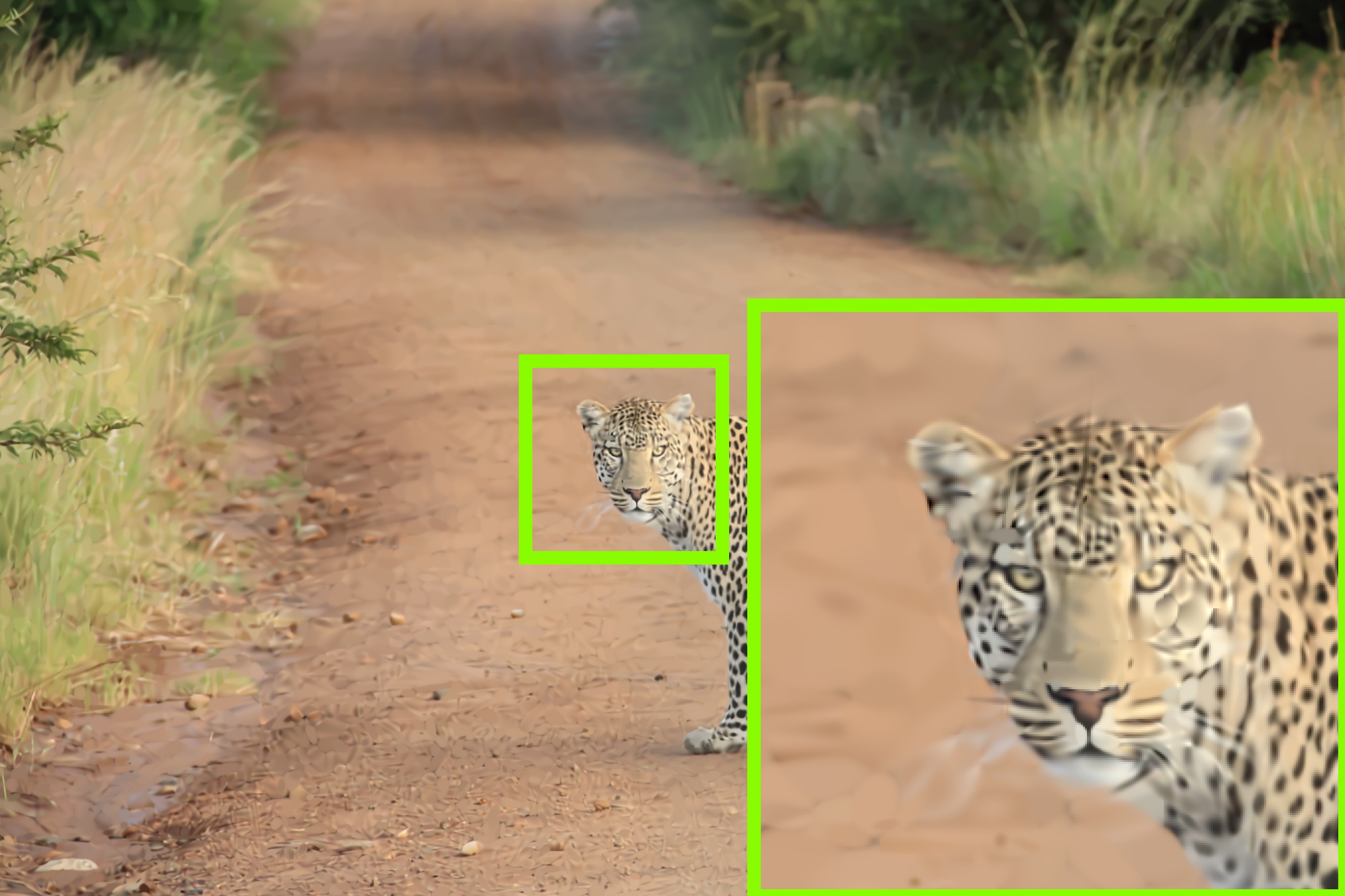}
  } \hspace{-0.18cm}
\subfloat{
    \includegraphics[width=\ImageCompCLICSuppRes]{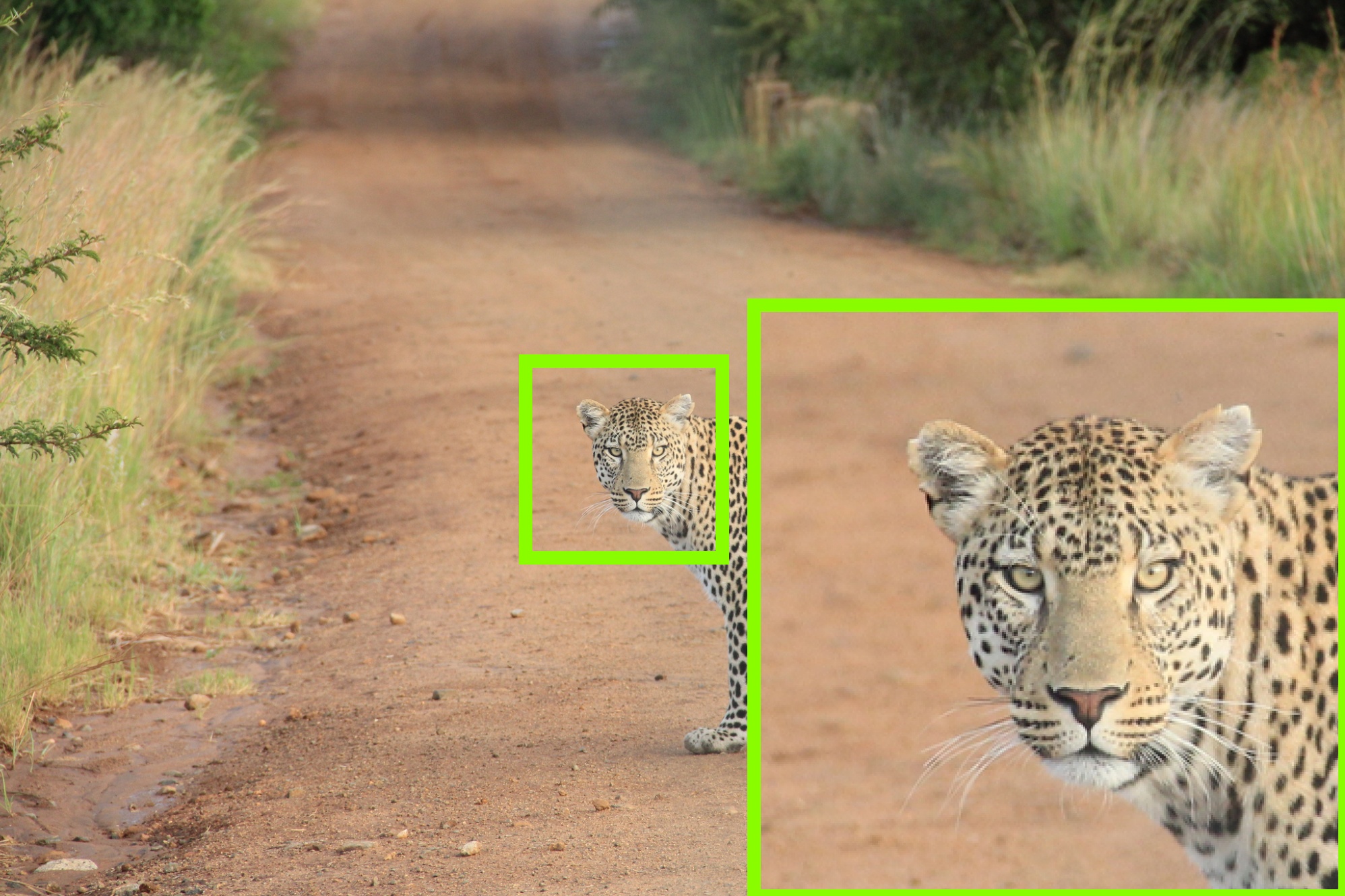}
  }
\vspace{0.2mm} \\
\subfloat{
    \includegraphics[width=\ImageCompCLICSuppRes]{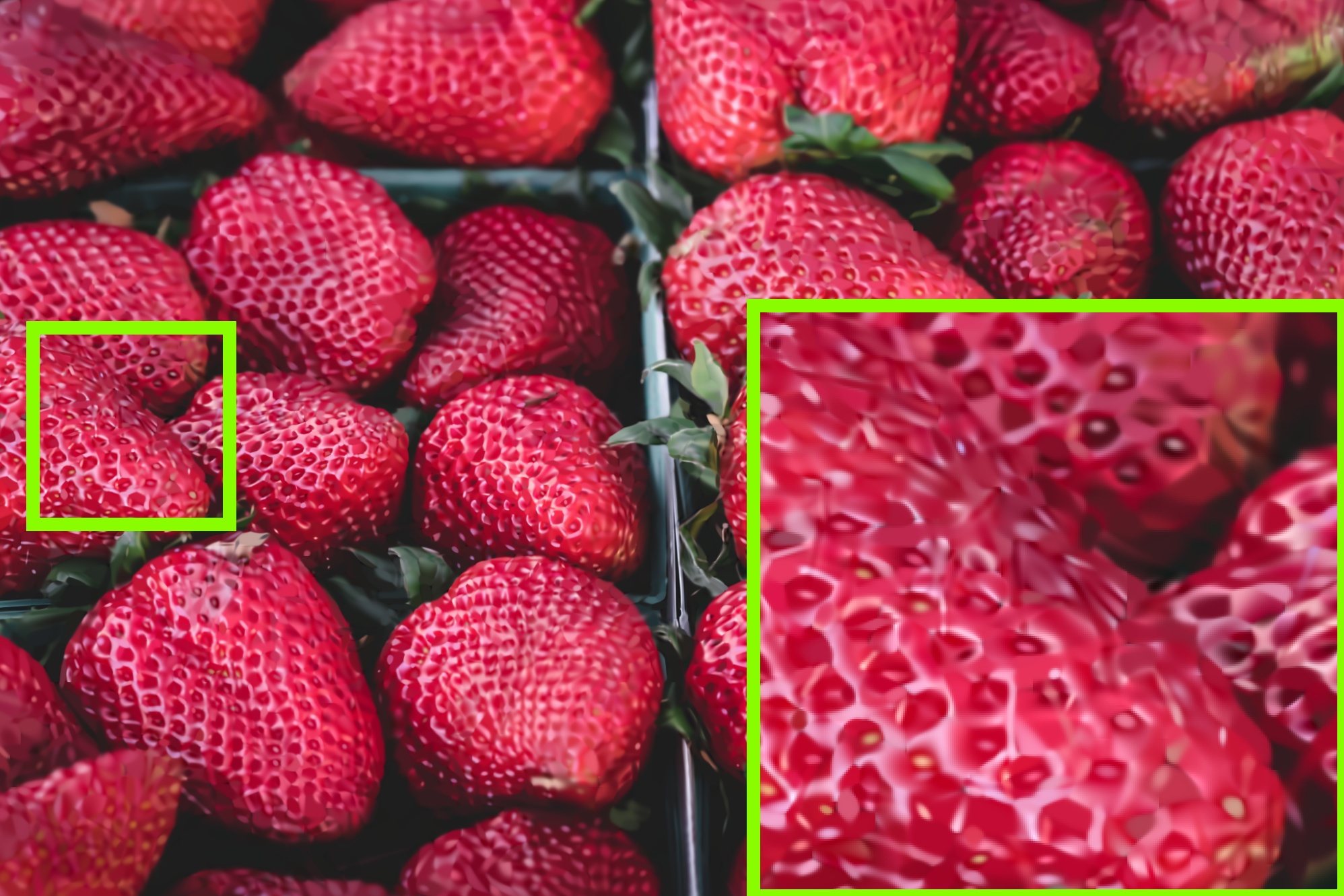}
  } \hspace{-0.18cm}
\subfloat{
    \includegraphics[width=\ImageCompCLICSuppRes]{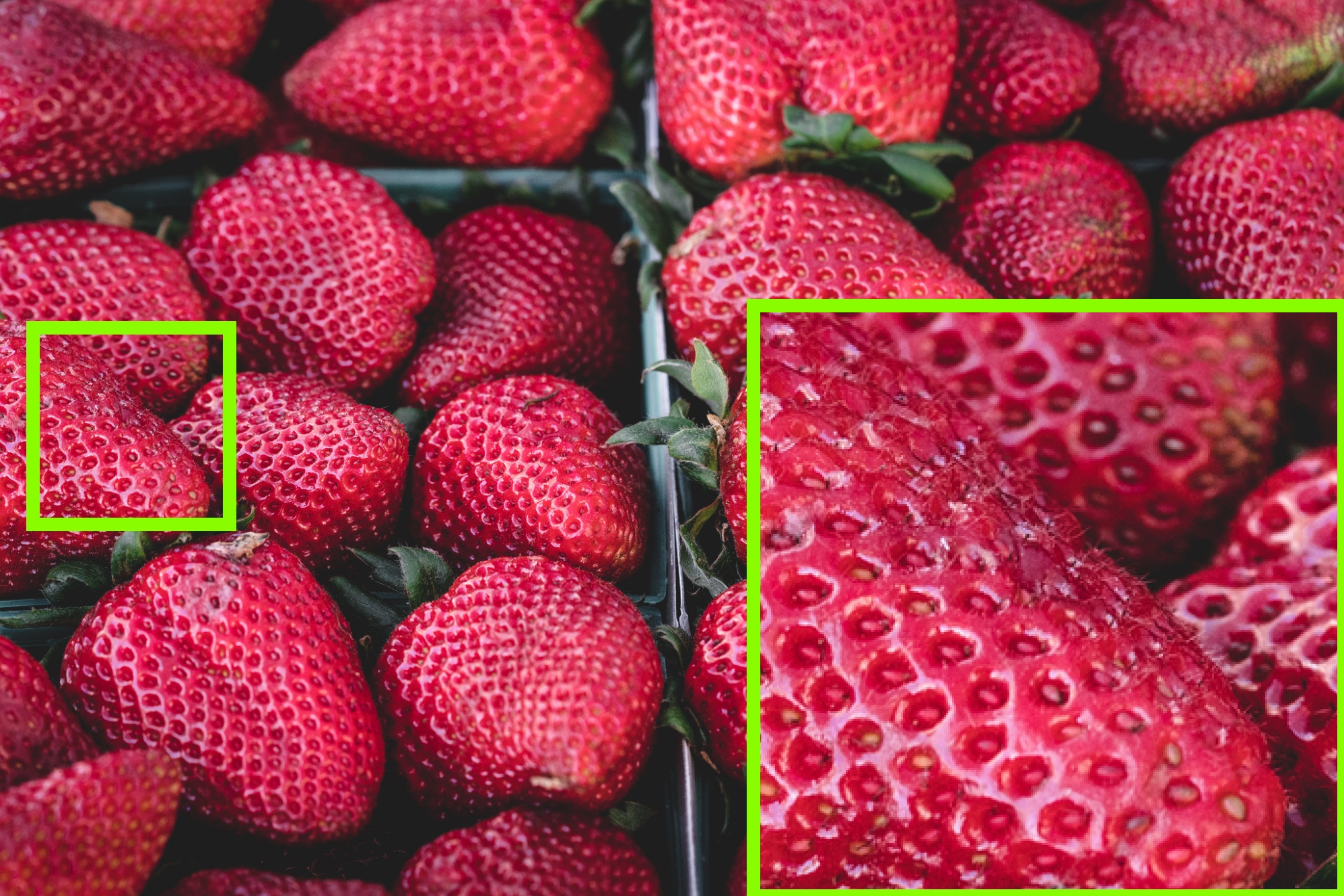}
  }
\vspace{0.2mm} \\
\setcounter{subfigure}{0}
\subfloat[Ours (0.50 bpp)]{
    \includegraphics[width=\ImageCompCLICSuppRes]{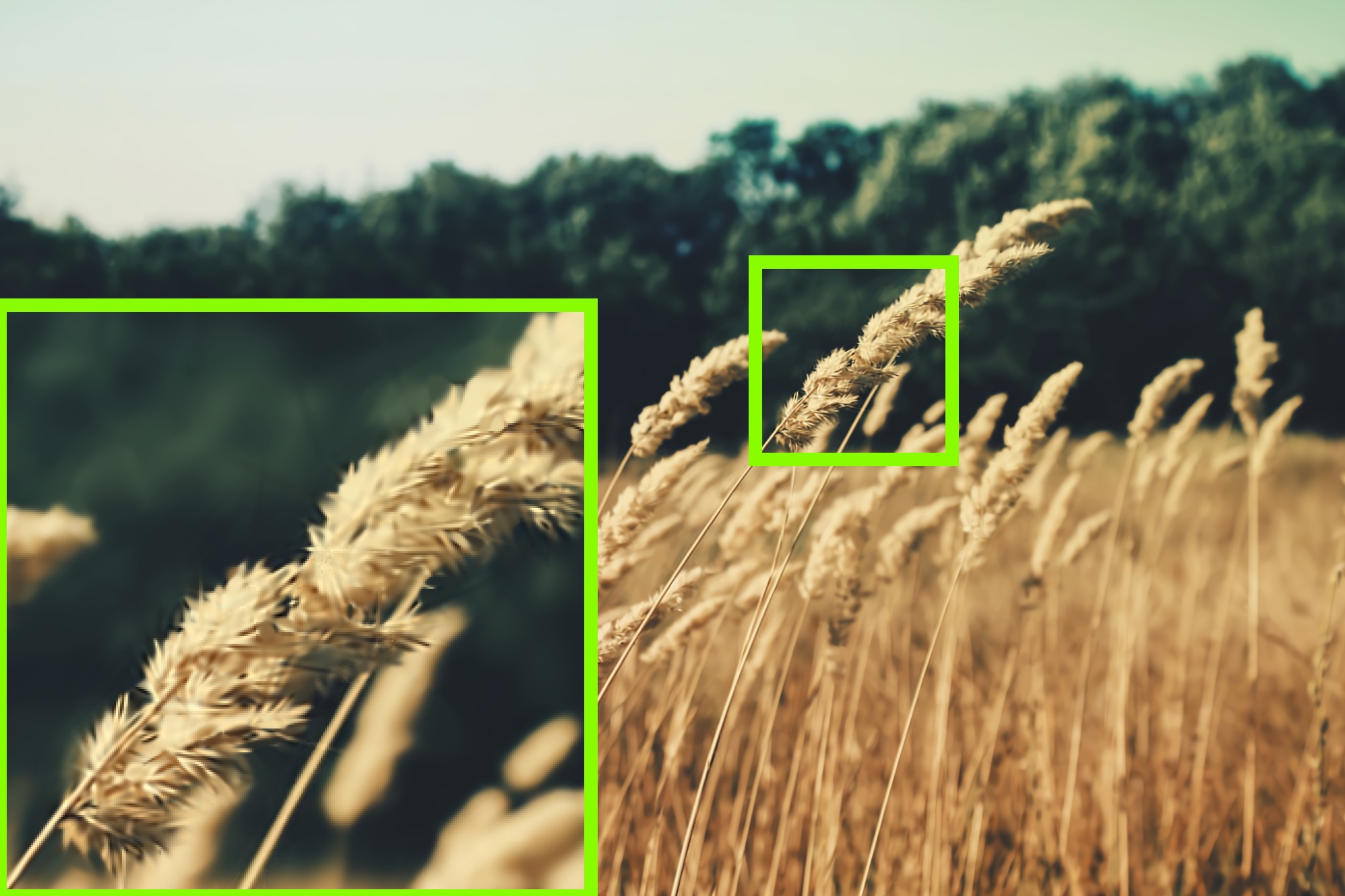}
  } \hspace{-0.18cm}
\subfloat[Reference]{
    \includegraphics[width=\ImageCompCLICSuppRes]{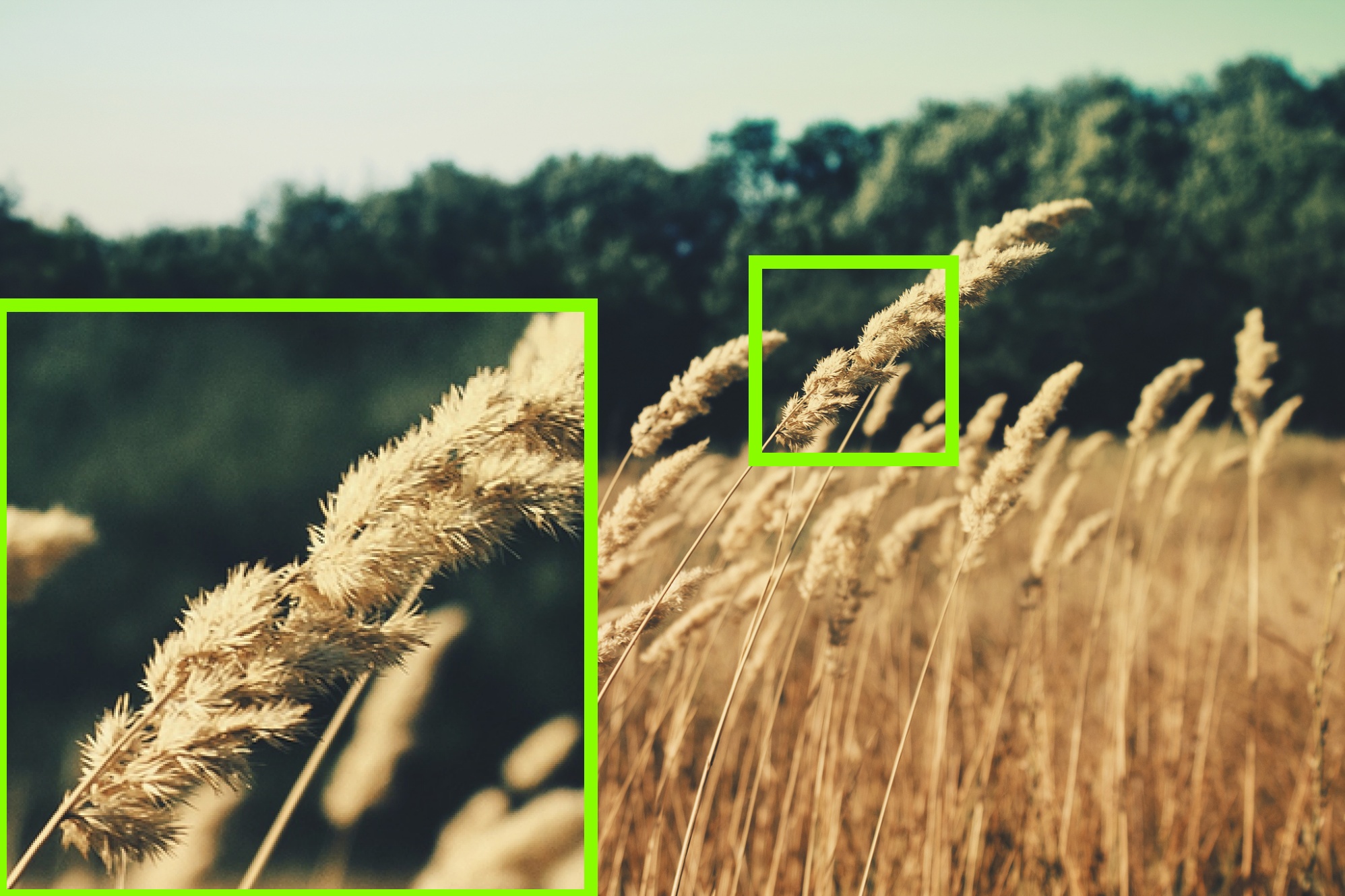}
  }
\Caption{\revise{Qualitative results on the professional validation split of the CLIC2020 dataset \cite{toderici2020workshop} (\Cref{sec:evaluation-image}).}}
{}
\end{figure*}
\clearpage
\section{Additional Joint Image Compression and Restoration Results}
\label{fig:application-image-restoration-supp-1}
\label{fig:application-image-restoration-supp-2}
\newcommand{\ImageRestorationSuppRes}{0.26\linewidth}
\begin{figure*}[h]
\centering
\subfloat{
    \includegraphics[width=\ImageRestorationSuppRes]{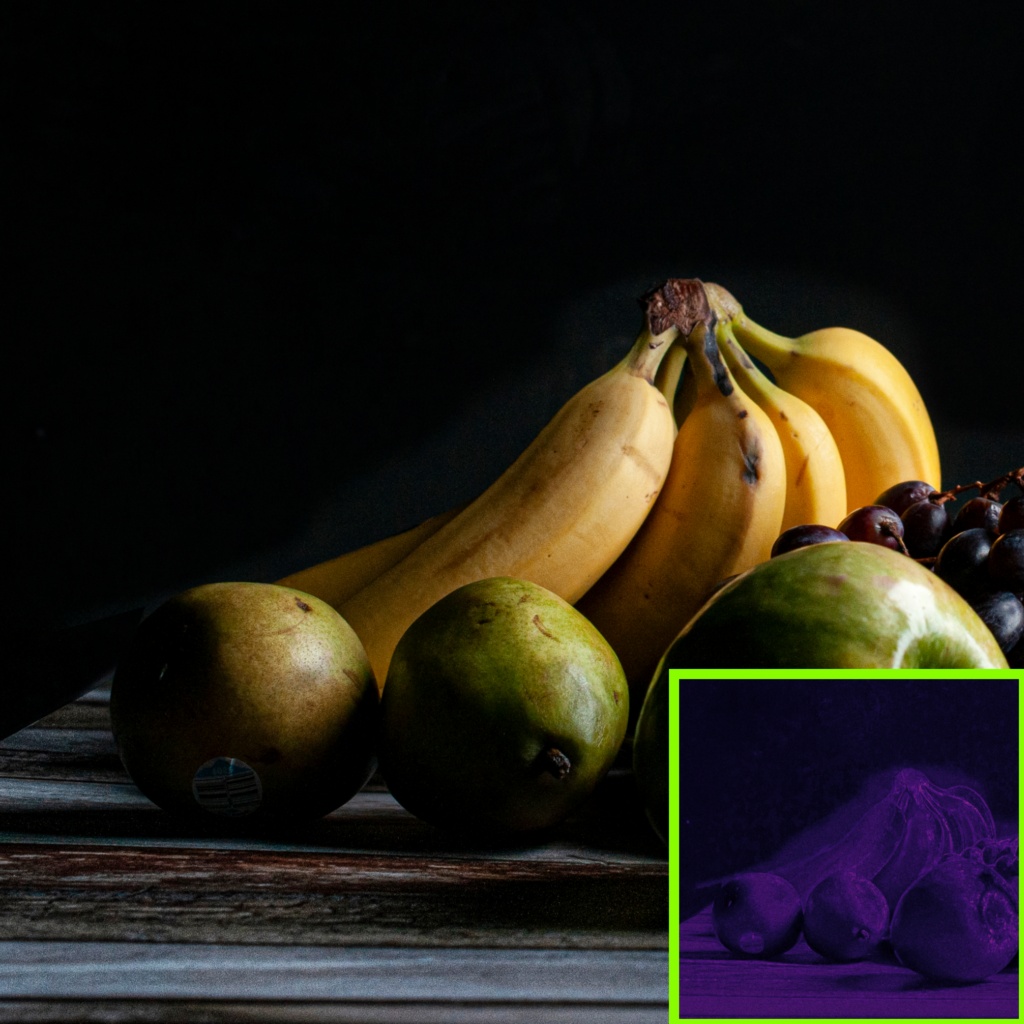}
  } \hspace{-0.18cm}
\subfloat{
    \includegraphics[width=\ImageRestorationSuppRes]{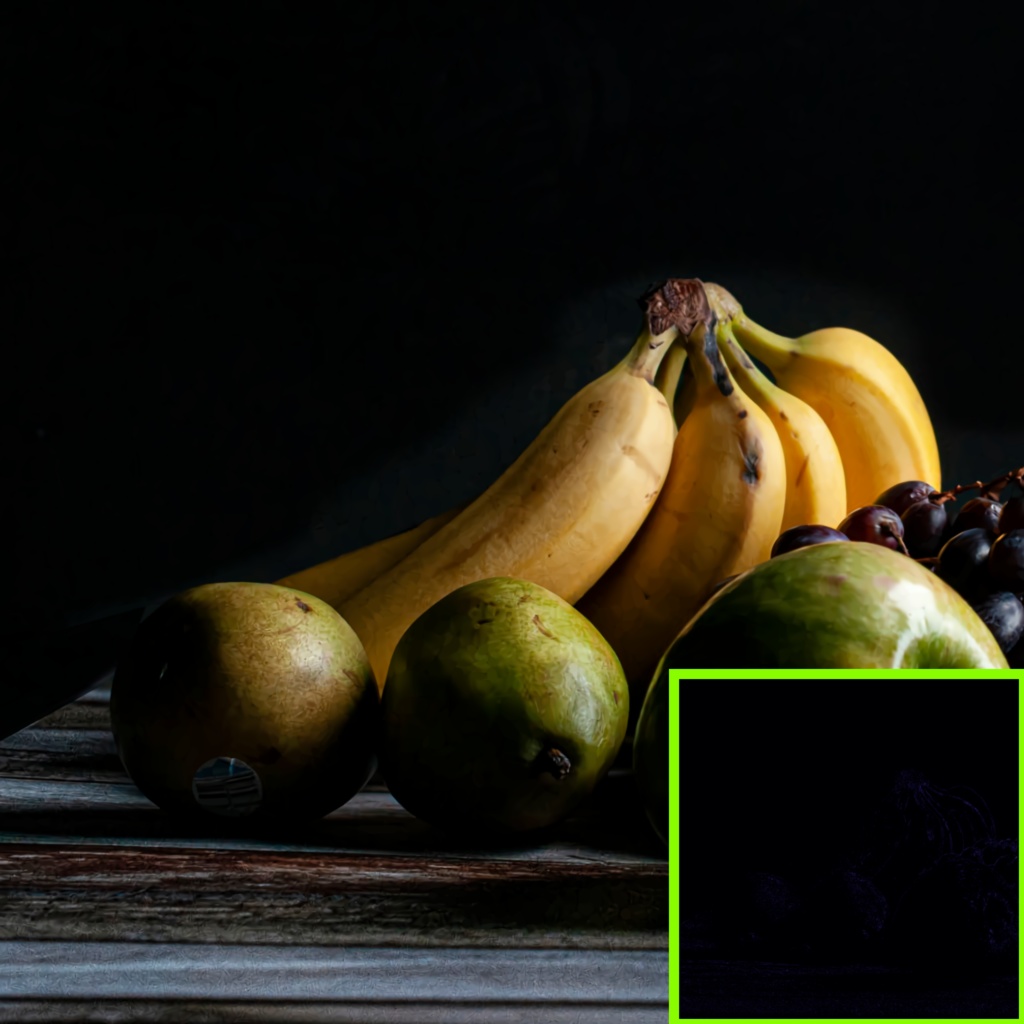}
  } \hspace{-0.18cm}
\subfloat{
    \includegraphics[width=\ImageRestorationSuppRes]{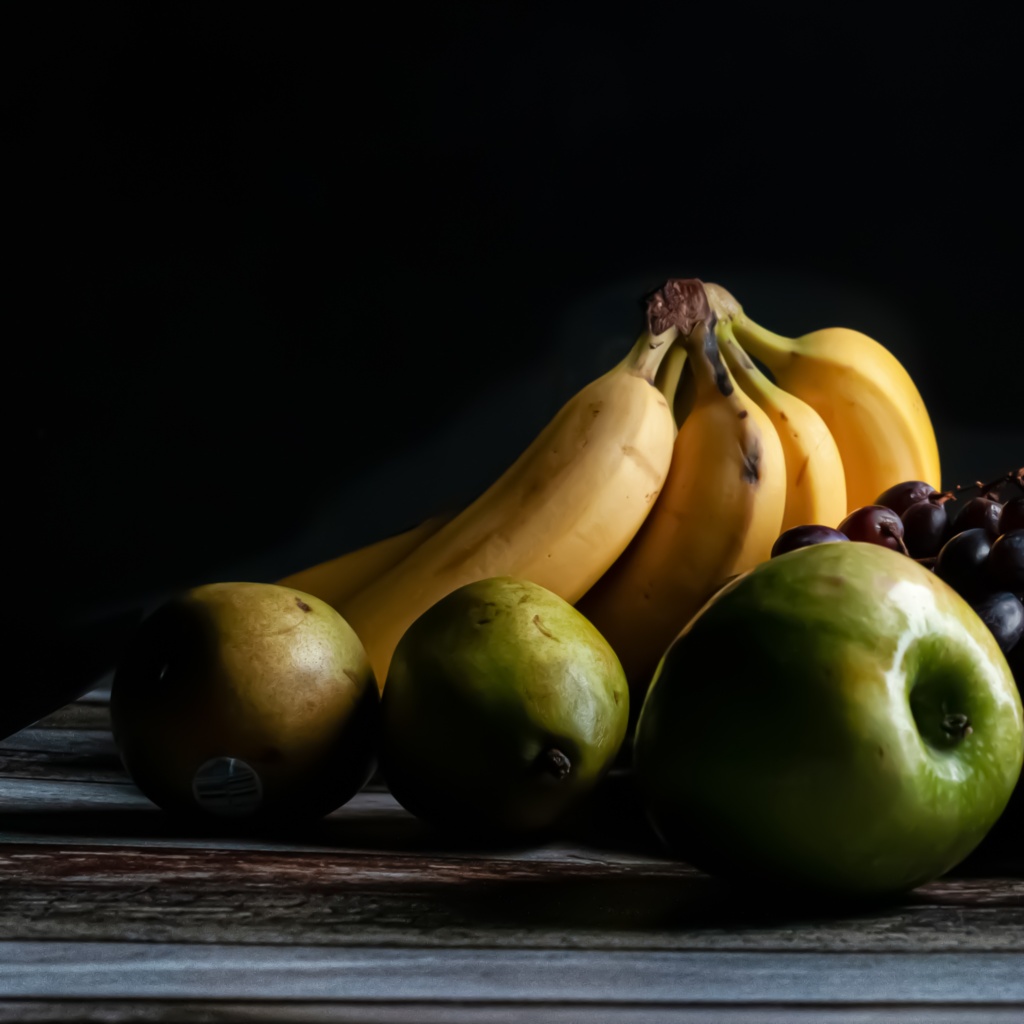}
  }
\vspace{0.2mm} \\
\subfloat{
    \includegraphics[width=\ImageRestorationSuppRes]{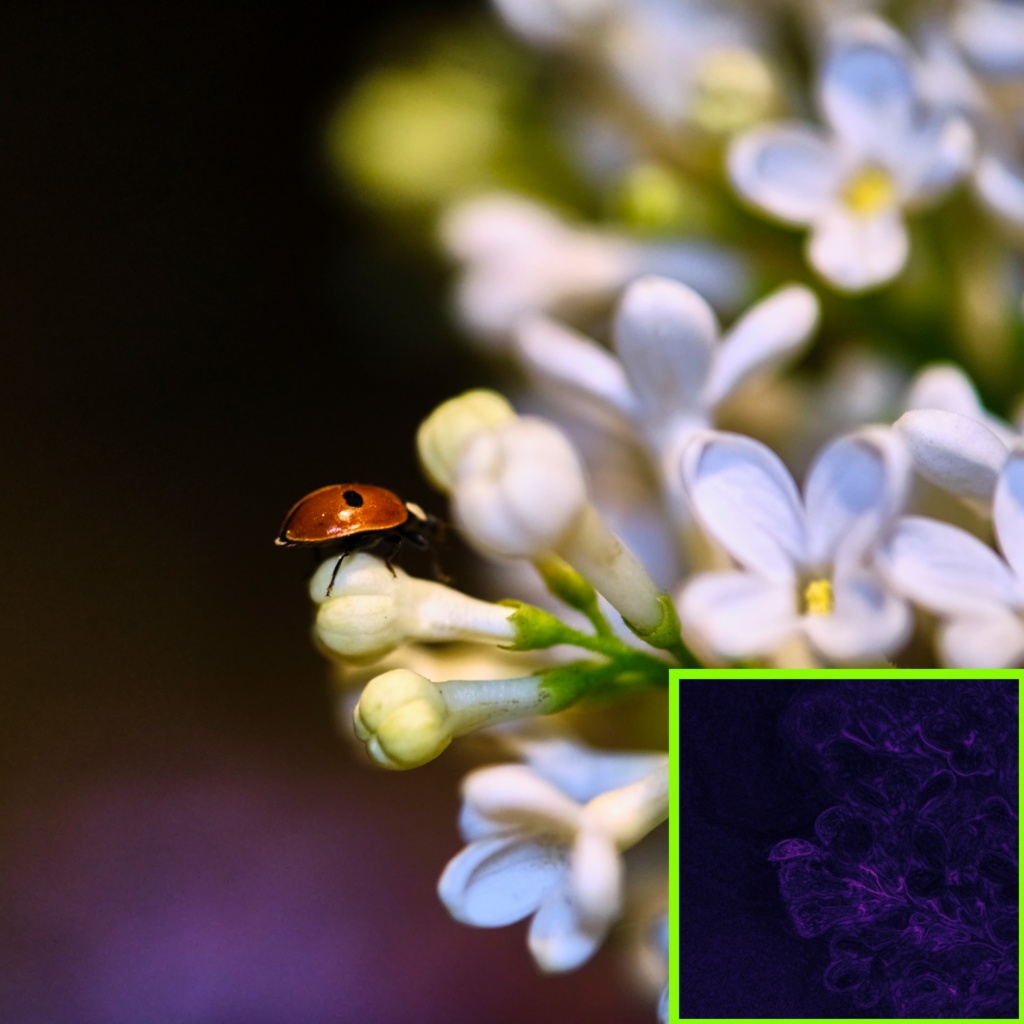}
  } \hspace{-0.18cm}
\subfloat{
    \includegraphics[width=\ImageRestorationSuppRes]{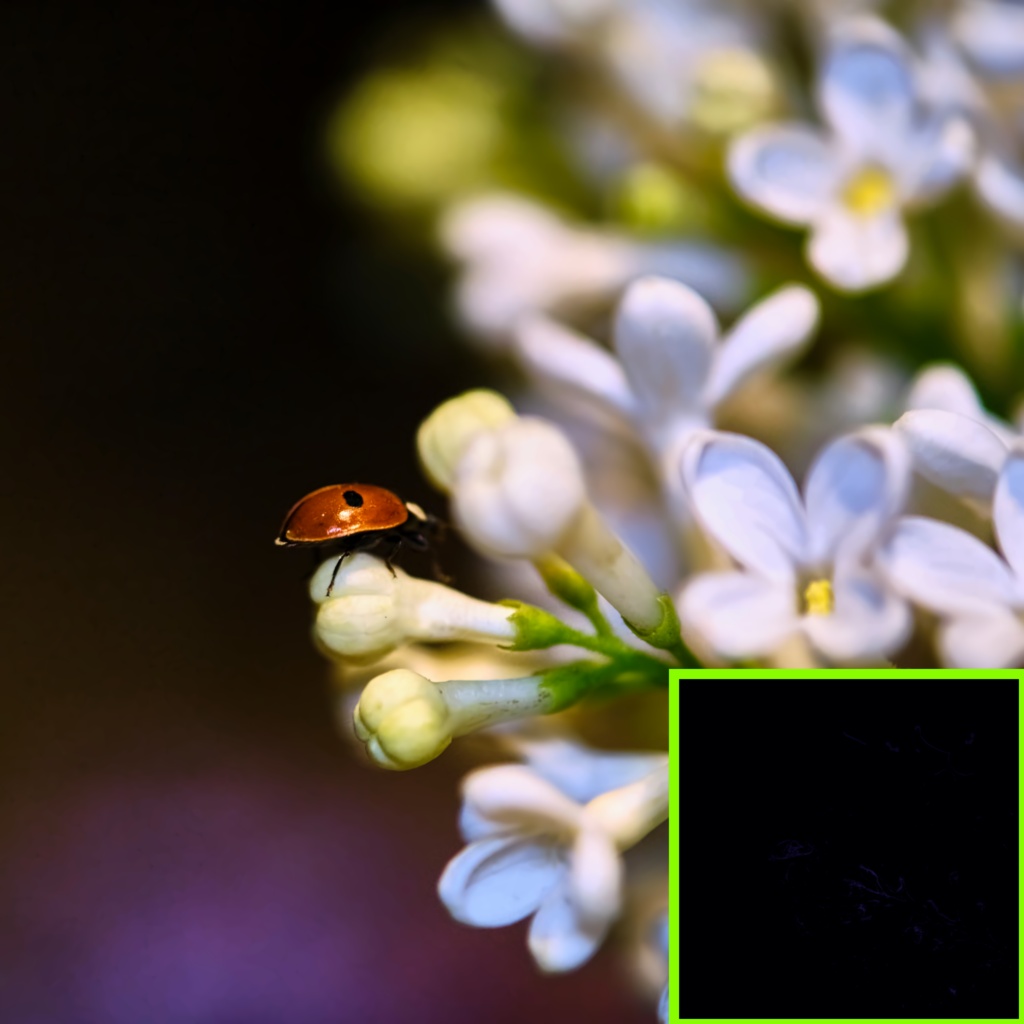}
  } \hspace{-0.18cm}
\subfloat{
    \includegraphics[width=\ImageRestorationSuppRes]{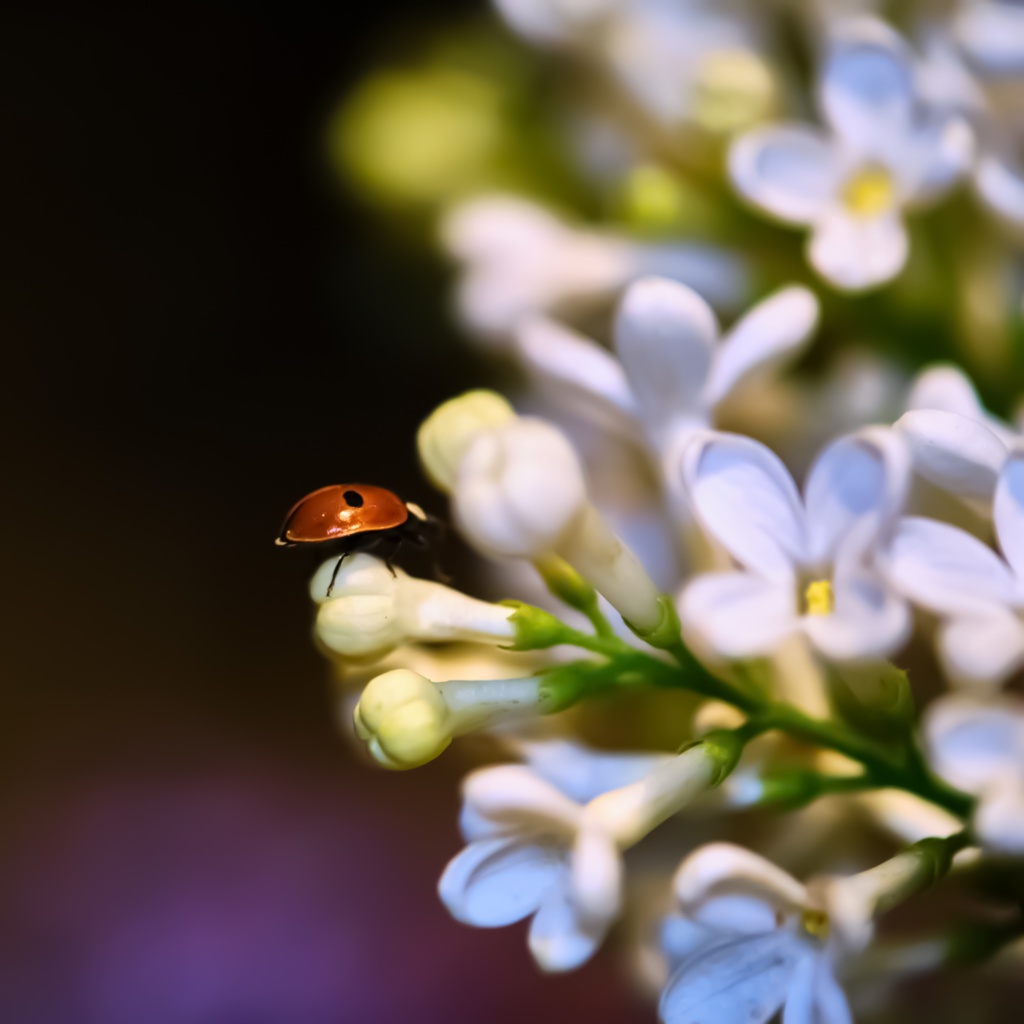}
  }
\vspace{0.2mm} \\
\subfloat{
    \includegraphics[width=\ImageRestorationSuppRes]{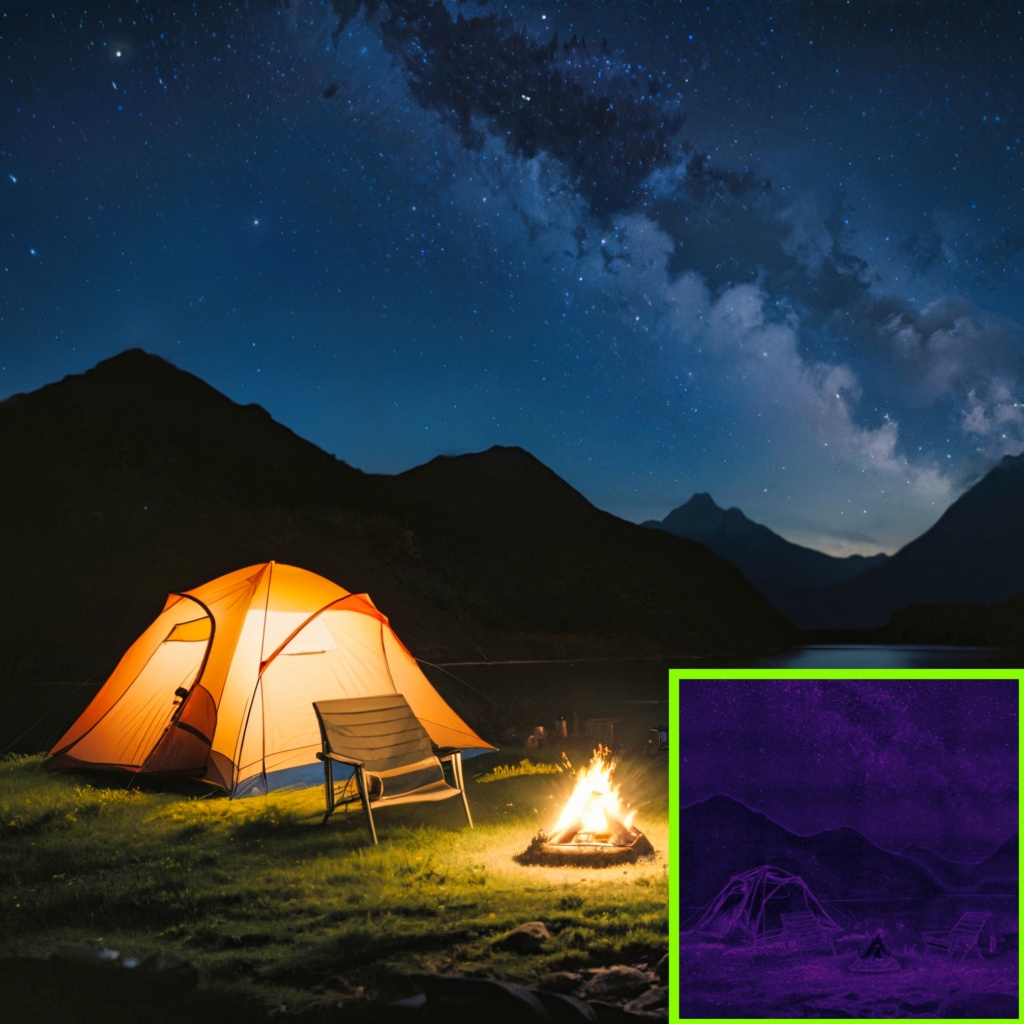}
  } \hspace{-0.18cm}
\subfloat{
    \includegraphics[width=\ImageRestorationSuppRes]{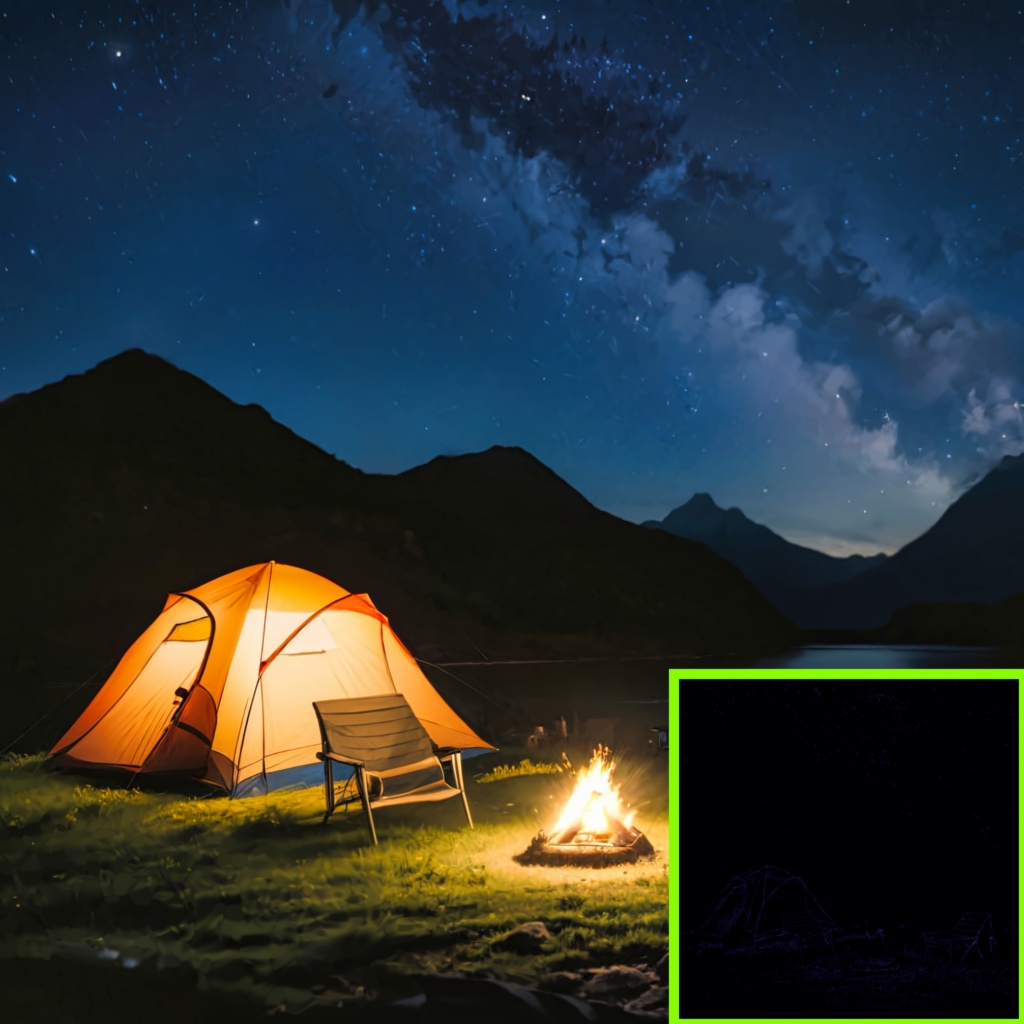}
  } \hspace{-0.18cm}
\subfloat{
    \includegraphics[width=\ImageRestorationSuppRes]{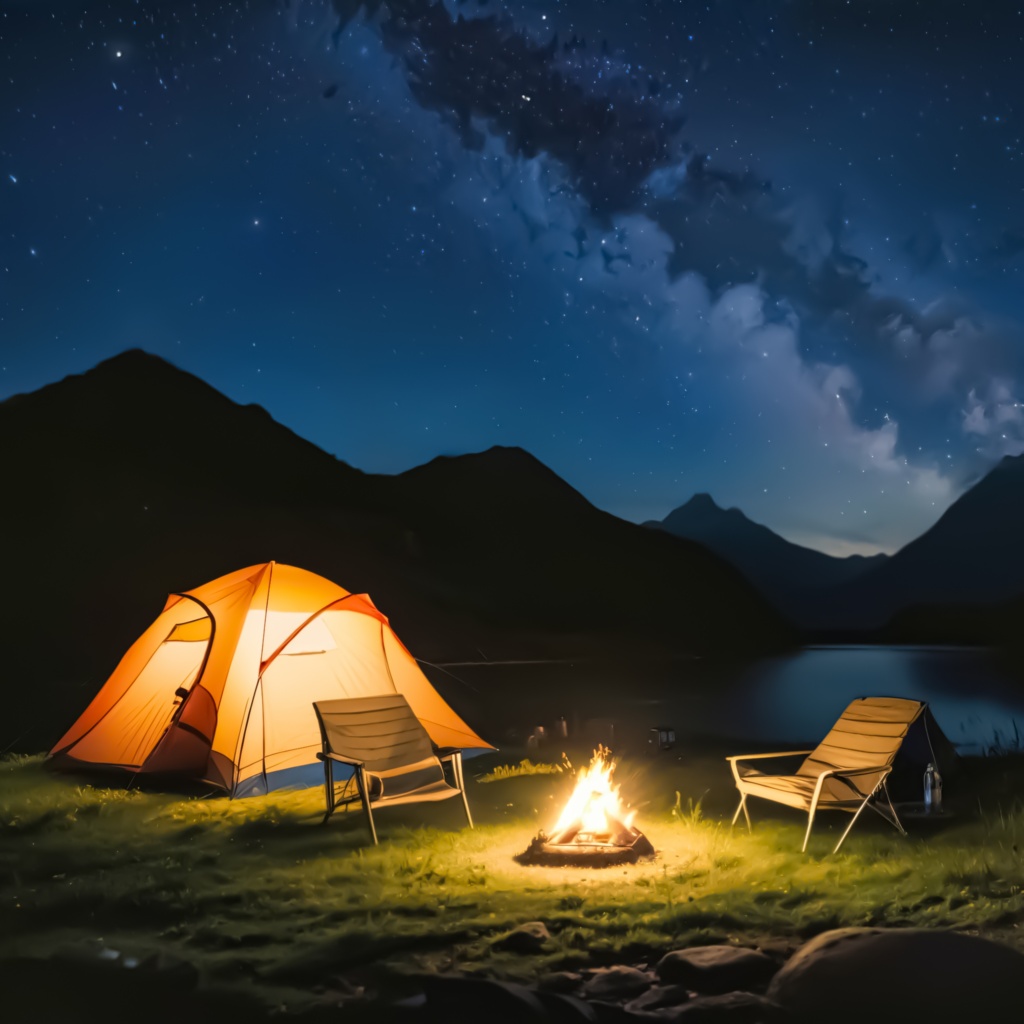}
  }
\vspace{0.2mm} \\
\setcounter{subfigure}{0}
\subfloat[Image with camera sensor noise]{
    \includegraphics[width=\ImageRestorationSuppRes]{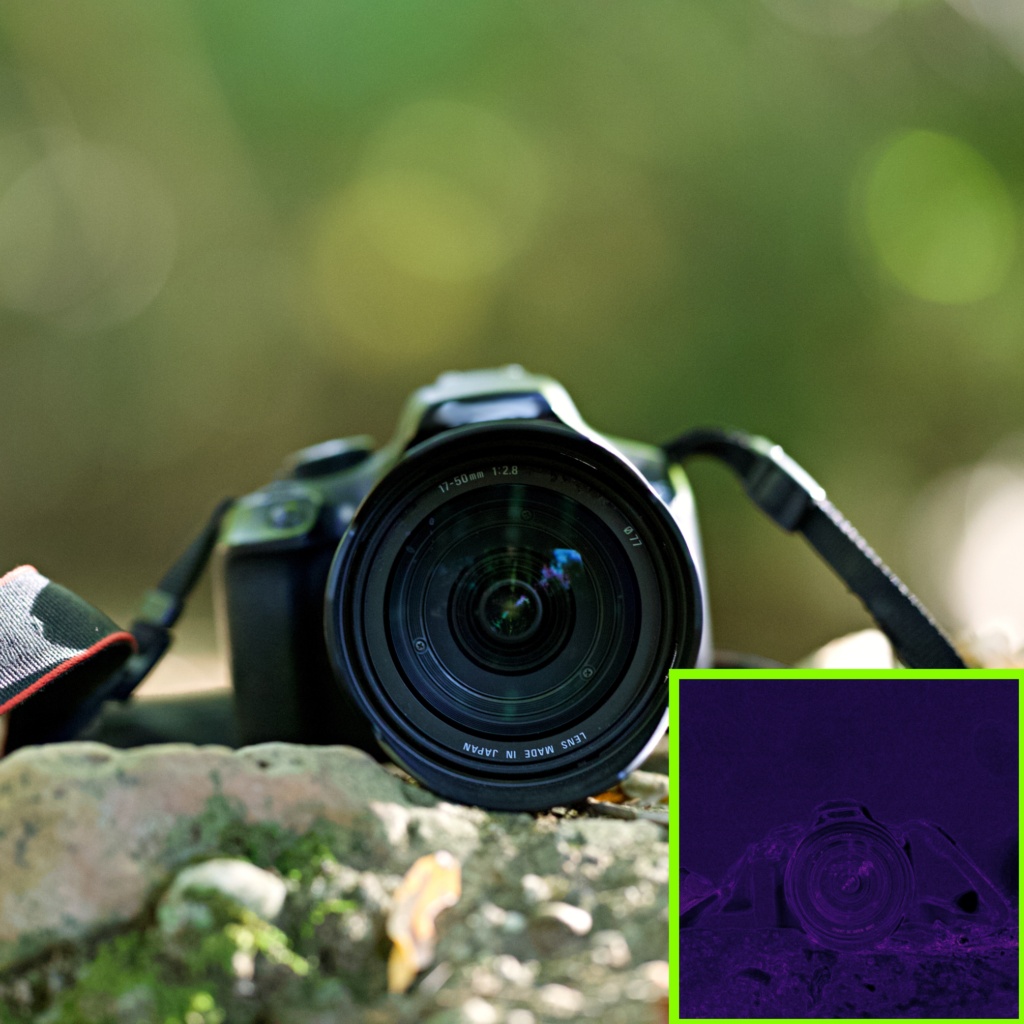}
  } \hspace{-0.18cm}
\subfloat[Image restored by \methodName]{
    \includegraphics[width=\ImageRestorationSuppRes]{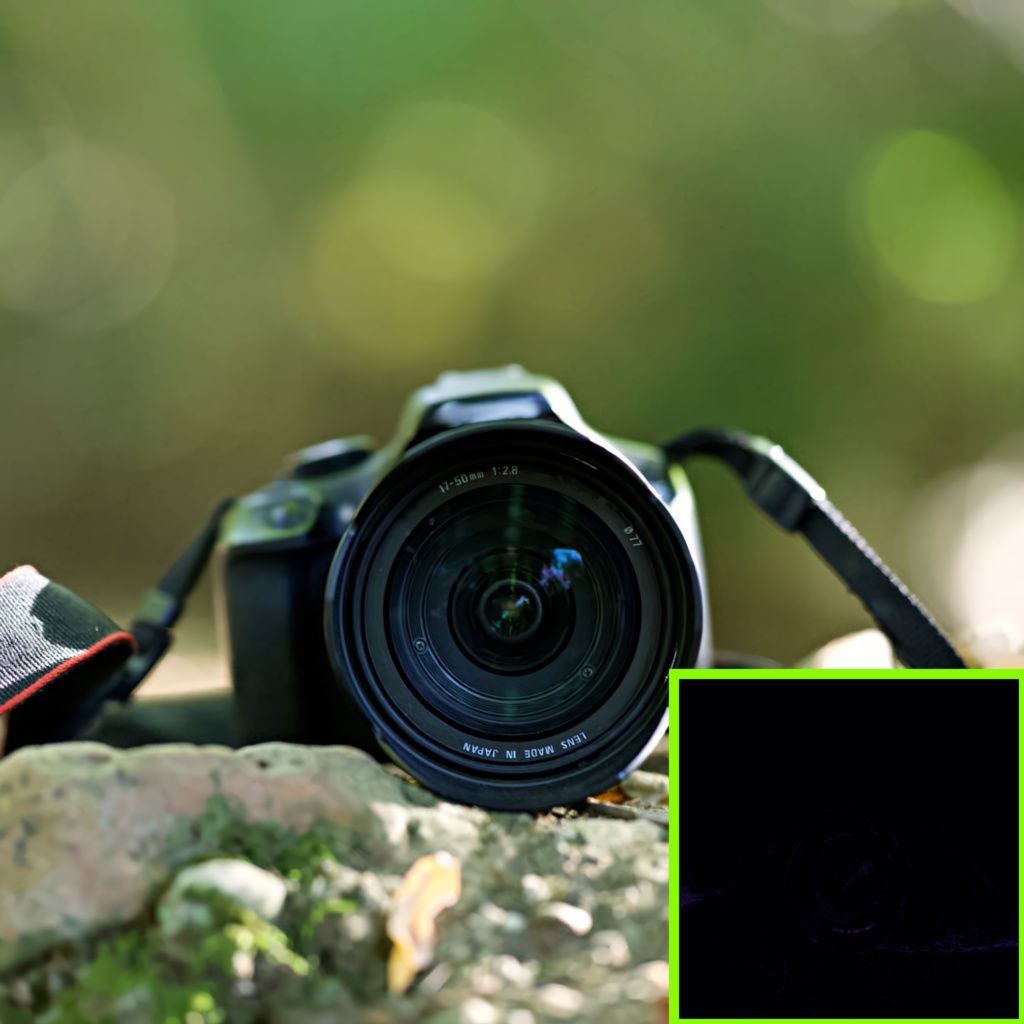}
  } \hspace{-0.18cm}
\subfloat[AI-denoised reference image]{
    \includegraphics[width=\ImageRestorationSuppRes]{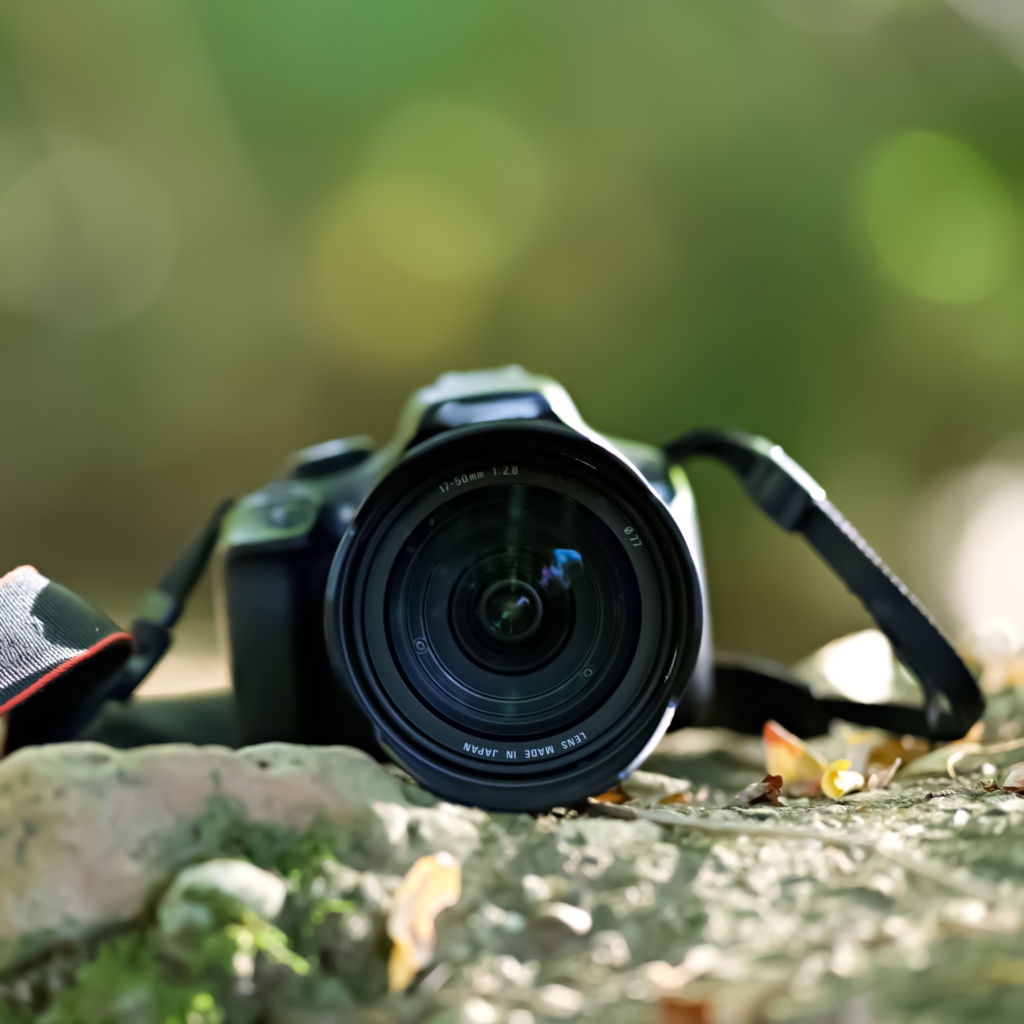}
  }
\Caption{Joint image compression and restoration (\Cref{sec:application-image-restoration}).}
{\revise{At low bitrates, \methodName effectively removes high-frequency artifacts from the input images while preserving detailed semantic content therein.}}
\end{figure*}
\begin{figure*}[h]
\centering
\subfloat{
    \includegraphics[width=\ImageRestorationSuppRes]{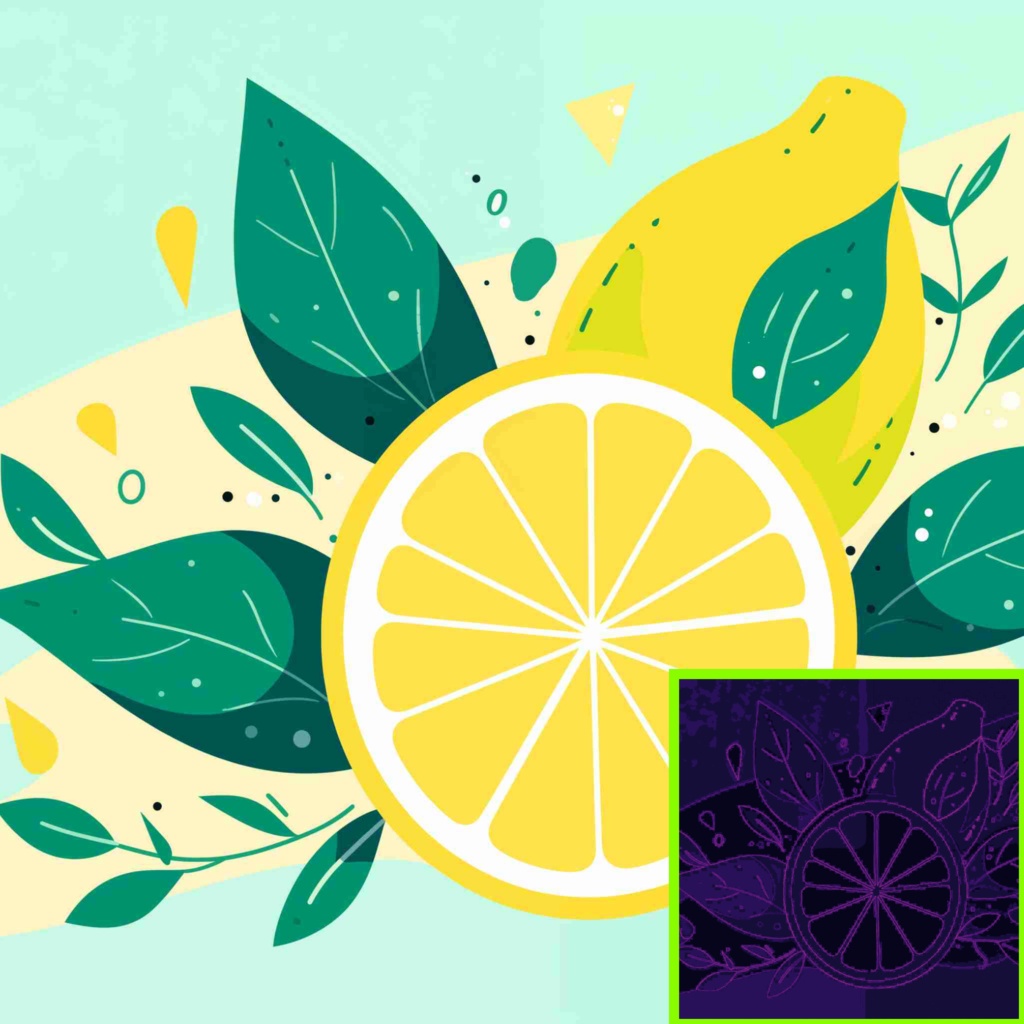}
  } \hspace{-0.18cm}
\subfloat{
    \includegraphics[width=\ImageRestorationSuppRes]{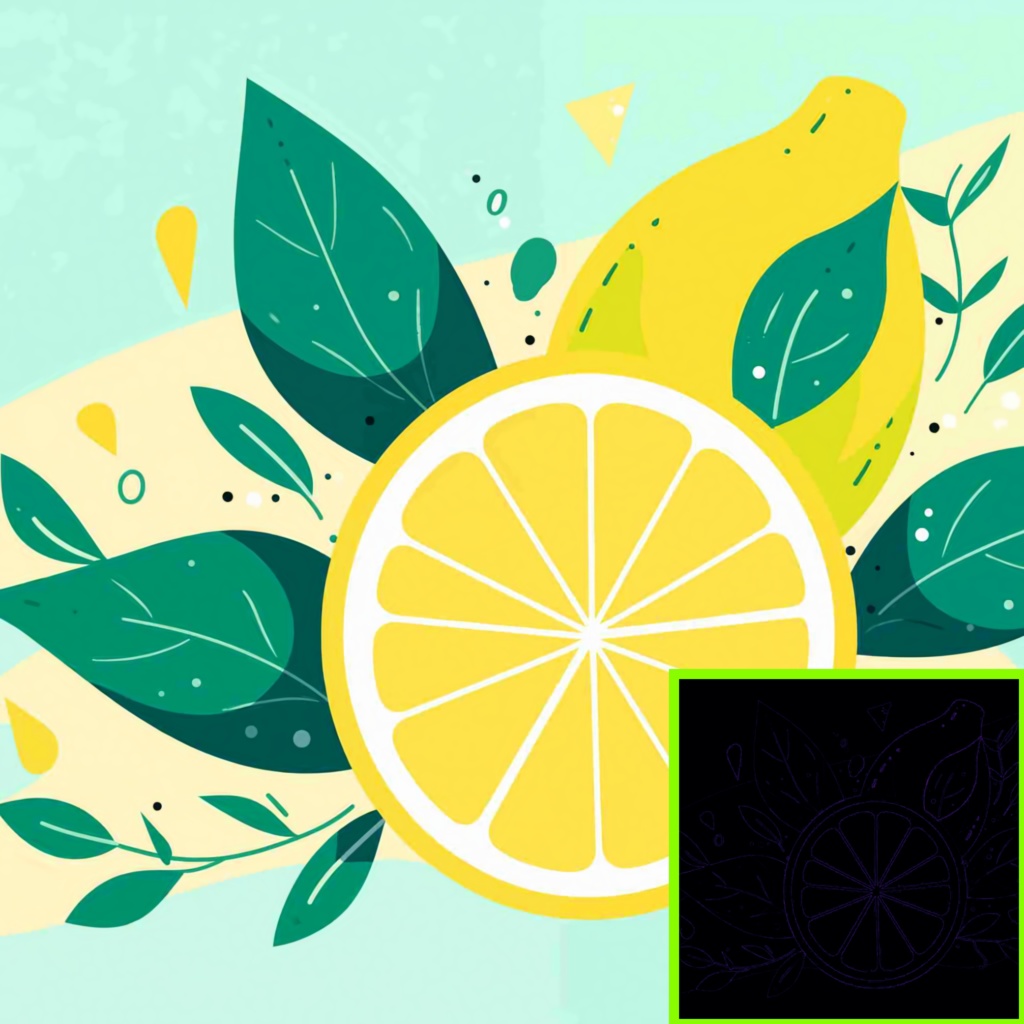}
  } \hspace{-0.18cm}
\subfloat{
    \includegraphics[width=\ImageRestorationSuppRes]{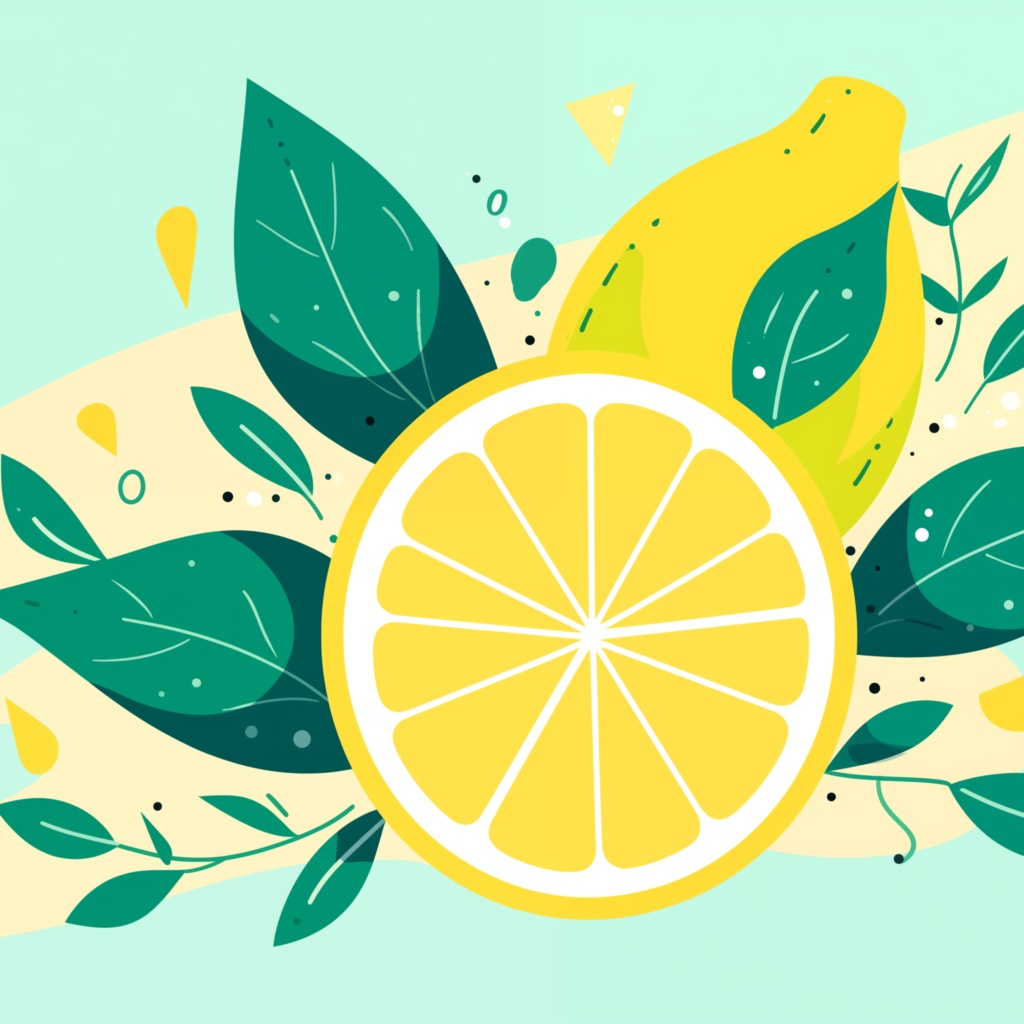}
  }
\vspace{0.2mm} \\
\subfloat{
    \includegraphics[width=\ImageRestorationSuppRes]{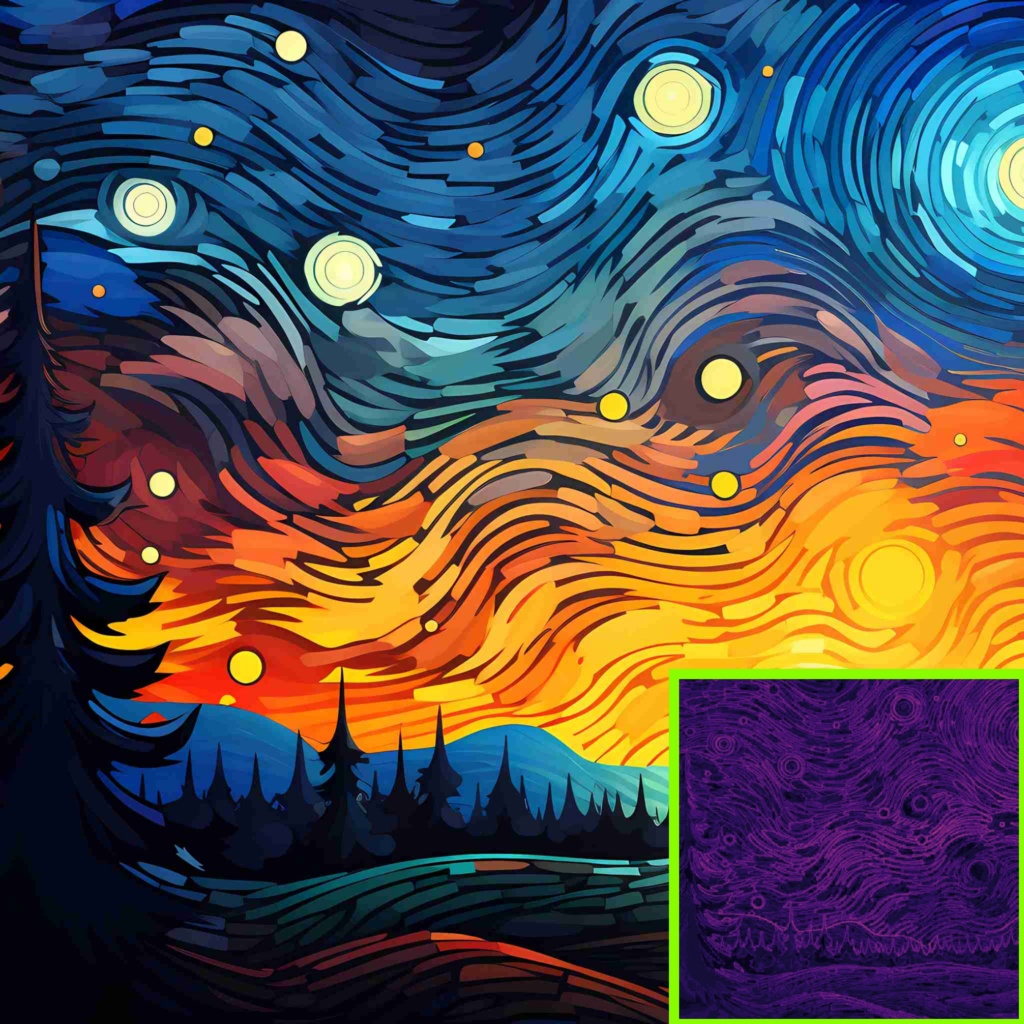}
  } \hspace{-0.18cm}
\subfloat{
    \includegraphics[width=\ImageRestorationSuppRes]{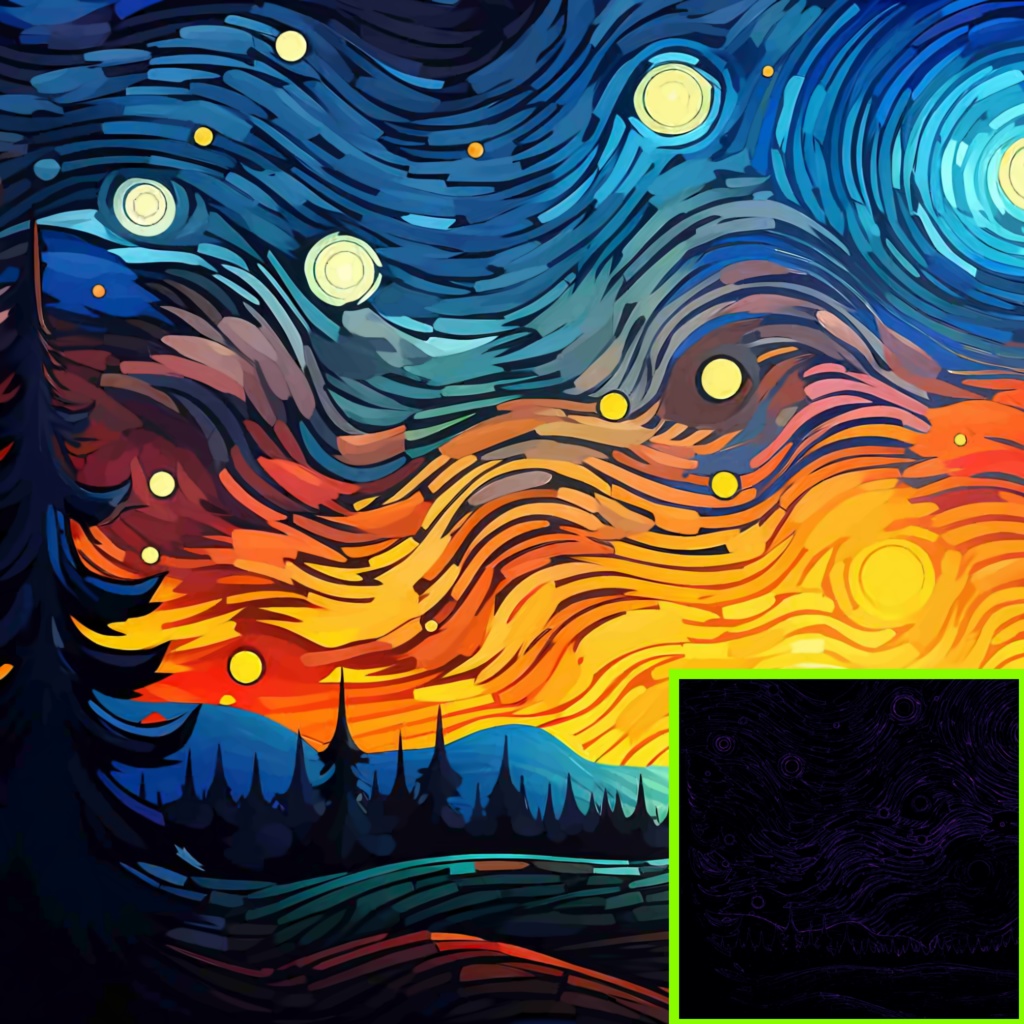}
  } \hspace{-0.18cm}
\subfloat{
    \includegraphics[width=\ImageRestorationSuppRes]{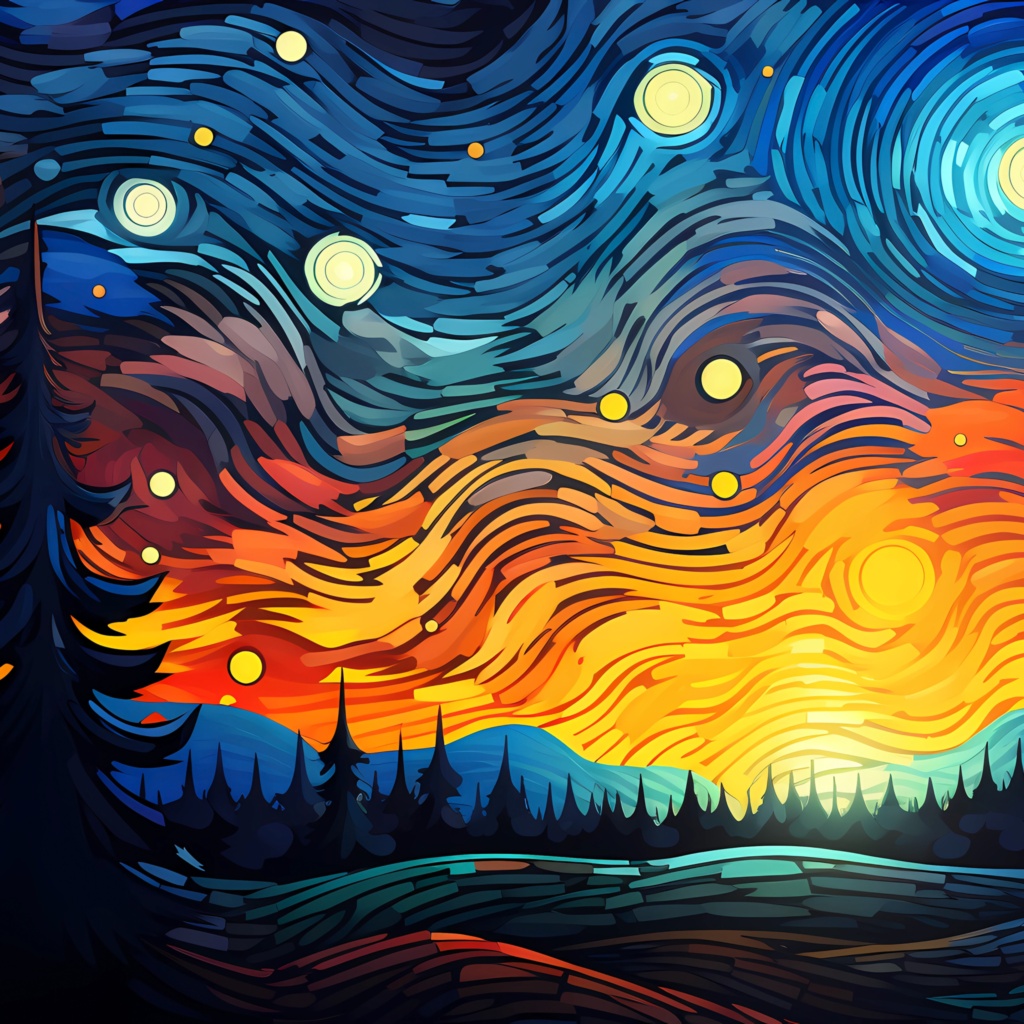}
  }
\vspace{0.2mm} \\
\subfloat{
    \includegraphics[width=\ImageRestorationSuppRes]{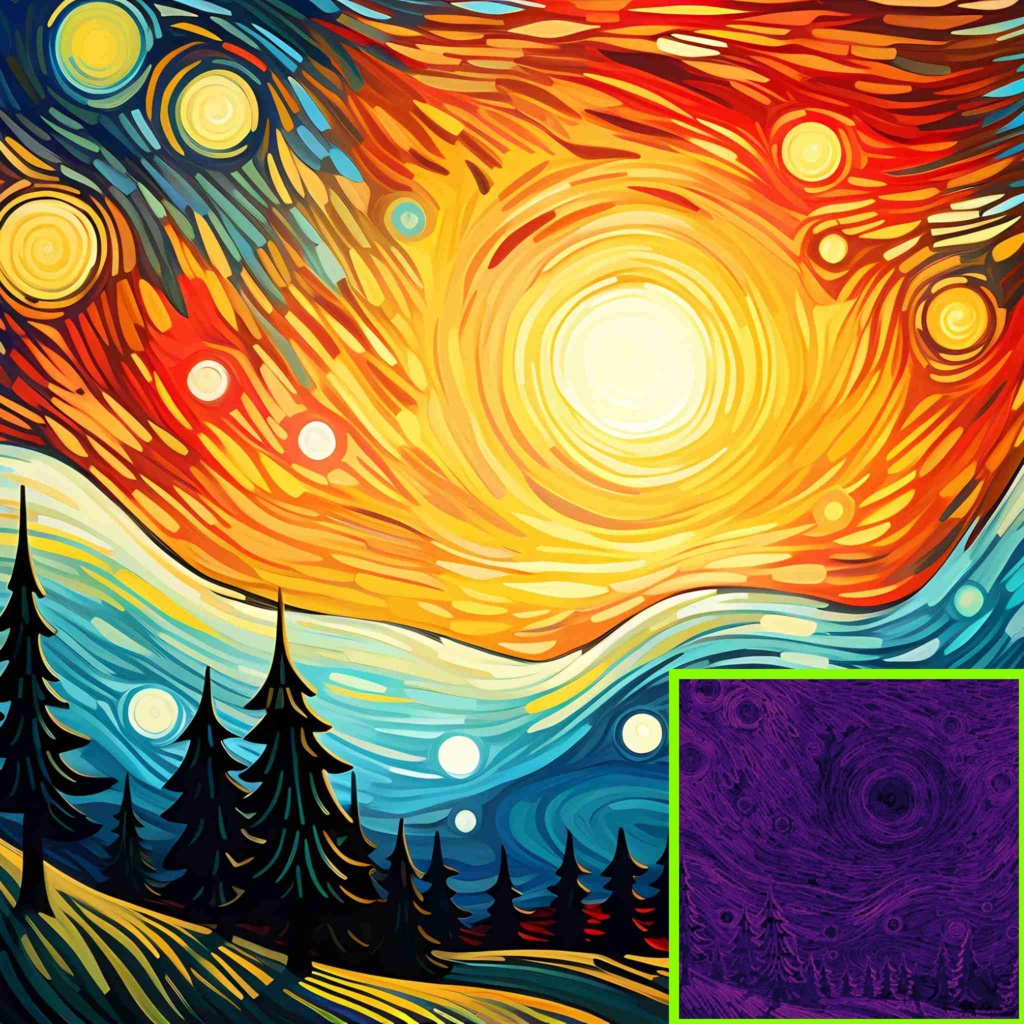}
  } \hspace{-0.18cm}
\subfloat{
    \includegraphics[width=\ImageRestorationSuppRes]{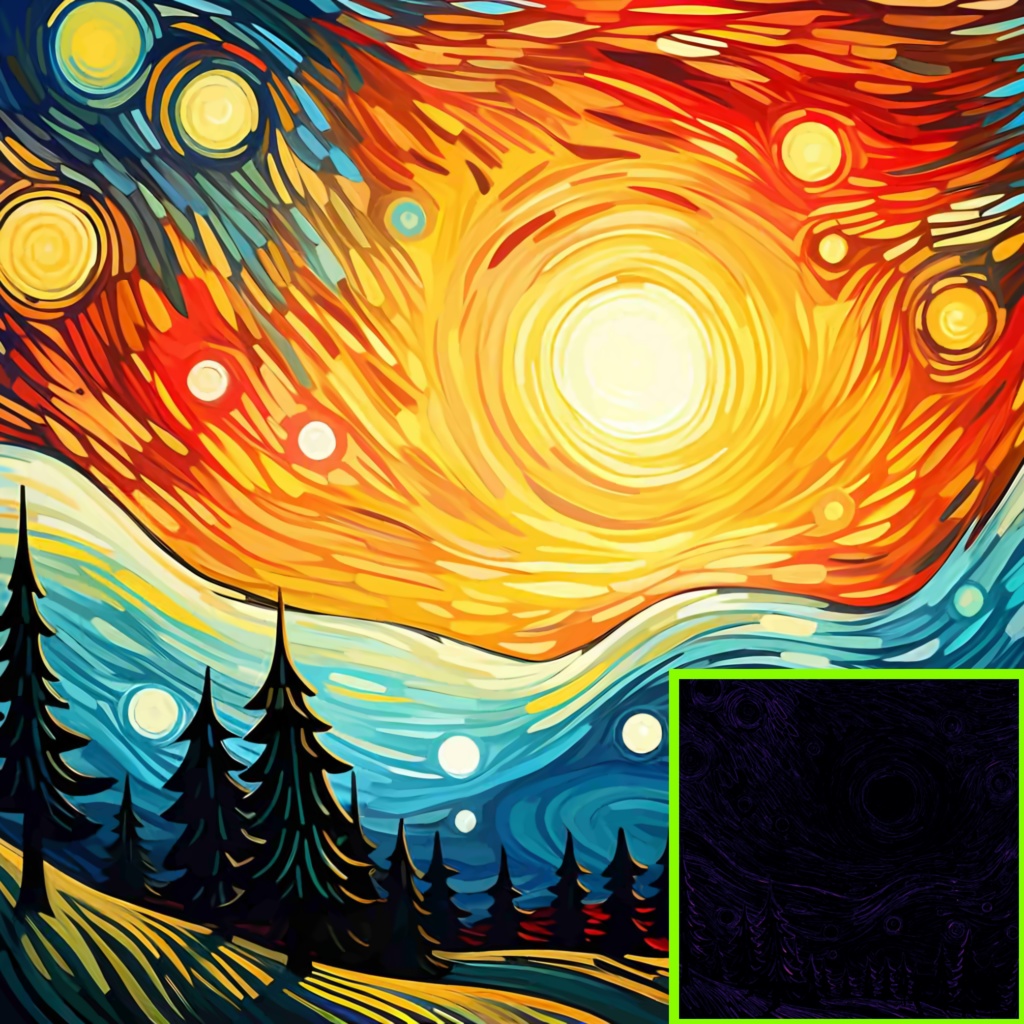}
  } \hspace{-0.18cm}
\subfloat{
    \includegraphics[width=\ImageRestorationSuppRes]{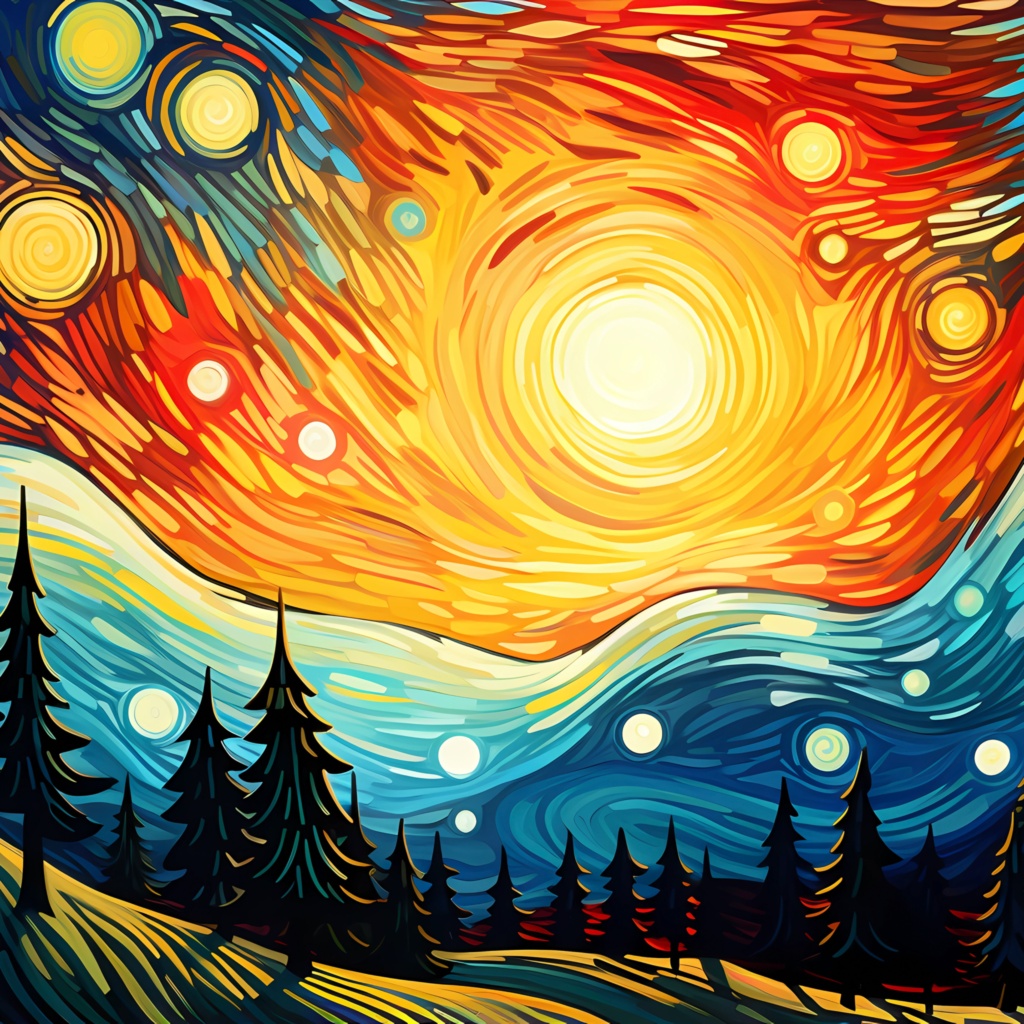}
  }
\vspace{0.2mm} \\
\setcounter{subfigure}{0}
\subfloat[Image compressed by JPEG]{
    \includegraphics[width=\ImageRestorationSuppRes]{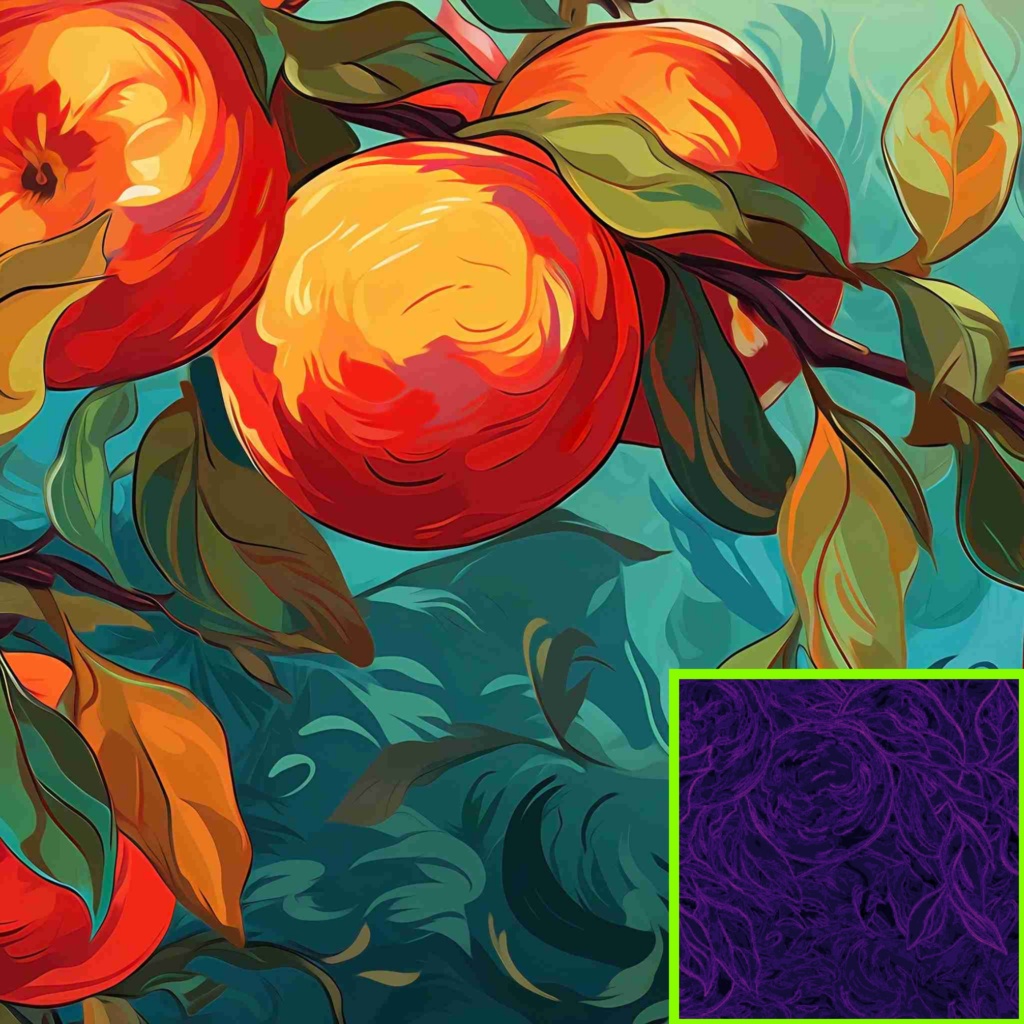}
  } \hspace{-0.18cm}
\subfloat[Image restored by \methodName]{
    \includegraphics[width=\ImageRestorationSuppRes]{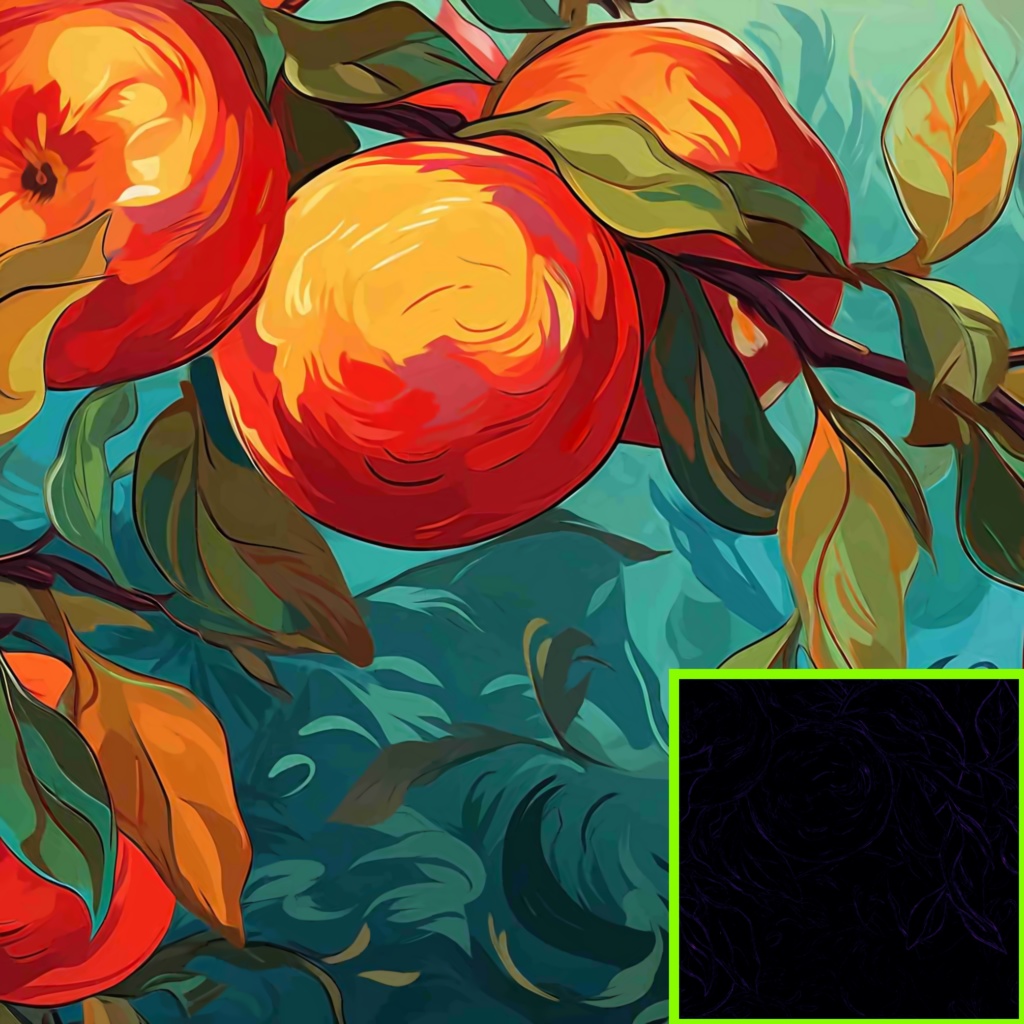}
  } \hspace{-0.18cm}
\subfloat[Uncompressed reference image]{
    \includegraphics[width=\ImageRestorationSuppRes]{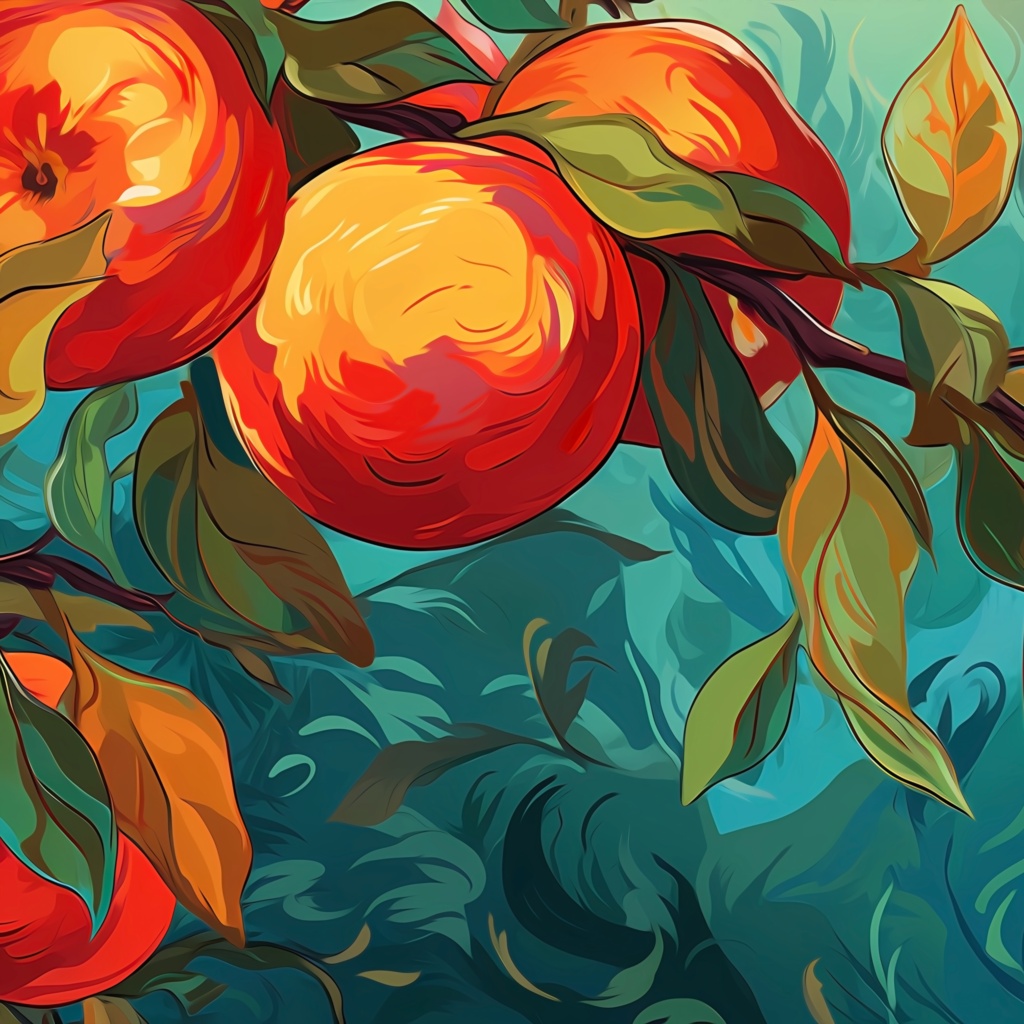}
  }
\Caption{Joint image compression and restoration (\Cref{sec:application-image-restoration}).}
{\revise{At low bitrates, \methodName effectively removes high-frequency artifacts from the input images while preserving detailed semantic content therein.}}
\end{figure*}
\clearpage
\section{\revise{Additional Texture Compression Results.}}
\label{fig:evaluation-texture-supp-1}
\label{fig:evaluation-texture-supp-2}
\label{fig:evaluation-texture-supp-3}
\newcommand{\TextureCompSuppRes}{0.176\linewidth}
\begin{figure*}[h]
\centering
\subfloat{
    \includegraphics[width=\TextureCompSuppRes]{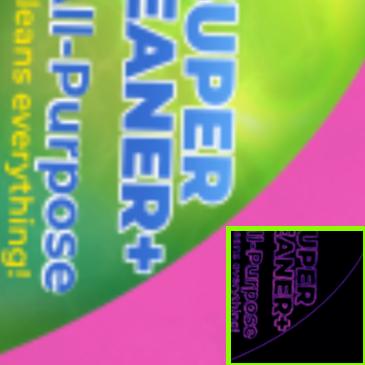}
  } \hspace{-0.18cm}
\subfloat{
    \includegraphics[width=\TextureCompSuppRes]{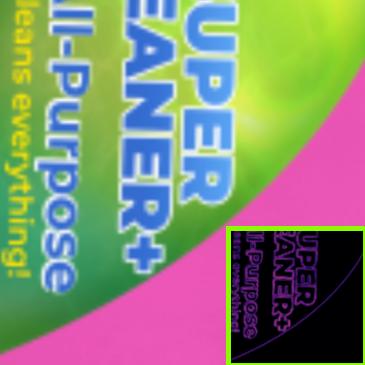}
  } \hspace{-0.18cm}
\subfloat{
    \includegraphics[width=\TextureCompSuppRes]{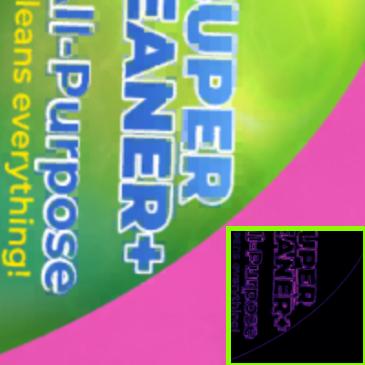}
  } \hspace{-0.18cm}
\subfloat{
    \includegraphics[width=\TextureCompSuppRes]{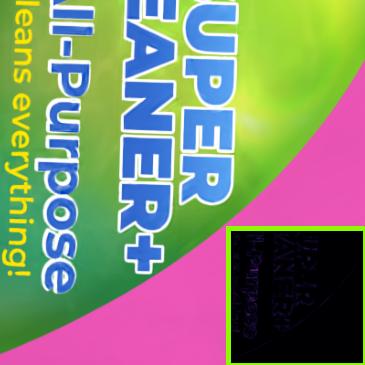}
  } \hspace{-0.18cm}
\subfloat{
    \includegraphics[width=\TextureCompSuppRes]{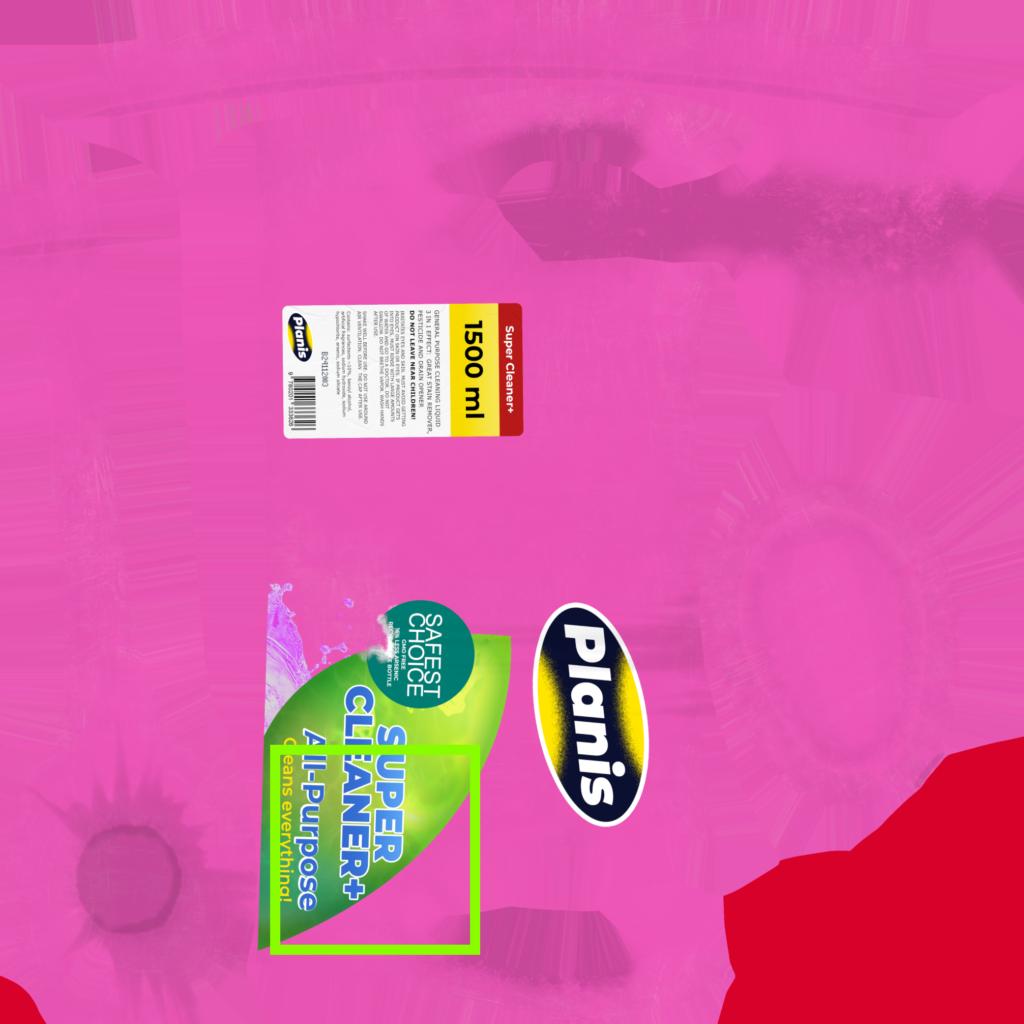}
  }
\vspace{0.2mm} \\
\subfloat{
    \includegraphics[width=\TextureCompSuppRes]{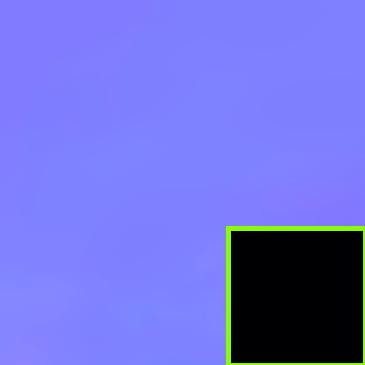}
  } \hspace{-0.18cm}
\subfloat{
    \includegraphics[width=\TextureCompSuppRes]{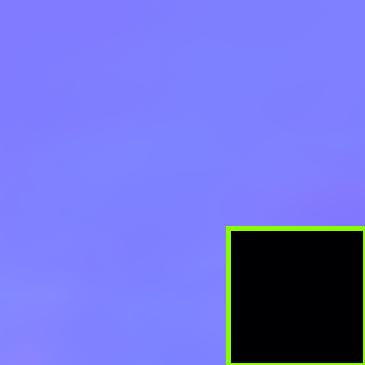}
  } \hspace{-0.18cm}
\subfloat{
    \includegraphics[width=\TextureCompSuppRes]{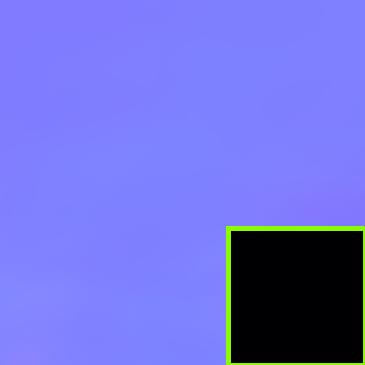}
  } \hspace{-0.18cm}
\subfloat{
    \includegraphics[width=\TextureCompSuppRes]{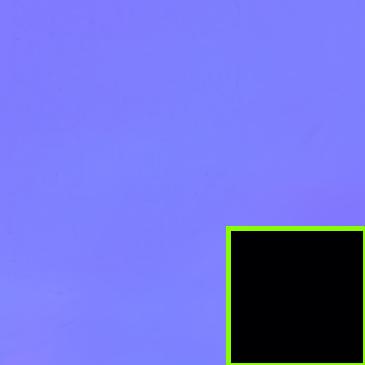}
  } \hspace{-0.18cm}
\subfloat{
    \includegraphics[width=\TextureCompSuppRes]{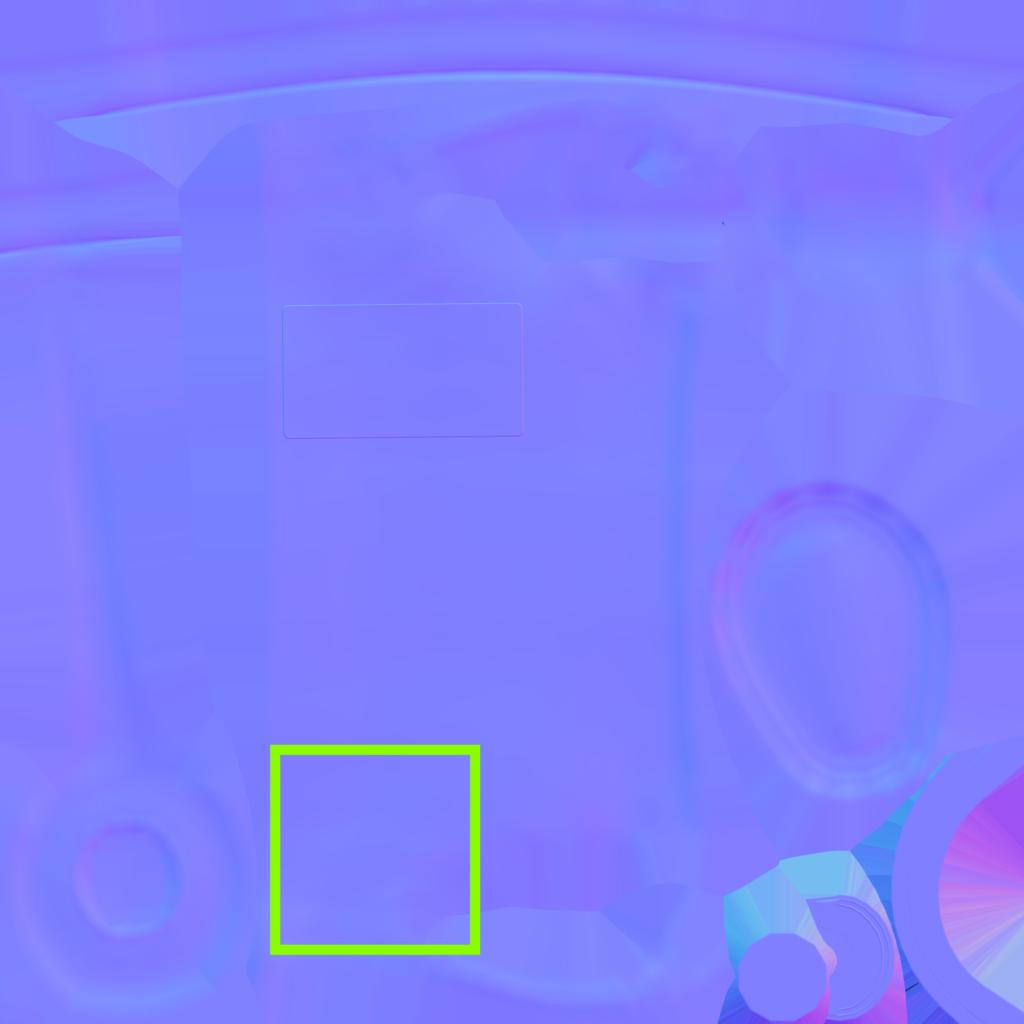}
  }
\vspace{0.2mm} \\
\subfloat{
    \includegraphics[width=\TextureCompSuppRes]{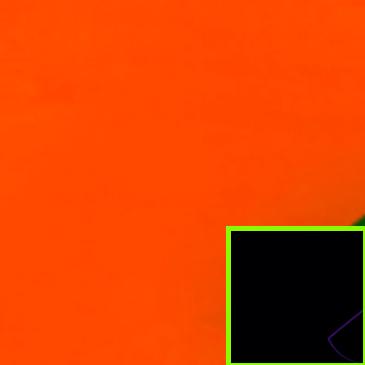}
  } \hspace{-0.18cm}
\subfloat{
    \includegraphics[width=\TextureCompSuppRes]{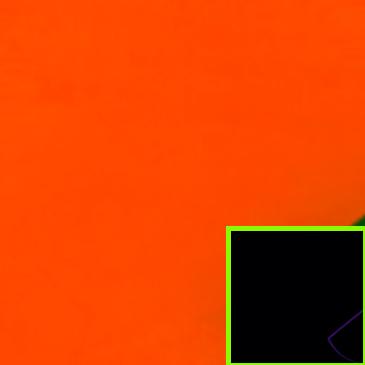}
  } \hspace{-0.18cm}
\subfloat{
    \includegraphics[width=\TextureCompSuppRes]{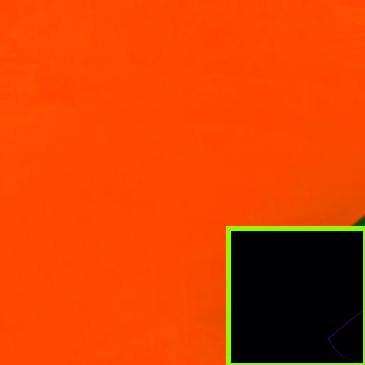}
  } \hspace{-0.18cm}
\subfloat{
    \includegraphics[width=\TextureCompSuppRes]{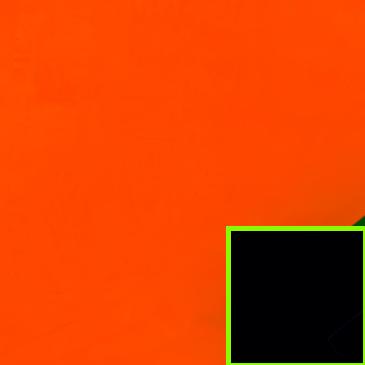}
  } \hspace{-0.18cm}
\subfloat{
    \includegraphics[width=\TextureCompSuppRes]{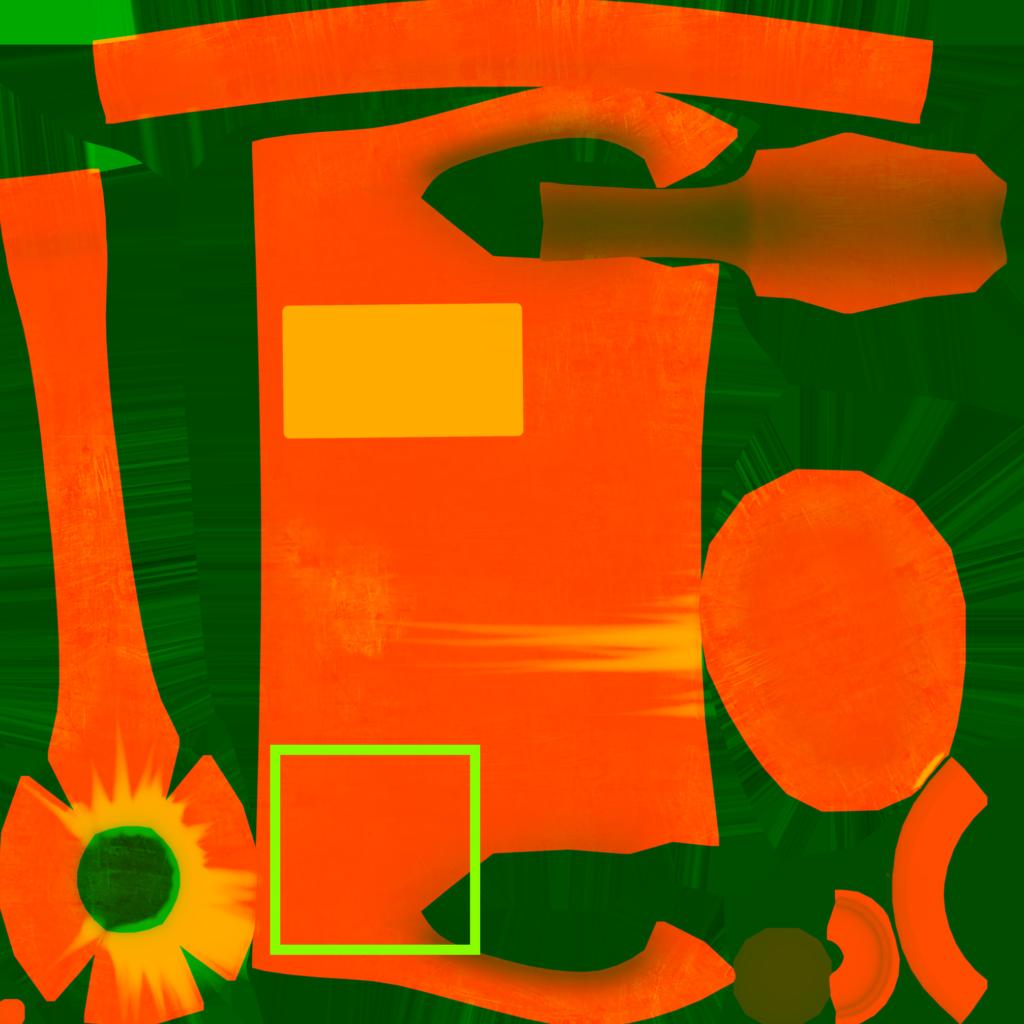}
  }
\vspace{0.2mm} \\
\subfloat{
    \includegraphics[width=\TextureCompSuppRes]{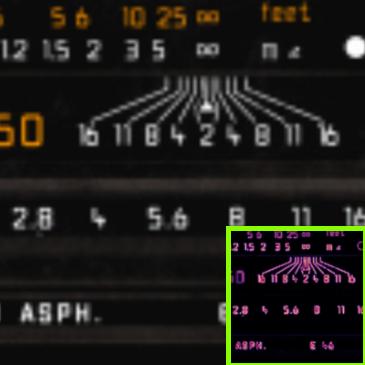}
  } \hspace{-0.18cm}
\subfloat{
    \includegraphics[width=\TextureCompSuppRes]{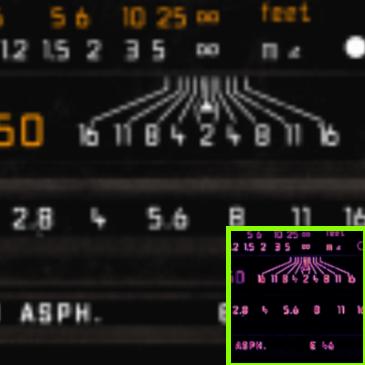}
  } \hspace{-0.18cm}
\subfloat{
    \includegraphics[width=\TextureCompSuppRes]{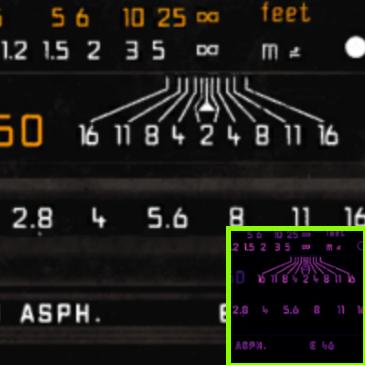}
  } \hspace{-0.18cm}
\subfloat{
    \includegraphics[width=\TextureCompSuppRes]{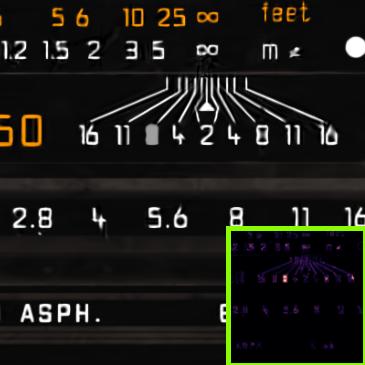}
  } \hspace{-0.18cm}
\subfloat{
    \includegraphics[width=\TextureCompSuppRes]{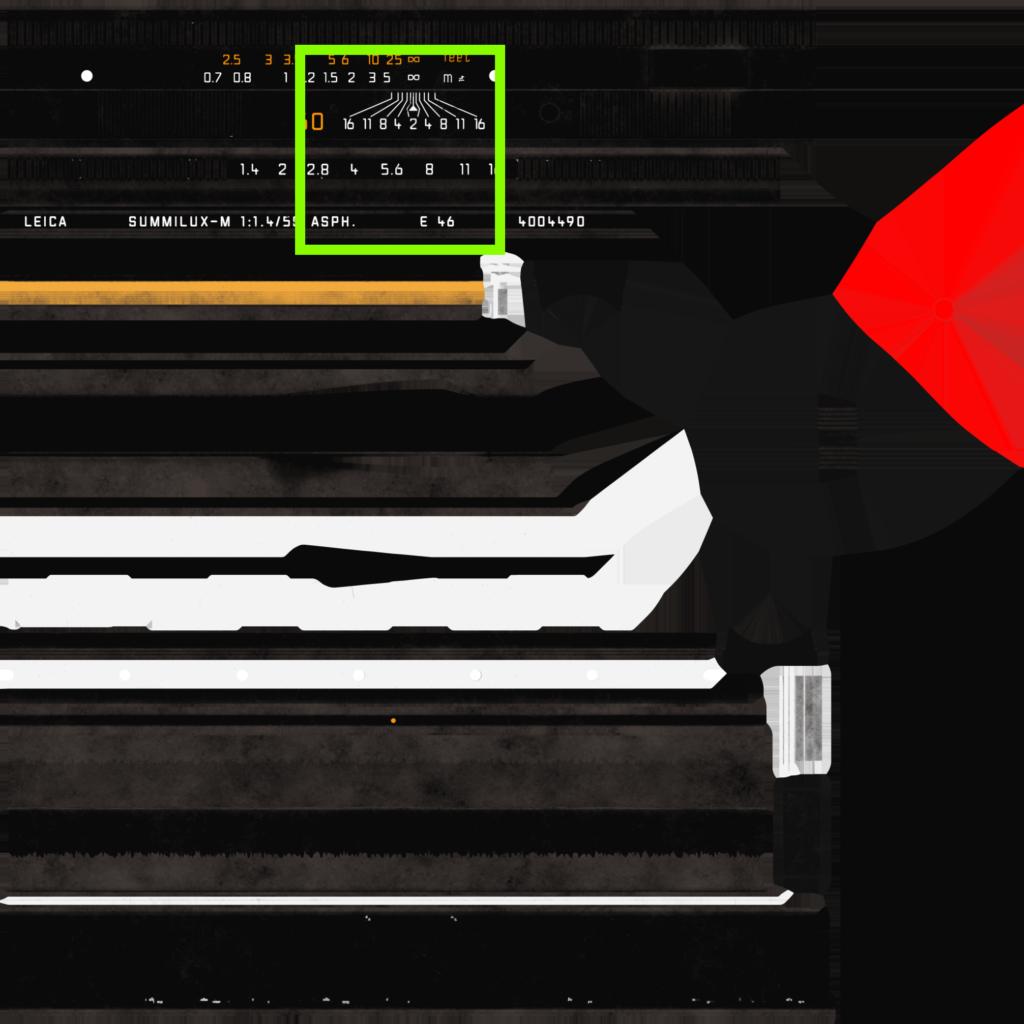}
  }
\vspace{0.2mm} \\
\subfloat{
    \includegraphics[width=\TextureCompSuppRes]{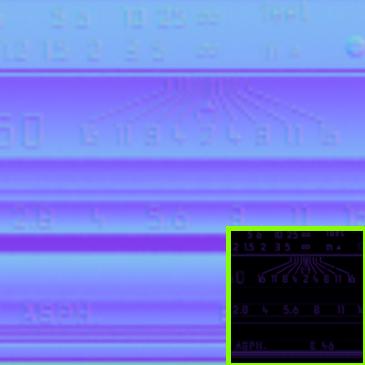}
  } \hspace{-0.18cm}
\subfloat{
    \includegraphics[width=\TextureCompSuppRes]{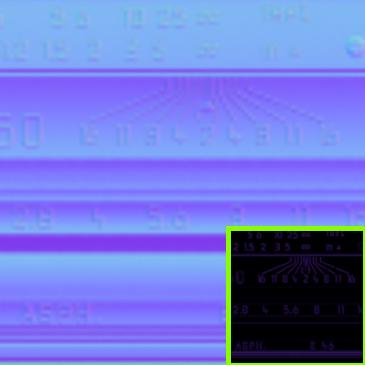}
  } \hspace{-0.18cm}
\subfloat{
    \includegraphics[width=\TextureCompSuppRes]{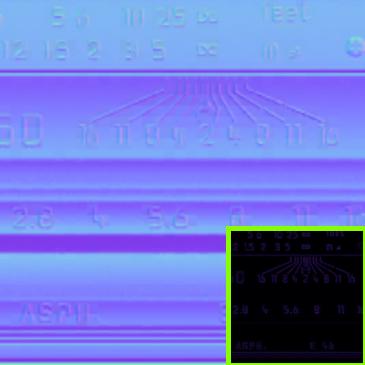}
  } \hspace{-0.18cm}
\subfloat{
    \includegraphics[width=\TextureCompSuppRes]{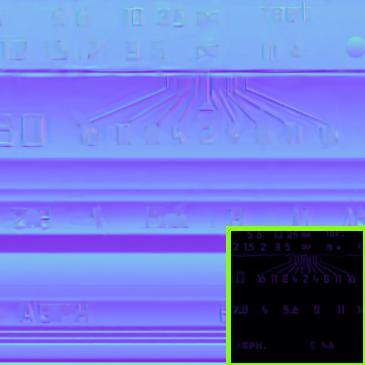}
  } \hspace{-0.18cm}
\subfloat{
    \includegraphics[width=\TextureCompSuppRes]{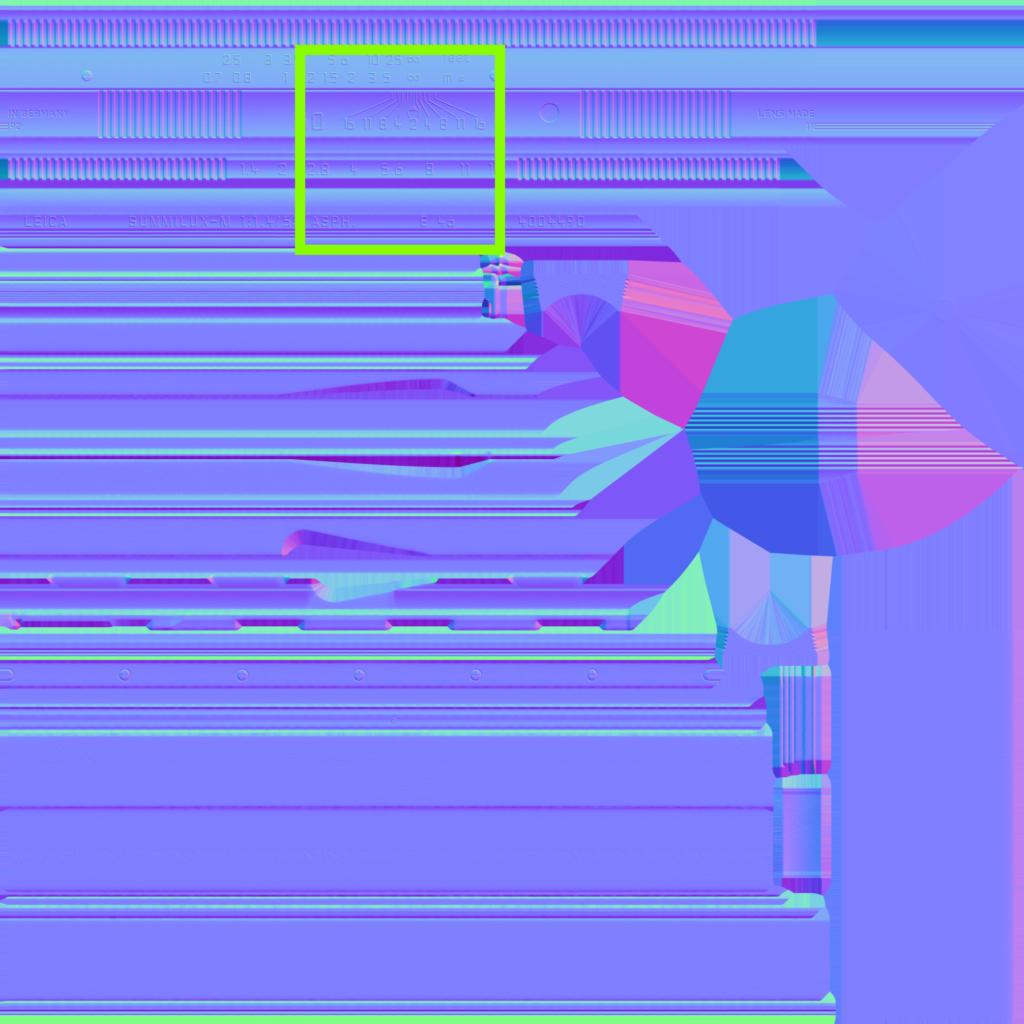}
  }
\vspace{0.2mm} \\
\setcounter{subfigure}{0}
\subfloat[BC1 (0.083 bppc)]{
    \includegraphics[width=\TextureCompSuppRes]{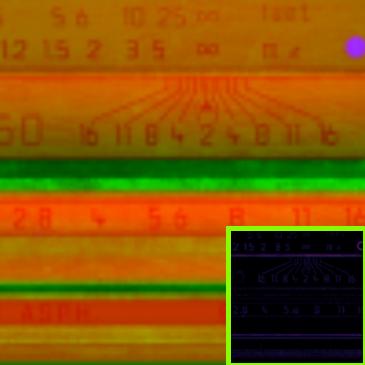}
  } \hspace{-0.18cm}
\subfloat[BC7 (0.167 bppc)]{
    \includegraphics[width=\TextureCompSuppRes]{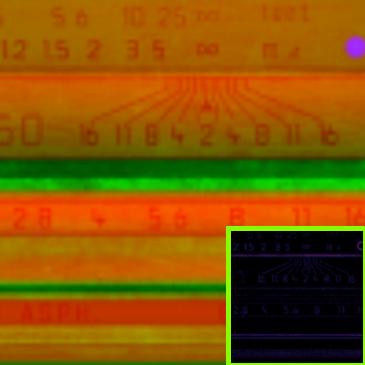}
  } \hspace{-0.18cm}
\subfloat[ASTC (0.074 bppc)]{
    \includegraphics[width=\TextureCompSuppRes]{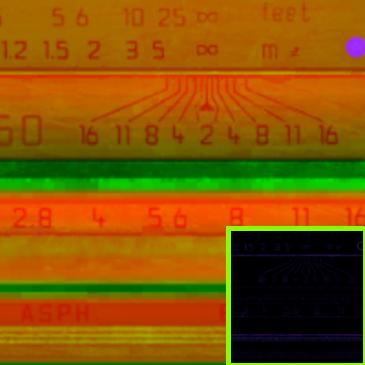}
  } \hspace{-0.18cm}
\subfloat[Ours (0.089 bppc)]{
    \includegraphics[width=\TextureCompSuppRes]{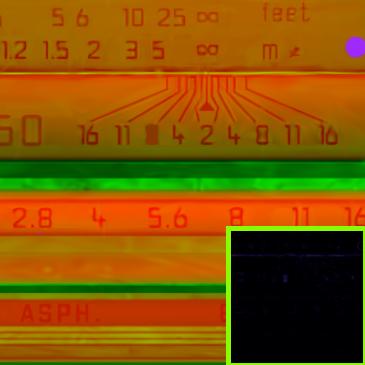}
  } \hspace{-0.18cm}
\subfloat[Reference]{
    \includegraphics[width=\TextureCompSuppRes]{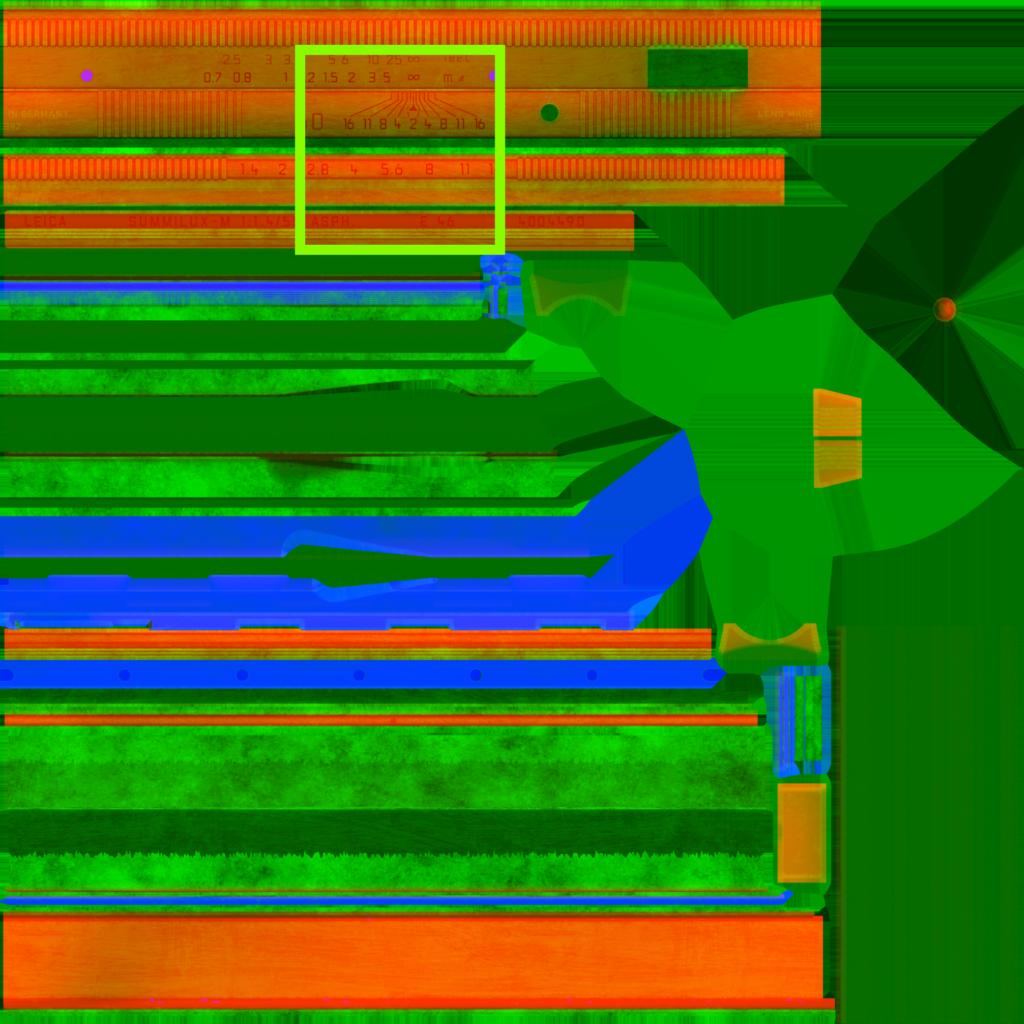}
  }
\Caption{\revise{Qualitative comparison against industry-standard GPU texture compression algorithms (\Cref{sec:evaluation-texture}).}}
{}
\end{figure*}
\begin{figure*}[h]
\centering
\subfloat{
    \includegraphics[width=\TextureCompSuppRes]{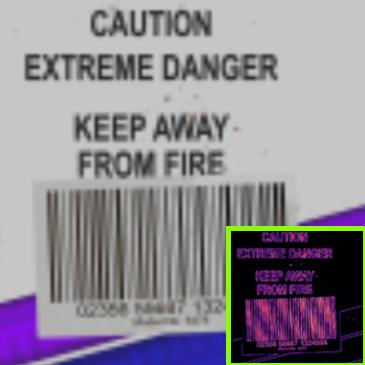}
  } \hspace{-0.18cm}
\subfloat{
    \includegraphics[width=\TextureCompSuppRes]{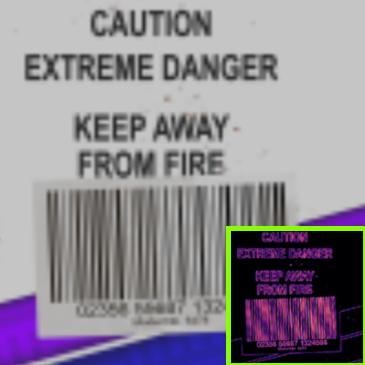}
  } \hspace{-0.18cm}
\subfloat{
    \includegraphics[width=\TextureCompSuppRes]{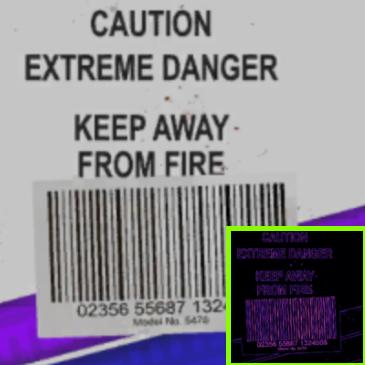}
  } \hspace{-0.18cm}
\subfloat{
    \includegraphics[width=\TextureCompSuppRes]{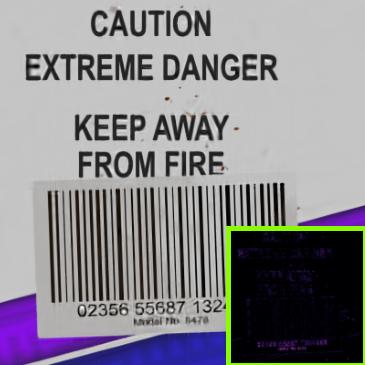}
  } \hspace{-0.18cm}
\subfloat{
    \includegraphics[width=\TextureCompSuppRes]{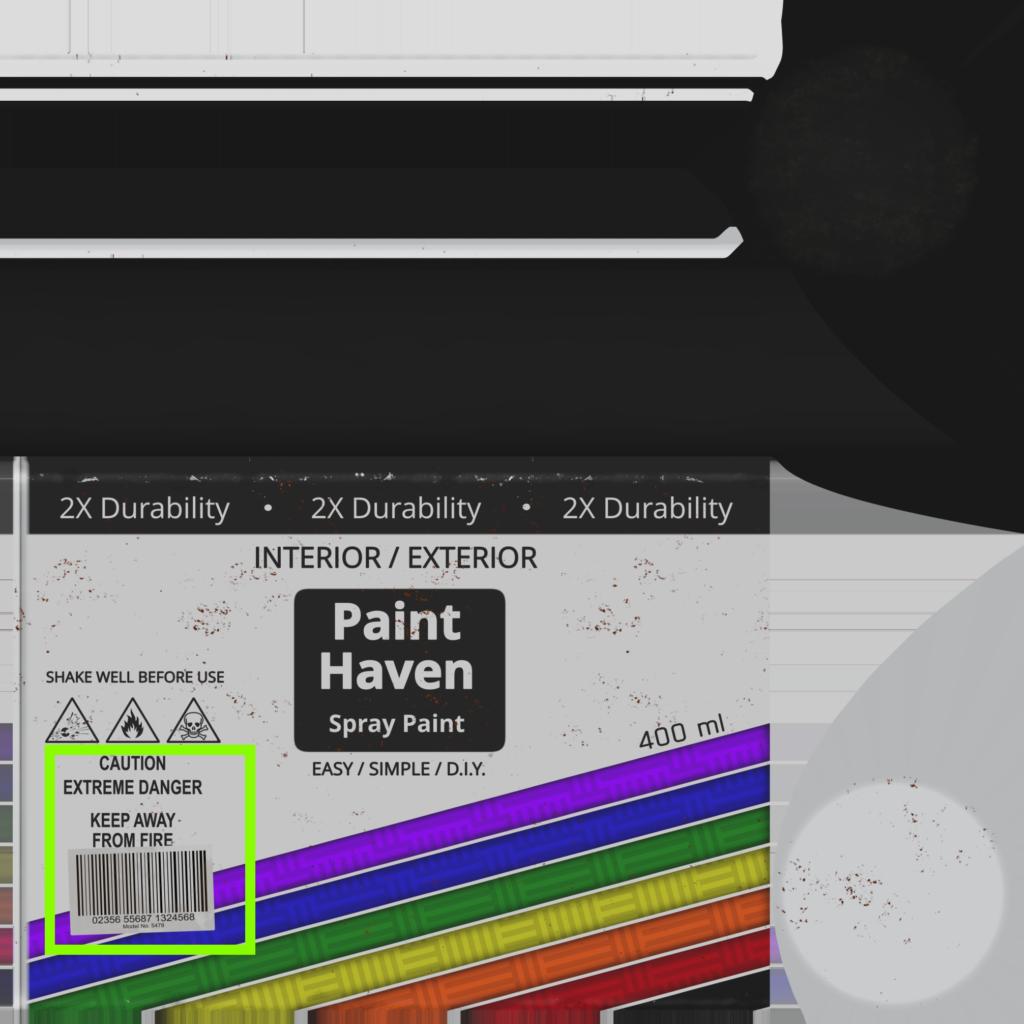}
  }
\vspace{0.2mm} \\
\subfloat{
    \includegraphics[width=\TextureCompSuppRes]{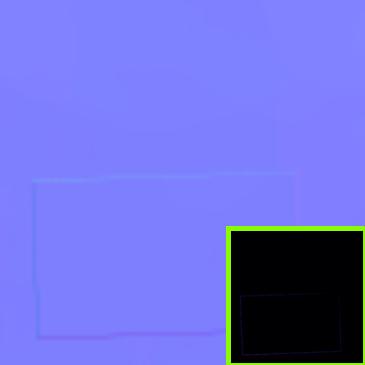}
  } \hspace{-0.18cm}
\subfloat{
    \includegraphics[width=\TextureCompSuppRes]{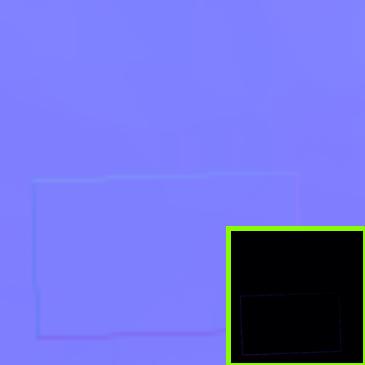}
  } \hspace{-0.18cm}
\subfloat{
    \includegraphics[width=\TextureCompSuppRes]{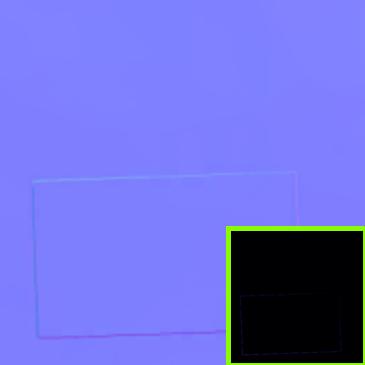}
  } \hspace{-0.18cm}
\subfloat{
    \includegraphics[width=\TextureCompSuppRes]{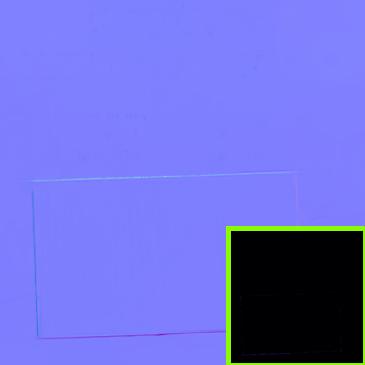}
  } \hspace{-0.18cm}
\subfloat{
    \includegraphics[width=\TextureCompSuppRes]{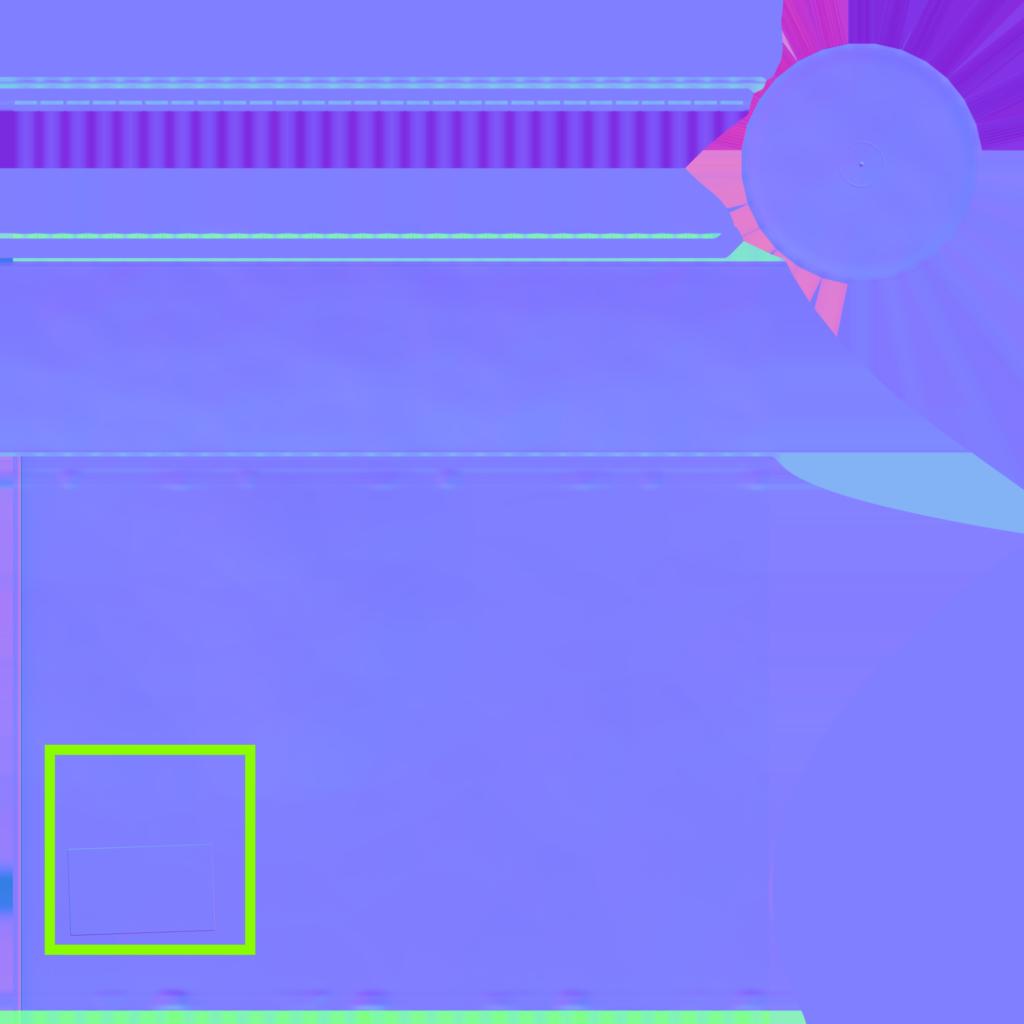}
  }
\vspace{0.2mm} \\
\subfloat{
    \includegraphics[width=\TextureCompSuppRes]{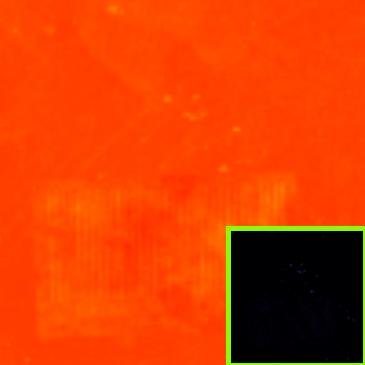}
  } \hspace{-0.18cm}
\subfloat{
    \includegraphics[width=\TextureCompSuppRes]{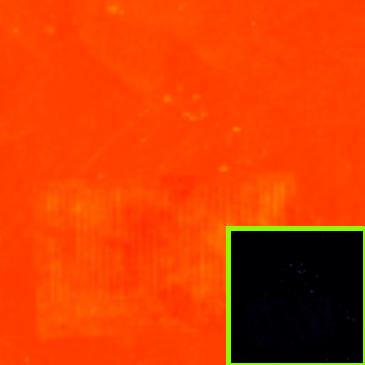}
  } \hspace{-0.18cm}
\subfloat{
    \includegraphics[width=\TextureCompSuppRes]{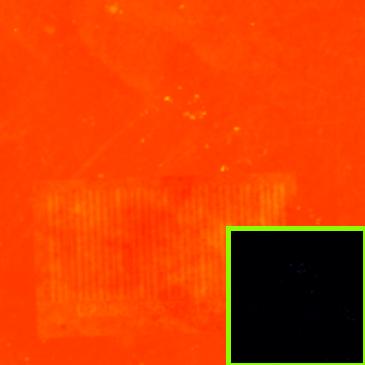}
  } \hspace{-0.18cm}
\subfloat{
    \includegraphics[width=\TextureCompSuppRes]{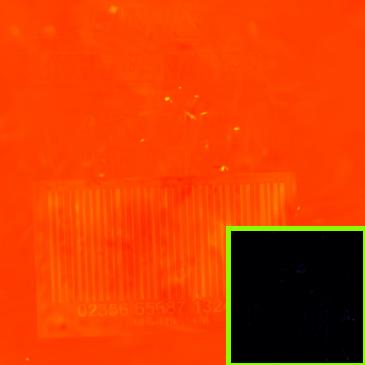}
  } \hspace{-0.18cm}
\subfloat{
    \includegraphics[width=\TextureCompSuppRes]{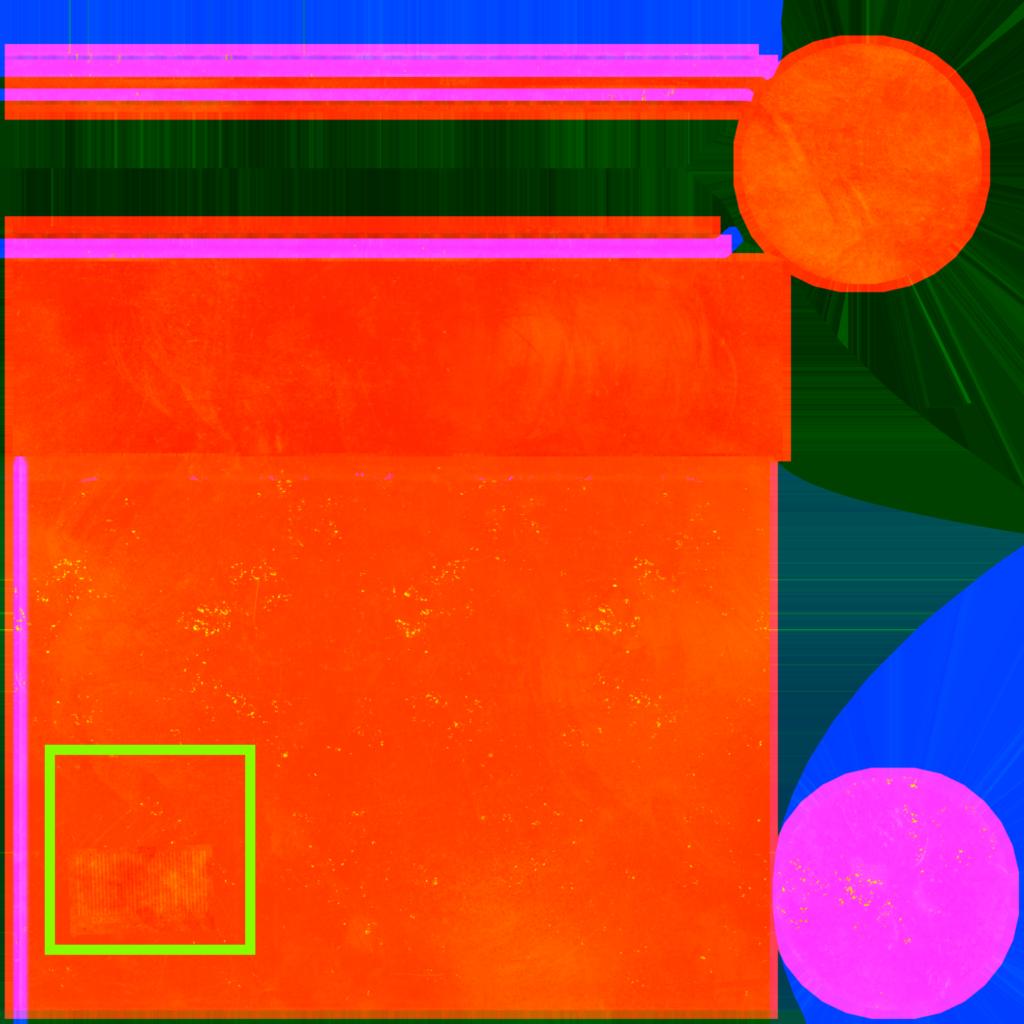}
  }
\vspace{0.2mm} \\
\subfloat{
    \includegraphics[width=\TextureCompSuppRes]{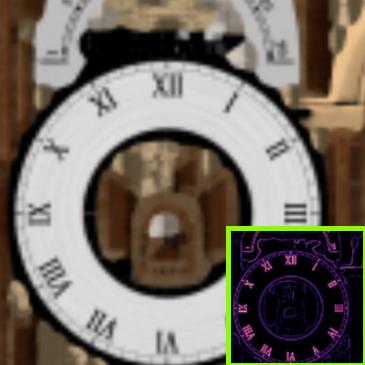}
  } \hspace{-0.18cm}
\subfloat{
    \includegraphics[width=\TextureCompSuppRes]{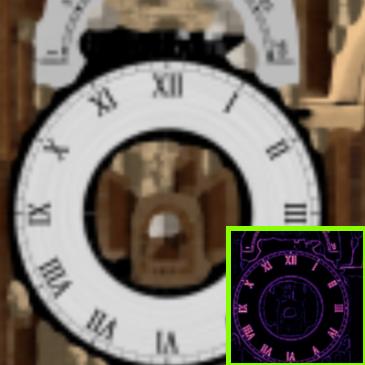}
  } \hspace{-0.18cm}
\subfloat{
    \includegraphics[width=\TextureCompSuppRes]{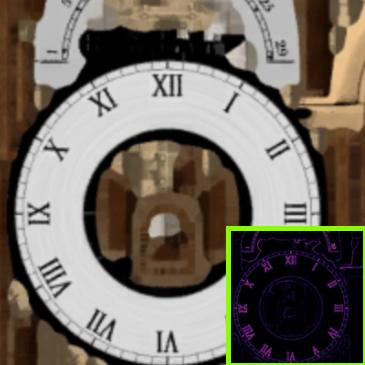}
  } \hspace{-0.18cm}
\subfloat{
    \includegraphics[width=\TextureCompSuppRes]{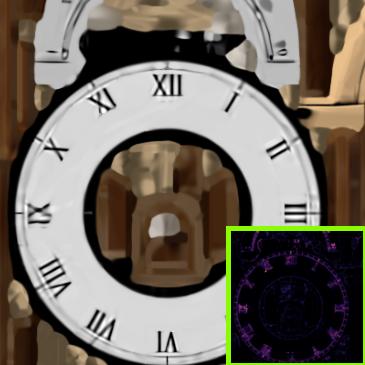}
  } \hspace{-0.18cm}
\subfloat{
    \includegraphics[width=\TextureCompSuppRes]{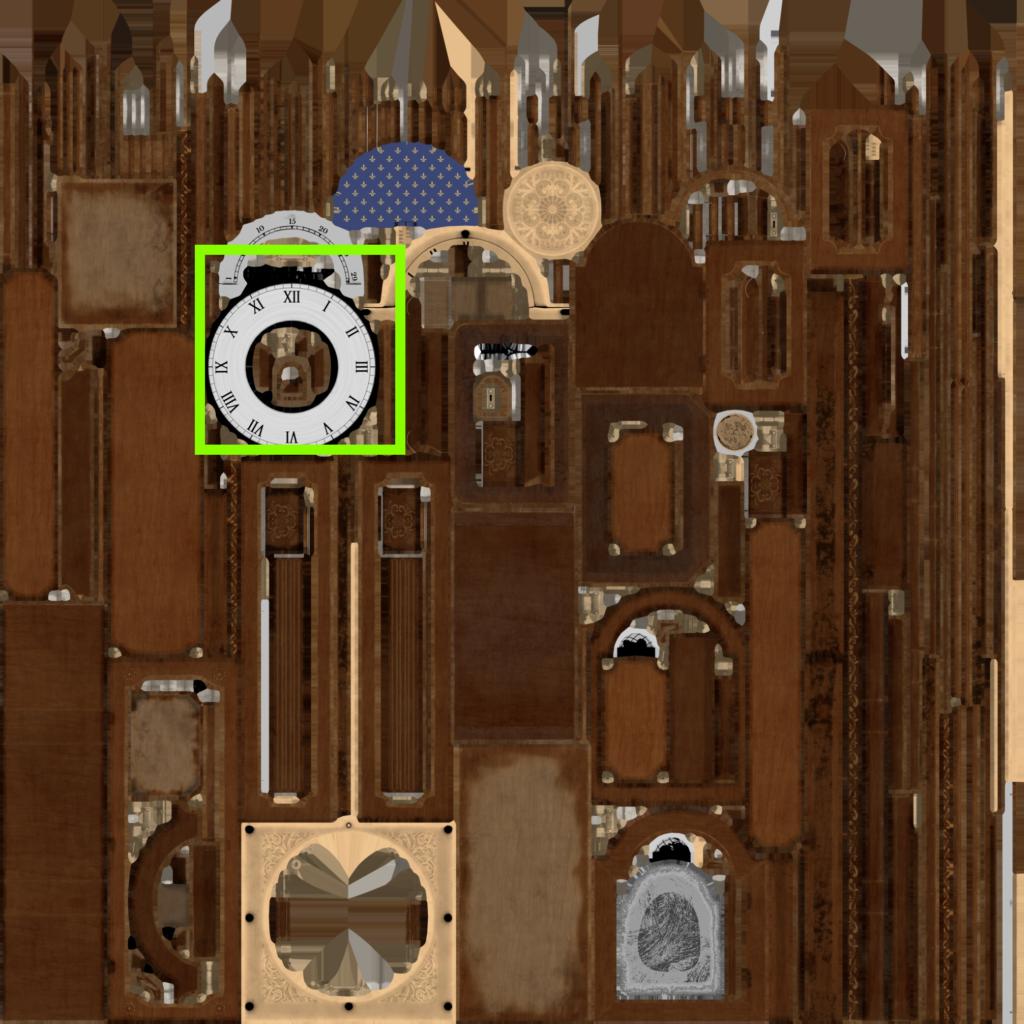}
  }
\vspace{0.2mm} \\
\subfloat{
    \includegraphics[width=\TextureCompSuppRes]{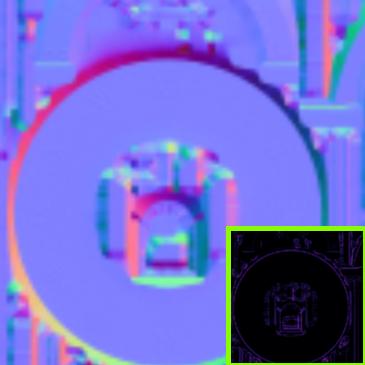}
  } \hspace{-0.18cm}
\subfloat{
    \includegraphics[width=\TextureCompSuppRes]{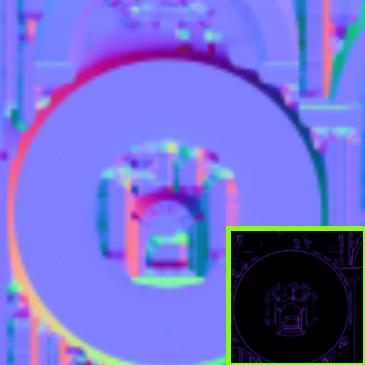}
  } \hspace{-0.18cm}
\subfloat{
    \includegraphics[width=\TextureCompSuppRes]{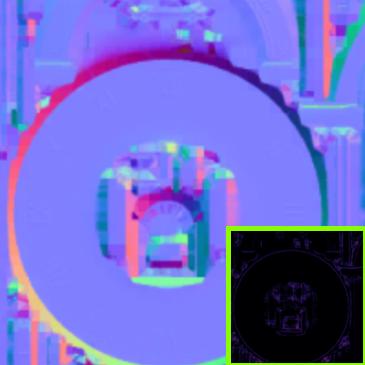}
  } \hspace{-0.18cm}
\subfloat{
    \includegraphics[width=\TextureCompSuppRes]{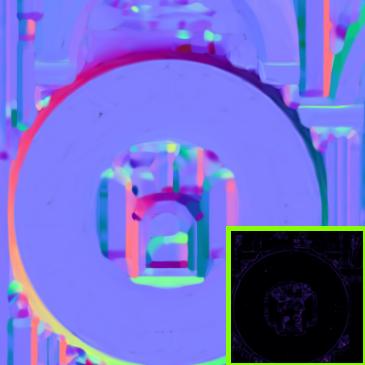}
  } \hspace{-0.18cm}
\subfloat{
    \includegraphics[width=\TextureCompSuppRes]{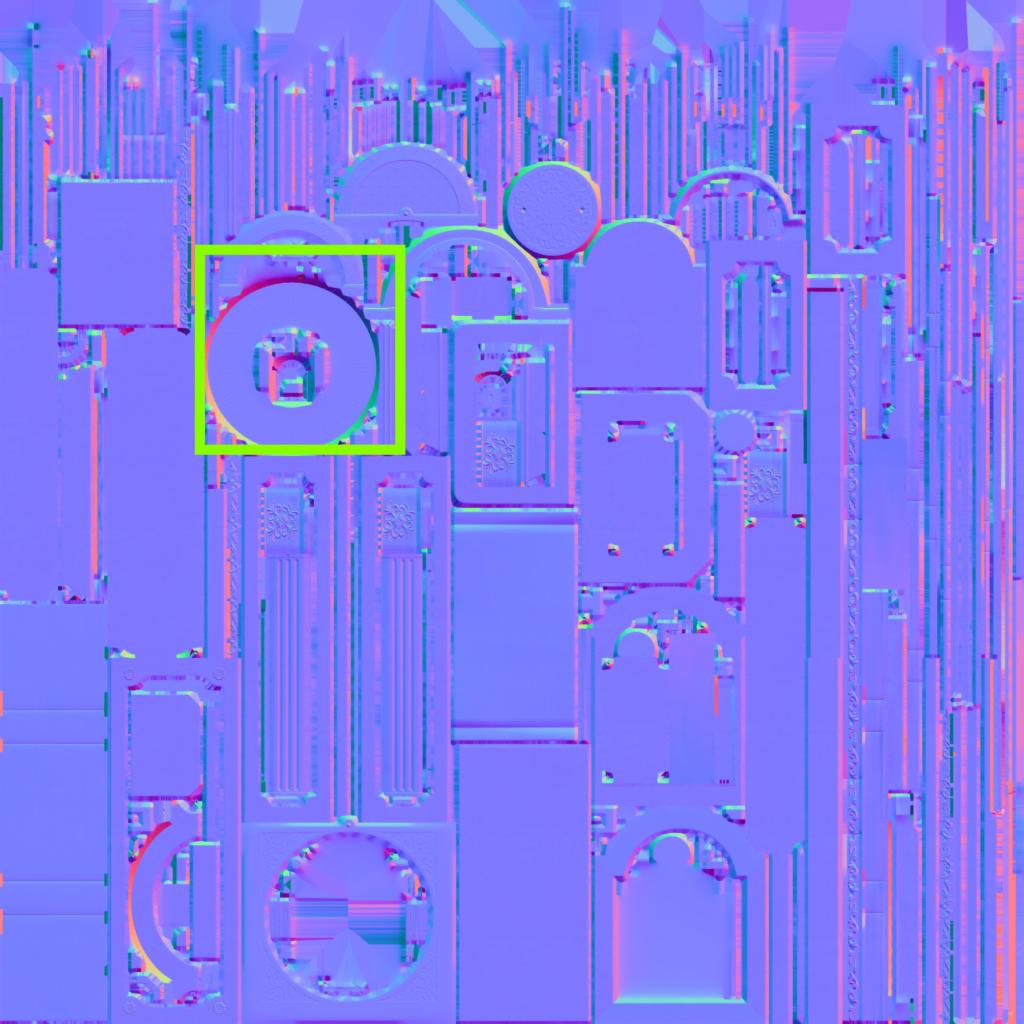}
  }
\vspace{0.2mm} \\
\setcounter{subfigure}{0}
\subfloat[BC1 (0.083 bppc)]{
    \includegraphics[width=\TextureCompSuppRes]{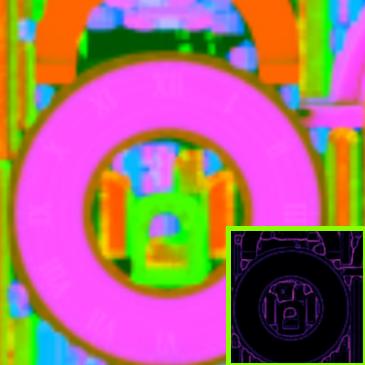}
  } \hspace{-0.18cm}
\subfloat[BC7 (0.167 bppc)]{
    \includegraphics[width=\TextureCompSuppRes]{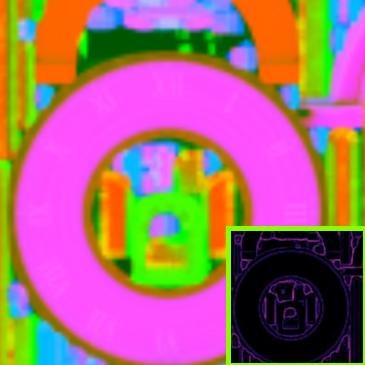}
  } \hspace{-0.18cm}
\subfloat[ASTC (0.074 bppc)]{
    \includegraphics[width=\TextureCompSuppRes]{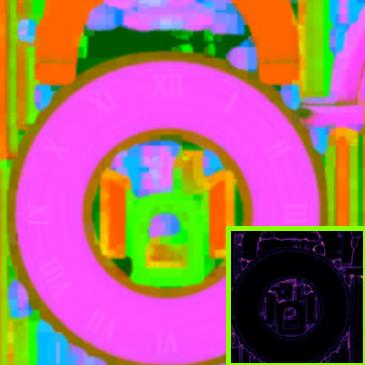}
  } \hspace{-0.18cm}
\subfloat[Ours (0.089 bppc)]{
    \includegraphics[width=\TextureCompSuppRes]{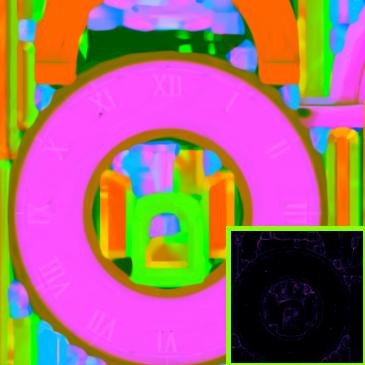}
  } \hspace{-0.18cm}
\subfloat[Reference]{
    \includegraphics[width=\TextureCompSuppRes]{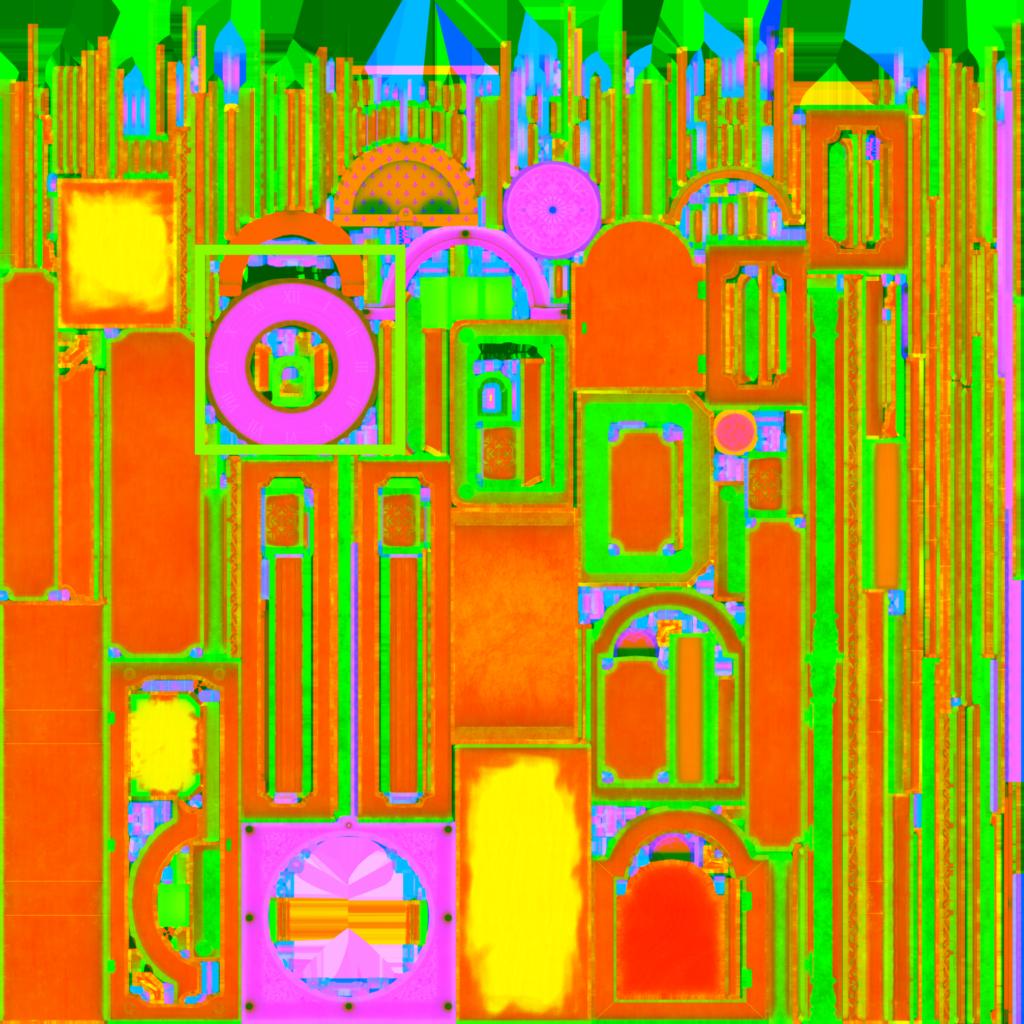}
  }
\Caption{\revise{Qualitative comparison against industry-standard GPU texture compression algorithms (\Cref{sec:evaluation-texture}).}}
{}
\end{figure*}
\begin{figure*}[h]
\centering
\subfloat{
    \includegraphics[width=\TextureCompSuppRes]{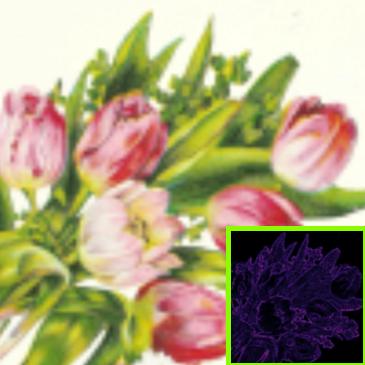}
  } \hspace{-0.18cm}
\subfloat{
    \includegraphics[width=\TextureCompSuppRes]{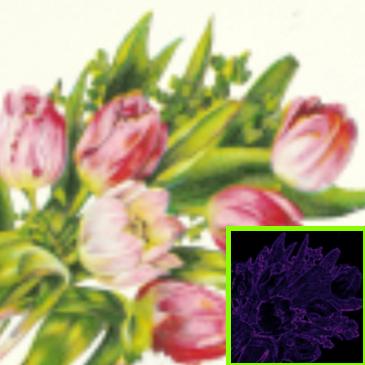}
  } \hspace{-0.18cm}
\subfloat{
    \includegraphics[width=\TextureCompSuppRes]{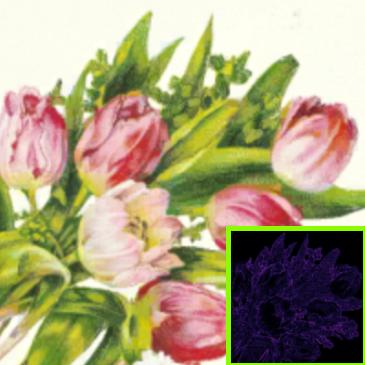}
  } \hspace{-0.18cm}
\subfloat{
    \includegraphics[width=\TextureCompSuppRes]{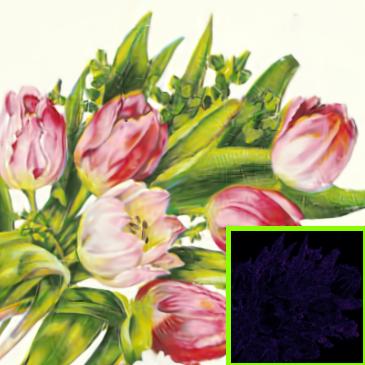}
  } \hspace{-0.18cm}
\subfloat{
    \includegraphics[width=\TextureCompSuppRes]{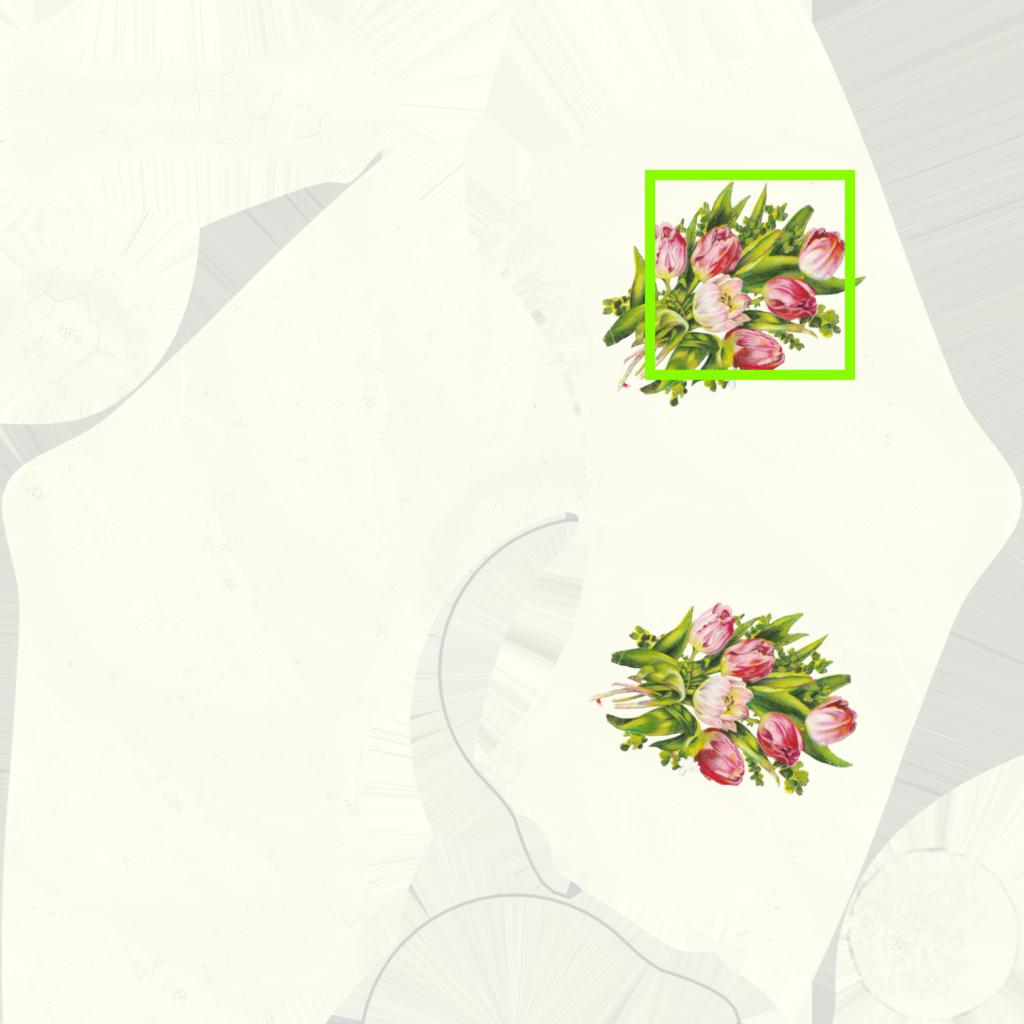}
  }
\vspace{0.2mm} \\
\subfloat{
    \includegraphics[width=\TextureCompSuppRes]{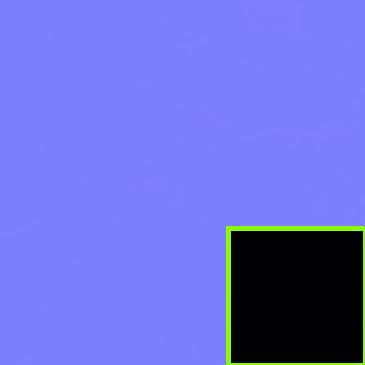}
  } \hspace{-0.18cm}
\subfloat{
    \includegraphics[width=\TextureCompSuppRes]{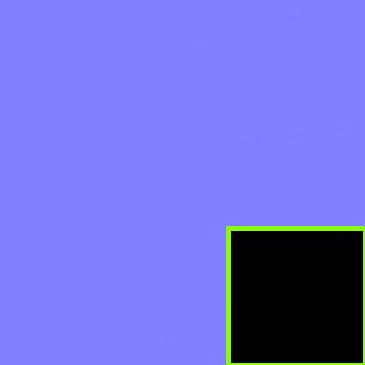}
  } \hspace{-0.18cm}
\subfloat{
    \includegraphics[width=\TextureCompSuppRes]{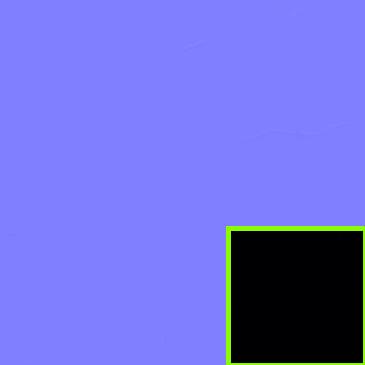}
  } \hspace{-0.18cm}
\subfloat{
    \includegraphics[width=\TextureCompSuppRes]{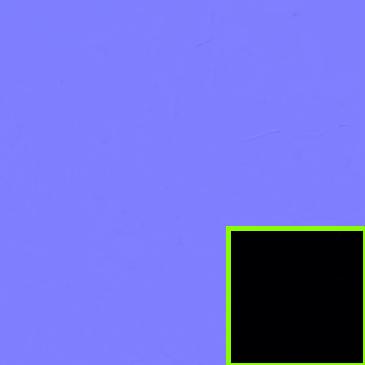}
  } \hspace{-0.18cm}
\subfloat{
    \includegraphics[width=\TextureCompSuppRes]{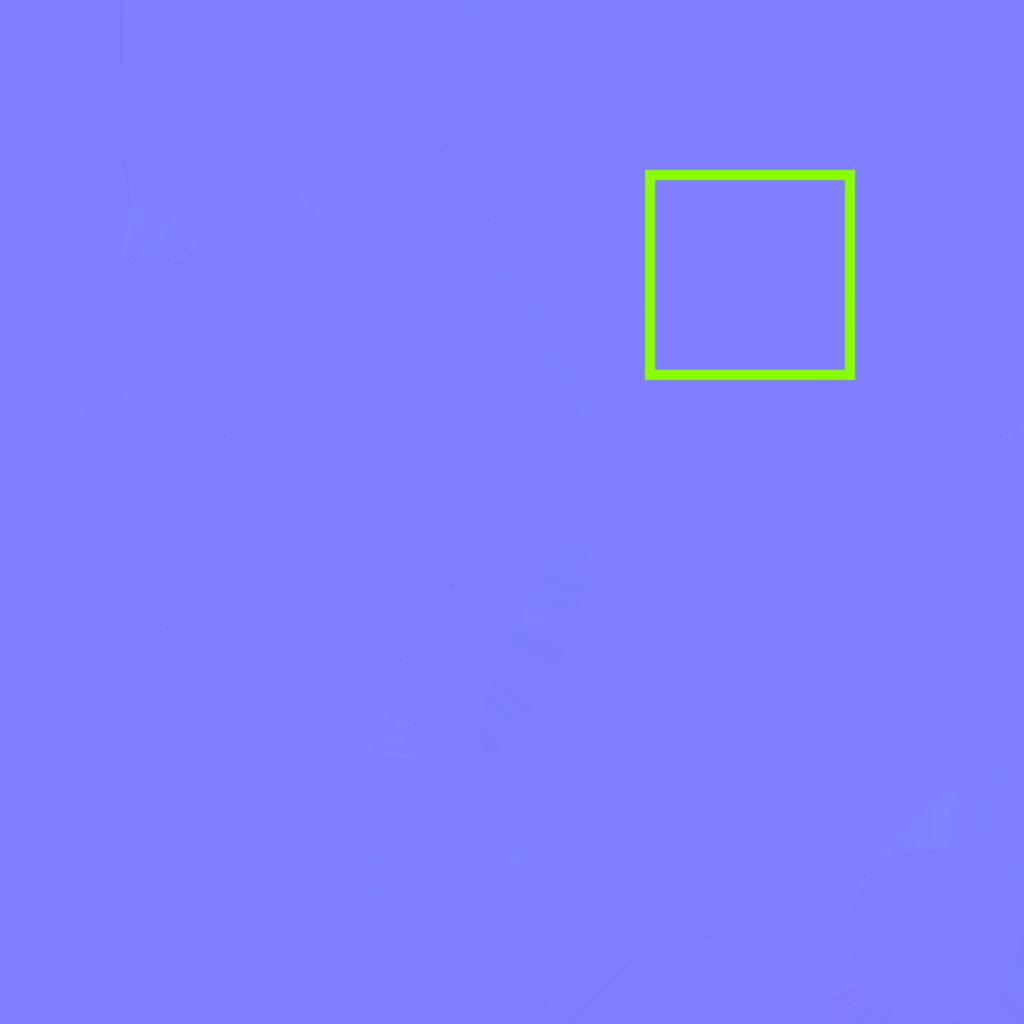}
  }
\vspace{0.2mm} \\
\subfloat{
    \includegraphics[width=\TextureCompSuppRes]{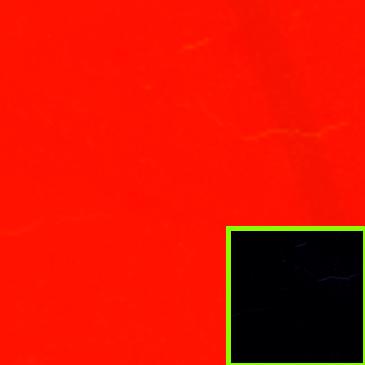}
  } \hspace{-0.18cm}
\subfloat{
    \includegraphics[width=\TextureCompSuppRes]{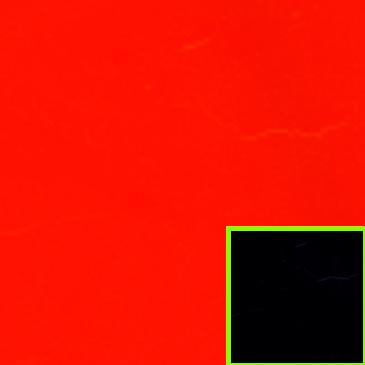}
  } \hspace{-0.18cm}
\subfloat{
    \includegraphics[width=\TextureCompSuppRes]{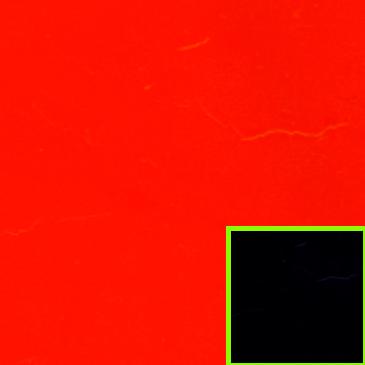}
  } \hspace{-0.18cm}
\subfloat{
    \includegraphics[width=\TextureCompSuppRes]{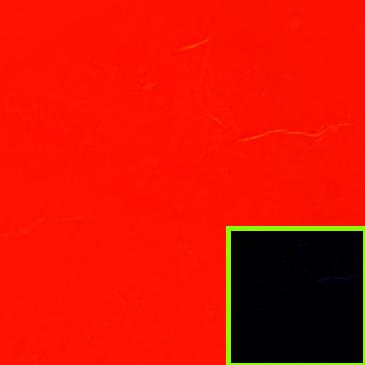}
  } \hspace{-0.18cm}
\subfloat{
    \includegraphics[width=\TextureCompSuppRes]{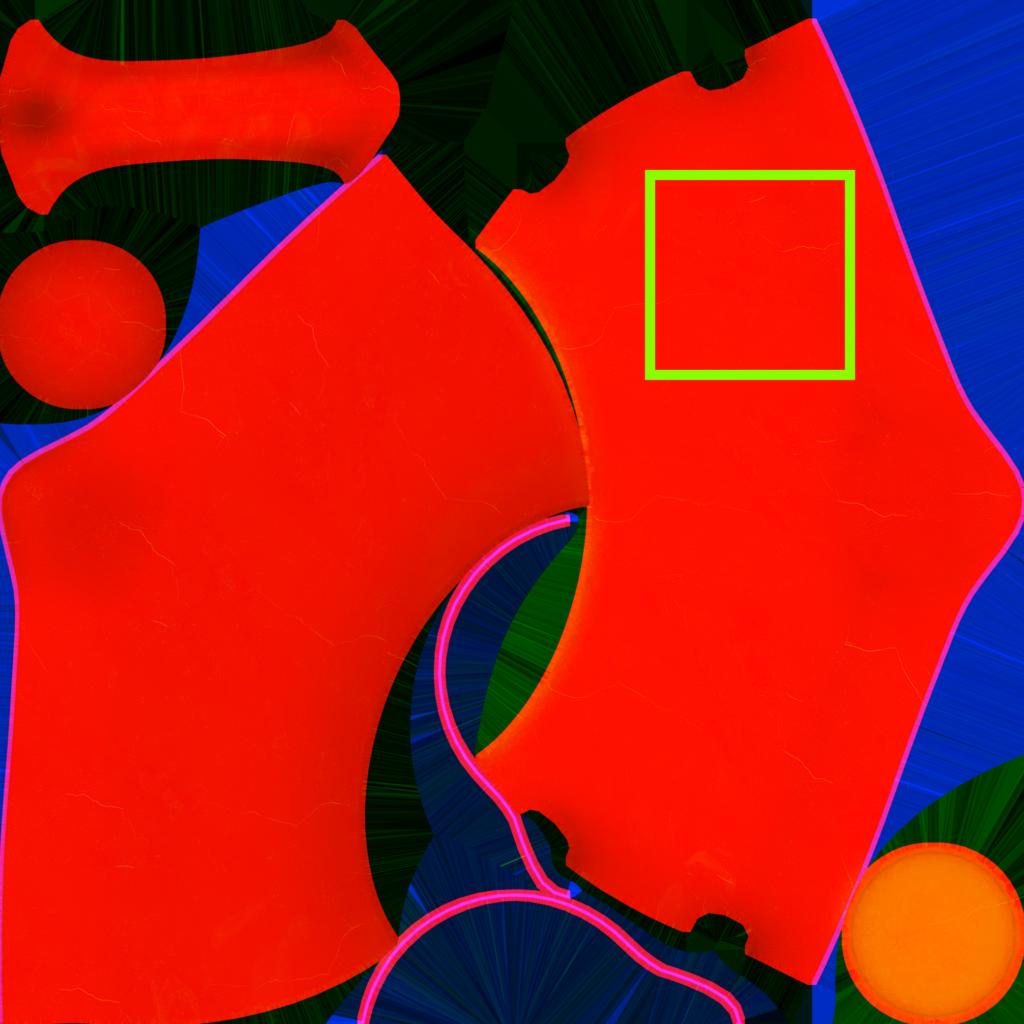}
  }
\vspace{0.2mm} \\
\subfloat{
    \includegraphics[width=\TextureCompSuppRes]{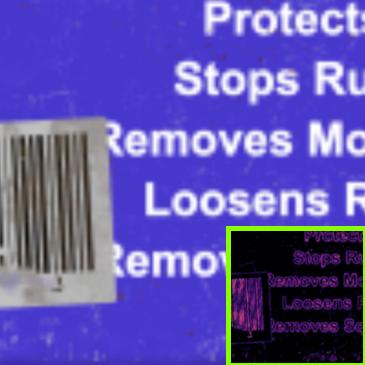}
  } \hspace{-0.18cm}
\subfloat{
    \includegraphics[width=\TextureCompSuppRes]{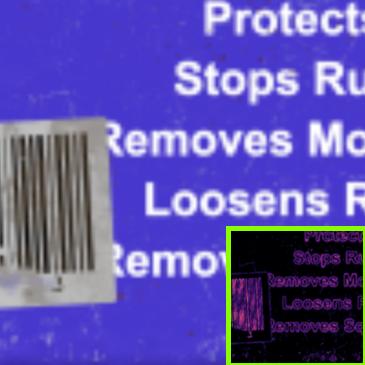}
  } \hspace{-0.18cm}
\subfloat{
    \includegraphics[width=\TextureCompSuppRes]{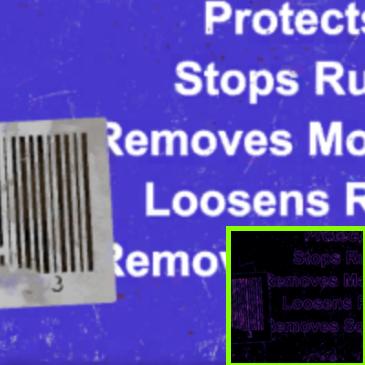}
  } \hspace{-0.18cm}
\subfloat{
    \includegraphics[width=\TextureCompSuppRes]{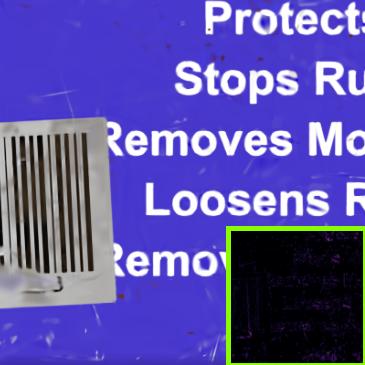}
  } \hspace{-0.18cm}
\subfloat{
    \includegraphics[width=\TextureCompSuppRes]{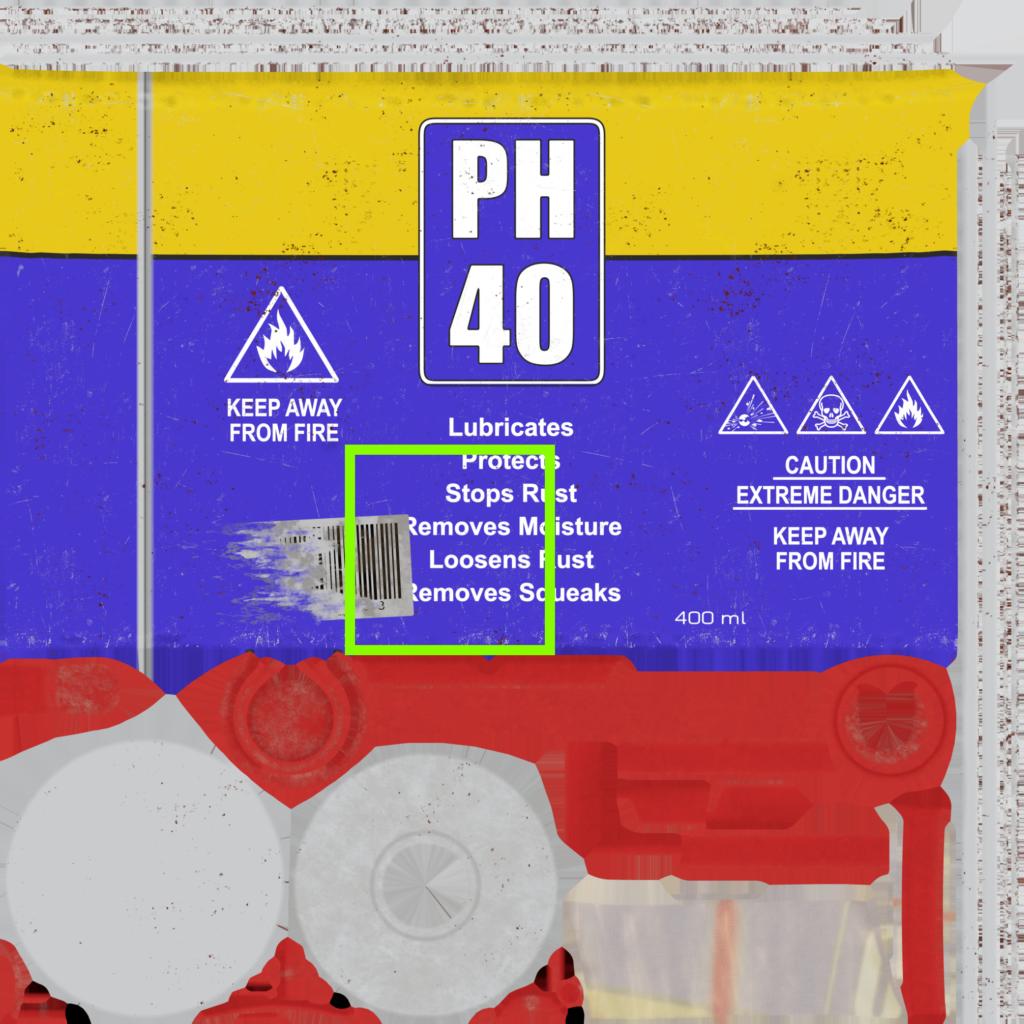}
  }
\vspace{0.2mm} \\
\subfloat{
    \includegraphics[width=\TextureCompSuppRes]{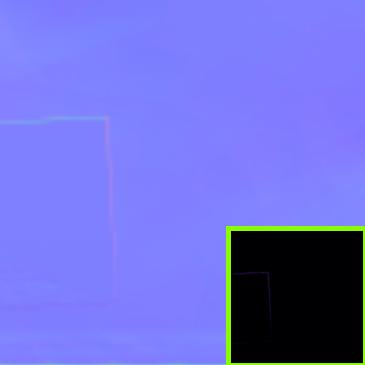}
  } \hspace{-0.18cm}
\subfloat{
    \includegraphics[width=\TextureCompSuppRes]{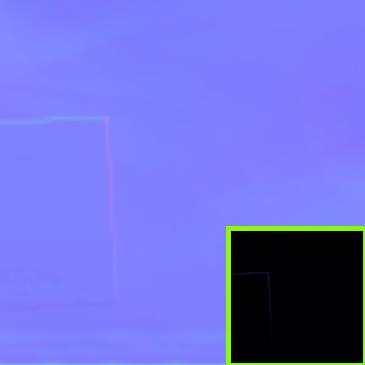}
  } \hspace{-0.18cm}
\subfloat{
    \includegraphics[width=\TextureCompSuppRes]{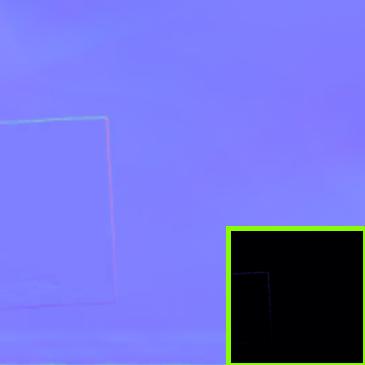}
  } \hspace{-0.18cm}
\subfloat{
    \includegraphics[width=\TextureCompSuppRes]{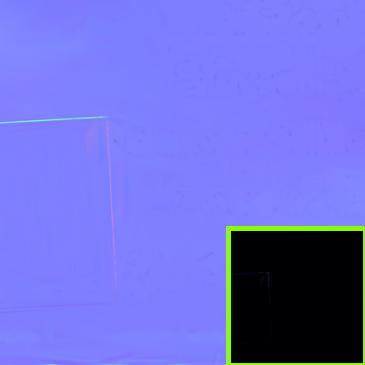}
  } \hspace{-0.18cm}
\subfloat{
    \includegraphics[width=\TextureCompSuppRes]{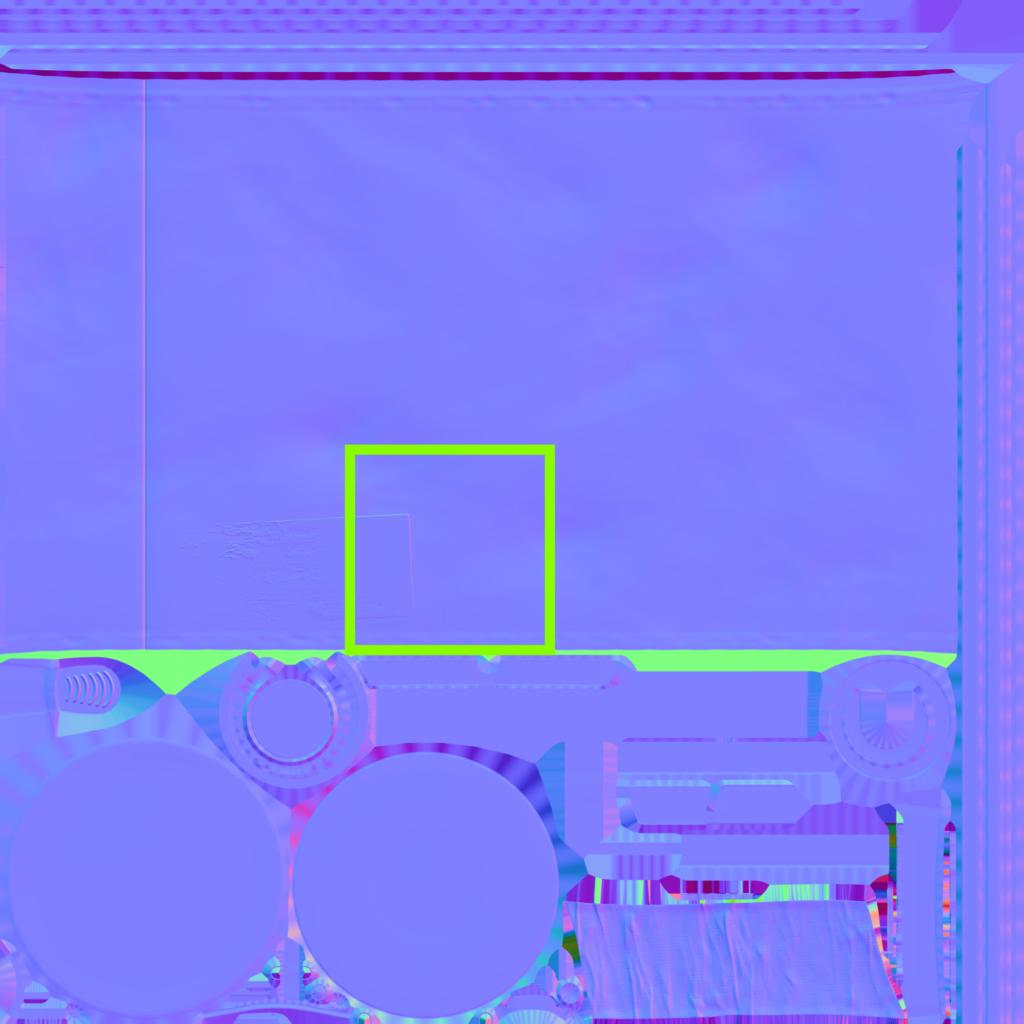}
  }
\vspace{0.2mm} \\
\setcounter{subfigure}{0}
\subfloat[BC1 (0.083 bppc)]{
    \includegraphics[width=\TextureCompSuppRes]{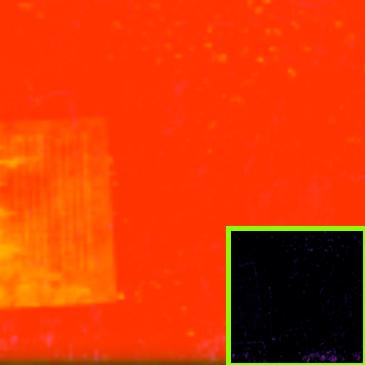}
  } \hspace{-0.18cm}
\subfloat[BC7 (0.167 bppc)]{
    \includegraphics[width=\TextureCompSuppRes]{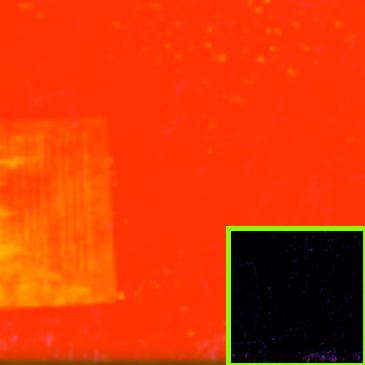}
  } \hspace{-0.18cm}
\subfloat[ASTC (0.074 bppc)]{
    \includegraphics[width=\TextureCompSuppRes]{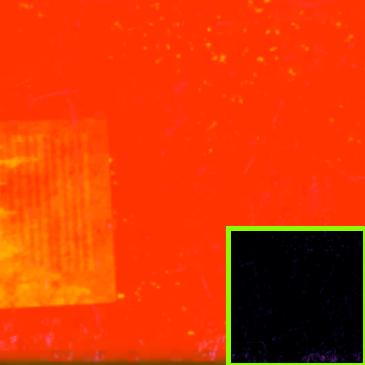}
  } \hspace{-0.18cm}
\subfloat[Ours (0.089 bppc)]{
    \includegraphics[width=\TextureCompSuppRes]{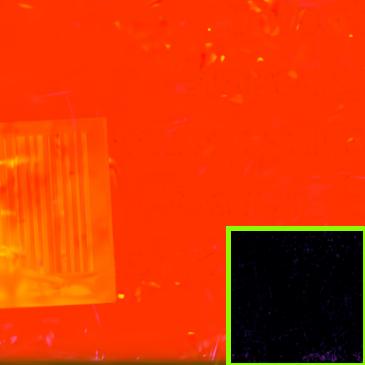}
  } \hspace{-0.18cm}
\subfloat[Reference]{
    \includegraphics[width=\TextureCompSuppRes]{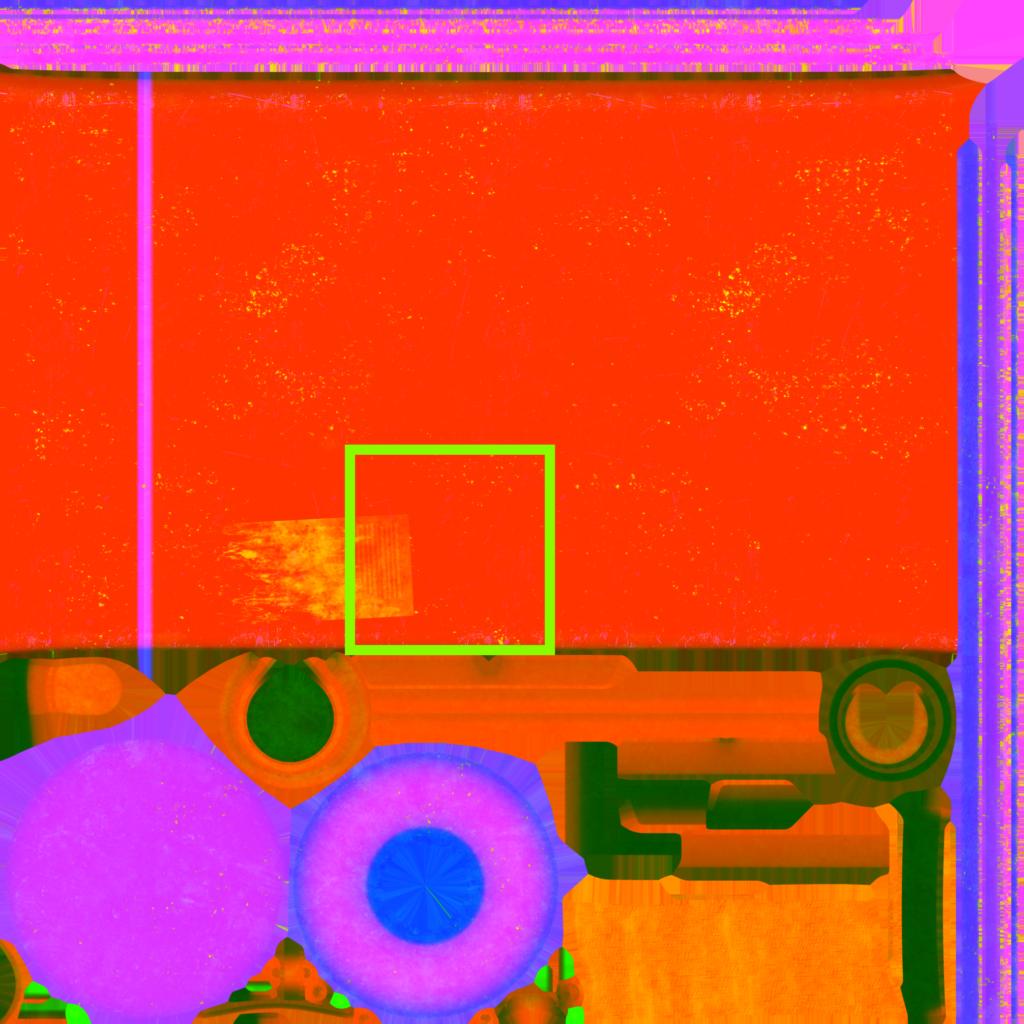}
  }
\Caption{\revise{Qualitative comparison against industry-standard GPU texture compression algorithms (\Cref{sec:evaluation-texture}).}}
{}
\end{figure*}

\end{appendices}

\end{document}